\documentclass[lettersize,journal]{IEEEtran}
\usepackage{amsmath,amsfonts}
\usepackage{bbding}
\usepackage{algorithmic}
\usepackage{algorithm}
\usepackage{array}
\usepackage[caption=false,font=normalsize,labelfont=sf,textfont=sf]{subfig}
\usepackage{textcomp}
\usepackage{stfloats}
\usepackage{url}
\usepackage{verbatim}
\usepackage{graphicx}
\usepackage{cite}
\usepackage{makecell}
\usepackage{bm}
\usepackage{soul} 
\usepackage{color, xcolor}
\soulregister{\cite}7 
\soulregister{\ref}7 
\soulregister{\eqref}7 

\graphicspath{{Figures/}}

\begin{document}
	
	\title{Direct Visual Servoing Based on Discrete Orthogonal Moments}
	
	\author{Yuhan Chen, \IEEEmembership{Member, IEEE,} Max Q.-H. Meng, \IEEEmembership{Fellow, IEEE,} and Li Liu, \IEEEmembership{Member, IEEE}
		\thanks{This work was supported in part by Shenzhen Key Laboratory of Robotics Perception and Intelligence (ZDSYS20200810171800001), and in part by the Hong Kong RGC GRF under Grant 14220622 and 14204321. (Corresponding author: Li Liu.)}
		\thanks{Yuhan Chen, Max Q.-H. Meng, and Li Liu are with the Department of Electronics and Electrical Engineering, Southern University of Science and Technology, Shenzhen 518055, China. Li Liu is also with the Department of Electronic Engineering, the Chinese University of Hong Kong, Hong Kong 999077, China. (e-mail: chenyh7@sustech.edu.cn; max.meng@ieee.org; liu6@sustech.edu.cn)}}

	\maketitle
	
	\begin{abstract}
		This paper proposes a new approach to achieve direct visual servoing (DVS) based on discrete orthogonal moments (DOMs).
		DVS is performed in such a way that the extraction of geometric primitives, matching, and tracking steps in the conventional feature-based visual servoing pipeline can be bypassed. 
		Although DVS enables highly precise positioning, it suffers from a limited convergence domain and poor robustness due to the extreme nonlinearity of the cost function to be minimized and the presence of redundant data between visual features. 
		To tackle these issues, we propose a generic and augmented framework that considers DOMs as visual features. 
		By using the Tchebichef, Krawtchouk, and Hahn moments as examples, we not only present the strategies for adaptively tuning the parameters and order of the visual features but also exhibit an analytical formulation of the associated interaction matrix.
		Simulations demonstrate the robustness and accuracy of our approach, as well as its advantages over the state-of-the-art.
		Real-world experiments have also been performed to validate the effectiveness of our approach.
	\end{abstract}
	

	\section{Introduction}
	\IEEEPARstart{V}{isual} servoing (VS) refers to the use of the vision sensor data to control the motion of a robot \cite{fathian2018camera}.
	In a typical VS pipeline, two closely linked themes are subjects of active research \cite{bakthavatchalam2014improving}: the design of visual features associated with the robotic task to be realized and the control scheme with the chosen visual features such that the desired features are obtained during VS.
	The latter adopts the control scheme of ensuring an exponential decoupled decrease in error.
	The former employs the geometric primitives (points, straight lines, ellipses, and cylinders) as the visual features in image-based VS \cite{cong2022visual, shi2018adaptive, 7580657} or reconstructs the camera pose from geometric primitives as inputs for position-based VS \cite{shi2018collision, chen2022closed, heshmati2020self, 4084568}.
	The above approaches subject the image stream to an ensemble of measurement processes, including image processing, feature matching, and visual tracking steps, from which the visual features are determined \cite{bakthavatchalam2018direct}.
	Alternatively, a current method that bypasses these steps, namely Direct Visual Servoing (DVS), has been proposed over decade \cite{collewet2008visual, collewet2011photometric}.
	It simply employs the luminosity intensity of the overall image to perform the VS pipeline.
	The DVS technique has shown highly accurate positioning even for approximated depths, partial occlusions, and specular and low-textured environments. 
	Nevertheless, it suffers from a limited convergence domain and poor robustness due to the extreme nonlinearity of the cost function to be minimized and the presence of redundant data between visual features.
	
	Several methods have been reported to enhance the performance of the DVS approach, which are generally divided into two categories: learning-based and model-based.
	Typical learning-based DVS methods are presented in \cite{marchand2019subspace} and \cite{felton2022visual}.
	The scheme proposed in \cite{marchand2019subspace} projects the image onto an orthogonal basis derived from the Principal Component Analysis (PCA) algorithm.
	Recently, \cite{felton2022visual} developed a novel framework to perform VS in the latent space learned from a convolutional autoencoder (AE).
	AE has been revealed for its ability to compress redundant information into a compact code with better reconstruction than PCA-based techniques.
	However, these methods involve an offline learning process, e.g., \cite{marchand2019subspace} and \cite{felton2022visual} require learning the eigenspace and the encoded information, respectively.
  In addition, portability, data dependence, and lack of interpretability also serve as major limitations of learning-based methods.
	Instead, the model-based VS methods can bypass the above problems, and they include algorithms based on histogram \cite{bateux2016histograms, li2021kullback}, photometric moments \cite{bakthavatchalam2018direct}, photometric Gaussian mixtures (PGM) \cite{crombez2018visual}, and  Discrete Cosine Transform (DCT) \cite{marchand2020direct}, etc.
	All these methods extract global features by directly calculating the luminosity intensities of the overall image and have demonstrated superior VS results.
	Specifically, \cite{bateux2016histograms} reported an approach considering histograms as visual features.
	However, the method depicted in \cite{li2021kullback} can converge successfully with a faster convergence rate than \cite{bateux2016histograms}.
	However, the robustness of the proposed method in \cite{li2021kullback} is to be investigated.
	\cite{bakthavatchalam2018direct} proposes a general model for the photometric moments enhanced with spatial weighting to tackle the issue of the appearance and disappearance of portions of the scene from the camera field of view during the servo.
	Although this method can obtain a satisfactory camera trajectory, the methods proposed in \cite{crombez2018visual} and \cite{marchand2020direct} are superior to it in terms of robustness.
	The objective of PGM-based VS (PGM-VS) \cite{crombez2018visual} is to minimize the difference between the desired Gaussian mixture and the Gaussian mixture computed from the current image varying over time. 
	Although the robustness of this method has been demonstrated in numerous experiments, the fundamental parameters $\lambda_{gi}$ in the method rely heavily on empirical determinations, which limits its application.
	The method presented in \cite{marchand2020direct} is to transform, via the DCT, the image from the spatial to the frequency domain and then use the coefficients of the DCT to establish a new control law.
	The DCT is a discrete orthogonal basis, which is helpful for image compression and filtering.
	Hence, the DCT-based VS (DCT-VS) has higher robustness;
	nevertheless, this technique is not flexible enough to only consider global features without focusing on local information.
	In other words, it cannot be adjusted adaptively according to the various images.
	Therefore, we will propose a VS scheme with a large convergence region, strong robustness, and flexible parameter selection.
	
	Inspired by DCT-VS, we propose a generic and augmented DVS framework by taking discrete orthogonal moments (DOMs) as visual features into consideration. 
	Strictly speaking, both DCT and PCA belong to the subclass DOMs.
	DOMs are essentially the projection of the image on a discrete orthonormal basis.
	It is well noted that there is a large amount of redundant data between neighboring pixels, which can be effectively eliminated by orthogonal moments;
	however, the computation of continuous orthogonal moments, such as Legendre and Zernike \cite{lakshminarayanan2011zernike}, requires a coordinate transformation and a suitable approximation of the continuous integrals, thereby leading to further computational complexity and discretization errors \cite{mukundan2001discrete}.
	This is because DOMs can sufficiently address the above issues, such that they are employed to represent visual features.
        DOMs are subdivided into orthogonal on the non-uniform lattice and orthogonal on the uniform lattice \cite{Nikiforov1991}.
        The latter can be directly defined on the image grid, but the former needs to introduce an intermediate, non-uniform lattice \cite{zhu2007image}.
        Hence, We deployed the latter into the VS, such as Tchebichef (Chebyshev), Krawtchouk, and Hahn moments (TMs, KMs, and HMs).
	Such three types of moments have similar properties to DCT.
	In particular, HMs are more flexible in parameter tuning to consider global features and local information.
	
	The main contributions of this paper are as follows:
	\begin{itemize}
		\item{we propose a generic and augmented DVS framework by considering DOMs as visual features and provide an approach to calculate the order of moments in the VS;}
		\item{we present an analytical formulation of the associated interaction matrix;}
		\item{we indicate how to determine the relevant parameters when KMs and HMs are utilized for VS;}
		\item{we confirm through various simulations and robotic VS experiments that these methods allow for large displacements and a satisfactory decrease of the error norm.}
	\end{itemize}
	
	The rest of the paper is organized in the following manner. 
	Section \ref{sec: Visual_Feature} presents the formulation of the DOMs and the associated VS features.
	Section \ref{sec: Adaptive_Parameter} provides an adaptive selection of the associated parameters for KMs and HMs when employed as visual features. 
	The order of DOMs is also investigated when it is used as visual features.
	Afterward, Section \ref{sec: Model_Interaction} elaborates on the derivation of the interaction matrix related to the DOMs feature.
	Subsequently, Section \ref{sec: Simulation_and_Experiment} validates the DOMs-based VS (DOM-VS) control scheme through various experiments conducted on both simulations and a real robotic arm platform.
	Finally, conclusions and future work are given in Section \ref{sec: Conclusion}.
	
	\section{DOMs as Visual Features} \label{sec: Visual_Feature}
	A set of DOMs computed from a digital image represents the global characteristics of the image shape and exhibits a large amount of information regarding the different geometric features of the image \cite{1998Moment}.
	Therefore, this section elaborates on the DOM representation of DVS as visual features.
	Sections \ref{sec: Discrete_Orthogonal_Polynomials} and \ref{sec: Computation_Polynomials} review the definitions and computations of the discrete orthogonal polynomials, namely Tchebichef, Krawtchouk, and Hahn polynomials, respectively.
	Then, Section \ref{sec: Relationship_polynomials} introduces the relations between these three polynomials.
	Finally, Section \ref{sec: Discrete_Orthogonal_Moment} elaborates on the DOM as a current compact representation of visual features. 
	
	\subsection{Discrete Orthogonal Polynomials} \label{sec: Discrete_Orthogonal_Polynomials}
	The set of polynomials that are orthogonal on the uniform lattice  $\{u=0,1,2,..., N-1\}$, Tchebichef, Krawtchouk, and Hahn polynomials, are discussed in this subsection.
	
	The discrete orthogonal polynomials $p_n(u)$ are defined as the polynomial solutions of the following difference equation
	\begin{equation}	\label{eq: discrete_orthogonal_polynomials_diff}
		\sigma(u) \Delta \nabla p_n(u) + \tau(u)  \Delta p_n(u) +\lambda_n  p_n(u) = 0,
	\end{equation}
	where $\Delta p_n(u) = p_n(u+1) - p_n(u) $ and $\nabla p_n(u) = p_n(u) - p_n(u-1)$ denote the forward and backward first-order difference operators, respectively.
	$\sigma(u)$ and $\tau(u)$ are the functions of the second and first degree, respectively. 
	$\lambda_n$ is an appropriate constant (see \cite{zhu2010general} for more details).
	The set of polynomials $\{p_n(u)\}$ with weight $w(u)$ and norm $\rho_n$ satisfies an orthogonality condition
	\begin{equation} \label{eq: orthogonality_condition}
		\sum_{u=0}^{s} p_n(u) p_m(u) w(u) = \rho(n) \delta_{nm},  \quad 0 \leq n, m \leq s,
	\end{equation}
	where $s$ is $N-1$ for discrete Tchebichef, Hahn polynomials and $N$ for Krawtchouk polynomials, and $\delta_{mn}$ denotes the Dirac function.
	Subsequently, the normalized discrete orthogonal polynomials can be obtained by appropriate weighting 
	\begin{equation} \label{eq: normalized orthogonal polynomials}
		\tilde{p}_n(u) = p_n(u) \sqrt{\frac{w(u)}{\rho(n)}}.
	\end{equation}
	Hence, the orthogonality condition in \eqref{eq: orthogonality_condition} can be re-expressed as 
	\begin{equation} \label{eq: orthogonality_condition_norm}
		\sum_{u=0}^{s} \tilde{p}_n(u) \tilde{p}_m(u) = \delta_{nm},  \quad 0 \leq n, m \leq s.
	\end{equation}
	The computation of normalized polynomials $\tilde{p}_n(u)$ is elaborated below.
	
	\subsection{Computation of Normalized Discrete Orthogonal Polynomials} \label{sec: Computation_Polynomials}
	
	The computation of normalized polynomials $\tilde{p}_n(u)$ has consistently been a significant concern \cite{mukundan2001image, yap2003image,  mukundan2004some}.
	The numerical instability can, therefore, quickly occur in evaluating such polynomials if the recurrence relations are not correctly used.
	The $u$ recurrence relation  is more advantageous than the $n$ recurrence relation in avoiding error accumulation in the result \cite{yap2007image}. 
	Hence, the following will introduce the recurrence relations concerning $u$ for these three polynomials: the normalized Tchebichef ($\tilde{t}_n$), Krawtchouk ($\tilde{k}_n$), and Hahn ($\tilde{h}_n$) polynomials.
	
	According to \eqref{eq: discrete_orthogonal_polynomials_diff} and \eqref{eq: normalized orthogonal polynomials}, the recurrence relations with respect to $u$ can be expressed as 
	\begin{equation} \label{eq: norm_polynomials}
		\begin{split}
			\tilde{p}_n(u)  =\frac{1}{\sigma(u-1) + \tau(u-1)} \Bigg(  \left( 2\sigma(u-1) + \tau(u-1) - \lambda_n \right)   \\ 
			\left. \sqrt{\frac{w(u)}{w(u-1)}} \tilde{p}_n(u-1)  - \sigma(u-1) \sqrt{\frac{w(u)}{w(u-2)}} \tilde{p}_n(u-2)\right) .
		\end{split}
	\end{equation}
	Referring to \cite{nikiforov1988special}, we present some information that facilitates the computation of \eqref{eq: norm_polynomials} in Table \ref{tab: polynomials_information} for normalized Tchebichef, Krawtchouk, and Hahn polynomials.
	And the initial values of the normalized polynomials, $\tilde{p}_n(0)$ and $\tilde{p}_n(1)$, are listed in Table \ref{tab: polynomials_initial}.
	
	\begin{table*} 
		\begin{center}
			\caption{Computational  information for the normalized Tchebichef $\tilde{t}_n(u; N)$, Krawtchouk $\tilde{k}_n(u; p, N)$, and Hahn $\tilde{h}_n(u; a, b, N)$ polynomials,\\
				($p \in (0,1)$ for Krawtchouk, and $a,b \in \mathbb{N} $ for Hahn).}
			\label{tab: polynomials_information}
			\begin{tabular}{| c | c | c | c |}
				\hline
				$\tilde{p}_n(u)$ & $\tilde{t}_n(u; N)$ &  $\tilde{k}_n(u; p, N)$ &$\tilde{h}_n(u; a, b, N)$ \\
				\hline
				$\sigma(u)$&  $u(N-u)$ &  $u$& $u(N+a-u)$\\
				\hline
				$\tau(u)$& $N - 1 - 2x$   &  $\frac{Np-u}{1-p}$& $(b+1)(N-1) - (a+b+2)u$\\ 
				\hline
				$\lambda_n$&  $n(n+1)$ & $\frac{n}{1-p}$& $n(a+b+n+1)$\\
				\hline 
				$\frac{w(u)}{w(u-1)}$& 1  & $\frac{p}{1-p} \frac{N-u+1}{u}$&$\frac{b+u}{u}\frac{N-u}{N+a-u}$ \\
				\hline			
				$\frac{w(u)}{w(u-2)}$& 1  &$\frac{p^2}{(1-p)^2} \frac{(N-u+1)(N-u+2)}{(u-1)u}$ &$\frac{(b+u-1)(b+u)}{(u-1)u} \frac{(N-u)(N-u+1)}{(N-u+a)(N-u+a+1)}$ \\
				\hline	
			\end{tabular}
		\end{center}
	\end{table*}
	
	\begin{table*} 
		\begin{center}
			\caption{Initial values of the recurrence relation concerning $u$ for the normalized Tchebichef, Krawtchouk, and Hahn polynomials,\\
				($p \in (0,1)$ for Krawtchouk and $a,b \in \mathbb{N}$ for Hahn).}
			\label{tab: polynomials_initial}
			\begin{tabular}{| c | c | c | }
				\hline
				$\tilde{p}_n(u)$ & $u=0$ &  $u=1$ \\
				\hline
				$\tilde{t}_n(u; N)$& \makecell[c]{$-\sqrt{\frac{N-n}{N+n}} \sqrt{\frac{2n+1}{2n-1}}\tilde{t}_{n-1}(0; N)$, \\$\tilde{t}_0(0; N) = \sqrt{\frac{1}{N}}$ } &$\left( 1-\frac{n(n+1)}{N-1}  \right) \tilde{t}_n(0; N) $   \\
				\hline
				$\tilde{k}_n(u; p, N)$&   \makecell[c]{$-\sqrt{\frac{N-n+1}{n}} \sqrt{\frac{p}{1-p}}\tilde{k}_{n-1}(0; p, N)$,\\ $\tilde{k}_{0}(0; p, N) = (1-p)^{N/2}$} &  $ \left( 1 - \frac{n}{Np}\right)  \sqrt{\frac{w(1)}{w(0)}} \tilde{k}_n(0; p, N)$ \\ 
				\hline
				$\tilde{h}_n(u; a, b, N)$ &\makecell[c]{$-\sqrt{\frac{(N-n)(n+b)(a+b+n)}{(a+n)(a+b+n+N)n}}\sqrt{\frac{a+b+2n+1}{a+b+2n-1}}\tilde{h}_{n-1}(0; a, b, N)$, \\  $\tilde{h}_{0}(0; a, b, N)=\sqrt{\prod_{i=1}^{b+1} \frac{a+i}{N+a+i-1}}$}   &$\left( 1 - \frac{n(n+a+b+1)}{(b+1)(N-1)}\right)  \sqrt{\frac{w(1)}{w(0)}} \tilde{h}_n(0; a, b, N)$  \\
				\hline 
			\end{tabular}
		\end{center}
	\end{table*}
	
	\subsection{Relationship among Normalized Tchebichef, Krawtchouk and Hahn polynomials} \label{sec: Relationship_polynomials} 
	
	It is noted that the normalized Tchebichef, Krawtchouk, and Hahn  polynomials are interrelated \cite{yap2007image}. 
	If we define 
	$p=b/(a+b)$ and $t = a + b$,  then the parameters in normalized Hahn polynomial can be expressed as 
	\begin{equation}
		\left\{\begin{array}{l}
			b=pt	\\ 
			a=(1-p)t.
		\end{array}\right.  \label{eq: ab}
	\end{equation}
	
	If $t \to 0$ or $t \to \infty$, the normalized Hahn polynomial is converted to normalized Tchebichef polynomial $(a=0, b=0)$  or Krawtchouk polynomial $(a+b \to \infty)$, respectively \cite{yap2007moments}
	\begin{align}
		\lim_{t \to 0} \tilde{h}_n(u; a, b, N) &= \tilde{t}_n(u; N),  \notag\\
		\lim_{t \to \infty} \tilde{h}_n(u; a, b, N) &= \tilde{k}_n(u; p, N).
	\end{align}
	
	\begin{figure}
		\centering 
		
		\subfloat[]{\includegraphics[width=0.33\hsize]{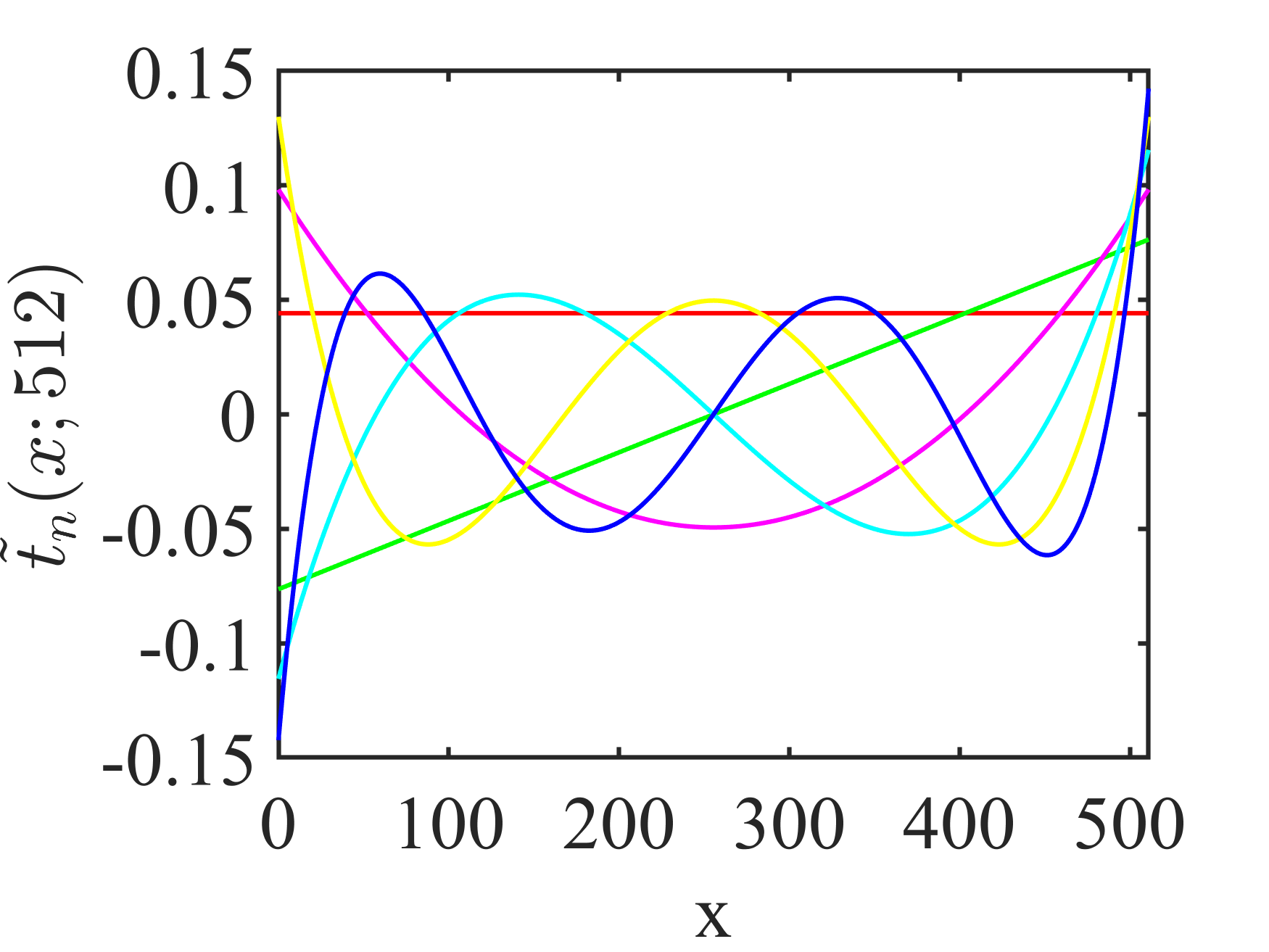}  \label{fig: Tchebichef_order_5_N_512}}	
		\subfloat[]{\includegraphics[width=0.33\hsize]{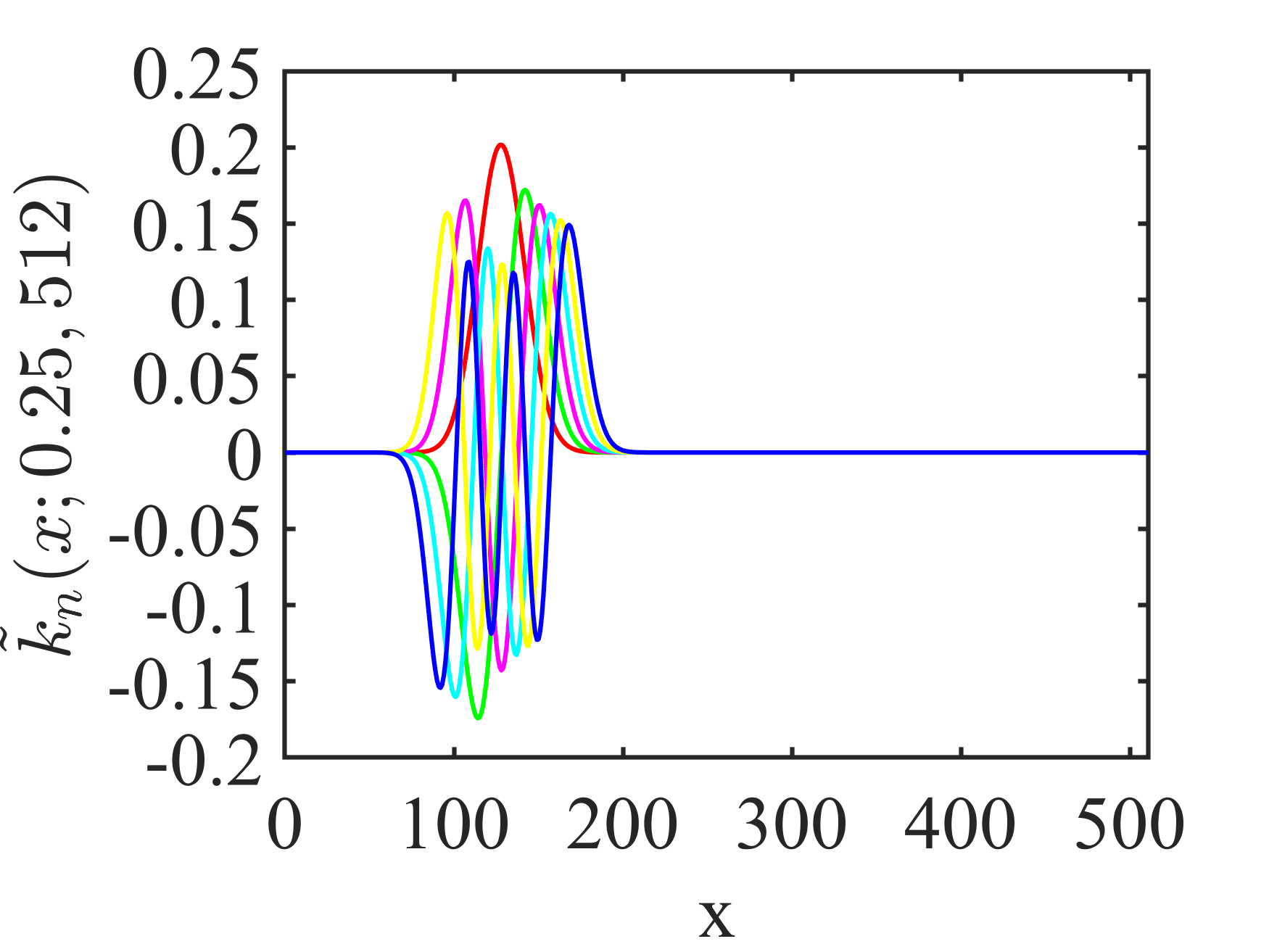}  \label{fig: Krawtchouk_order_5_N_512_p_quarter}}	
		\subfloat[]{\includegraphics[width=0.33\hsize]{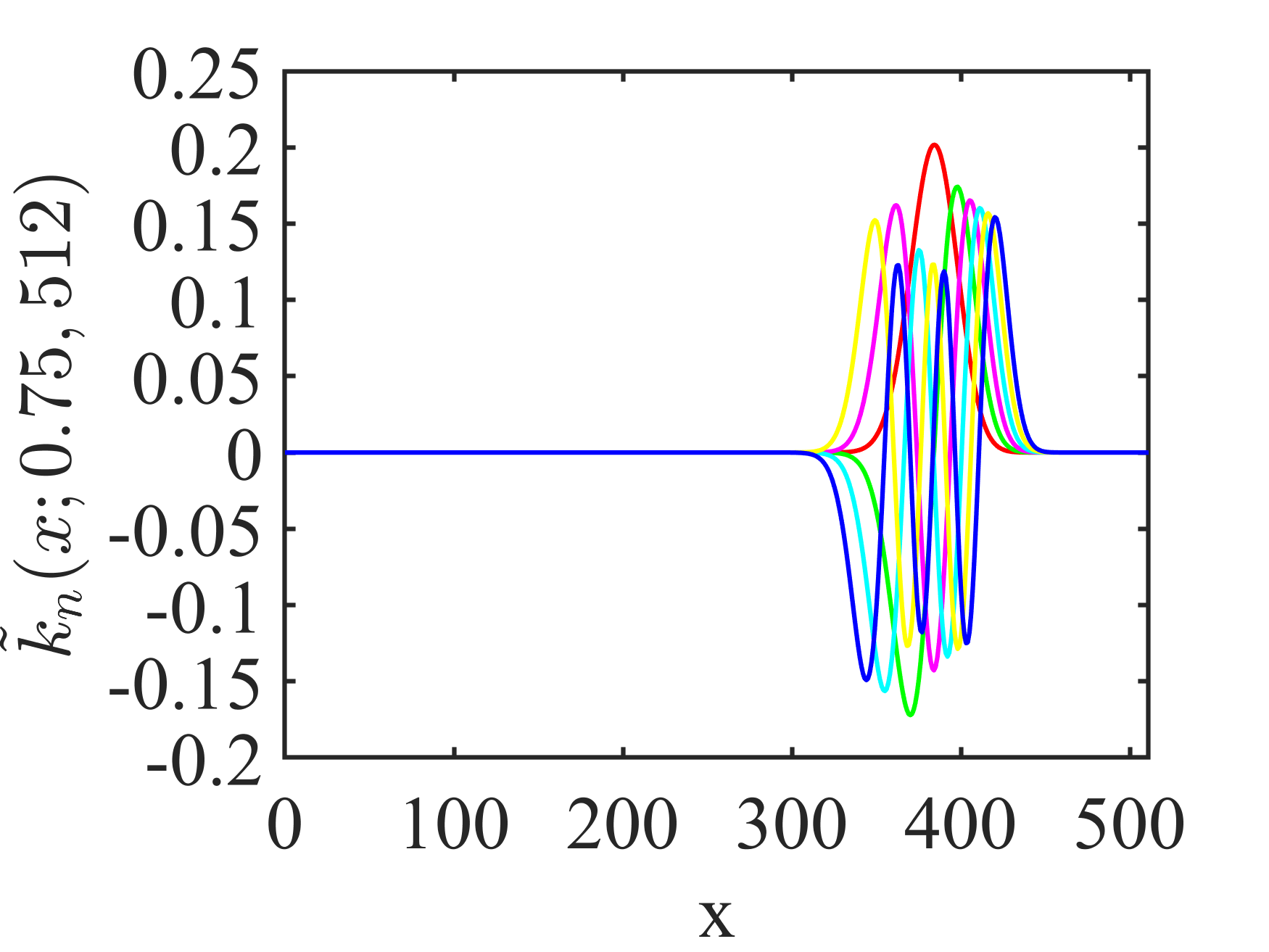}  \label{fig: Krawtchouk_order_5_N_512_p_3quarter}}

		\subfloat[]{\includegraphics[width=0.33\hsize]{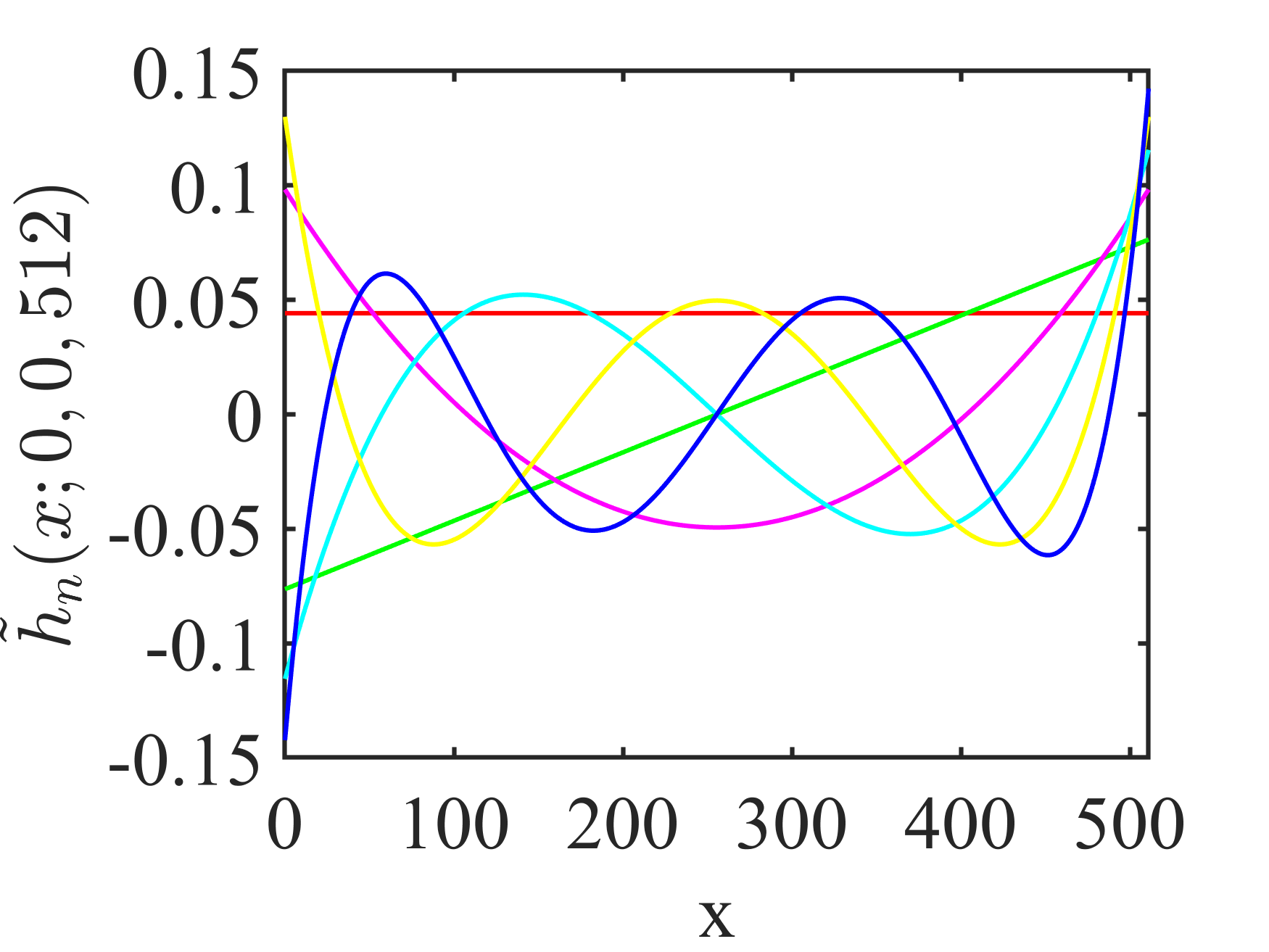}  \label{fig: Hahn_order_5_N_512_a_0_b_0}}
		\subfloat[]{\includegraphics[width=0.33\hsize]{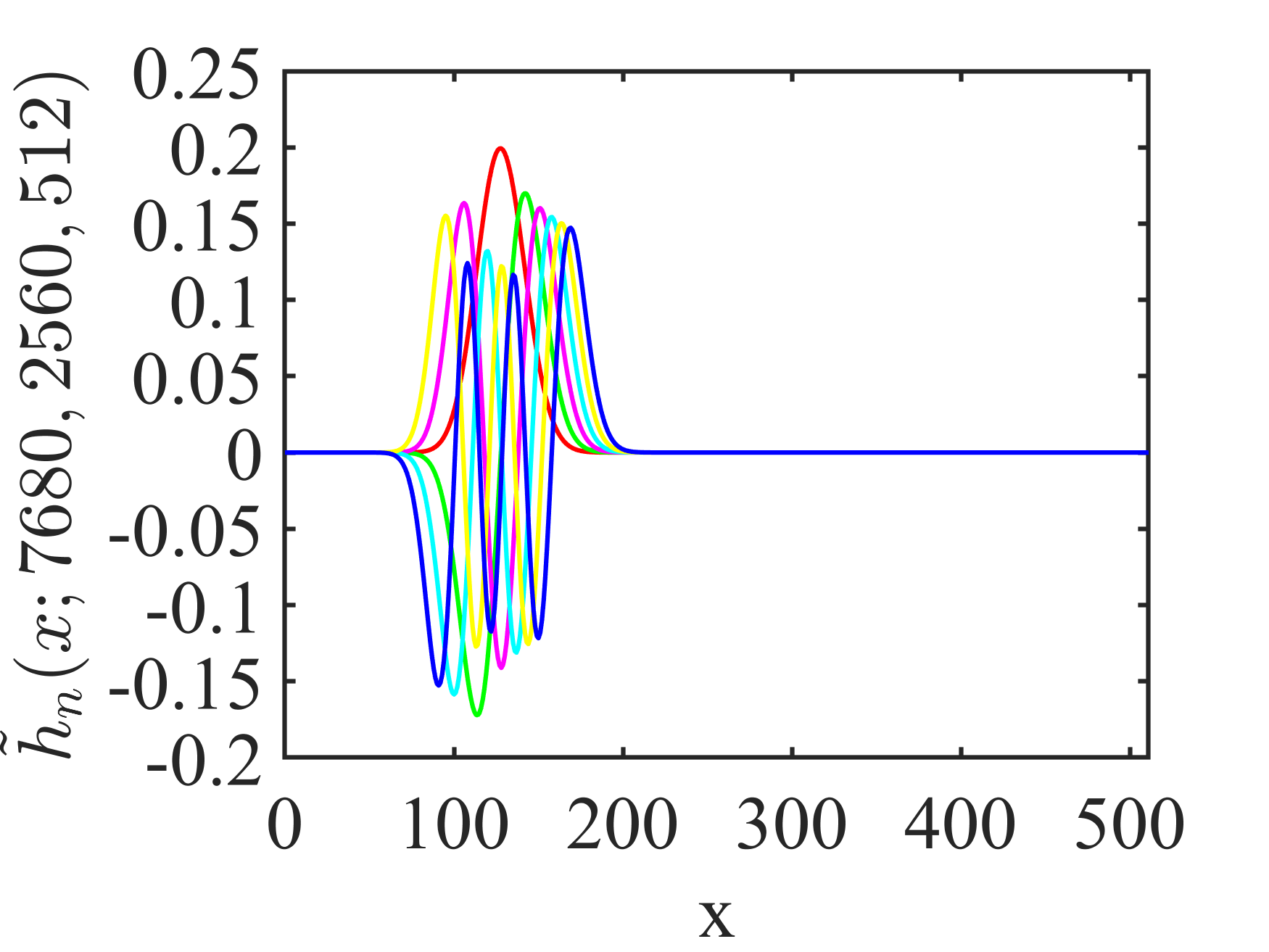}  \label{fig: Hahn_order_5_N_512_a_7680_b_2560}}
		\subfloat[]{\includegraphics[width=0.33\hsize]{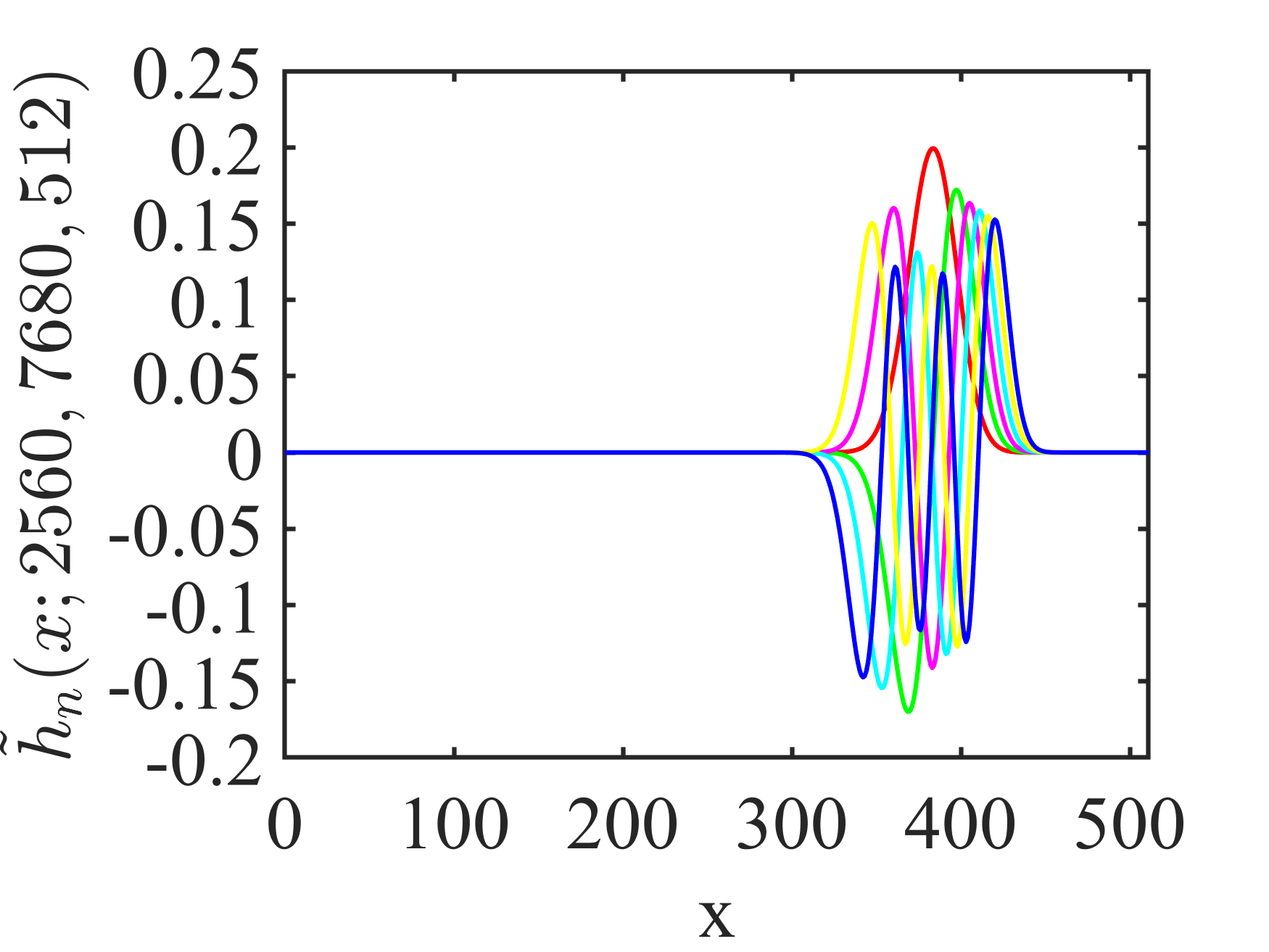}  \label{fig: Hahn_order_5_N_512_a_2560_b_7680}}
		
		\caption{Plots of normalized polynomials ($N=512$, $n={0,1,...,5}$). (a) Tchebichef polynomials. (b) Krawtchouk polynomials ($p=0.25$). (c) Krawtchouk polynomials ($p=0.75$). (d) Hahn polynomials ($a=0, b=0$). (e) Hahn polynomials ($a=7680, b=2560$). (f) Hahn polynomials ($a=2560, b=7680$).}
		
		\label{fig: normalized_polynomials}
	\end{figure}
	
	Fig. \ref{fig: normalized_polynomials} shows plots of the normalized polynomials ($N=512$, $n={0,1,...,5}$). 
	It can be observed from Figs. \ref{fig: Tchebichef_order_5_N_512} and \ref{fig: Hahn_order_5_N_512_a_0_b_0} that for $a=0, b=0$, the normalized Hahn polynomial is equivalent to the normalized Tchebichef polynomial.
	Moreover, the normalized Tchebichef polynomial satisfies the symmetry property 
	\begin{equation}
		\tilde{t}_n(N-1-u; N) = (-1)^n \tilde{t}_n(u; N), 
	\end{equation}
	which can be exploited to reduce the time required for computation significantly.
	It can be shown that if $t \gg 2N$ ($t=20N$), we can confirm that the normalized Hahn polynomial satisfactorily approximates the normalized Krawtchouk polynomial (see Figs. \ref{fig: Krawtchouk_order_5_N_512_p_quarter}, \ref{fig: Krawtchouk_order_5_N_512_p_3quarter}, \ref{fig: Hahn_order_5_N_512_a_7680_b_2560} and \ref{fig: Hahn_order_5_N_512_a_2560_b_7680}).
	The parameter of the normalized Krawtchouk polynomial $p \in (0,1)$ is used to shift the region-of-interest (ROI).
	If $p<0.5$, the ROI is on the left (see Fig. \ref{fig: Krawtchouk_order_5_N_512_p_quarter}), whilst the ROI is on the right if $p>0.5$ (see Fig. \ref{fig: Krawtchouk_order_5_N_512_p_3quarter}).
	The specific quantitative description is discussed in Section \ref{sec: Krawtchouk_Moments_Parameters}.
	From Fig. \ref{fig: normalized_polynomials}, it can be seen that the normalized Tchebichef polynomial holds the global information extraction capability, the normalized Krawtchouk polynomial holds the local information  extraction capability, and the normalized Hahn polynomial holds both of these capabilities.
	Hence, the latter is more suitable as a visual feature, whose verification  will be presented in Section \ref{sec: Simulation_and_Experiment}.
	
	\subsection{DOMs} \label{sec: Discrete_Orthogonal_Moment}
	This subsection discusses DOMs as novel compact visual features, which are derived from normalized polynomials $\tilde{p}_n(u)$. 
	
	Given a digital image $\mathbf{I}(u,v)$ with size $N \times M$, that is, $u \in [0, N-1]$ and $v \in [0, M-1]$, the $(n+m)$th order moments with a variable normalized orthogonal polynomials as the basis function for an image is defined as
	\begin{equation}
		P_{nm} = \sum_{u} \sum_{v} \mathbf{p}_{nm}(u,v) \mathbf{I}(u,v), \quad n, m = 0, 1, ..., s, \label{eq: Pnm}
	\end{equation}
	where orthogonal operators $\mathbf{p}_{nm}(u,v) = \tilde{p}_n(u) \tilde{p}_m(v)$.
	Hence, TMs, KMs, and HMs can be written as 
	\begin{align}
		&T_{nm} = \sum_{u} \sum_{v} \mathbf{t}_{nm}(u,v) \mathbf{I}(u,v), \label{eq: Tnm}\\
		&K_{nm}({}^{\alpha}p, {}^{\beta}p) =  \sum_{u} \sum_{v}  \mathbf{k}_{nm}(u,v,{}^{\alpha}p, {}^{\beta}p) \mathbf{I}(u,v), \label{eq: K_nm}\\
		&H_{nm}({}^{\alpha}a, {}^{\alpha}b, {}^{\beta}a, {}^{\beta}b) =  \notag \\ & \qquad \sum_{u} \sum_{v}\mathbf{h}_{nm}(u,v,{}^{\alpha}a, {}^{\alpha}b, {}^{\beta}a, {}^{\beta}b) \mathbf{I}(u,v),
		\label{eq: H_nm}
	\end{align}
	where TM, KM, and HM operators can be defined as 
	\begin{align}
		& \mathbf{t}_{nm}(u,v) = \tilde{t}_n(u; N) \tilde{t}_m(v; M), \notag\\
		&\mathbf{k}_{nm}(u,v,{}^{\alpha}p, {}^{\beta}p) =  \tilde{k}_n(u; {}^{\alpha}p, N) \tilde{k}_m(v; {}^{\beta}p, M), \notag \\
		&\mathbf{h}_{nm}(u,v,{}^{\alpha}a, {}^{\alpha}b, {}^{\beta}a, {}^{\beta}b) = \tilde{h}_n(u; {}^{\alpha}a, {}^{\alpha}b, N) \tilde{h}_m(v; {}^{\beta}a, {}^{\beta}b, M).\notag
	\end{align}
	
	DOMs have been widely adopted for image compression and filtering in the image processing domain \cite{mukundan2001image, yap2003image, yap2007image}, such that they achieve better image dimensionality reduction and robustness when used as image features in DVS.
	If we choose the order of orthogonal moments to be $l$, the VS features can be represented as 
	\begin{equation}
		\mathbf{s} = \left[  P_{00},  P_{10}, P_{01}, \cdots, P_{nm} \right]^\text{T}, \quad n+m \leq l, \label{eq: visual_feature}
	\end{equation}
	where $P_{nm}$ can be calculated from \eqref{eq: Pnm}.
	Hence, we propose three DOM-VS schemes, namely: TMs-based VS (TM-VS), KMs-based VS (KM-VS), and HMs-based VS (HM-VS).
	If we perform KM-VS or HM-VS, the parameters ${}^{\alpha}p, {}^{\beta}p$ in \eqref{eq: K_nm} and ${}^{\alpha}a, {}^{\alpha}b, {}^{\beta}a, {}^{\beta}b$ in \eqref{eq: H_nm} need to be determined.
	And the order of orthogonal moments $l$ also needs to be calculated.
	The following describes how to derive these parameters.
	
	\section{Adaptive Parameter Selection} \label{sec: Adaptive_Parameter}
	Articles \cite{yap2003image} and \cite{yap2007image} show that suitable parameters can effectively reduce the error of reconstructing images by KMs and HMs.
	This is because appropriate parameters can capture valuable information about the image.
	Inspired by this, adaptive parameter selection is highly critical to VS.
	Moreover, a reasonable parameter tuning mechanism helps us to obtain a concise interaction matrix (see Section \ref{sec: KM_HM_Le} for more details).
	In a VS phase, the current image $\mathbf{I}(u,v)$ and the desired image $\mathbf{I}^*(u,v)$ are known. 
	Therefore, this section will present the method for the adaptive selection of parameters
	${}^{\alpha}p, {}^{\beta}p, {}^{\alpha}a, {}^{\alpha}b, {}^{\beta}a, {}^{\beta}b$, and $l$ based on $\mathbf{I}(u,v)$ and $\mathbf{I}^*(u,v)$.
	
	\subsection{Selection of KM Parameters ${}^{\alpha}p$ and ${}^{\beta}p$} \label{sec: Krawtchouk_Moments_Parameters}
	This subsection describes the adaptive KM parameters selection method.
	We first define a superposition projection of an image in the $u$ and $v$ directions, respectively. They can be written as
	\begin{equation}
		{}^{\alpha}\mathbf{I}(u) = \sum_{v} \mathbf{I}(u,v),  \quad
		{}^{\beta}\mathbf{I}(v) = \sum_{u} \mathbf{I}(u,v).  \label{eq: I_alpha_beta}
	\end{equation}
	Then the intensity centroid $(u_c, v_c)$ of the image is calculated by
	\begin{equation}
		u_c = \frac{\sum_{u} u\ { }^{\alpha}\mathbf{I}(u) }{\sum_{u} {}^{\alpha}\mathbf{I}(u)}, \quad
		v_c = \frac{\sum_{v} v\ { }^{\beta}\mathbf{I}(v) }{\sum_{v} {}^{\beta}\mathbf{I}(v)}.
	\end{equation}
	Note that this is equivalent to calculating geometric moments \cite{chen2022closed}.
	Similarly, the intensity centroid $(u_c^*, v_c^*)$ of the desired image can also be obtained from $\mathbf{I}^*(u,v)$.
	Therefore, the intensity centroid of two images as a whole is defined as
	\begin{equation}
		\bar{u}_c = \frac{u_c + u_c^*}{2}, \quad \bar{v}_c = \frac{v_c + v_c^*}{2}, \label{eq: x_bar_c_y_bar_c}
	\end{equation}
	which will facilitate the calculation of the interaction matrix in Section \ref{sec: KM_HM_Le}.
	
	We can use the weighting functions ${}^\alpha w_K(u)$ and ${}^\beta w_K(v)$ to represent the importance of the KM on the $u$ and $v$ directions of the image, respectively, which are defined as
	\begin{equation}
		\begin{split}
			{}^\alpha w_K(u; {}^\alpha p, N) =  \binom{N}{u} ({}^\alpha p)^u (1-{}^\alpha p)^{N-u},\\
			{}^\beta w_K(v; {}^\beta p, M) =  \binom{M}{v} ({}^\beta p)^v (1-{}^\beta p)^{M-v}. \label{eq: omega_K_alpha_beta}
		\end{split}
	\end{equation}
	They are  the probability mass function (PMF) of a binomial distribution.
	So the mean of weighting functions are
	\begin{equation}
		{}^\alpha \mu_K =  {}^\alpha p N, \quad {}^\beta \mu_K =  {}^\beta p M.
	\end{equation}
	We set $\bar{u}_c = {}^\alpha \mu_K$ and $\bar{v}_c = {}^\beta \mu_K$, then ${}^\alpha p$ and ${}^\beta p$ can be calculated by
	\begin{equation}
		{}^\alpha p = \frac{\bar{u}_c}{N}, \quad {}^\beta p = \frac{\bar{v}_c}{M}. \label{eq: p_alpha_beta}
	\end{equation}

	\begin{figure}
		\centering 
		
		\subfloat[]{\includegraphics[width=0.45\hsize]{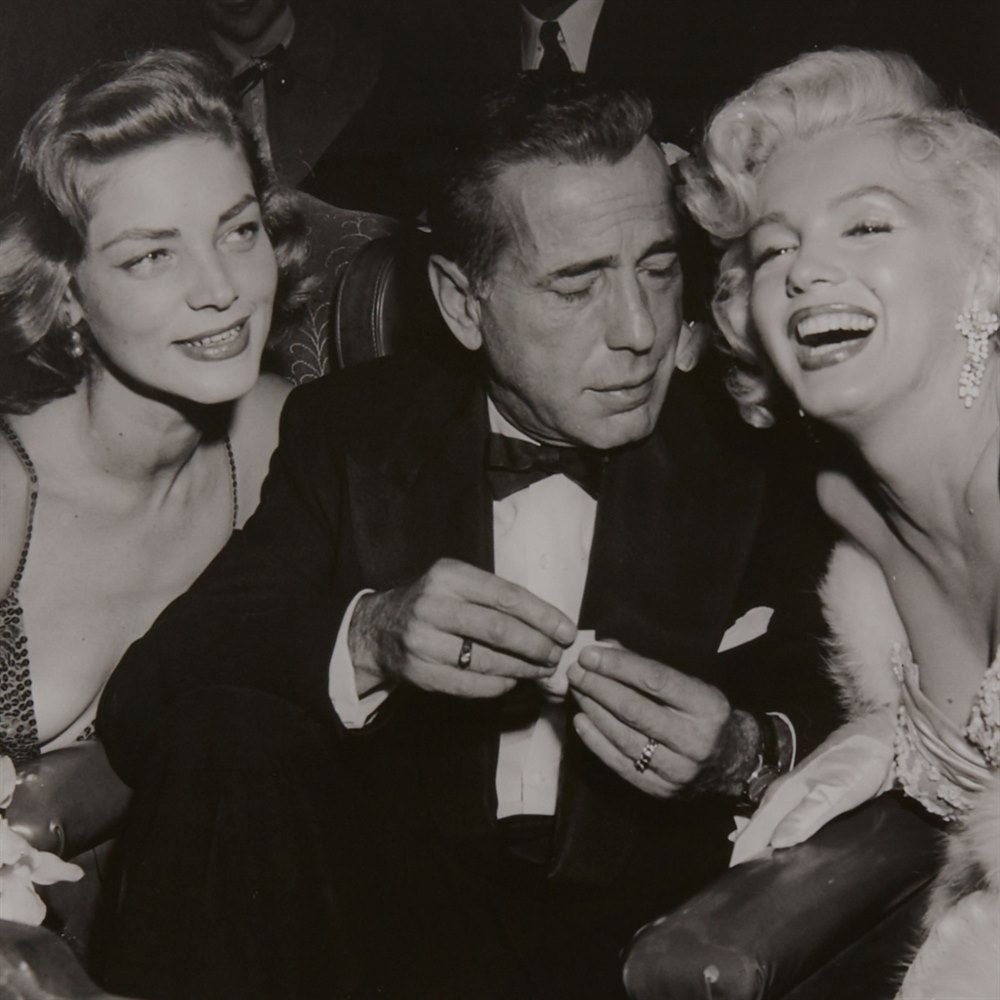}  \label{fig: Image_No_Zoom}}	
		\subfloat[]{\includegraphics[width=0.52\hsize]{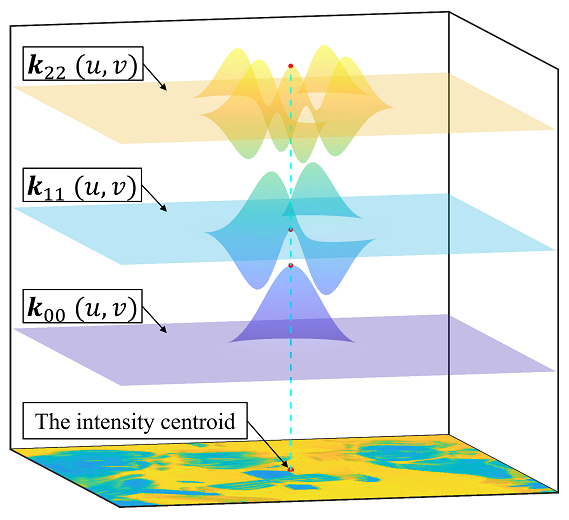}  \label{fig: Krawtchouk_operators}}
		
		\caption{Example of calculating the Krawtchouk moment parameters ($N=M=128$). (a) Example image (assuming the initial image is the same as the desired image). (b) Surface plots of the 0th, 2nd, and 4th order  Krawtchouk moment operators (${}^\alpha p = 0.5074 $ and ${}^\beta p = 0.4812$).}
		
		\label{fig: Krawtchouk_Example}
	\end{figure}
	
	Fig. \ref{fig: Krawtchouk_Example} shows an example of the calculation of the KM parameters.
	Assuming the same initial and desired images as in Fig. \ref{fig: Image_No_Zoom}, the intensity centroid of the image is $u_c =  64.9448$ and $v_c = 61.5981$. 
	Hence, KM parameters are ${}^\alpha p = 0.5074 $ and ${}^\beta p = 0.4812$.
	Fig. \ref{fig: Krawtchouk_operators}  shows the surface plots of the 0th, 2nd, and 4th order KM operators ($\mathbf{k}_{00}$, $\mathbf{k}_{11}$,  and $\mathbf{k}_{22}$).
	The ROI of the KM is only a part of the whole image, so some information is lost.
	
	In summary, ${}^{\alpha}p$ and ${}^{\beta}p$ affect the position of the ROI in the $u$ and $v$ directions during the VS, respectively, but the range of the ROI cannot be changed.

	\subsection{Selection of HM Parameters ${}^{\alpha}a, {}^{\alpha}b, {}^{\beta}a$, and ${}^{\beta}b$} \label{sec: HM Parameters}
	
	This subsection describes how to select the HM parameters adaptively.
	This method is inspired by \cite{yap2007image}.
	
	First, injecting \eqref{eq: p_alpha_beta} in \eqref{eq: ab}, we can obtain 
	\begin{equation}
		\left\{\begin{array}{l}
			{}^{\alpha}b(\mathbf{I}) = \frac{\bar{u}_c}{N} {}^{\alpha} \bar{t}(\mathbf{I})	\\ 
			{}^{\alpha}a(\mathbf{I})=(1 - \frac{\bar{u}_c}{N}) {}^{\alpha}\bar{t}(\mathbf{I}),
		\end{array}\right.  \quad
		\left\{\begin{array}{l}
			{}^{\beta}b(\mathbf{I}) = \frac{\bar{v}_c}{M} {}^{\beta}\bar{t}(\mathbf{I})	\\ 
			{}^{\beta}a(\mathbf{I})=(1 - \frac{\bar{v}_c}{M}) {}^{\beta}\bar{t}(\mathbf{I}).
		\end{array}\right.  \label{eq: ab_alpha_beta}
	\end{equation}
	Then we only need to determine the parameters, ${}^{\alpha}\bar{t}(\mathbf{I}), {}^{\beta}\bar{t}(\mathbf{I}) \in [0,\infty)$, which are related to  the dispersion of VS image. The dispersion are defined as ${}^{\alpha}\bar{d}(\mathbf{I}) \in [0, N-1]$ and ${}^{\beta}\bar{d}(\mathbf{I}) \in [0, M-1]$.
	And we can let 
	\begin{equation}
		{}^{\alpha}\bar{t}(\mathbf{I}) = e^{{}^{\alpha}\kappa {}^{\alpha}\bar{d}(\mathbf{I}) + {}^{\alpha}\varrho }, \quad
		{}^{\beta}\bar{t}(\mathbf{I}) = e^{{}^{\beta}\kappa {}^{\beta}\bar{d}(\mathbf{I}) + {}^{\beta}\varrho }, \label{eq: t_alpha_beta}
	\end{equation}
	where 
	\begin{equation}
		{}^{\alpha}\bar{d}(\mathbf{I}) = \frac{{}^{\alpha}d(\mathbf{I}) + {}^{\alpha}d^*(\mathbf{I}) }{2}, \quad {}^{\beta}\bar{d}(\mathbf{I}) = \frac{{}^{\beta}d(\mathbf{I}) + {}^{\beta}d^*(\mathbf{I}) }{2}, \notag
	\end{equation}
	where ${}^{\alpha}d(\mathbf{I}), {}^{\beta}d(\mathbf{I}), {}^{\alpha}d^*(\mathbf{I})$, and ${}^{\beta}d^*(\mathbf{I})$ denote the dispersion of the initial and desired image in the $u$ and $v$ directions, respectively. 
	The following will introduce how to calculate these parameters ${}^{\alpha}\kappa, {}^{\alpha}\varrho, {}^{\alpha}d(\mathbf{I}), {}^{\alpha}d^*(\mathbf{I}), {}^{\beta}\kappa, {}^{\beta}\varrho, {}^{\beta}d(\mathbf{I})$, and ${}^{\beta}d^*(\mathbf{I}) $.
	
	While the weighting function (\ref{eq: omega_K_alpha_beta}) of Krawtchouk polynomials is the PMF of a  binomial distribution, the mean is $\mu = Np$ and variance is $\sigma^2 = Np(1-p)$. 
	According to \eqref{eq: ab}, if $t \rightarrow \infty$, the weighting function of the Hahn polynomial is equivalent to the weighting function of the Krawtchouk polynomial.
	Thus, it is reasonable to assume that if $t \rightarrow \infty$, the variance of the weighting function of the Hahn polynomial is also $\sigma^2 = Np(1-p) $.
	
	So far, we can use the following constraints: 
	\begin{itemize}
		\item{if ${}^{\alpha}d(\mathbf{I})  = 3 {}^{\alpha}\sigma = 3\sqrt{\bar{u}_c (1 - \bar{u}_c / N)}$, ${}^{\alpha}t(\mathbf{I}) = \infty \approx 20N$;}
		\item{if ${}^{\alpha}d(\mathbf{I})  = N-1$, ${}^{\alpha}t(\mathbf{I}) = 0 \approx 0.01$;}
		\item{if ${}^{\beta}d(\mathbf{I})  = 3 {}^{\beta}\sigma = 3\sqrt{\bar{v}_c (1 - \bar{v}_c / M)}$, ${}^{\alpha}t(\mathbf{I}) = \infty \approx 20M$;}
		\item{if ${}^{\beta}d(\mathbf{I})  = M-1$, ${}^{\beta}t(\mathbf{I}) = 0 \approx 0.01$.}
	\end{itemize}
	The matrix for this constraint is expressed as
	\begin{equation}
		\mathbf{B} \bm{\chi} = \mathbf{C},
	\end{equation}
	where 
	\begin{equation}
		\mathbf{B} = \begin{bmatrix}
			3\sqrt{\bar{u}_c (1 - \bar{u}_c / N)}	&  1&  0& 0\\ 
			N-1&  1&  0& 0\\ 
			0&  0& 3\sqrt{\bar{v}_c (1 - \bar{v}_c / M)} & 1\\ 
			0&  0&  M-1& 1
		\end{bmatrix}, \notag
	\end{equation}
	\begin{equation}
		\bm{\chi} = \begin{bmatrix}
			{}^{\alpha}\kappa &  {}^{\alpha}\varrho& {}^{\beta}\kappa & {}^{\beta}\varrho
		\end{bmatrix}^\text{T},\notag
	\end{equation}
	\begin{equation}
		\mathbf{C} = \begin{bmatrix}
			\ln 20N&  \ln 0.01 & \ln 20M  & \ln 0.01 
		\end{bmatrix}^\text{T}.\notag
	\end{equation}
	It is easy to get $\bm{\chi} = \mathbf{B}^{-1} \mathbf{C}$.
	Finally, we only need to calculate the ${}^{\alpha}d(\mathbf{I})$ and ${}^{\beta}d(\mathbf{I})$ of the image in VS to obtain ${}^{\alpha}t(\mathbf{I})$ and ${}^{\beta}t(\mathbf{I})$.
	
	It is well known that the PMF of a binomial distribution approximates the probability density function (PDF) of a Gaussian distribution when $Np, N(1 - p)>5$ \cite{yap2007image}.
	$3\sigma$ rule can be expressed as $\text{P}_{3\sigma} = \text{Pr}(\mu - 3\sigma\leq u \leq \mu + 3\sigma) \approx 0.9974$ for the Gaussian distribution, where $\text{Pr}(\cdot)$ is the probability function.
	When $Np, N(1 - p)>5$ and $t \rightarrow \infty$,  the weighting function of the Hahn polynomial is the Gaussian-like distribution, whose $3\sigma$ rule can be expressed as $\text{P}_{3\sigma}^h \approx \text{P}_{3\sigma} \approx 0.9974$.
	The PMFs of the image with respect to $u$ and $v$ are
	\begin{align}
		{}^{\alpha}\mathbf{P}(u) = \frac{{}^{\alpha}\mathbf{I}(u)}{\sum_{u} {}^{\alpha}\mathbf{I}(u)}, \quad
		{}^{\beta}\mathbf{P}(v) = \frac{{}^{\beta}\mathbf{I}(v)}{\sum_{v} {}^{\beta}\mathbf{I}(v)},
	\end{align}
	where ${}^{\alpha}\mathbf{I}(u)$ and ${}^{\beta}\mathbf{I}(v)$ are defined by \eqref{eq: I_alpha_beta}.
	Then we can calculate ${}^{\alpha}d(\mathbf{I})$ and ${}^{\beta}d(\mathbf{I})$ as 
	\begin{equation}
		\begin{split}
			{}^{\alpha}d(\mathbf{I}) &= \arg_{d}  {}^{\alpha}\mathbf{P}(\bar{u}_c) \\
			& \qquad + \sum_{i=1}^{d} {}^{\alpha}\mathbf{P}(\bar{u}_c+i) +  {}^{\alpha}\mathbf{P}(\bar{u}_c-i) \geq  \text{P}_{3\sigma}^h,\\
			{}^{\beta}d(\mathbf{I}) &= \arg_{d}  {}^{\beta}\mathbf{P}(\bar{v}_c) \\ 
			& \qquad  +   \sum_{j=1}^{d} {}^{\beta}\mathbf{P}(\bar{v}_c+j) +  {}^{\beta}\mathbf{P}(\bar{v}_c-j) \geq  \text{P}_{3\sigma}^h .
		\end{split}
	\end{equation}
	Note that we specify that if $u<0$ or $u>N-1$, ${}^{\alpha}P(u) =0$; if $v<0$ or $v>M-1$, ${}^{\beta}P(v) = 0$.
	Similarly, ${}^{\alpha}d^*(\mathbf{I})$ and ${}^{\beta}d^*(\mathbf{I}) $ can also be determined.
	So far, the necessary parameters for calculating ${}^{\alpha}\bar{t}(\mathbf{I}) $ and ${}^{\beta}\bar{t}(\mathbf{I})$ have been determined. And we can calculate them  according to \eqref{eq: t_alpha_beta}.
	Finally, \eqref{eq: ab_alpha_beta} can be used to obtain the HM parameters ${}^{\alpha}a, {}^{\alpha}b, {}^{\beta}a$ and ${}^{\beta}b$.
	It can be seen that the above method uses the $3\sigma$ rule.
	When the image noise is large, the $2\sigma$ rule is also used to calculate these parameters.
	
\begin{figure}
		\centering 
		
		\subfloat[]{\includegraphics[width=0.49\hsize]{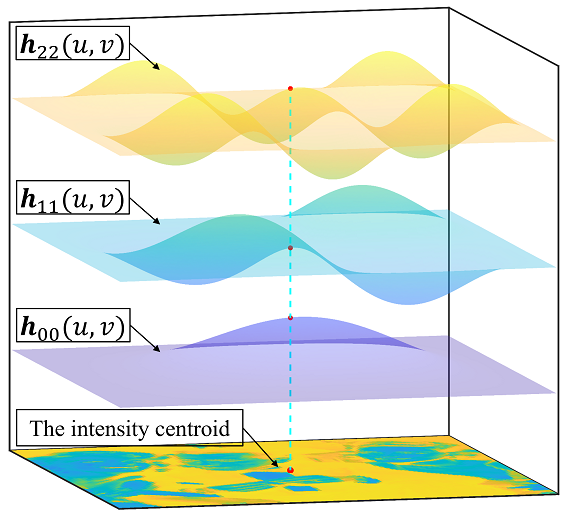}  \label{fig: Hahn_operators_No_Zoom}}
		\subfloat[]{\includegraphics[width=0.49\hsize]{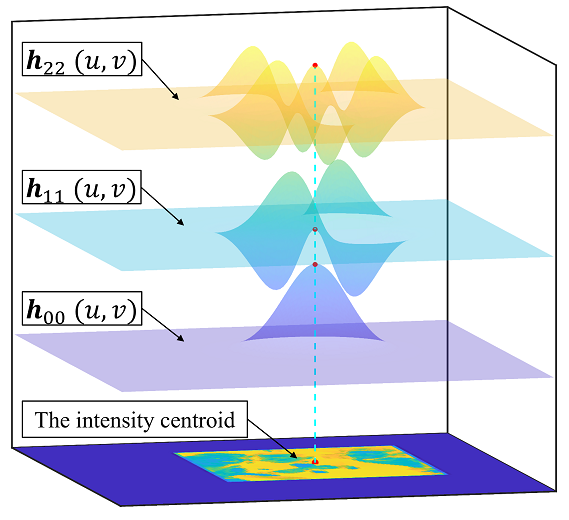}  \label{fig: Hahn_operators_Zoom}}
		
		\caption{Example of calculating the Hahn moment parameters ($N=M=128$). (a) Surface plots of the 0th, 2nd, and 4th order  Hahn moment operators for the origin image (${}^{\alpha}a=5, {}^{\alpha}b=6, {}^{\beta}a=7$ and ${}^{\beta}b=6$). (b) Surface plots of the 0th, 2nd, and 4th order Hahn moment operators for the transformed image(${}^{\alpha}a=175, {}^{\alpha}b=127, {}^{\beta}a=198$ and ${}^{\beta}b=136$).}
		
		\label{fig: Hahn_Example}
	\end{figure}

\begin{figure}
		\centering 
		
		\subfloat[]{\includegraphics[width=0.49\hsize]{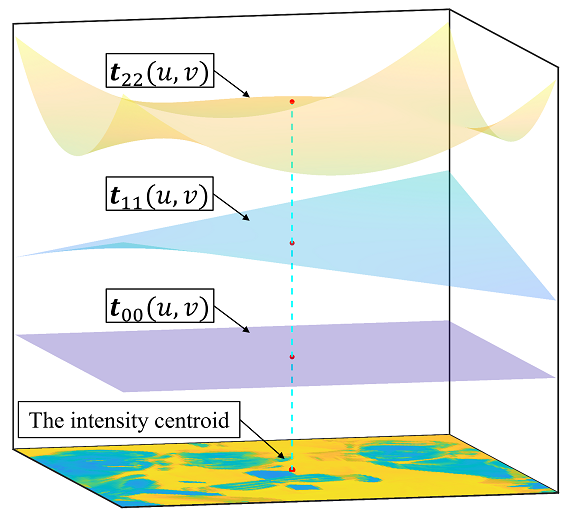}  \label{fig: Tchebichef_operators_No_Zoom}}
		\subfloat[]{\includegraphics[width=0.49\hsize]{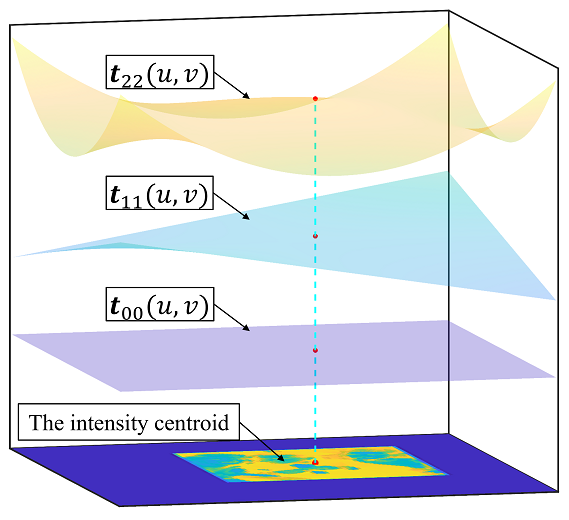}  \label{fig: Tchebichef_operators_Zoom}}
		
		\caption{Surface plots of the 0th, 2nd, and 4th order  Tchebichef moment operators  ($N=M=128$). (a) Calculation result of the original image. (b) Calculation result of the transformed image.}
		
		\label{fig: Tchebichef_Example}
	\end{figure}

	Fig. \ref{fig: Hahn_Example} shows an example of calculating the  HM parameters. 
	Still assuming that the initial and desired images are the same as Fig. \ref{fig: Image_No_Zoom}, the HM parameters obtained by calculation is ${}^{\alpha}a=5, {}^{\alpha}b=6, {}^{\beta}a=7$ and ${}^{\beta}b=6$.
	And surface plots of the 0th, 2nd, and 4th order HM operators ($\mathbf{h}_{00}$, $\mathbf{h}_{11}$, and $\mathbf{h}_{22}$) are showed in Fig. \ref{fig: Hahn_operators_No_Zoom}, whose ROI is significantly larger than the ROI of KM operators in Fig. \ref{fig: Krawtchouk_operators}.
	The images in Fig. \ref{fig: Image_No_Zoom} are scaled and translated to better illustrate the adaptability of the method proposed in this subsection.
	The HM parameters calculated from the transformed image are ${}^{\alpha}a=175, {}^{\alpha}b=127, {}^{\beta}a=198$ and ${}^{\beta}b=136$. 
	And surface plots are shown in Fig. \ref{fig: Hahn_operators_Zoom}.
	As a comparison, Fig. \ref{fig: Tchebichef_Example} shows surface plots of the 0th, 2nd, and 4th order TM operators ($\mathbf{t}_{00}$, $\mathbf{t}_{11}$, and $\mathbf{t}_{22}$). 
	Because Tchebichef polynomials do not have any parameters to be adjusted, the TM operators are the same for both the original image and the transformed image.
	The adaptive HM operators have excellent performance for both the original and the transformed images, which is not true for the KM and TM operators.
	
	In short, $b/(a+b)$ and $a+b$ affect the position and range of the ROI during the VS. Thus, HM-VS can consider both global and local information of the image through flexible parameter tuning.
	
	\subsection{Selection of DOM Order $l$}  \label{sec: order}
	If the order of the orthogonal moments is determined, then the VS features can be obtained according to \eqref{eq: visual_feature}.
	When the order of the orthogonal moments is small, the VS feature mainly considers the  coarse information of the image, which has the advantage of better image filtering and compression properties. 
	However, since it does not consider the detailed information of the image, this makes the VS procedure prone to local minima.
	In contrast, when the order of the orthogonal moments is large, the VS features mainly consider the detailed information of the image, which has the advantage of excellent convergence accuracy.
	However, the VS process  converges slowly due to its excessive attention to detailed information.
	The following introduces a method for selecting the orthogonal moments order $l$ to exploit their advantages while avoiding shortcomings.
	
	We intend to approach the target object quickly with a small order when it is far from the target pose and converges with high accuracy with a large order when it is close to the target pose.
	First, we define the minimum and maximum orders ($l_{\text{min}}$ and $l_{\text{max}}$), which are empirically fixed values.
	The required order can then be expressed as
	\begin{equation}
		l = (l_{\text{max}} - l_{\text{min}}) \eta +  l_{\text{min}}, \quad \eta \in [0,1], \quad l \in \mathbb{N}, \label{eq: l}
	\end{equation}
	where $\eta$ is defined as
	\begin{equation}
		\eta = \frac{\bar{e}_{\text{I}^{o}}}{\bar{e}_{\text{I}^{o}} + \lambda_\eta \bar{e}_\text{I}}, \label{eq: eta}
	\end{equation}
	where $\bar{e}_\text{I}$ and $\bar{e}_{I^{o}}$ are the mean square error of the current and  initial images, respectively, and are defined as 
	\begin{align}
		\bar{e}_\text{I} &= \frac{\sum_{u} \sum_{v} \left( \mathbf{I}(u,v) - \mathbf{I}^*(u,v) \right) ^2}{N \times M}, \notag  \\
		\bar{e}_{\text{I}^{o}} &= \frac{\sum_{u} \sum_{v} \left( \mathbf{I}^o(u,v) - \mathbf{I}^*(u,v)\right) ^2 }{N \times M}, \notag
	\end{align}
	where $\mathbf{I}^o(u,v), \mathbf{I}(u,v)$ and $\mathbf{I}^*(u,v)$ are the initial, current, and desired images, respectively.
	Based on the control law of exponentially decreasing feature error (ideal case) introduced in Section \ref{sec: Model_Interaction}, it is reasonable to assume that the  $\bar{e}_\text{I}$ also decreases exponentially, i.e., $\bar{e}_\text{I} = \bar{e}_{\text{I}^{o}} e^{-\lambda_o t}$.
	Hence, \eqref{eq: eta} can be rewritten as
	\begin{equation}
		\eta = \frac{1}{1 + \lambda_\eta e^{-\lambda_o t}}, \notag
	\end{equation}
	where $\eta$ is the sigmoid function with an "S"-shaped curve, which is exactly what we need.
	In the following, we describe how to calculate $\lambda_\eta$ in \eqref{eq: eta}.
	
	We normally consider VS convergence when $\bar{e}_\text{I} = \epsilon$ is satisfied, where $\epsilon$ is a fixed value and $\bar{e}_{\text{I}^{o}} \gg \epsilon$.
	$\lambda_\eta$ is designed as a linear function of $\bar{e}_\text{I}$.
	Therefore, we can use the following constraints: 
	\begin{itemize}
		\item{if $\bar{e}_{\text{I}} = \bar{e}_{\text{I}^{o}}$, $\lambda_\eta = \bar{e}_{\text{I}^{o}} / \epsilon$;}
		\item{if $\bar{e}_{\text{I}} = \epsilon$, $\lambda_\eta =0$.}
	\end{itemize}
	For the former, we have $\eta = \frac{1}{1 + \bar{e}_{\text{I}^{o}} / \epsilon } \approx 0, l = l_{\text{min}}$; for the latter, we have $\eta = 1, l = l_{\text{max}}$.
	Based on the above constraints, $\lambda_\eta$ can be calculated as
	\begin{equation}
		\lambda_\eta = \frac{ \bar{e}_{\text{I}^{o}}}{\epsilon( \bar{e}_{\text{I}^{o}} - \epsilon)} (\bar{e}_{\text{I}} -\epsilon). \label{eq: lambda_eta}
	\end{equation}
	Injecting \eqref{eq: lambda_eta} in \eqref{eq: eta}, $\eta$ can be expressed  as
	\begin{equation}
		\eta = \frac{1}{1 + \frac{\bar{e}_{\text{I}}}{\epsilon} \frac{\bar{e}_{\text{I}} - \epsilon}{ \bar{e}_{\text{I}^{o}} - \epsilon}}.
	\end{equation}
	So far, $l$ can be calculated by \eqref{eq: l}.
	More details about the choice of the $l_{\text{min}}$ and $l_{\text{max}}$ are discussed in Section \ref{sec: Discussion}.
	
\begin{figure*}
		\centering 
		
		\subfloat[]{\includegraphics[width=0.15\hsize]{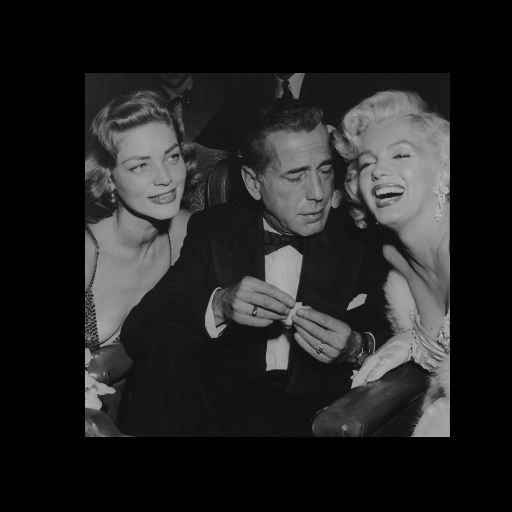}  \label{fig: Loss_Landscape_image}}
		\subfloat[]{\includegraphics[width=0.15\hsize]{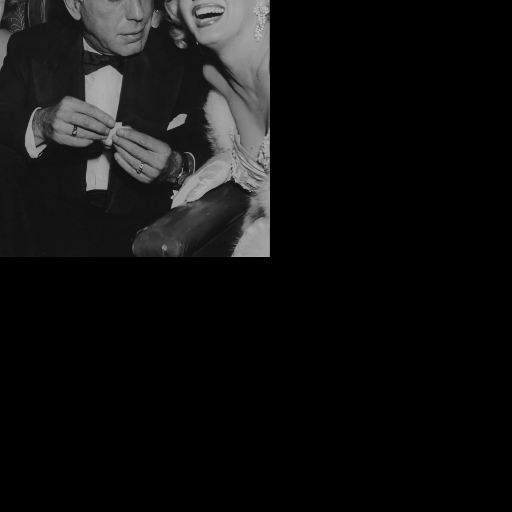}  \label{fig: Loss_Landscape_image_1}}
  	\subfloat[]{\includegraphics[width=0.15\hsize]{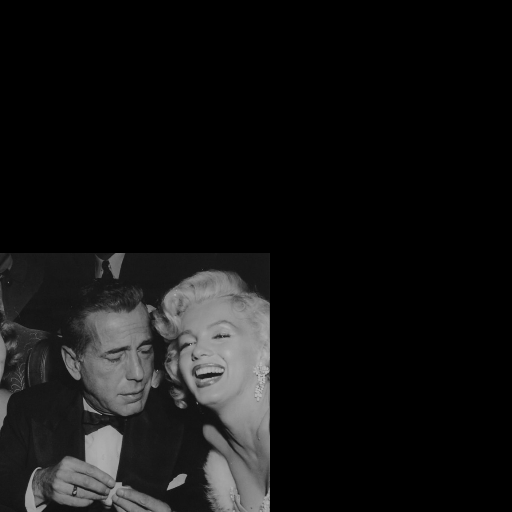}  \label{fig: Loss_Landscape_image_2}}
        \subfloat[]{\includegraphics[width=0.15\hsize]{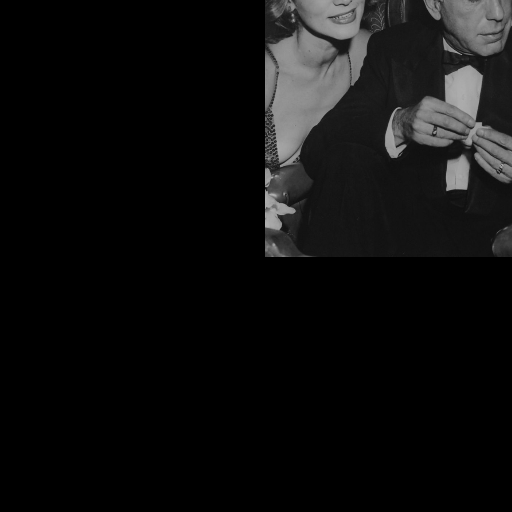}  \label{fig: Loss_Landscape_image_3}}
        \subfloat[]{\includegraphics[width=0.15\hsize]{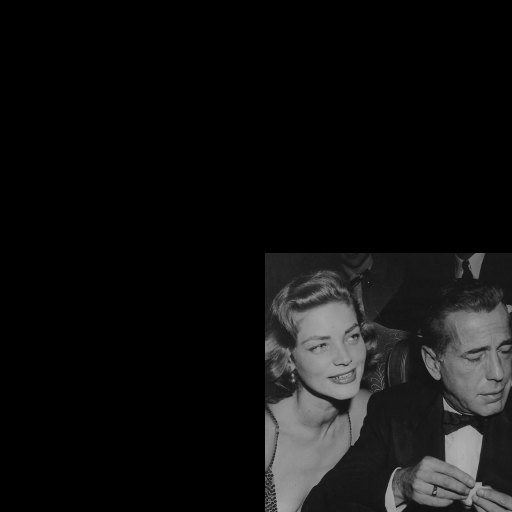}  \label{fig: Loss_Landscape_image_4}}

		\subfloat[]{\includegraphics[width=0.25\hsize]{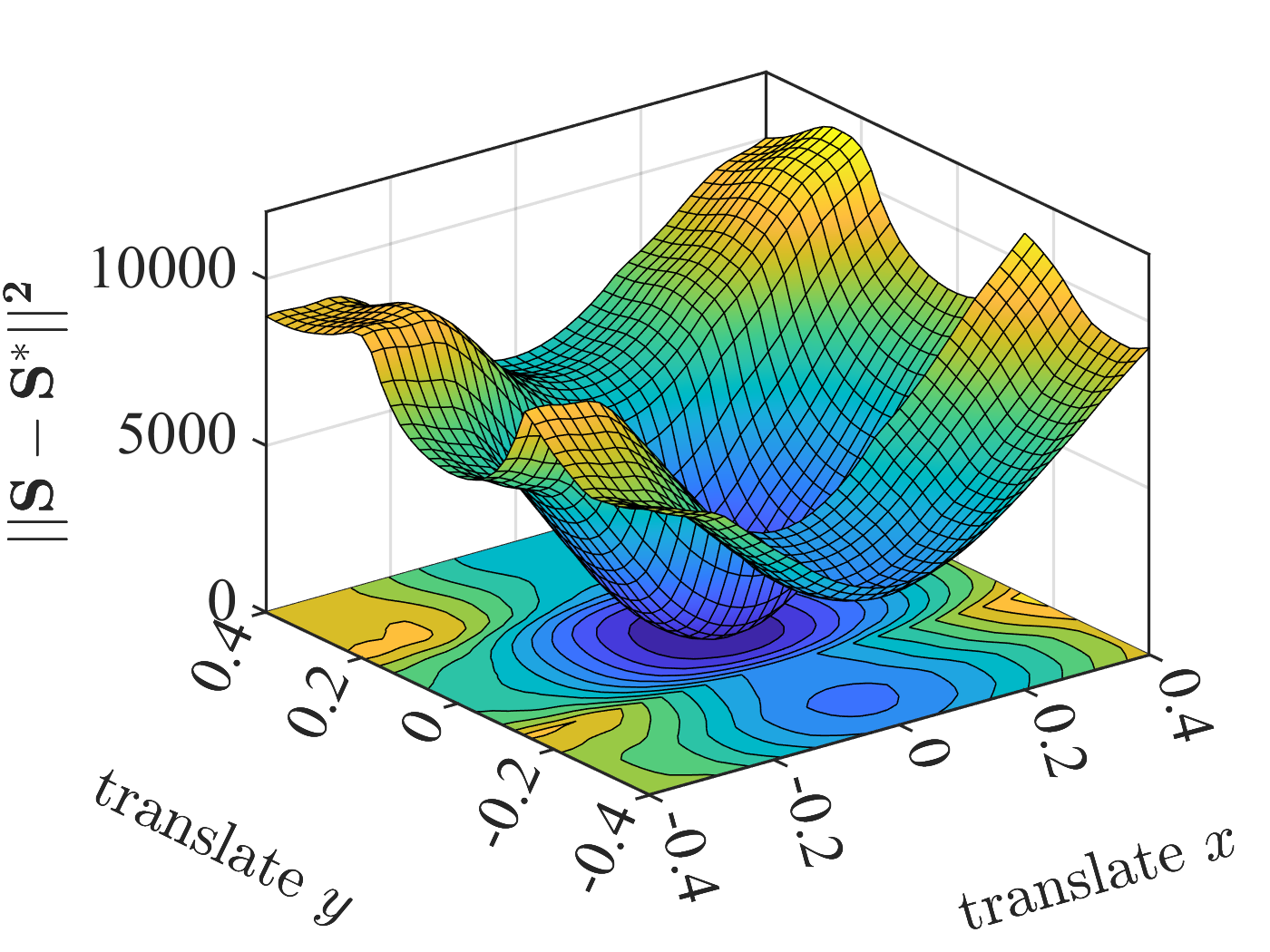}  \label{fig: Loss_Landscape_Tchebichef}}
		\subfloat[]{\includegraphics[width=0.25\hsize]{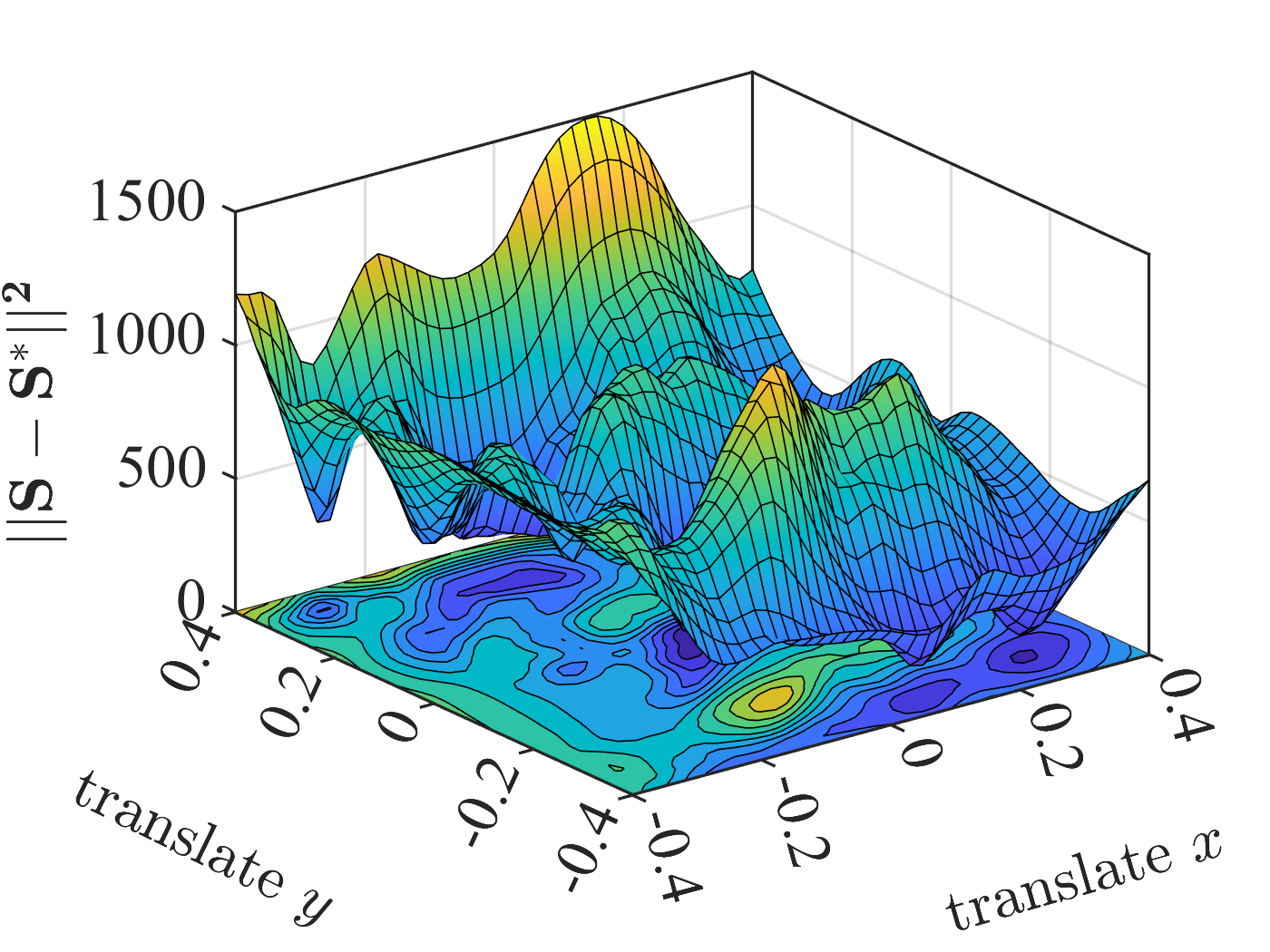}  \label{fig: Loss_Landscape_Krawtchouk}}
		\subfloat[]{\includegraphics[width=0.25\hsize]{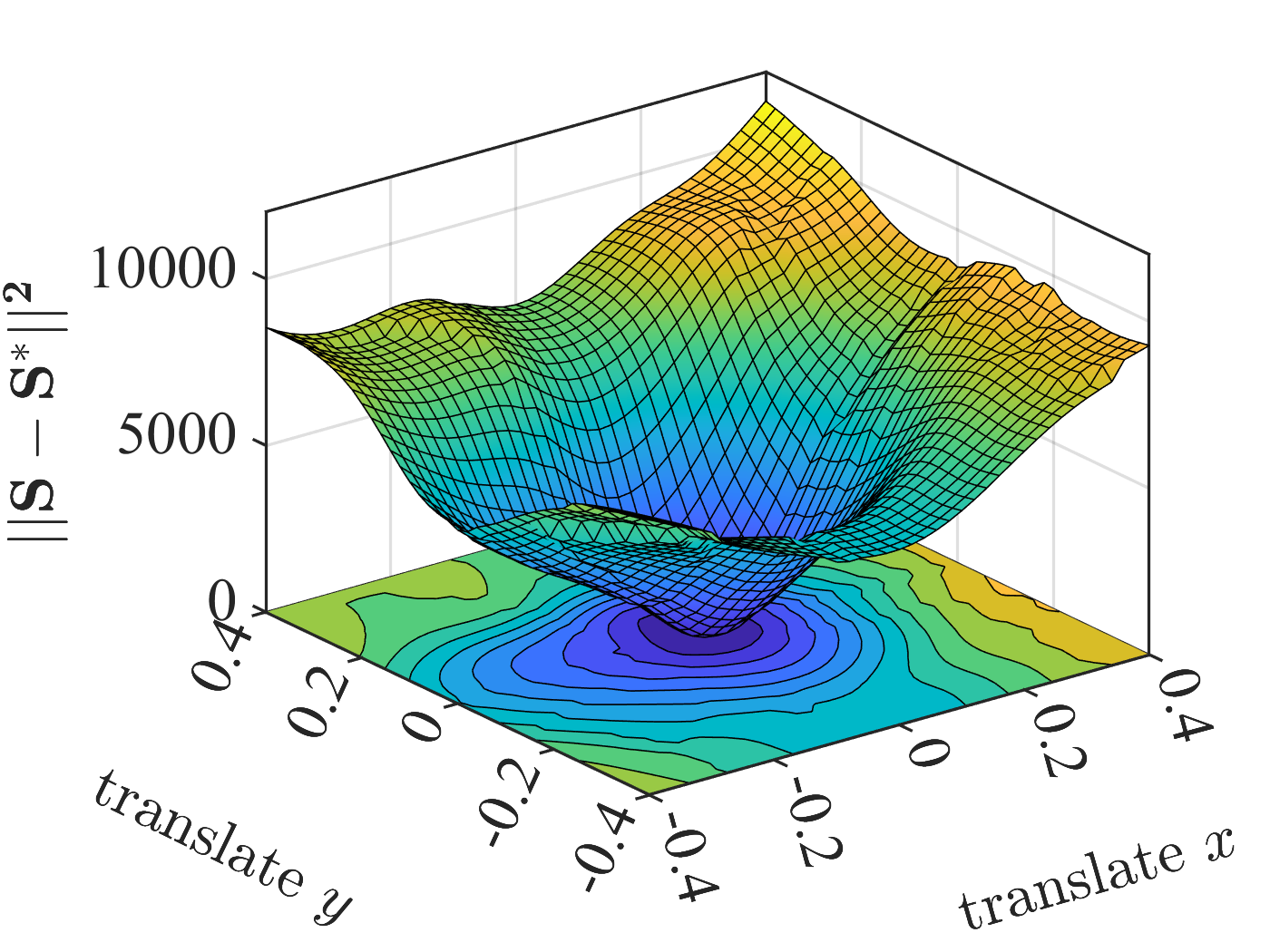}  \label{fig: Loss_Landscape_Hahn}}
		
		\caption{VS loss landscape on an x/y translation motion around the desired pose. (a) Desired image. (b)-(e) Images of the boundary position in the x,y direction. (f) TM-VS. (g) KM-VS. (h) HM-VS.}
		
		\label{fig: Loss_Landscape}
	\end{figure*}

\begin{figure*}
		\centering 
  
		\subfloat[]{\includegraphics[width=0.25\hsize]{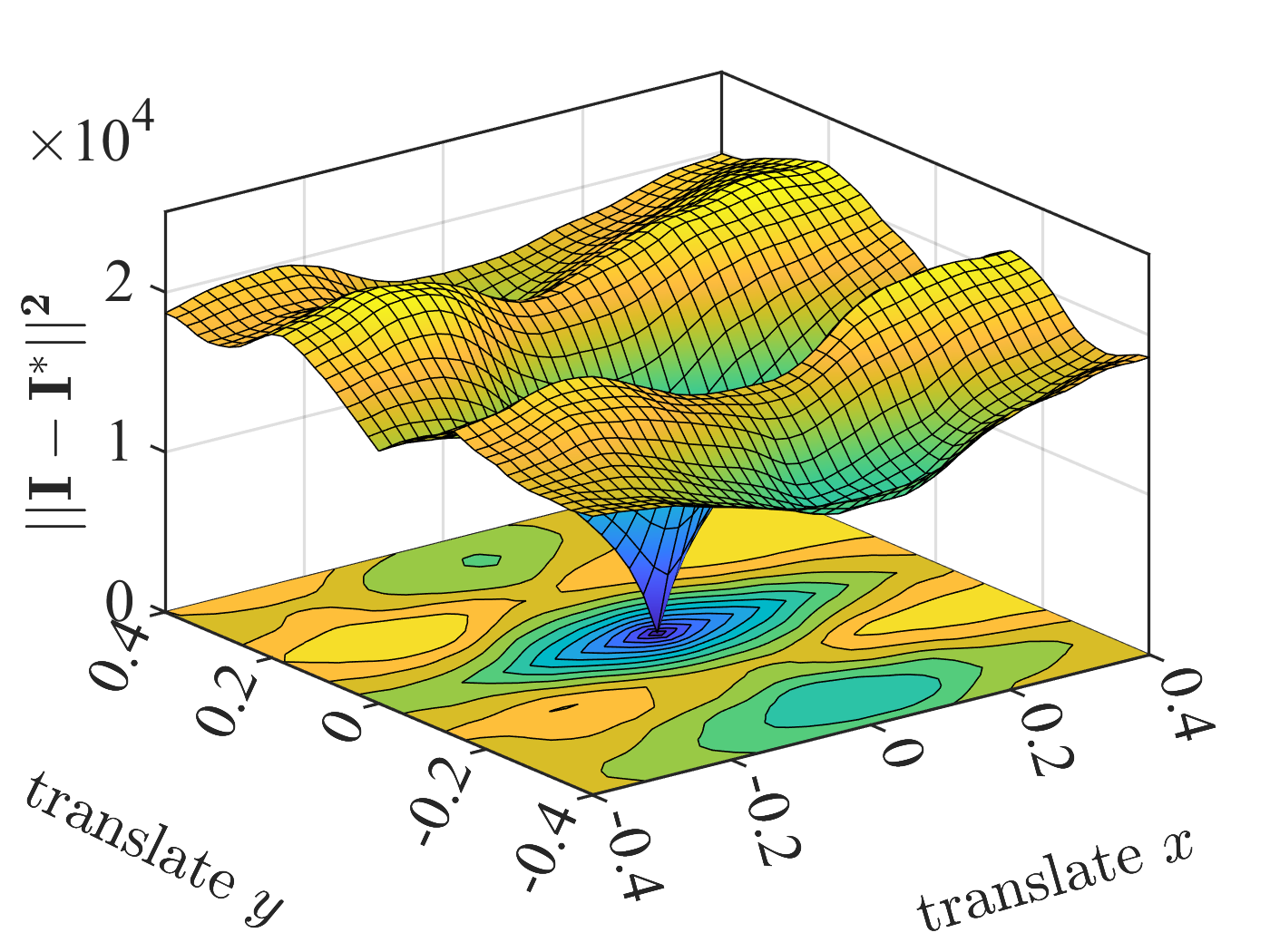}  \label{fig: Loss_Landscape_DVS}}
		\subfloat[]{\includegraphics[width=0.25\hsize]{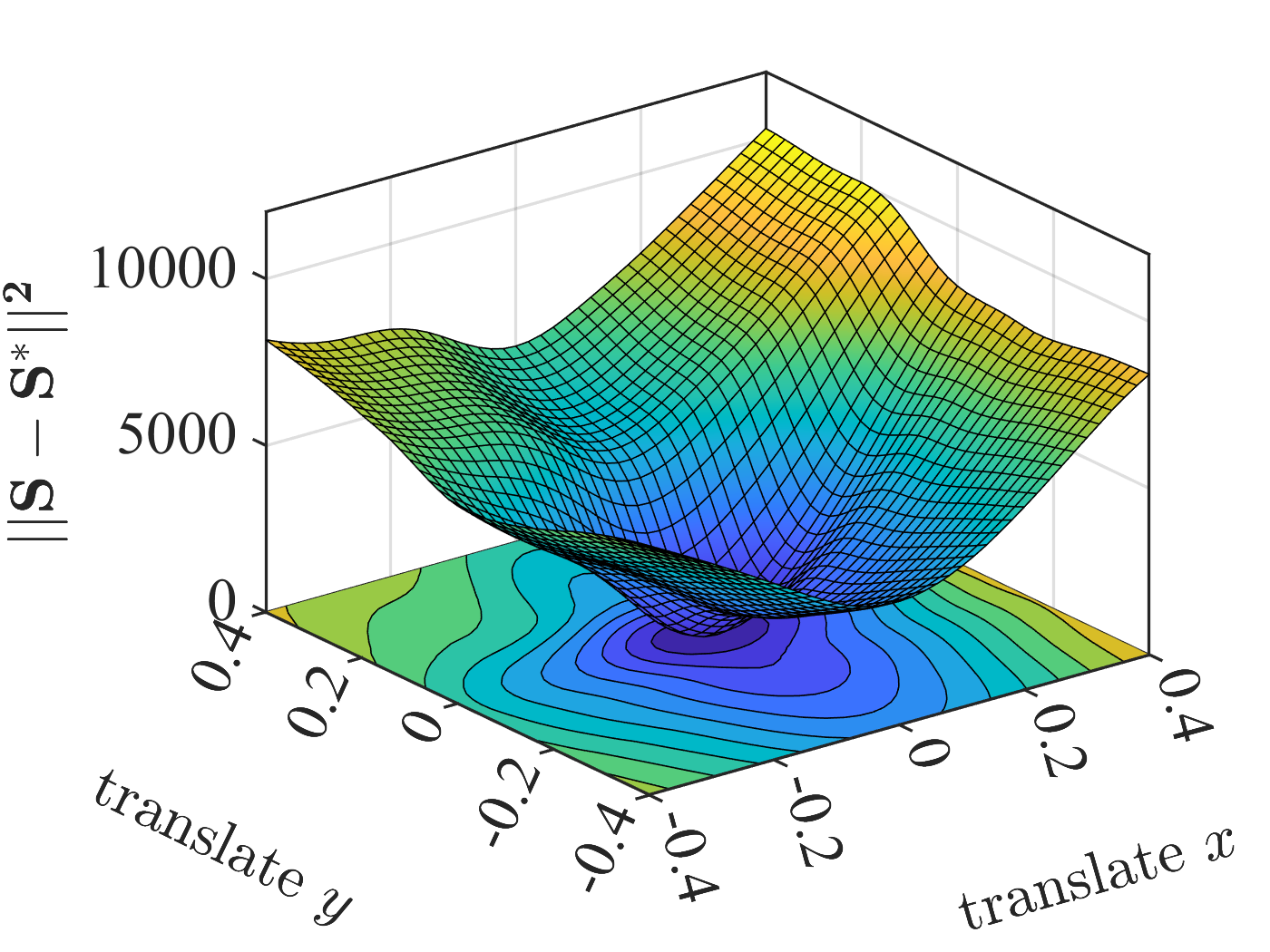}  \label{fig: Loss_Landscape_Hahn_order_3}}
		\subfloat[]{\includegraphics[width=0.25\hsize]{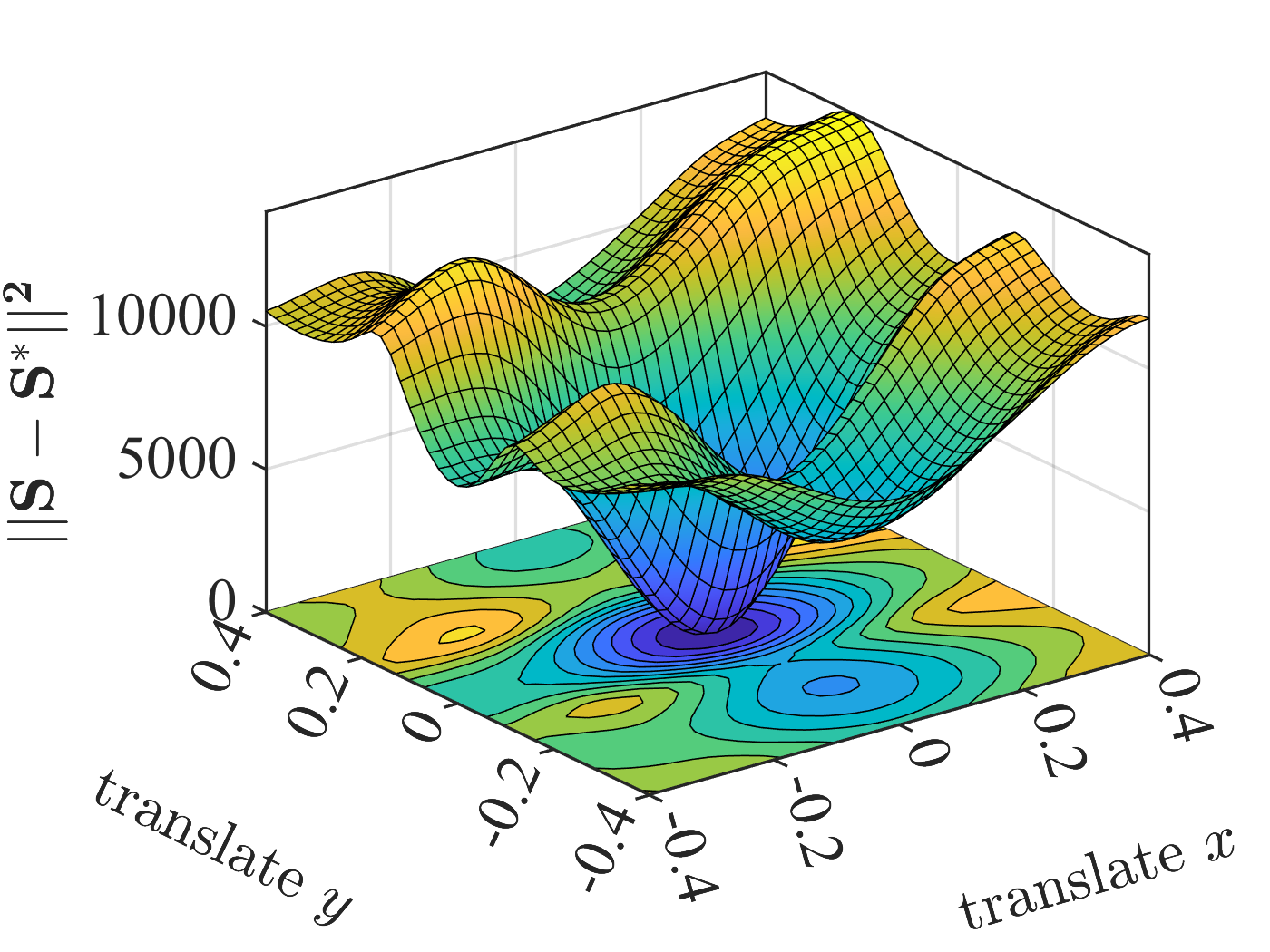}  \label{fig: Loss_Landscape_Hahn_order_6}}
  
  	\subfloat[]{\includegraphics[width=0.25\hsize]{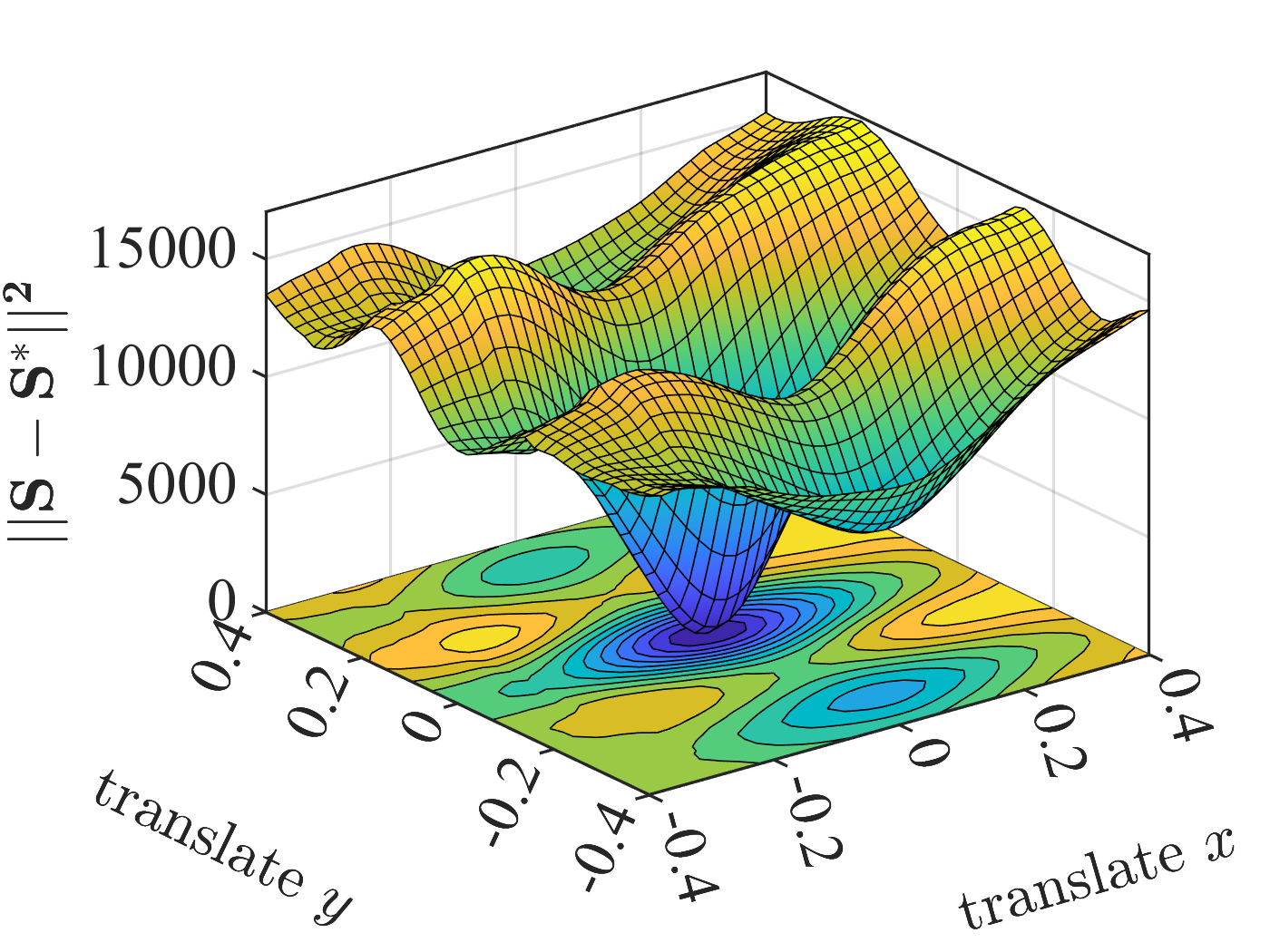}  \label{fig: Loss_Landscape_Hahn_order_10}}
		\subfloat[]{\includegraphics[width=0.25\hsize]{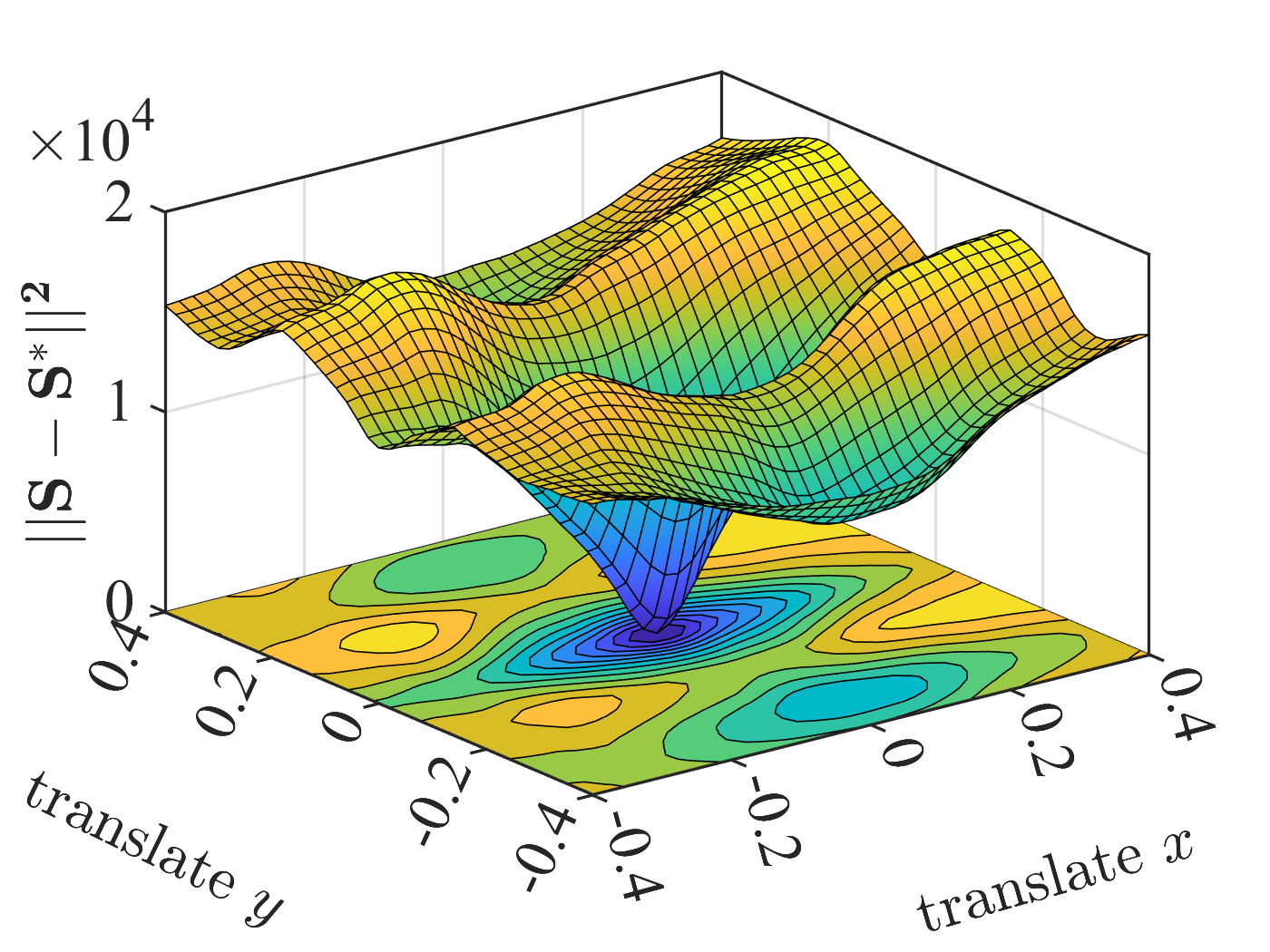}  \label{fig: Loss_Landscape_Hahn_order_20}}
		\subfloat[]{\includegraphics[width=0.25\hsize]{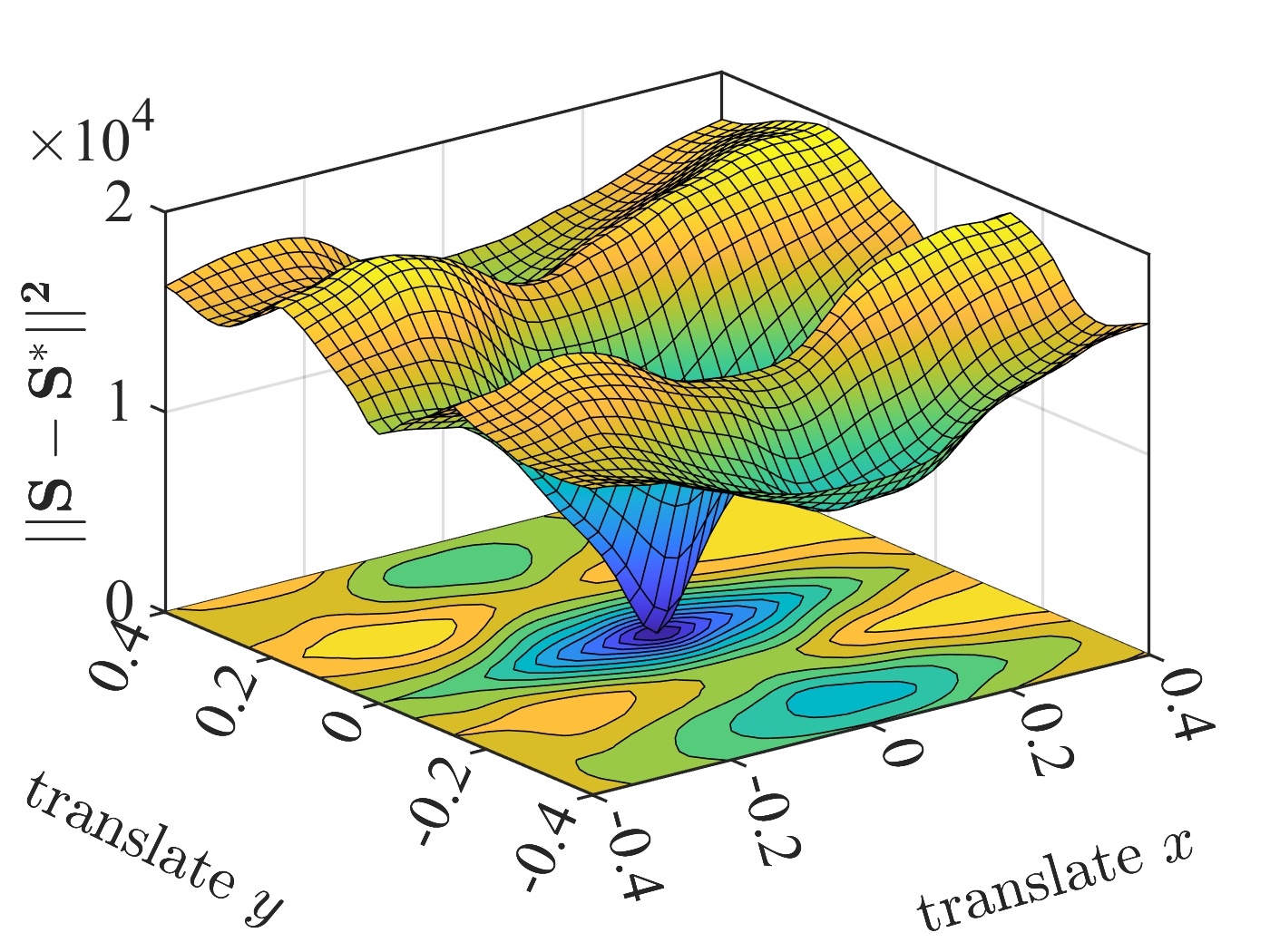}  \label{fig: Loss_Landscape_Hahn_order_30}}  
		
		\caption{Comparison of the HM-VS loss landscapes shape obtained for different $l$. (a) Photometric dense features. (b)-(f) HM-VS for, respectively, $l = \{3, 6, 10, 20, 30\}$.}
		
		\label{fig: Loss_Landscape_HM_order}
	\end{figure*}

	Take $l=4$ as an example, Fig. \ref{fig: Loss_Landscape} shows the VS loss landscape of the error function on an x/y translation motion around the desired pose.
        The desired image and the images in the x/y boundary position are illustrated in Figs. \ref{fig: Loss_Landscape_image}-\ref{fig: Loss_Landscape_image_4}, respectively.
        The HM-VS loss landscape (see Fig. \ref{fig: Loss_Landscape_Hahn}) is the most satisfactory, and the TM-VS loss landscape (see Fig. \ref{fig: Loss_Landscape_Tchebichef}) is better than the KM-VS (see Fig. \ref{fig: Loss_Landscape_Krawtchouk}).
 This is mainly because the ROI of the KM operators is not the whole image (see Fig. \ref{fig: Krawtchouk_operators}), and the large translations around the desired pose in this experiment resulted in parts of the image being out of  field-of-view.
	Let's compare the HM-VS loss landscapes shape obtained for different $l$.
 Fig. \ref{fig: Loss_Landscape_DVS} presents the loss landscape of the DVS. 
Figs. \ref{fig: Loss_Landscape_Hahn_order_3}-\ref{fig: Loss_Landscape_Hahn_order_30} exhibit the HM-VS loss landscapes for different order $l$ values.
It can be seen that the lower $l$ is, the larger the convex domain; the higher $l$ is, the faster the convergence rate near the ideal pose.
 In addition, if $l = N+M-2$, the visual features are  the set of all DOMs $\mathbf{s} = \left[  P_{00},  P_{10}, P_{01}, \cdots, P_{N-1M-1} \right]^\text{T}$.
	Hence, the TM-VS and HM-VS schemes proposed in this study are equivalent to the DVS \cite{collewet2008visual, marchand2020direct}.

	It is worth mentioning that although the order $l$ varies during VS, it does not affect the stability of the proposed method.
    Since we ensure that the same order is used to compute the visual features for both the initial and desired images each time.
    For example, if $l$ is assumed constant, $l=\{3,6,10,20,30\}$ corresponds to five visual servoing methods. 
    Our proposed method is similar to a combination of these methods.
    As long as each method is stable, our proposed method does not affect stability.
	
	\section{Interaction Matrix of DOMs} \label{sec: Model_Interaction}
	
	The aim of VS is to minimize the feature error $\mathbf{e}(t)$, which is typically defined by
	\begin{equation}
		\mathbf{e}(t) = \mathbf{s} - \mathbf{s}^*,
	\end{equation}
	where $\mathbf{s}^*$ is the desired value of visual features $\mathbf{s}$ to be reached in the image \cite{franccois2006visual}. 
	The key of VS is the interaction matrix $\mathbf{L}_e$ that links the time variation of feature error to the camera velocity $\mathbf{v}$ \cite{crombez2018visual}
	\begin{equation}
		\dot{\mathbf{e}} = \dot{\mathbf{s}} - \dot{\mathbf{s}}^* = \mathbf{L}_e \mathbf{v}. \label{eq: de}
	\end{equation}
	
	To ensure an exponential decoupled decrease of the feature error\cite{franccois2006visual}, the control law is designed as
	\begin{equation}
		\mathbf{v} = -\lambda \widehat{\mathbf{L}}_e^\dagger (\mathbf{s} - \mathbf{s}^*),
	\end{equation}
	where $\lambda$ is a positive scalar, $\widehat{\mathbf{L}}_e$ is an estimation or an approximation of $\mathbf{L}_e$ and $(\cdot)^\dagger$ is the Moore-Penrose  pseudo-inverse.
	The following will describe how to calculate $\widehat{\mathbf{L}}_e$.
	
	Based on \eqref{eq: visual_feature}, the visual features $\mathbf{s}$ and $ \mathbf{s}^*$ can be written as
	\begin{equation}
		\begin{split}
			\mathbf{s} &= P_{nm} = \sum_{u} \sum_{v} \mathbf{p}_{nm}(u,v) \mathbf{I}(u,v),\\
			\mathbf{s}^* &= P^*_{nm} = \sum_{u} \sum_{v} \mathbf{p}_{nm}(u,v) \mathbf{I}^*(u,v). 
		\end{split} \label{eq: s}
	\end{equation}
	The time variation of  visual features in \eqref{eq: s} can be calculated as
	\begin{equation}
		\begin{split}
			\dot{\mathbf{s}} &= \sum_{u} \sum_{v} \left( \mathbf{p}_{nm}(u,v) \dot{\mathbf{I}}(u,v) + \dot{\mathbf{p}}_{nm}(u,v) \mathbf{I}(u,v)\right) , \\
			\dot{\mathbf{s}}^* &= \sum_{u} \sum_{v} \dot{\mathbf{p}}_{nm}(u,v) \mathbf{I}^*(u,v).
		\end{split}
	\end{equation}
	Hence, \eqref{eq: de} can be expressed as
	\begin{align}
		\dot{\mathbf{e}} &= \sum_{u} \sum_{v}  \left( \mathbf{p}_{nm}(u,v) \dot{\mathbf{I}}(u,v) \right.\notag\\
  & \qquad \qquad \quad + \dot{\mathbf{p}}_{nm}(u,v) \left( \mathbf{I}(u,v) - \mathbf{I}^*(u,v)\right) \Big ) .  \label{eq: dot_e}
	\end{align}
	
	\subsection{Interaction Matrix of TMs}
	
	This subsection describes how to calculate the interaction matrix of TMs.
	It is clear from Section \ref{sec: Discrete_Orthogonal_Moment} that the TM operators are not time-varying ($\dot{\mathbf{t}}_{nm}=0$).
	So \eqref{eq: dot_e} can be simplified as
	\begin{equation}
		\dot{\mathbf{e}} = \sum_{u} \sum_{v} \mathbf{t}_{nm}(u,v) \dot{\mathbf{I}}(u,v) .   \label{eq: dot_e_TM}
	\end{equation}
	
	We introduce the calculation of $\dot{\mathbf{I}}$.
	The basic hypothesis assumes the temporal constancy of the brightness for a  physical point between two successive images. This hypothesis leads to the so-called optical flow constraint equation that links the temporal variation of the luminance $I$ to the image motion at pixel point $\mathbf{u} = (u,v)$ \cite{collewet2008visual, bakthavatchalam2018direct}: 
	\begin{equation}
		\nabla  I^\text{T} \dot{\mathbf{u}} + \dot{I} = 0, \label{eq: optical_flow}
	\end{equation}
	where $\nabla I^\text{T} =  [\nabla I_u, \nabla I_v]$ is the spatial gradient at the pixel point $\mathbf{u}$, where $\nabla I_u$  and $\nabla I_v$ are the components along $u$ and $v$ of the image gradient.
	Further,  the relationship linking the time variation in the coordinates of a pixel point in the image with the camera velocity is 
	\begin{equation}
		\dot{\mathbf{u}} = \mathbf{L}_\mathbf{u} \mathbf{v}, \label{eq: dot_x}
	\end{equation}
	where
	\begin{align}
		\mathbf{L}_\mathbf{u} &= \mathbf{L}_\kappa  \mathbf{L}_\mathbf{x} \notag \\
		&=\begin{bmatrix}
			\kappa_u	& 0 \\ 
			0&\kappa_v
		\end{bmatrix}
		\begin{bmatrix}
			-\frac{1}{Z}	& 0 & \frac{x}{Z} & xy & -(1+x^2) & y\\ 
			0& -\frac{1}{Z} & \frac{y}{Z} & 1+y^2 & -xy & -x 
		\end{bmatrix}
		\notag
	\end{align}
	where $\kappa_u$ and $\kappa_v$ are the horizontal and vertical scale factors of the camera intrinsic matrix,
	and $\mathbf{L}_\mathbf{x}$ is the interaction matrix related to a image point $\mathbf{x}=(x,y)$ \cite{franccois2006visual}.
	According to \eqref{eq: optical_flow} and \eqref{eq: dot_x}, we can obtain that
	\begin{equation}
		\dot{I} = \mathbf{L}_\text{I} \mathbf{v},  \label{eq: dot_I}
	\end{equation}
	where
	\begin{equation}
		\mathbf{L}_\text{I} = - \nabla  \mathbf{I}^\text{T} \mathbf{L}_\mathbf{u}. \notag
	\end{equation}	
	
	By plugging \eqref{eq: dot_I} into \eqref{eq: dot_e_TM}, the time variation of  the feature error becomes
	\begin{equation}
		\dot{\mathbf{e}} = \mathbf{L}_{e} \mathbf{v}, 
	\end{equation}
	where the interaction matrix with respect to $\mathbf{e}$ is
	\begin{equation}
		\mathbf{L}_{e} = \sum_{u} \sum_{v} \mathbf{t}_{nm}(u,v) 	\mathbf{L}_\text{I}. \notag
	\end{equation}
	Finally, the $\widehat{\mathbf{L}}_{e}$  can be designed as
	\begin{align}
		\widehat{\mathbf{L}}_{e}& = \frac{1}{2} \left( \mathbf{L}_{e}  + \mathbf{L}_{{e^{*}}}\right) \notag\\
		&= \sum_{u} \sum_{v}  \mathbf{t}_{nm}(u,v)  \frac{\mathbf{L}_\text{I} + \mathbf{L}_\text{I$^*$}}{2},
		\label{eq: TM_Le}
	\end{align}
	since it was efficient for large camera displacements \cite{malis2004improving}.
	
	\subsection{Interaction Matrix of KMs and HMs} \label{sec: KM_HM_Le}
	
	This subsection describes how the KM and HM interaction matrices are calculated.
	Both the KM and HM operators are time-varying ($\dot{\mathbf{k}}_{nm}\neq0, \dot{\mathbf{h}}_{nm} \neq0,$).
	Hence, $\dot{\mathbf{p}}_{nm}(u,v) $ in \eqref{eq: dot_e} needs to be calculated.
	In the remainder of the paper, we will omit the subscript $nm$ and the arguments $(u,v)$ for clarity.
	
	Following Sections \ref{sec: Krawtchouk_Moments_Parameters} and \ref{sec: HM Parameters}, the KM and HM parameters are adjusted to ensure the ROI of the operator varies with the image.
	Therefore, it is reasonable to formulate the hypothesis of temporal constancy for the KM and HM operators.
	Based on \eqref{eq: x_bar_c_y_bar_c}, we see that the rate of change of the operators $\mathbf{p}$ is half that of the image $\mathbf{I}$. 
	We can get that
	\begin{equation}
		p(\mathbf{u}, t) = p(\mathbf{u} + \frac{\Delta \mathbf{u}}{2}, t+\Delta t), \label{eq: p_constancy}
	\end{equation}
 where KM parameters (${}^{\alpha}p$ and ${}^{\beta}p$) or HM parameters (${}^{\alpha}a$, ${}^{\alpha}b$, ${}^{\beta}a$, and ${}^{\beta}b$) are omitted for compactness.
	A first-order Taylor development of \eqref{eq: p_constancy} gives
	\begin{equation}
		\frac{1}{2} \frac{\partial p}{\partial \mathbf{u}}  \mathbf{\dot{u}} + \frac{\partial p}{\partial t}   = \frac{1}{2} \nabla p^T \dot{\mathbf{u}} + \dot{p} = 0,
	\end{equation}
	where $\nabla p^\text{T} =  [\nabla p_u, \nabla p_v]$ is the spatial gradient of the operators $p$ and $\dot{p}$ is its time derivation.
	So $\dot{p}$ can be expressed as
	\begin{equation}
		\dot{p} = -\frac{1}{2} \left( \nabla p_u \dot{u} + \nabla p_v \dot{v} \right). \label{eq: dot_p}
	\end{equation}
	
	By plugging \eqref{eq: dot_p} into \eqref{eq: dot_e}, the time variation of  the feature error becomes
	\begin{equation}
		\begin{split}
			\dot{\mathbf{e}} &= \sum_{u} \sum_{v}  \left( \mathbf{p} \dot{\mathbf{I}} - \frac{1}{2} \left( \nabla \mathbf{p}_u \dot{u} + \nabla \mathbf{p}_v \dot{v} \right) \left( \mathbf{I} - \mathbf{I}^*\right)\right) \\
			&=\sum_{u} \sum_{v} \left(  \mathbf{p} \dot{\mathbf{I}} - \frac{1}{2} \left( \nabla \mathbf{p}_u \mathbf{I} \dot{u} + \nabla \mathbf{p}_v  \mathbf{I}\dot{v} \right) \right.\\
			& \qquad \qquad \qquad \left. +  \frac{1}{2} \left( \nabla \mathbf{p}_u \mathbf{I}^* \dot{u} + \nabla \mathbf{p}_v  \mathbf{I}^*\dot{v} \right)\right) .
		\end{split} \label{eq: dot_e_}
	\end{equation}
	Green's theorem can be expressed as
	\begin{equation}
		\sum_{u} \sum_{v}  \left( \frac{\partial Q}{\partial u} - \frac{\partial P}{\partial v} \right) = \sum_{\partial u} P + \sum_{\partial v} Q. \label{eq: Green}
	\end{equation}
	We define $Q=pI$ and $P=0$, then
	\begin{equation}
		\frac{\partial Q}{\partial u} = \nabla p_u I + p \nabla I_u, \quad \frac{\partial P}{\partial v} = 0. \label{eq: Green_u}
	\end{equation}
	Substituting \eqref{eq: Green_u} in \eqref{eq: Green}, we get
	\begin{equation}
		\sum_{u} \sum_{v} \nabla p_u I = \sum_{\partial v} pI - \sum_{u} \sum_{v} p \nabla I_u. \label{eq: Green_u_}
	\end{equation}
	It is reasonable to assume that the $pI$ lying on the border are all zero.
        Note this assumption is superior to the information persistence assumption in \cite{bakthavatchalam2018direct, crombez2018visual}, which requires that a uniformly colored $black^2$ background surrounds the acquired image.
        Another case in our assumption is $p$ lying on the border are zero, which is often satisfied (see Figs. \ref{fig: Krawtchouk_Example} and \ref{fig: Hahn_Example}).
	Based on our assumption, the term $\sum_{\partial v} pI$ in \eqref{eq: Green_u_} equals zero.
	We, therefore, can obtain
	\begin{align}
		\sum_{u} \sum_{v} \nabla p_u I &= - \sum_{u} \sum_{v} p \nabla I_u, \notag \\
		\sum_{u} \sum_{v} \nabla p_u I^* &= - \sum_{u} \sum_{v} p \nabla I^*_u. \label{eq: puI}
	\end{align}
	Similarly, if we define $Q=0$ and $P=pI$, then we can get
	\begin{align}
		\sum_{u} \sum_{v} \nabla p_v I &= - \sum_{u} \sum_{v} p \nabla I_v, \notag \\
		\sum_{u} \sum_{v} \nabla p_v I^* &= - \sum_{u} \sum_{v} p \nabla I^*_v. \label{eq: pvI}
	\end{align}
	By plugging \eqref{eq: puI} and \ref{eq: pvI} into \eqref{eq: dot_e_}, we obtain
	\begin{equation}
		\begin{split}
			\dot{\mathbf{e}} &=\sum_{u} \sum_{v} \left(  \mathbf{p} \dot{\mathbf{I}} + \frac{1}{2} \left( \mathbf{p}  \nabla \mathbf{I}_u \dot{u} + \mathbf{p}   \nabla \mathbf{I}_v \dot{v} \right) \right.\\
			& \qquad \qquad \qquad \left. - \frac{1}{2} \left(  \mathbf{p} \nabla \mathbf{I}^*_u \dot{u} +  \mathbf{p} \nabla \mathbf{I}^*_v  \dot{v} \right)\right).
		\end{split} \label{eq: dot_e_temp}
	\end{equation}
	Based on \eqref{eq: optical_flow}, \eqref{eq: dot_e_temp} can be simplified as
	\begin{equation}
		\begin{split}
			\dot{\mathbf{e}} &=\sum_{u} \sum_{v} \left(  \mathbf{p} \dot{\mathbf{I}} - \frac{1}{2}\mathbf{p} \dot{\mathbf{I}} + \frac{1}{2}\mathbf{p} \dot{\mathbf{I}} ^*\right) \\
			&= \sum_{u} \sum_{v} \mathbf{p} \frac{\dot{\mathbf{I}} +\dot{\mathbf{I}} ^*}{2} =  \mathbf{L}_{e} \mathbf{v},
		\end{split}
	\end{equation}
	where the interaction matrix with respect to $\mathbf{e}$ is
	\begin{equation}
		\widehat{\mathbf{L}}_{e} = \sum_{u} \sum_{v}  \mathbf{p} \frac{\mathbf{L}_\text{I} + \mathbf{L}_\text{I$^*$}}{2}.
		\label{eq: KM_HM_Le}
	\end{equation}
	
	It is essential to note that the \eqref{eq: TM_Le} and \eqref{eq: KM_HM_Le} are identical, so the interaction matrix for TMs, KMs, and HMs has a unified form.
	Also, the $\dot{\mathbf{p}}$ is not needed anymore to compute the interaction matrix.

	\section{Experimental Results} \label{sec: Simulation_and_Experiment}
	
	In this section, we evaluate the proposed control scheme by combining simulation and experimental results.
	Since the generic framework to consider DOMs as visual features is proposed for the first time, the VS schemes (TM-VS, KM-VS, and HM-VS) are compared in Section \ref{sec: Comparisons}.
	Then, Section \ref{sec: 3D_sim} presents results for challenging experiments highlighting the contribution of using DOMs as visual features.
	In Section \ref{sec: Robustness}, we investigate the robustness of the proposed VS scheme when some noise is added to the images. 
	The HM-VS scheme is compared with a baseline method and two state-of-the-art methods in Section \ref{sec: state-of-art}.
	Section \ref{sec: 3D_exp} shows experiments conducted with a robot in real environments. 
	Finally, the minimum and maximum orders ($l_{\text{min}}$ and $l_{\text{max}}$) are discussed in Section \ref{sec: Discussion}.
	
	\subsection{VS in Classical Simulation Environments} \label{sec: Comparisons}
	Simulation results were first performed  to compare TM-VS, KM-VS, and HM-VS.
	Given a vision sensor and a target object as examples,
	the co-simulation was performed on the MATLAB 2021b
	and CoppeliaSim 4.2 platforms.
	In the following simulations, the image size is $512 \times 512$, and the minimum and maximum orders are $l_{\text{min}}=4$ and $l_{\text{max}}=8$, respectively.
	
\begin{figure}
		\centering 
		
		\subfloat[]{\includegraphics[width=0.45\hsize]{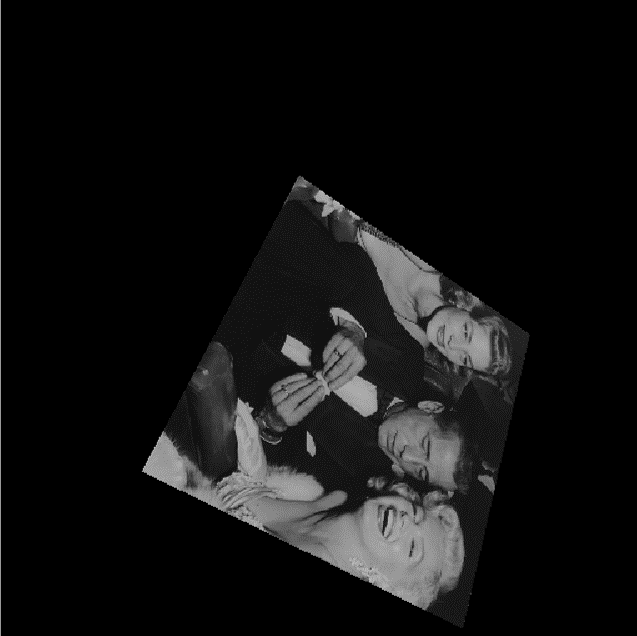}  \label{fig: In_Field_image_new}}	
		\subfloat[]{\includegraphics[width=0.45\hsize]{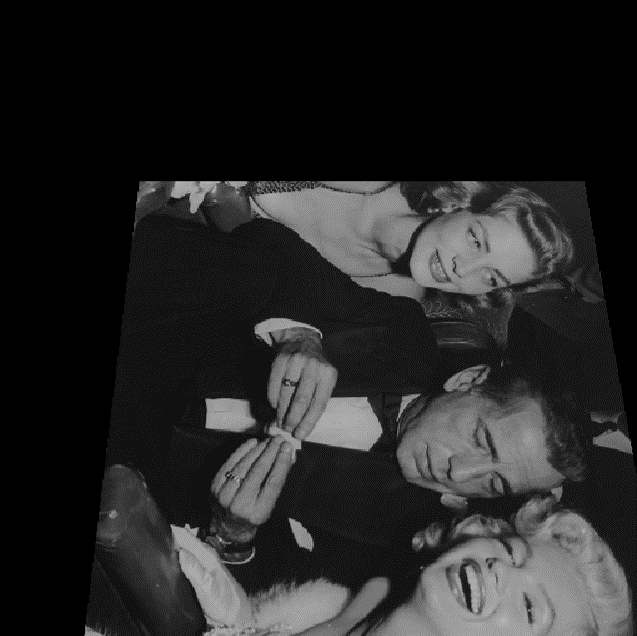}  \label{fig: In_Field_image_old}}		
		
		\subfloat[]{\includegraphics[width=0.49\hsize]{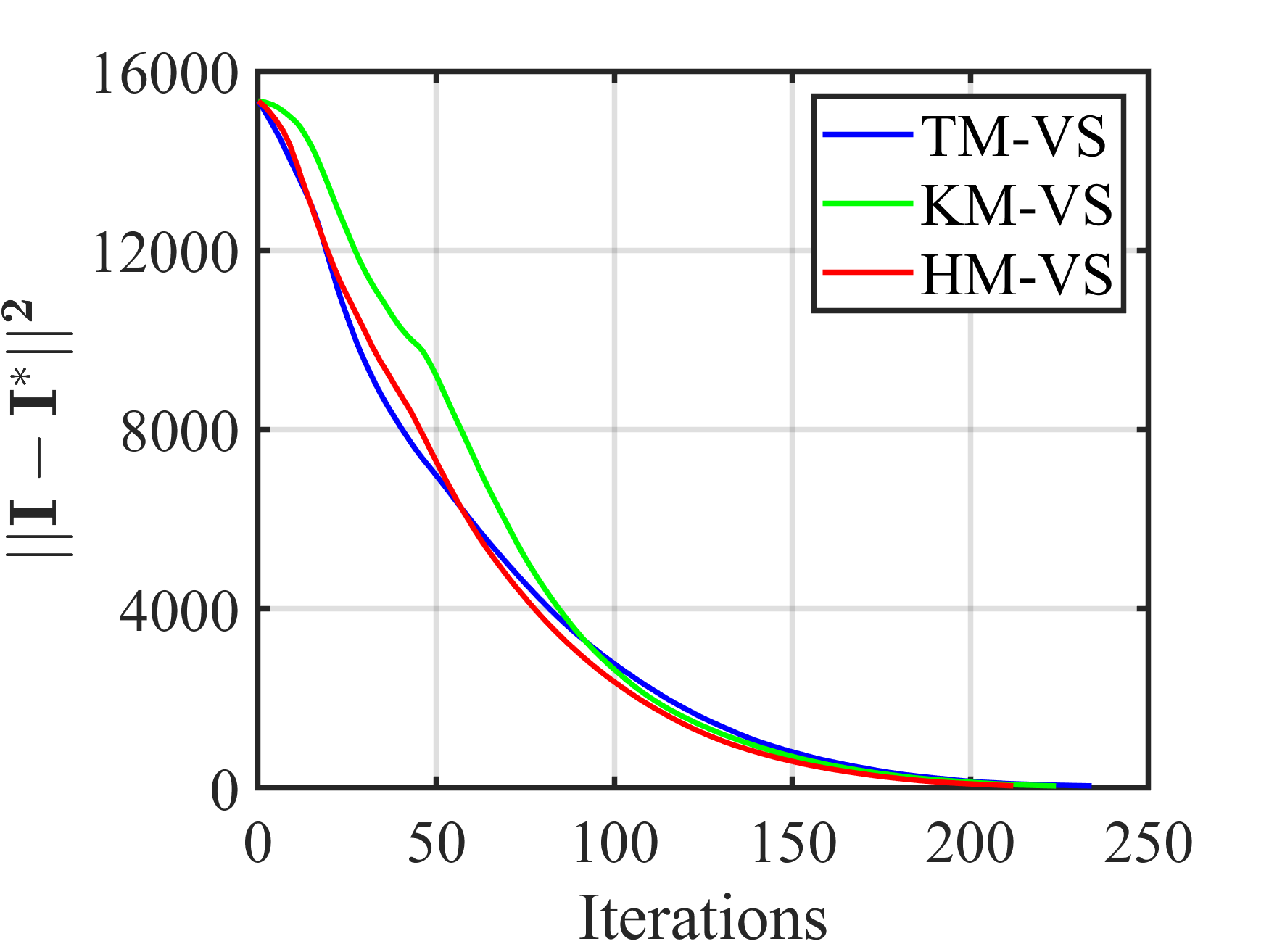}  \label{fig: In_Field_I}}	
		\subfloat[]{\includegraphics[width=0.49\hsize]{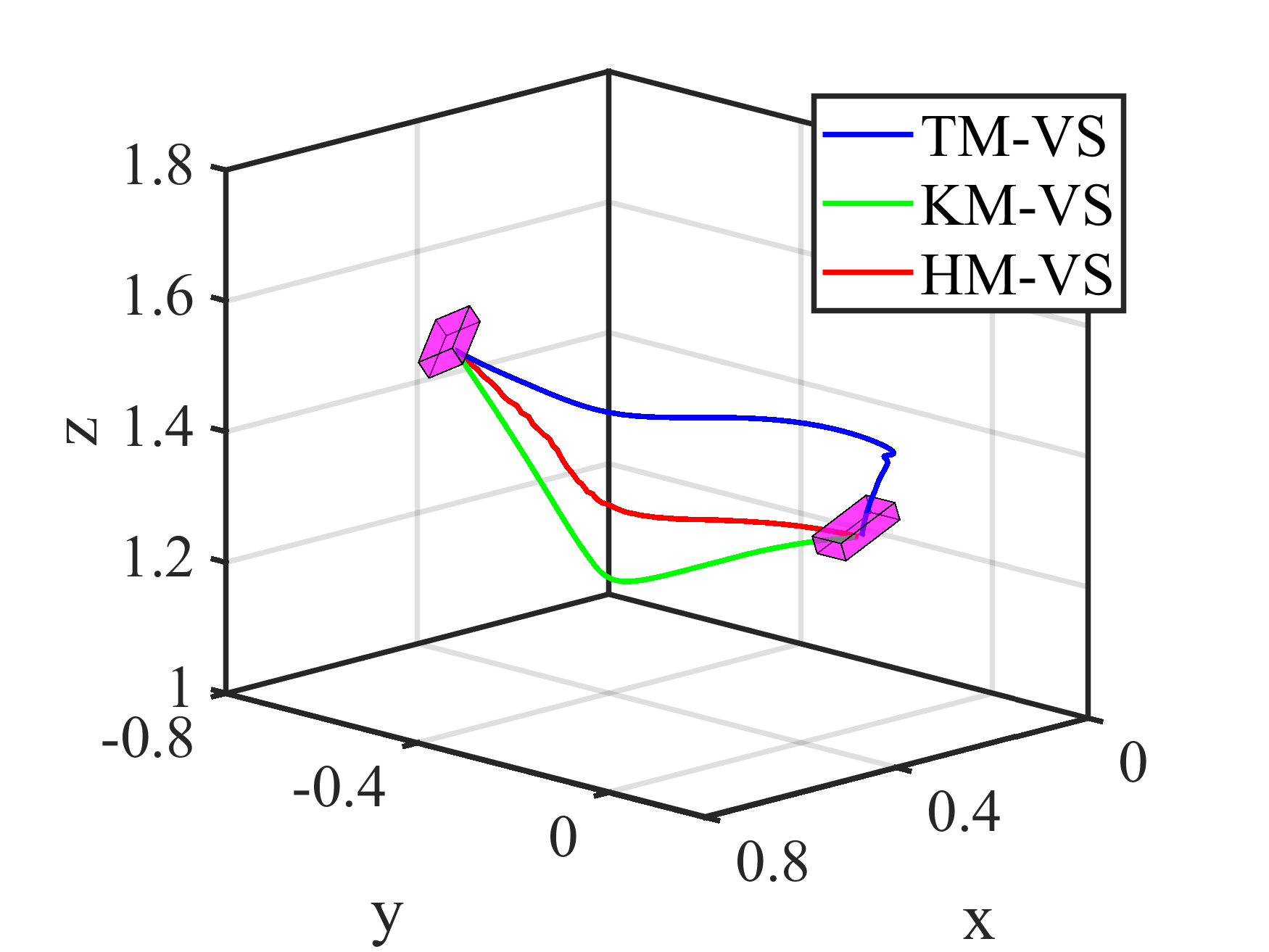}  \label{fig: In_Field_trajections}}		
		
		\caption{Experiment \#1: Comparison between TM-VS, KM-VS, and HM-VS in a classic scene. (a) Initial image. (b) Desired image. (c) Pixel errors. (d) Camera trajectories (in m).}
		
		\label{fig: In_Field}
	\end{figure}

	\subsubsection*{Experiment \#1 (see Fig. \ref{fig: In_Field})}
	
	This experiment has been carried out using a classic scene and controlling 6-DoF.
        The VS uses the true depth value for each pixel.
	Figs. \ref{fig: In_Field_image_new} and \ref{fig: In_Field_image_old} show the initial and desired images, respectively. 
	The error between the initial and desired pose is given by $(0.48\text{m}$, $-0.26\text{m}$, $-0.37\text{m}$, $-4.54^{\circ}$, $-17.06^{\circ}$, $-30.37^{\circ})$.
	Since the visual features computed by the TM, KM, and HM are different from each other, we compare the convergence of the three schemes by pixel errors $||\mathbf{I} - \mathbf{I}^*||^2$, and the results are shown in Fig. \ref{fig: In_Field_I}.
	For TM-VS and KM-VS, HM-VS has a faster convergence rate ($213$ iterations vs. $235$ and $225$).
	Although we cannot control the camera's trajectory in Cartesian space, it is clear from Fig. \ref{fig: In_Field_trajections} that the trajectory of HM-VS is better than those of TM-VS and KM-VS.
	The control results for each method are analyzed below.

	\begin{figure}
		\centering 
		
		\subfloat[]{\includegraphics[width=0.33\hsize]{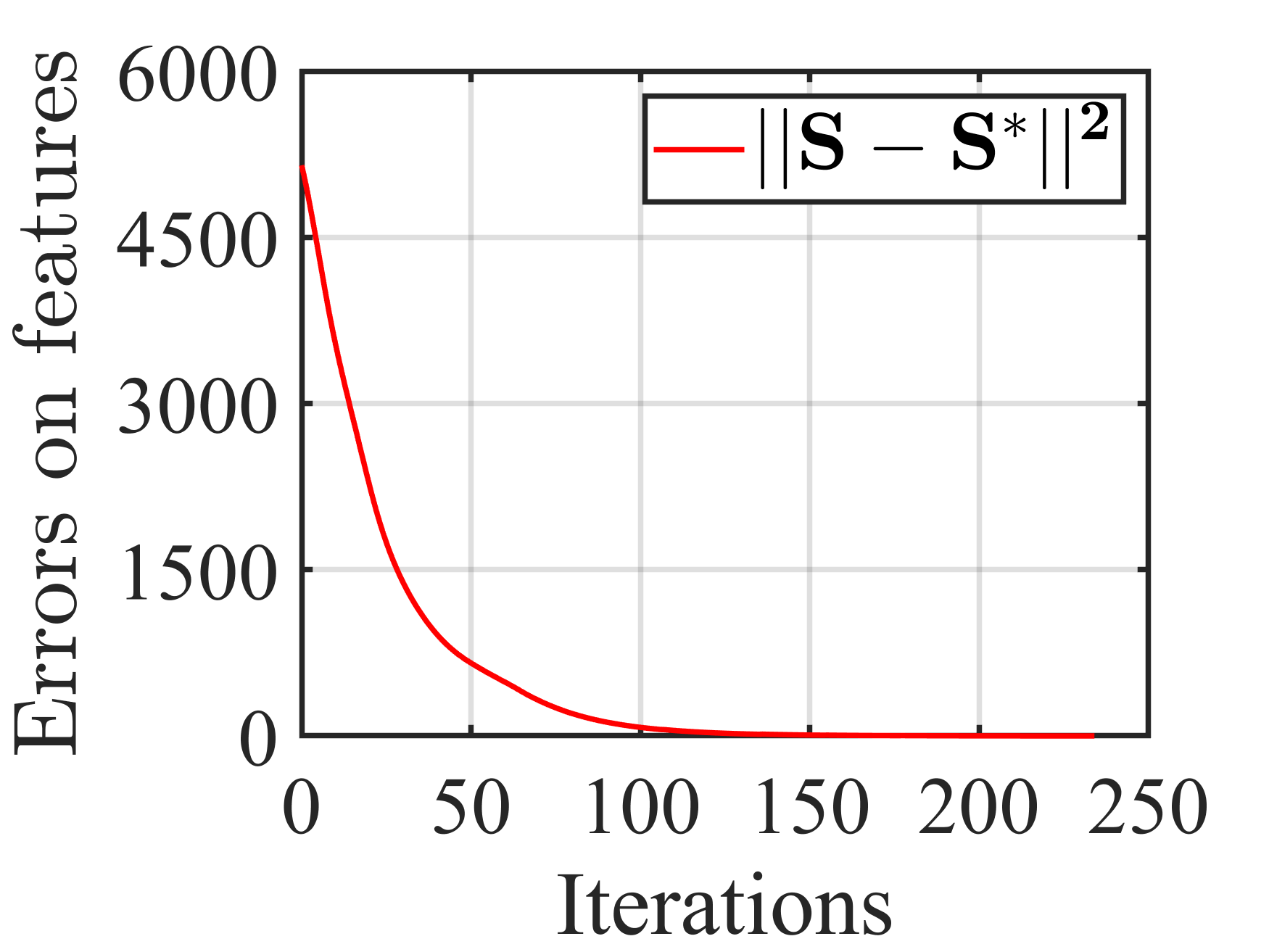}  \label{fig: In_Field_TMVS_feature_error}}	
		\subfloat[]{\includegraphics[width=0.33\hsize]{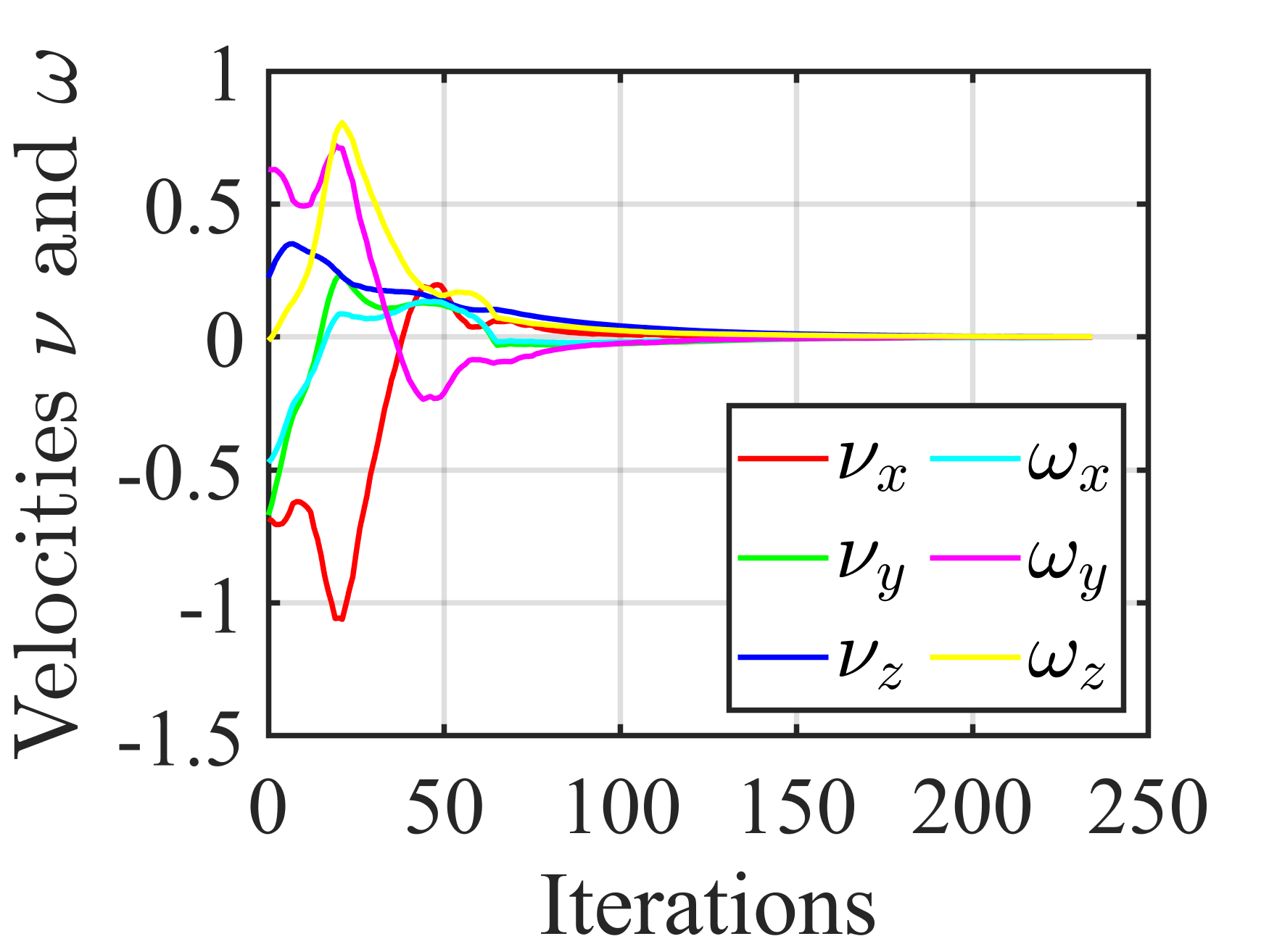}  \label{fig: In_Field_TMVS_velocity}}		
		\subfloat[]{\includegraphics[width=0.33\hsize]{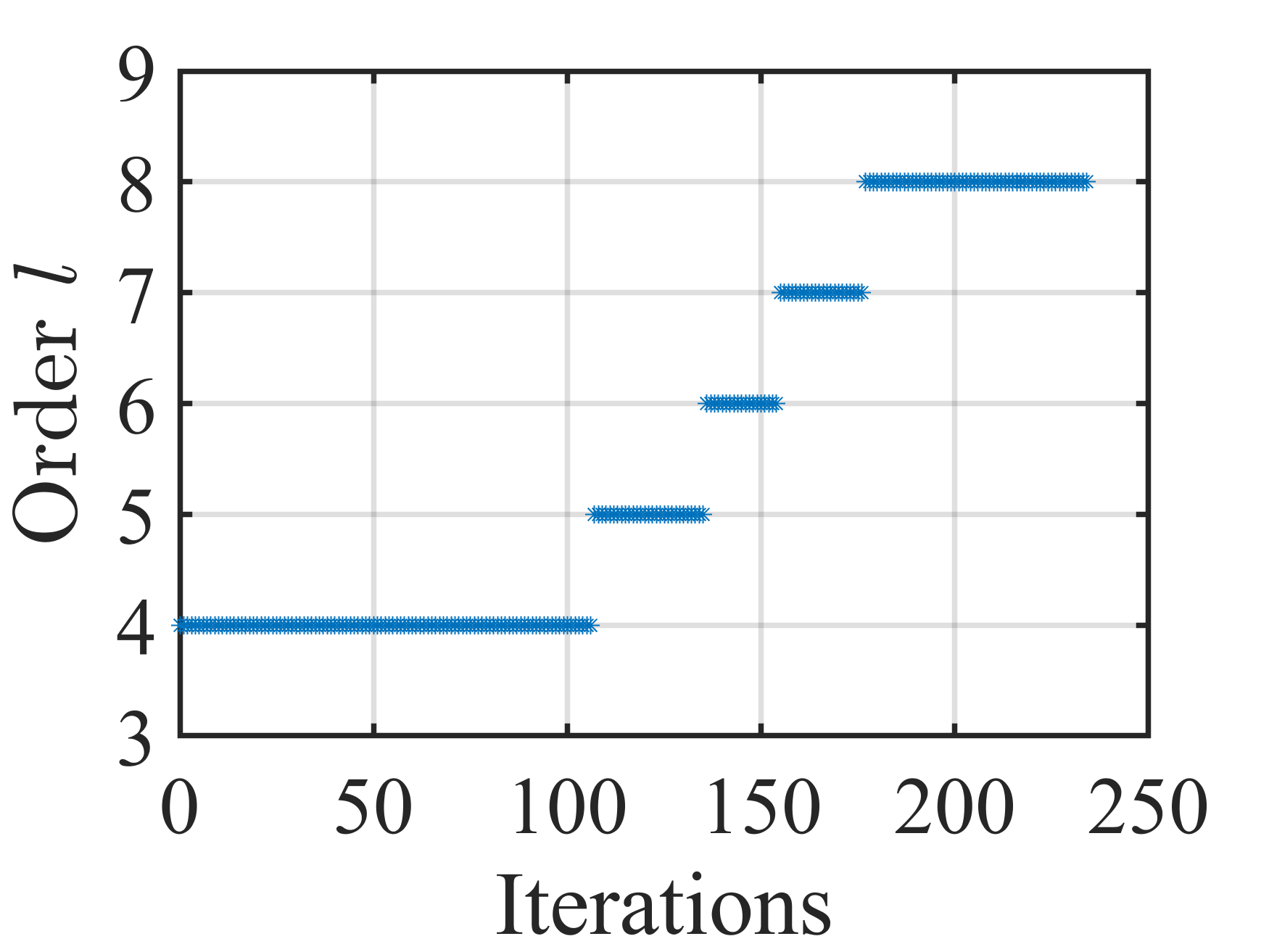}  \label{fig: In_Field_TMVS_order}}		
		
		\caption{Results for TM-VS in Experiment \#1. (a) Errors on features. (b) Camera velocities (in m/s and rad/s). (c) Order of  DOMs as visual features.}
		
		\label{fig: In_Field_TMVS}
	\end{figure}
	
	\begin{figure}
		\centering 
		
		\subfloat[]{\includegraphics[width=0.33\hsize]{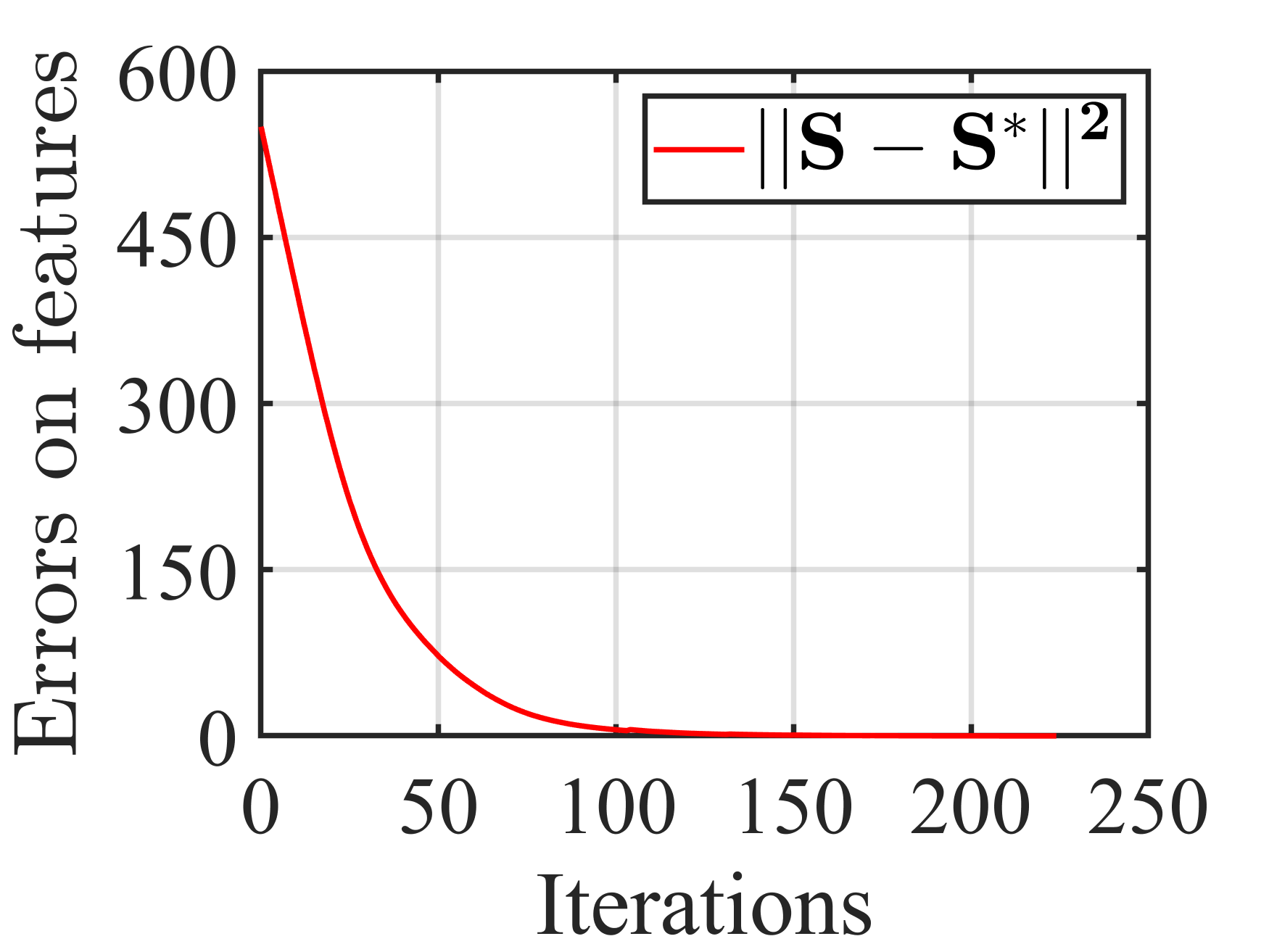}  \label{fig: In_Field_KMVS_feature_error}}	\qquad
		\subfloat[]{\includegraphics[width=0.33\hsize]{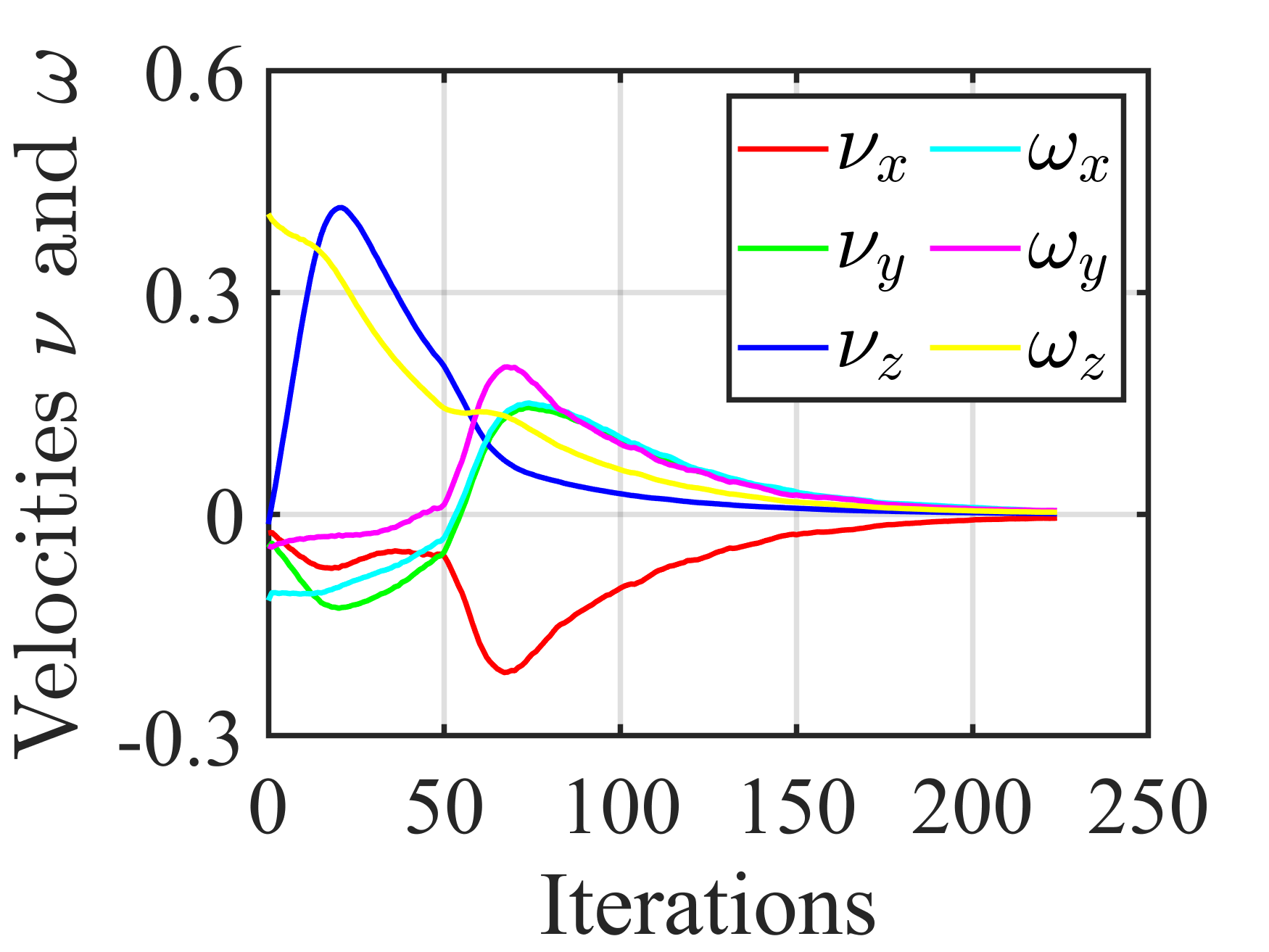}  \label{fig: In_Field_KMVS_velocity}}		
		
		\subfloat[]{\includegraphics[width=0.33\hsize]{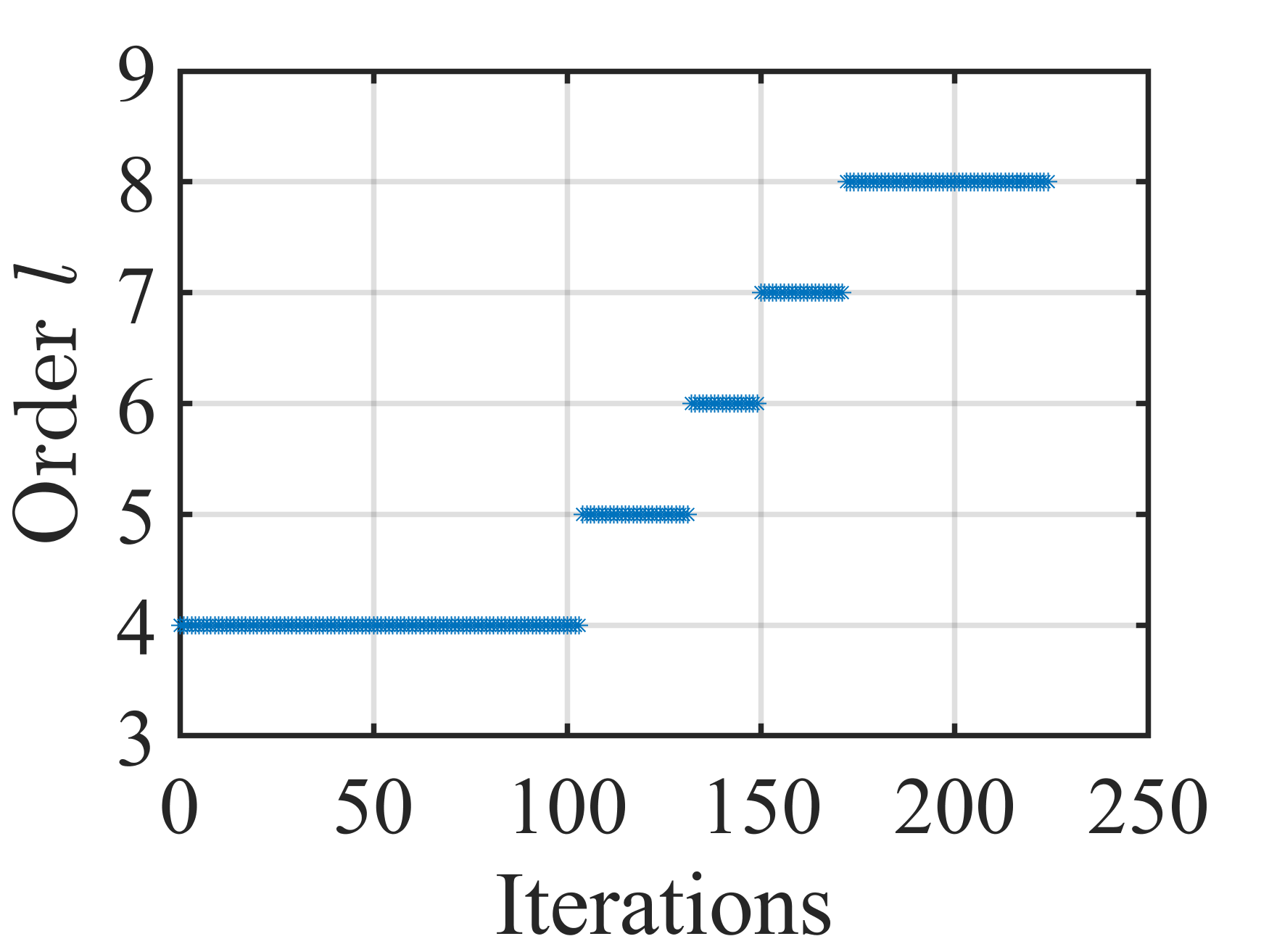}  \label{fig: In_Field_KMVS_order}}		\qquad
		\subfloat[]{\includegraphics[width=0.33\hsize]{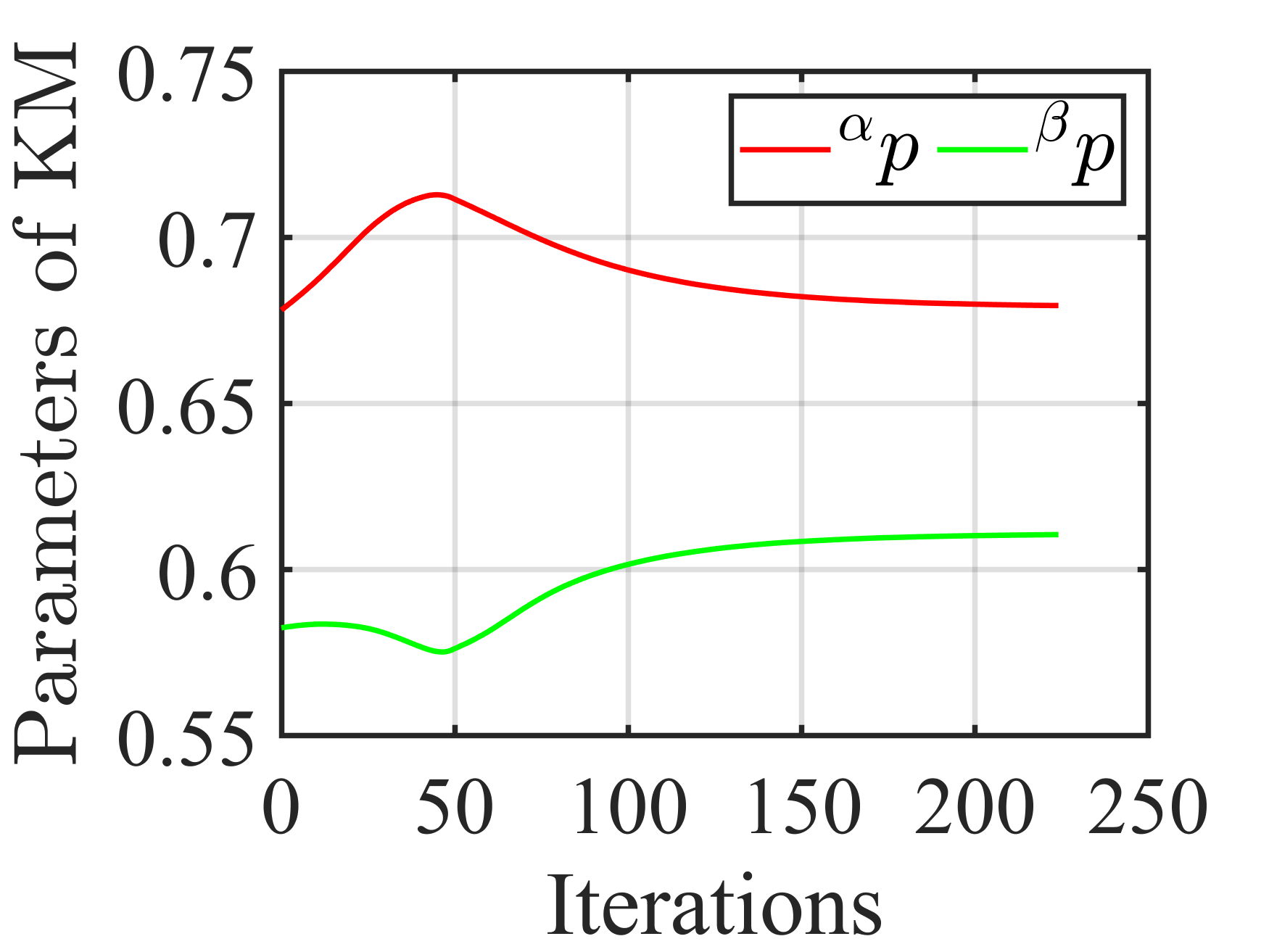}  \label{fig: In_Field_KMVS_p}}				
		
		\caption{Results for KM-VS in Experiment \#1. (a) Errors on features. (b) Camera velocities (in m/s and rad/s). (c) Order of  DOMs as visual features. (d) Parameters of KMs.}
		
		\label{fig: In_Field_KMVS}
	\end{figure}
	
	\begin{figure}
		\centering 
		
		\subfloat[]{\includegraphics[width=0.33\hsize]{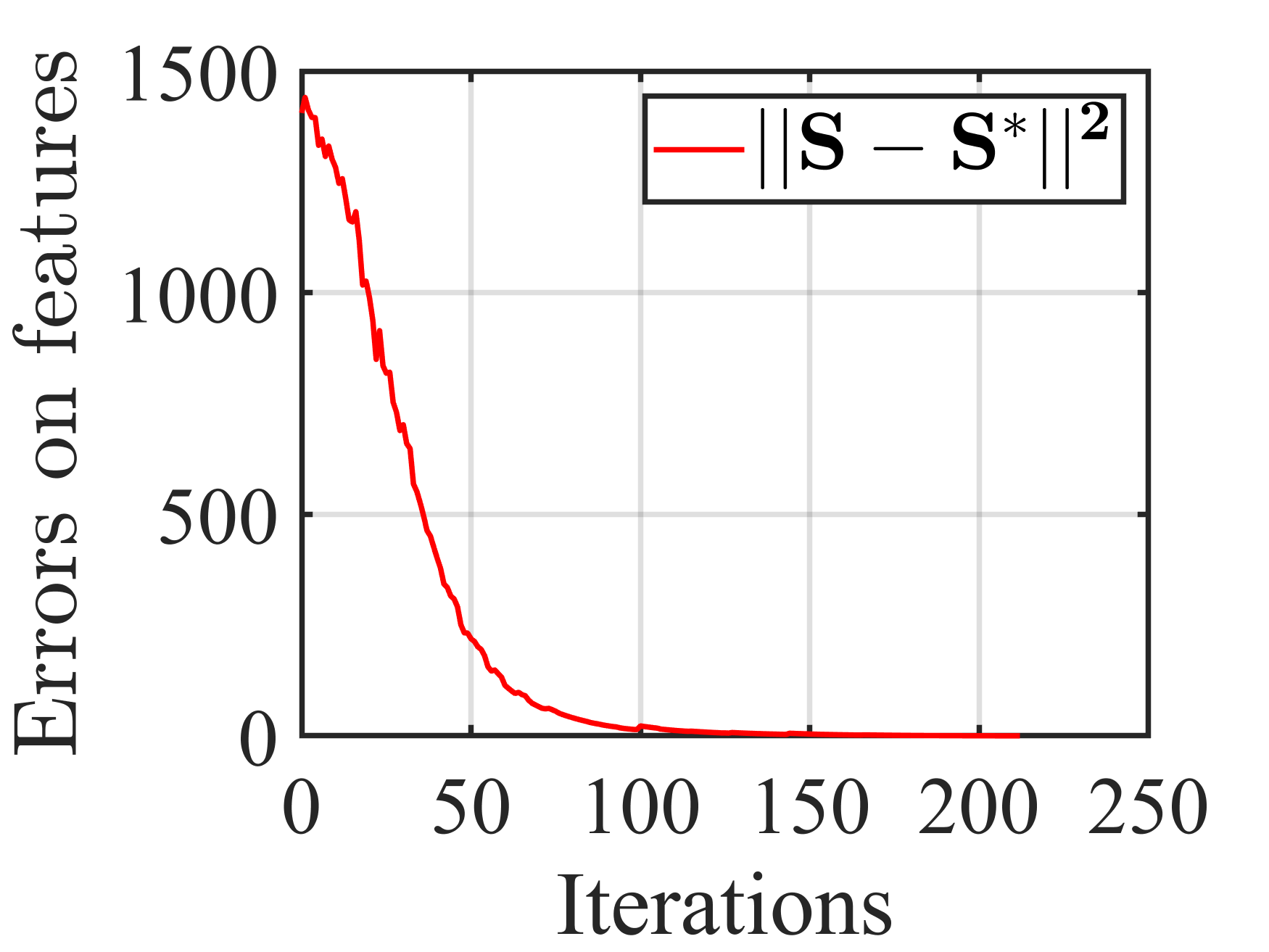}  \label{fig: In_Field_HMVS_feature_error}}	\qquad
		\subfloat[]{\includegraphics[width=0.33\hsize]{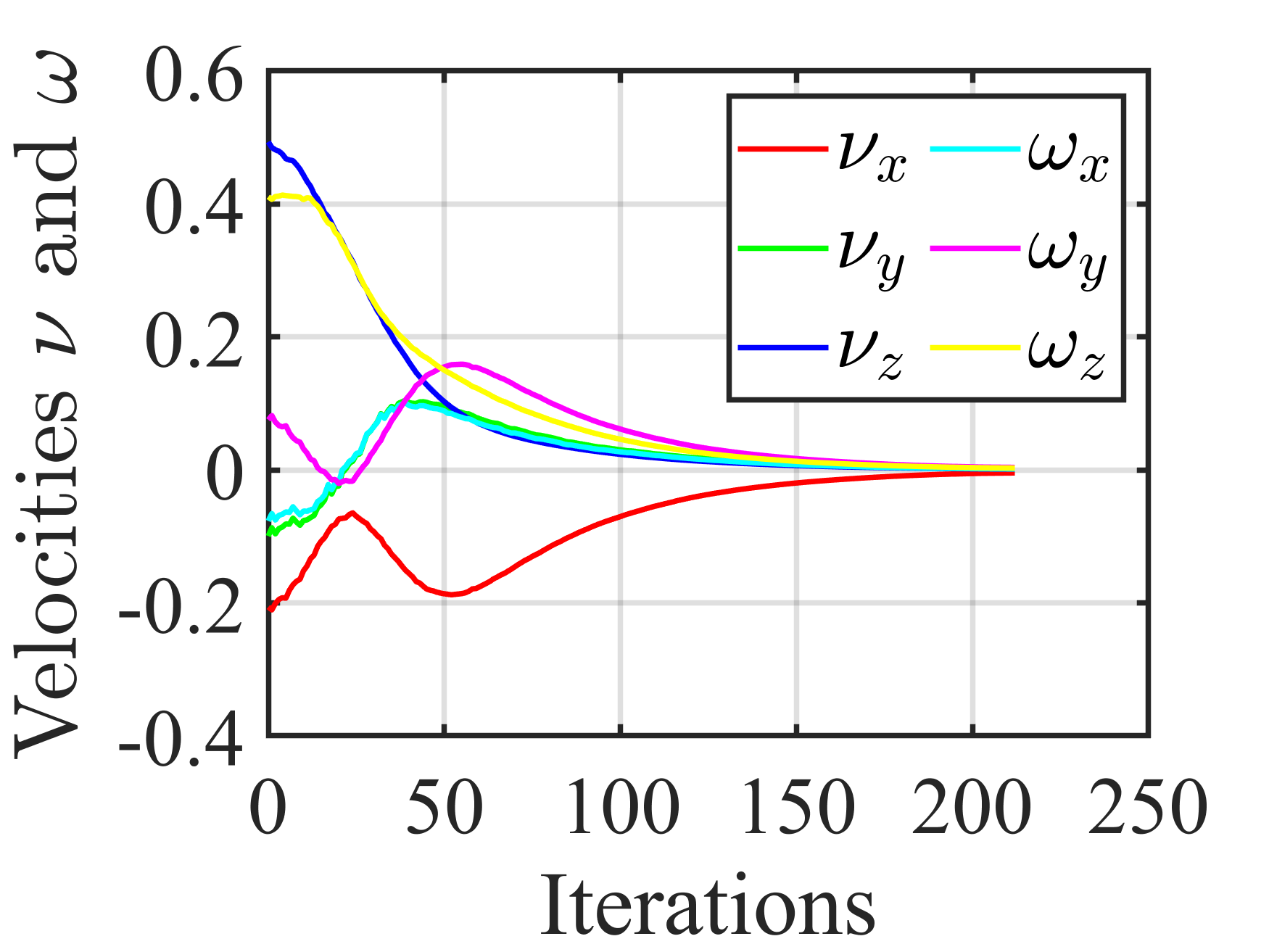}  \label{fig: In_Field_HMVS_velocity}}		
		
		\subfloat[]{\includegraphics[width=0.33\hsize]{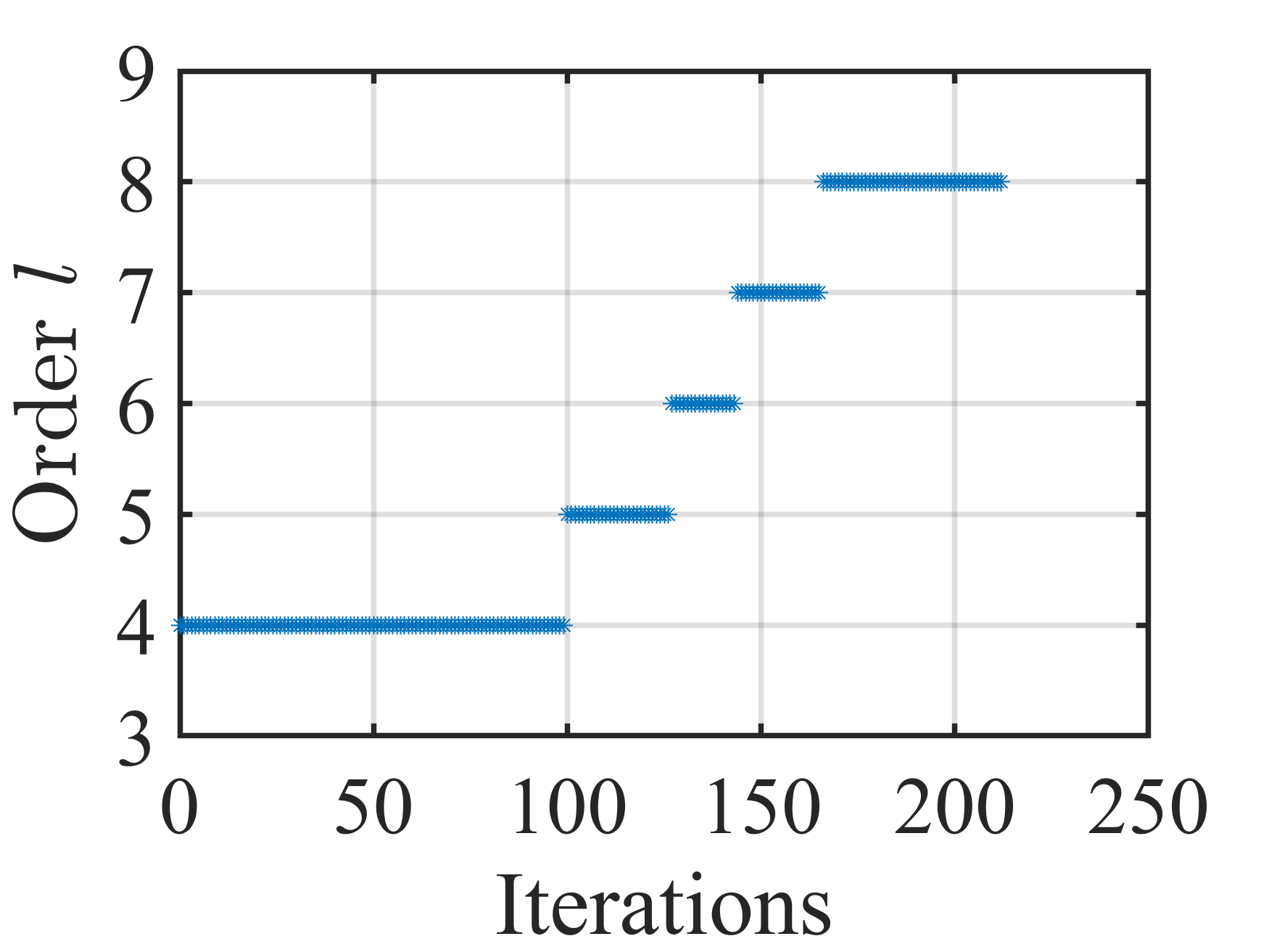}  \label{fig: In_Field_HMVS_order}}		\qquad
		\subfloat[]{\includegraphics[width=0.33\hsize]{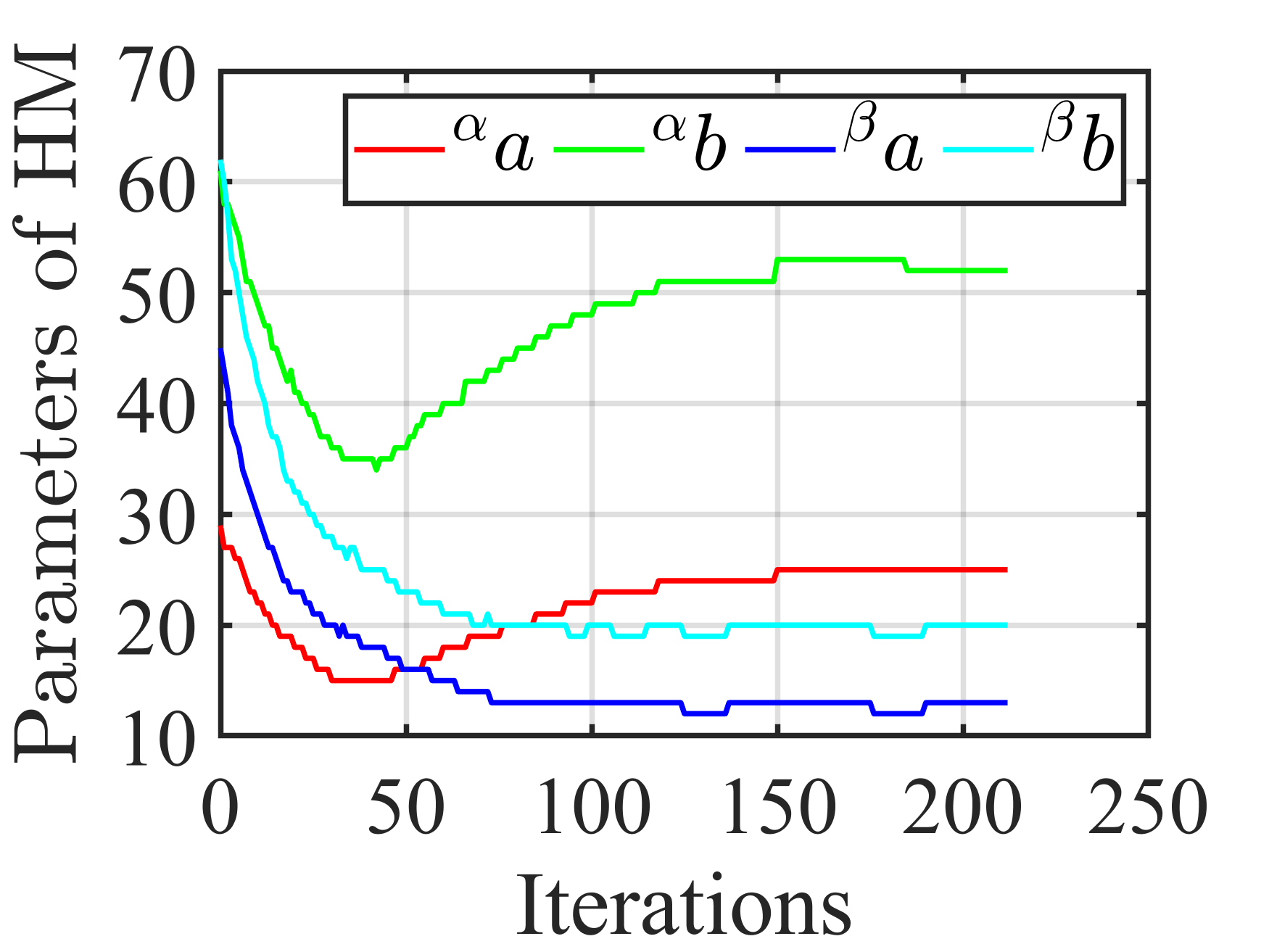}  \label{fig: In_Field_HMVS_ab}}				
		
		\caption{Results for HM-VS in Experiment \#1. (a) Errors on features. (b) Camera velocities (in m/s and rad/s). (c) Order of  DOMs as visual features. (d) Parameters of HMs.}
		
		\label{fig: In_Field_HMVS}
	\end{figure}
	
	Figs. \ref{fig: In_Field_TMVS}, \ref{fig: In_Field_KMVS}, and \ref{fig: In_Field_HMVS} show the results for TM-VS, KM-VS, and HM-VS, respectively.
	The exponentially decreasing feature errors validate the control law we designed in Section \ref{sec: Model_Interaction} (see Figs. \ref{fig: In_Field_TMVS_feature_error}, \ref{fig: In_Field_KMVS_feature_error}, and \ref{fig: In_Field_HMVS_feature_error}).
	The perturbation of the velocity plots (see Figs. \ref{fig: In_Field_TMVS_velocity}, \ref{fig: In_Field_KMVS_velocity}, and \ref{fig: In_Field_HMVS_velocity}) is due
	to the nonlinearity and discontinuity of the cost function, as well as to the changes in the image caused by the appearance and disappearance of portions of the scene from the camera field-of-view.
	Nevertheless, these VS schemes have successfully converged to the desired pose. 
	The order of the DOM during VS is shown in Figs. \ref{fig: In_Field_TMVS_order}, \ref{fig: In_Field_KMVS_order}, and \ref{fig: In_Field_HMVS_order} follows the same trend as we expected in Section \ref{sec: order}.
	The "S"-shaped curve completes the trade-off between convergence rate and accuracy.
	In the KM-VS scheme, the parameters ${}^{\alpha}p$ and ${}^{\beta}p$ of the KM are calculated online using the method proposed in Section \ref{sec: Krawtchouk_Moments_Parameters} to ensure successful control (see Fig. \ref{fig: In_Field_KMVS_p}).
	Similarly, the parameters ${}^{\alpha}a, {}^{\alpha}b, {}^{\beta}a$ and ${}^{\beta}b$ of HMs are calculated online using the approach proposed in Section \ref{sec: HM Parameters} to properly adjust the ROI  in the HM-VS scheme (see Fig. \ref{fig: In_Field_HMVS_ab}).
	
	In summary, the proposed TH-VS, KM-VS, and HM-VS schemes are all effective for the classical scenario.

	\subsection{VS in Complex 2-D and 3-D Simulation Environments} \label{sec: 3D_sim}

\begin{figure}
		\centering 
		
		\subfloat[]{\includegraphics[width=0.45\hsize]{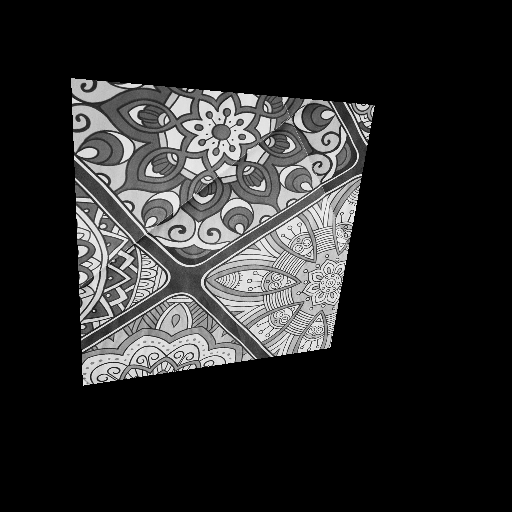}  \label{fig: Out_Field_image_new}}	
		\subfloat[]{\includegraphics[width=0.45\hsize]{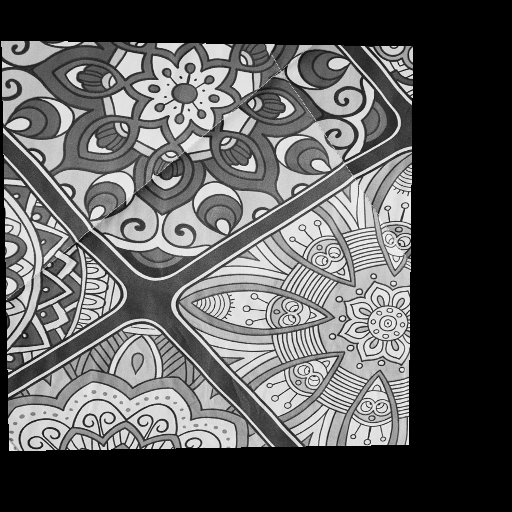}  \label{fig: Out_Field_image_old}}		
		
		\subfloat[]{\includegraphics[width=0.49\hsize]{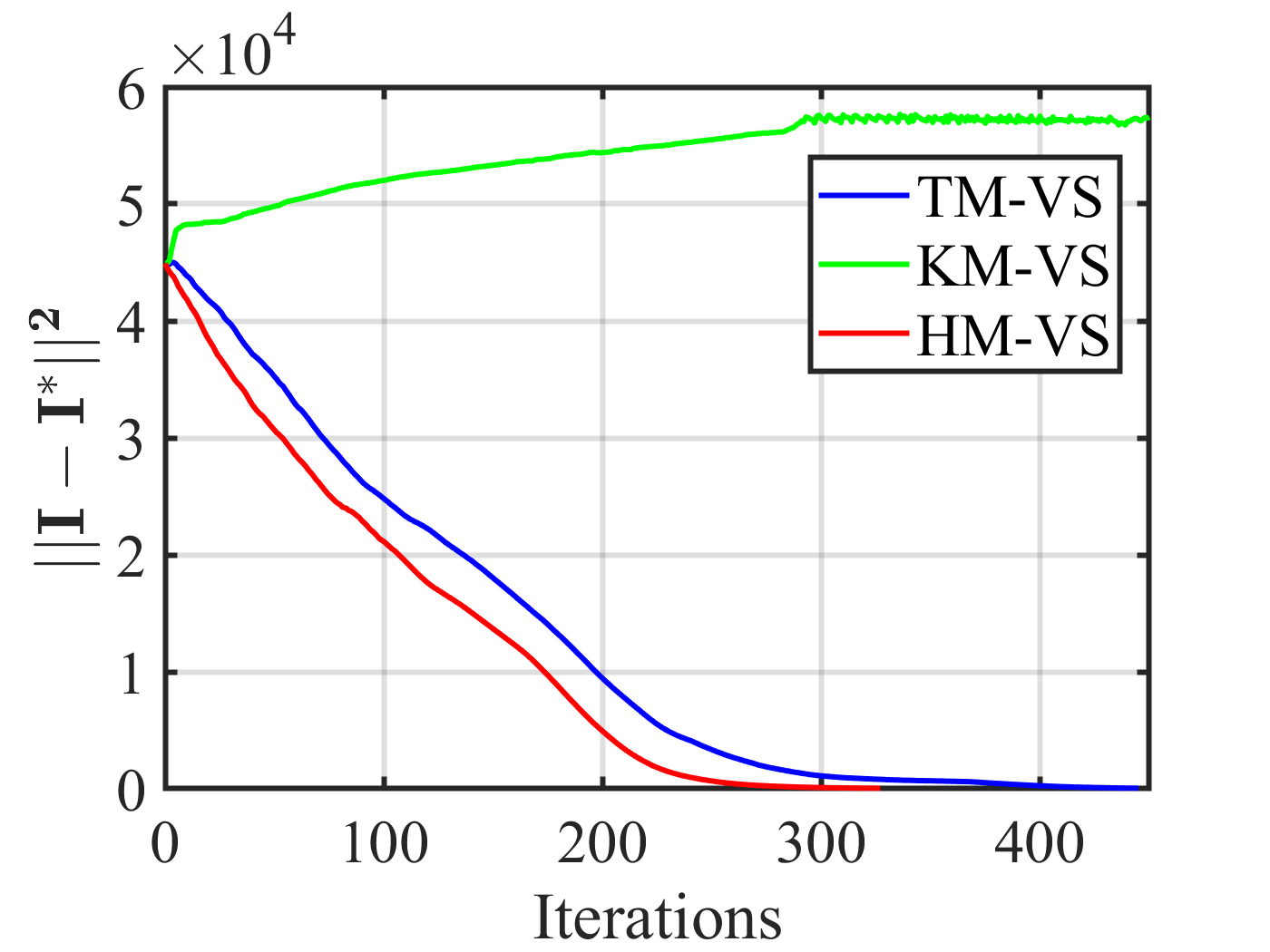}  \label{fig: Out_Field_I}}	
		\subfloat[]{\includegraphics[width=0.49\hsize]{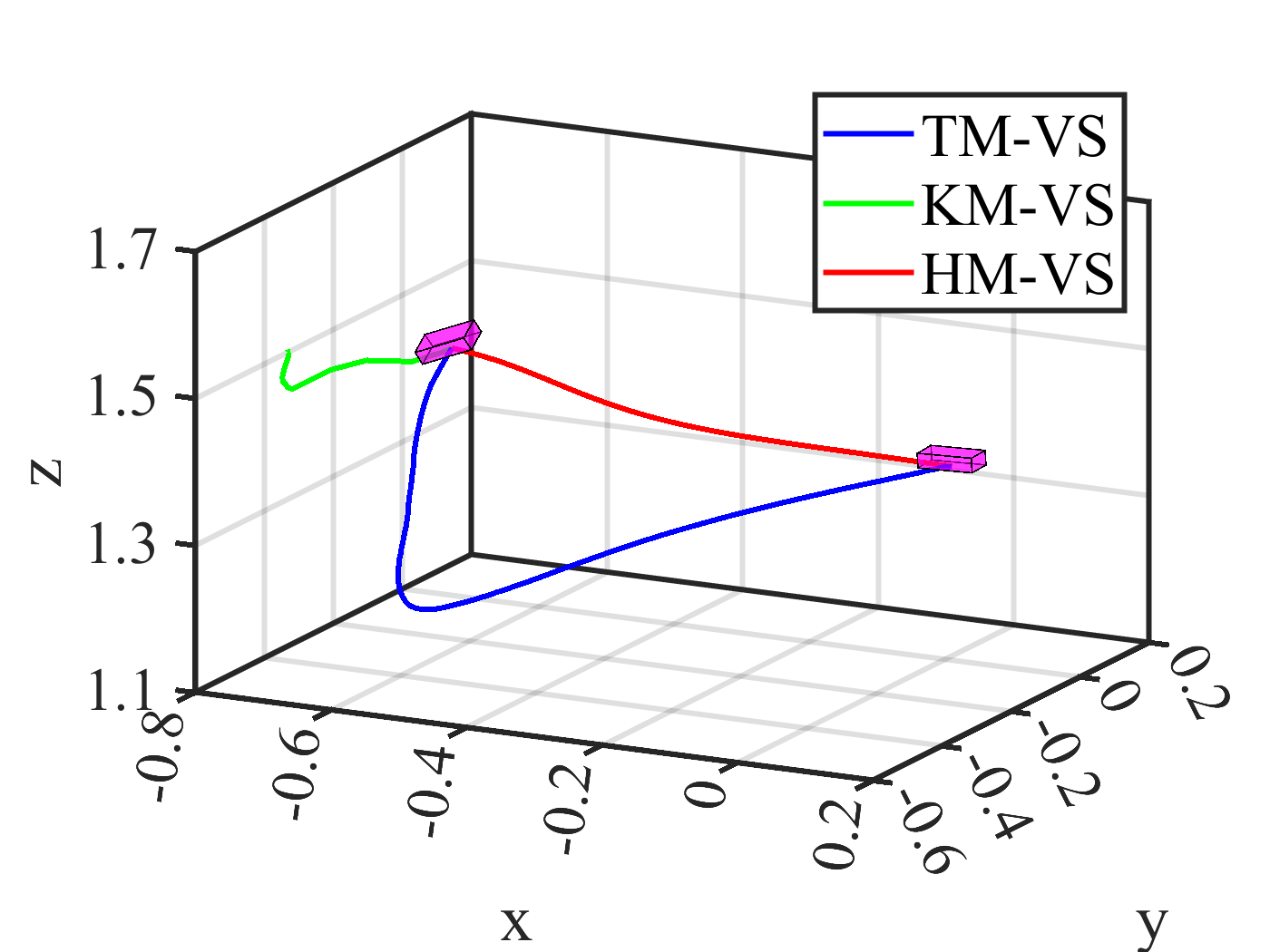}  \label{fig: Out_Field_trajections}}		
		
		\caption{Experiment \#2: Comparison between TM-VS, KM-VS, and HM-VS in a complex 2-D scene. (a) Initial image. (b) Desired image. (c) Pixel errors. (d) Camera trajectories  (in m).}
		
		\label{fig: Out_Field}
	\end{figure}

	\subsubsection*{Experiment \#2 (see Fig. \ref{fig: Out_Field})}
	
\begin{figure}
		\centering 
		
		\subfloat[]{\includegraphics[width=0.33\hsize]{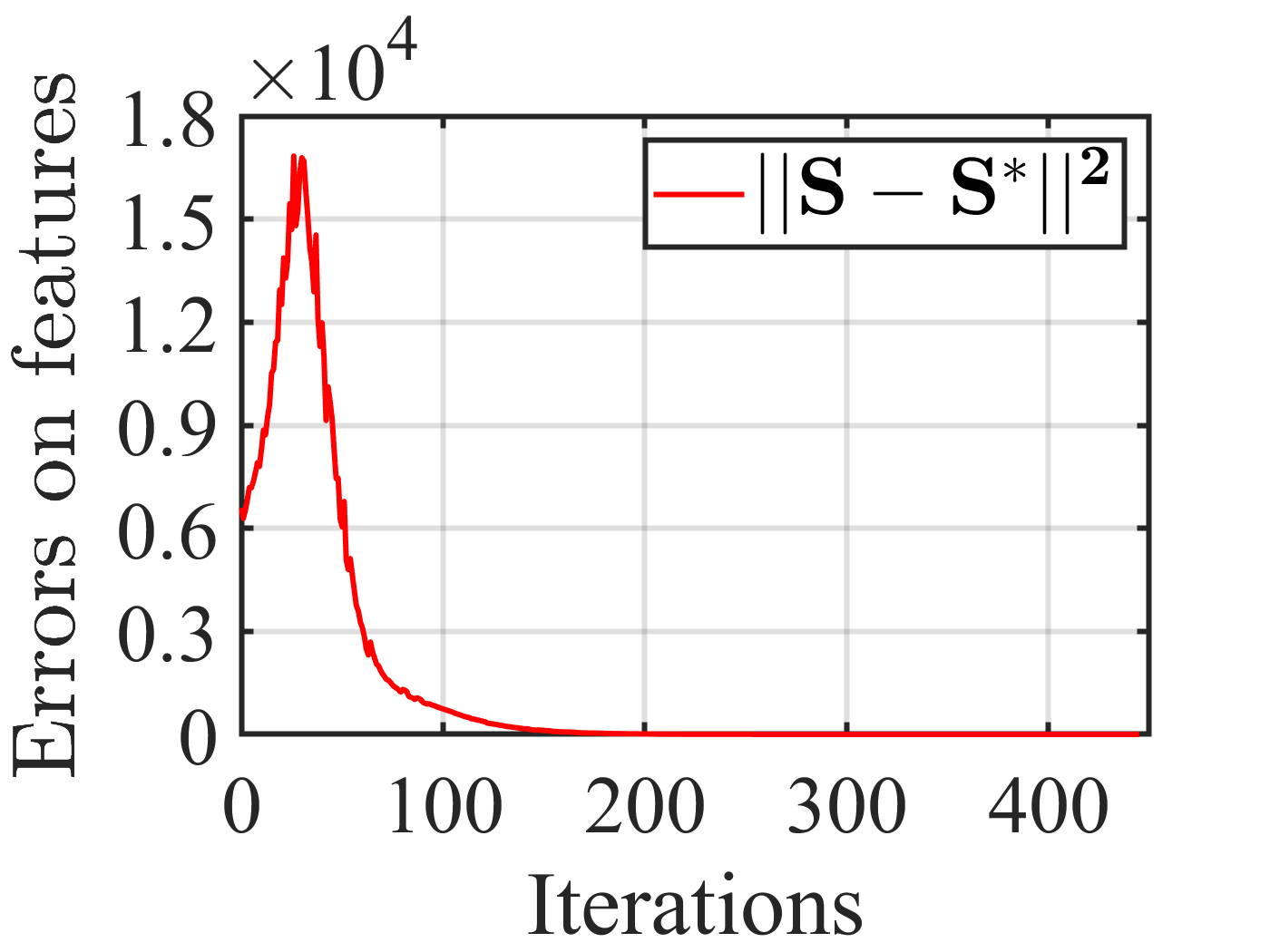}  \label{fig: Out_Field_TMVS_feature_error}}
		\subfloat[]{\includegraphics[width=0.33\hsize]{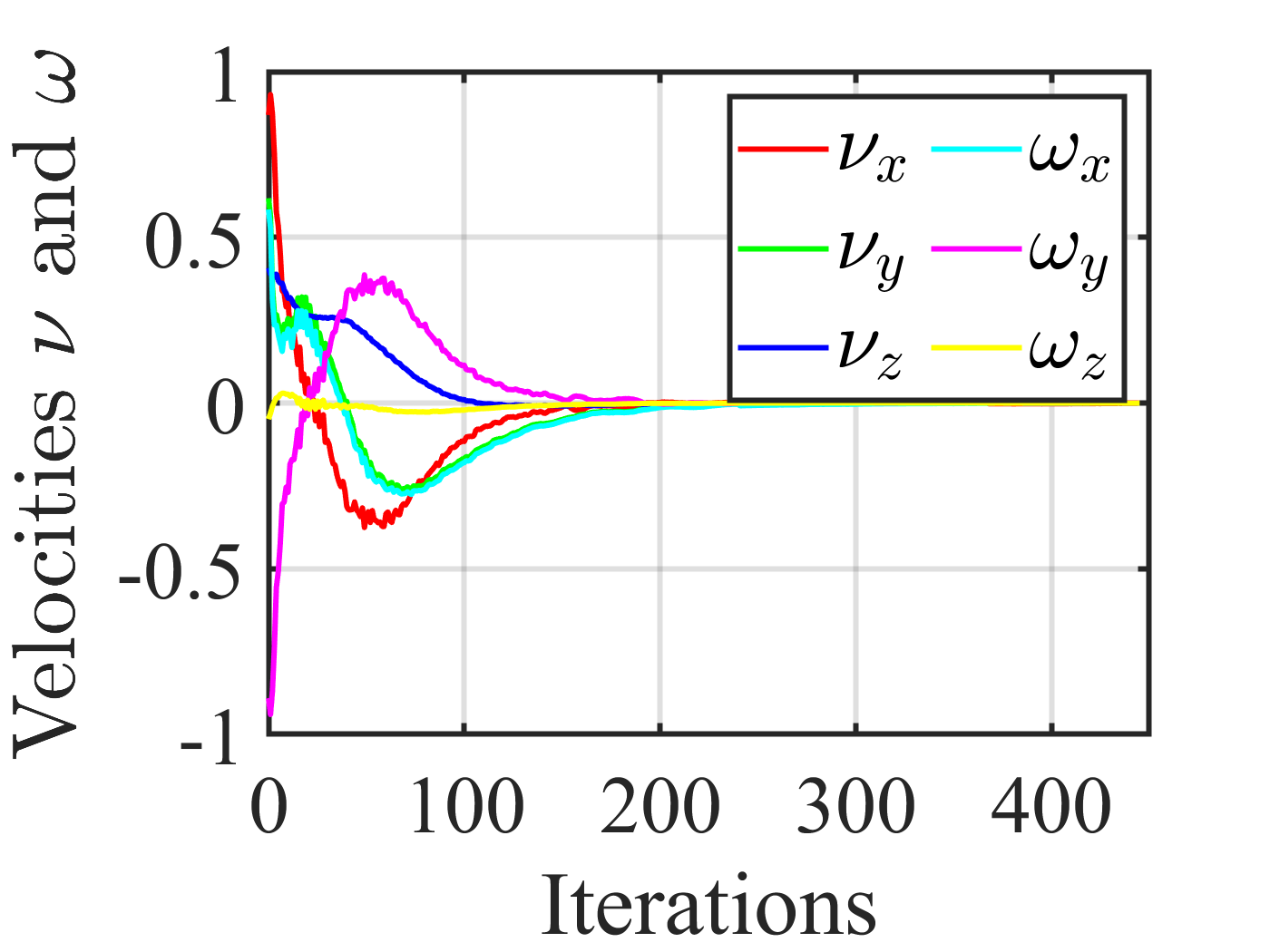}  \label{fig: Out_Field_TMVS_velocity}}		
		\subfloat[]{\includegraphics[width=0.33\hsize]{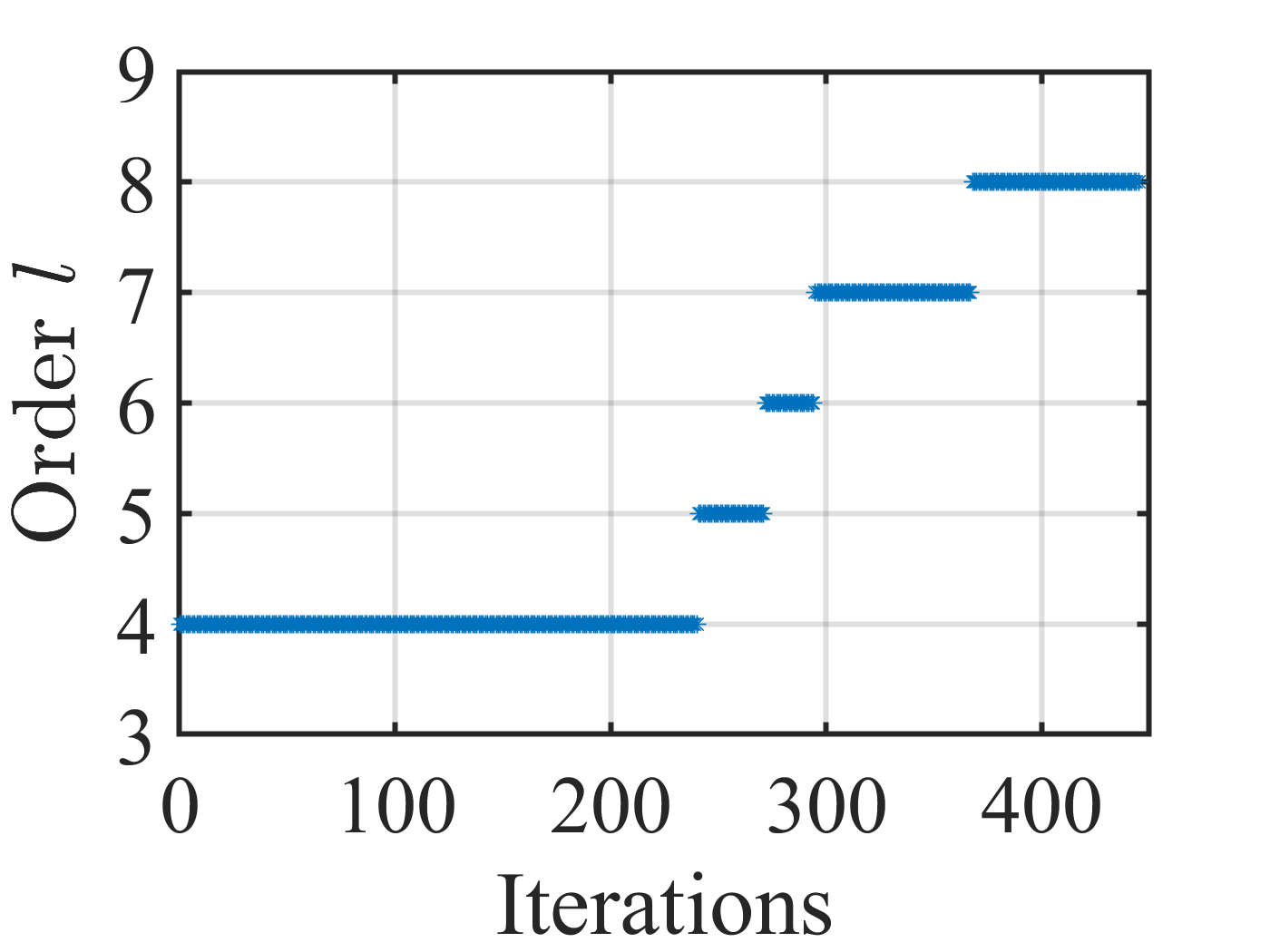}  \label{fig: Out_Field_TMVS_order}}				
		
		\caption{Results for TM-VS in Experiment \#2. (a) Errors on features. (b) Camera velocities (in m/s and rad/s). (c) Order of  DOMs as visual features.}
		
		\label{fig: Out_Field_TMVS}
	\end{figure}

\begin{figure}
		\centering 
		
		\subfloat[]{\includegraphics[width=0.33\hsize]{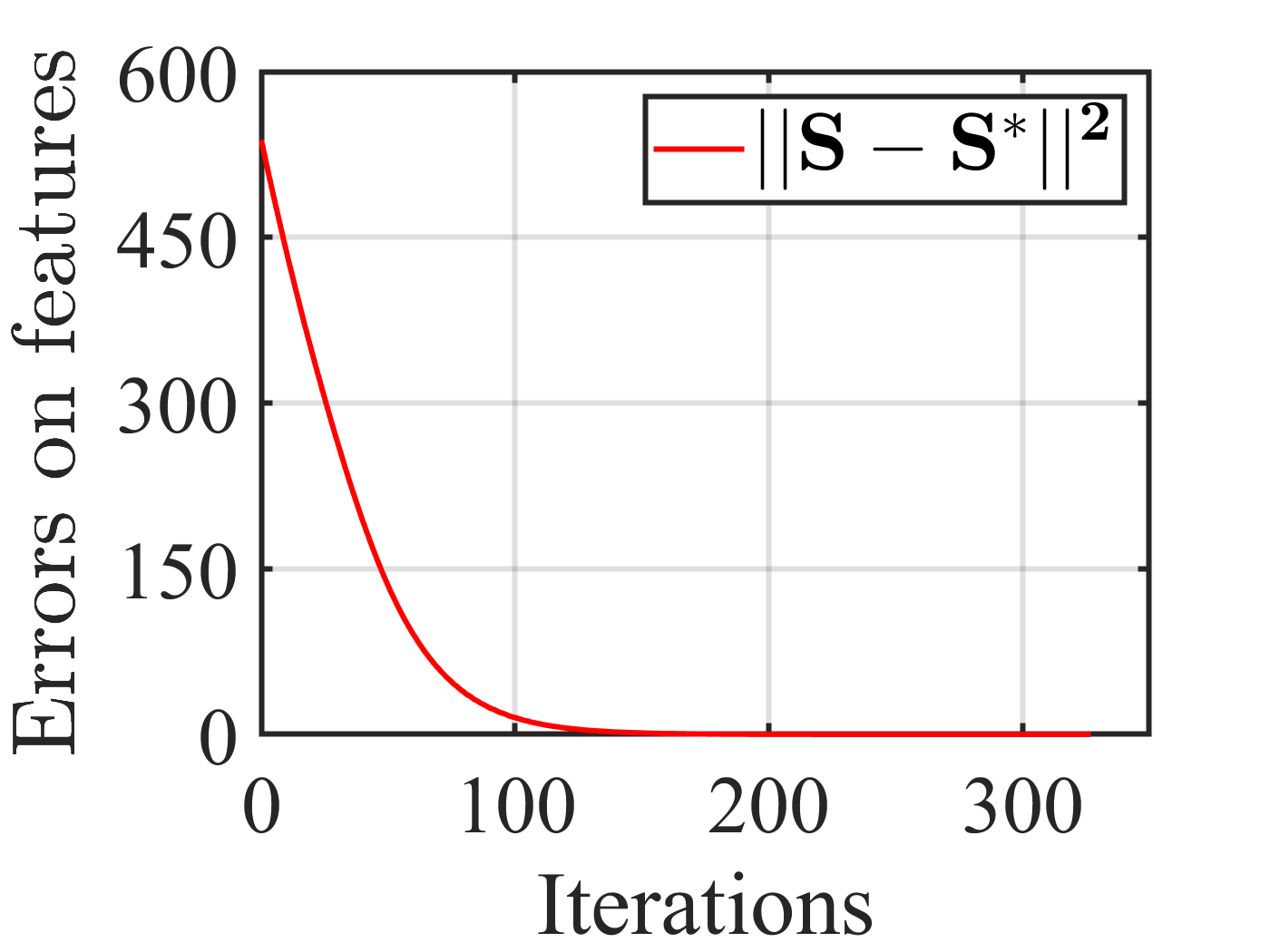}  \label{fig: Out_Field_HMVS_feature_error}}	\qquad
		\subfloat[]{\includegraphics[width=0.33\hsize]{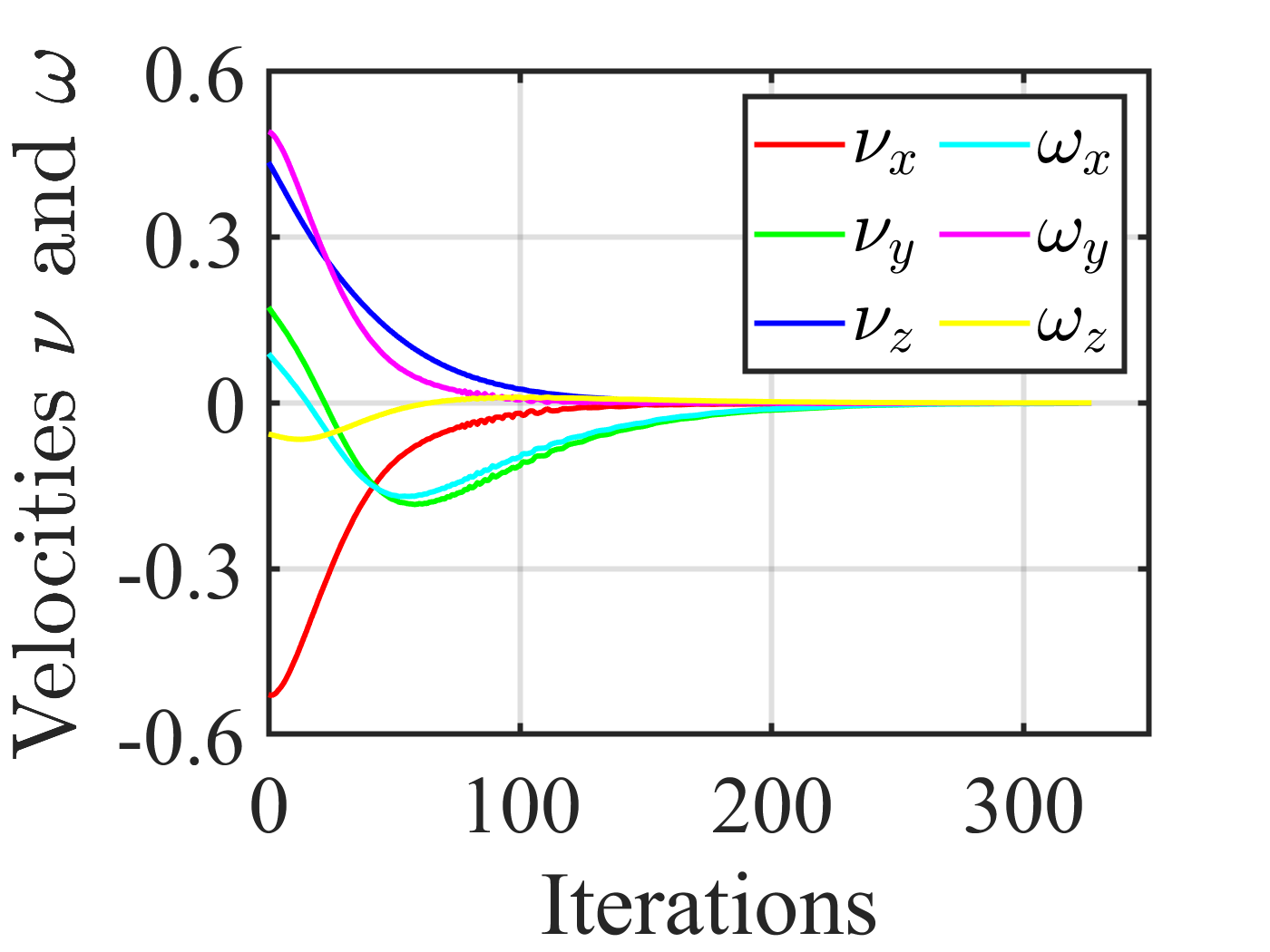}  \label{fig: Out_Field_HMVS_velocity}}		
		
		\subfloat[]{\includegraphics[width=0.33\hsize]{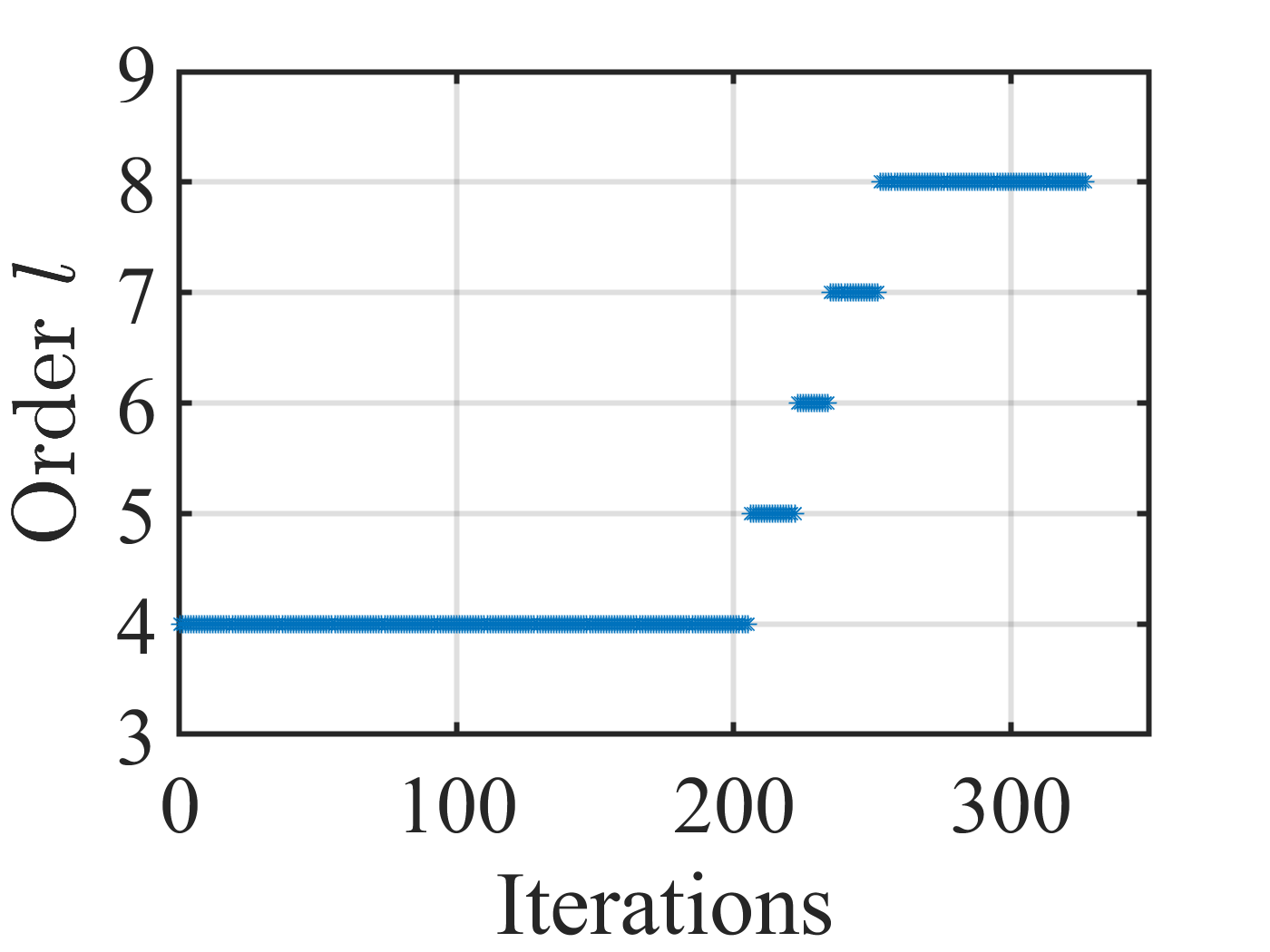}  \label{fig: Out_Field_HMVS_order}}		\qquad
		\subfloat[]{\includegraphics[width=0.33\hsize]{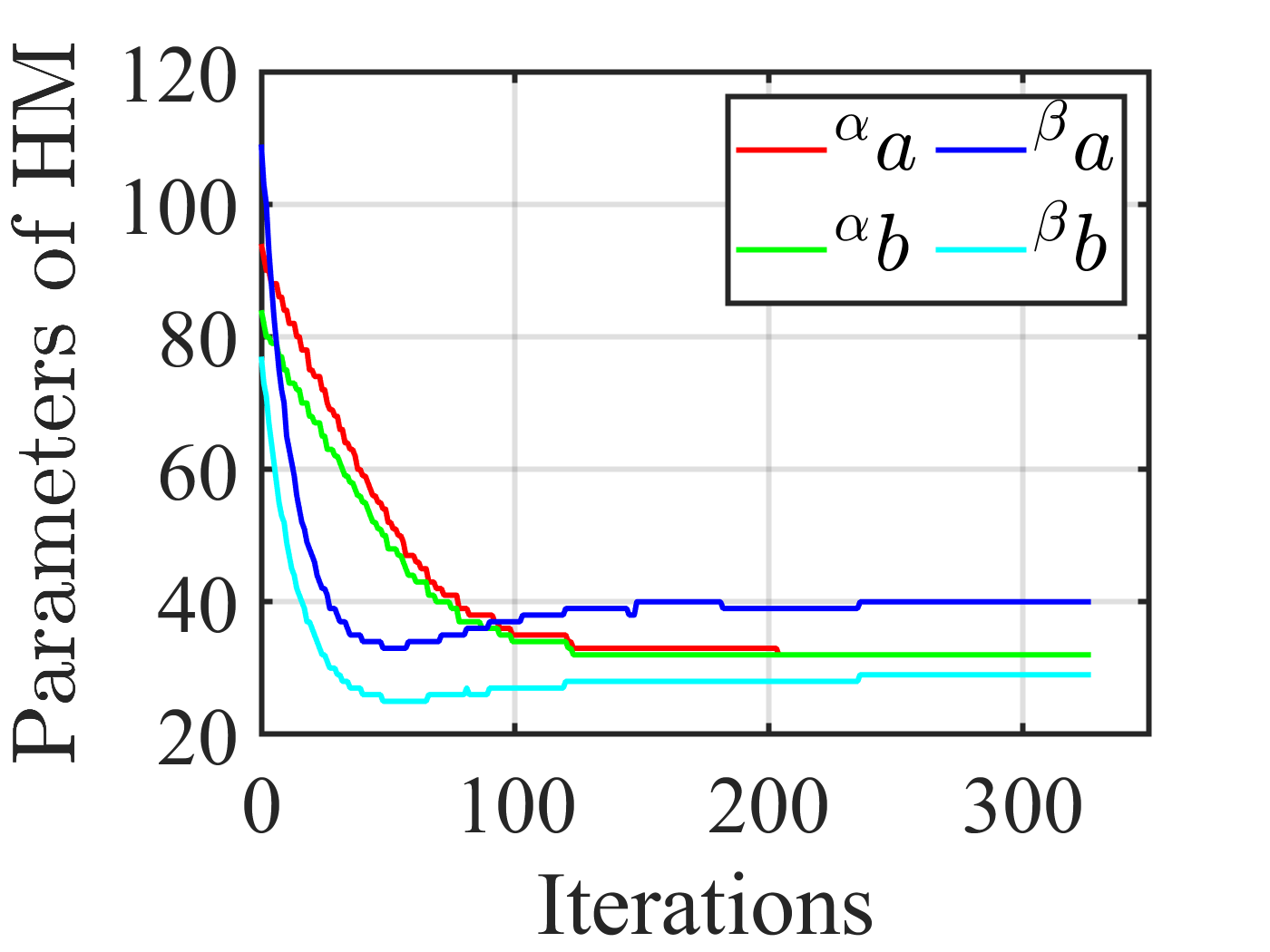}  \label{fig: Out_Field_HMVS_ab}}				
		
		\caption{Results for HM-VS in Experiment \#2. (a) Errors on features. (b) Camera velocities (in m/s and rad/s). (c) Order of  DOMs as visual features. (d) Parameters of HMs.}
		
		\label{fig: Out_Field_HMVS}
	\end{figure}

	The scenario with a complex textured plane and control of 6-DoF was used for this experiment.
        We suppose that the scene's depth is unknown. 
        The same depth value ($Z = 1m$) as an approximation for every pixel is used in the VS.
        The initial and desired images are illustrated in Figs. \ref{fig: Out_Field_image_new} and \ref{fig: Out_Field_image_old}, respectively.
        The large displacement between the initial and desired pose is given by $(0.54\text{m}$, $-0.45\text{m}$, $-0.27\text{m}$, $20.04^{\circ}$, $-21.54^{\circ}$, $-1.42^{\circ})$.
        The pixel errors and camera trajectories for the TM-VS, KM-VS, and HM-VS methods are shown in Figs. \ref{fig: Out_Field_I} and \ref{fig: Out_Field_trajections}, respectively.
        Both TM-VS and HM-VS have been successful in this task.
        For TM-VS ($446$ iterations), an accuracy of $(0.50mm, 0.60mm, 0.22mm)$ in translation and $(0.030^{\circ}, 0.023^{\circ}, 0.008^{\circ})$ in rotation is obtained.
        For HM-VS ($328$ iterations), an accuracy of $(0.40mm,0.06mm,0.12mm)$ in translation and $(0.005^{\circ}, 0.025^{\circ}, 0.002^{\circ})$ in rotation is obtained.
        For KM-VS, it ultimately fails because the textured plane is all outside the camera field-of-view.
        The details of TM-VS and HM-VS are illustrated in Figs. \ref{fig: Out_Field_TMVS} and \ref{fig: Out_Field_HMVS}, respectively.
        The feature error of the HM-VS is better than that of the TM-VS (see Figs. \ref{fig: Out_Field_TMVS_feature_error} and \ref{fig: Out_Field_HMVS_feature_error}).
        It is for the above reason that the HM-VS has fewer iterations.
        Since the pixel error is a non-exponential decrease (see Fig. \ref{fig: Out_Field_I}), the low-order components of the "S"-shaped curve account for the central part (see Figs. \ref{fig: Out_Field_TMVS_order} and \ref{fig: Out_Field_HMVS_order}).

\begin{figure}
		\centering 
		
		\subfloat[]{\includegraphics[width=0.45\hsize]{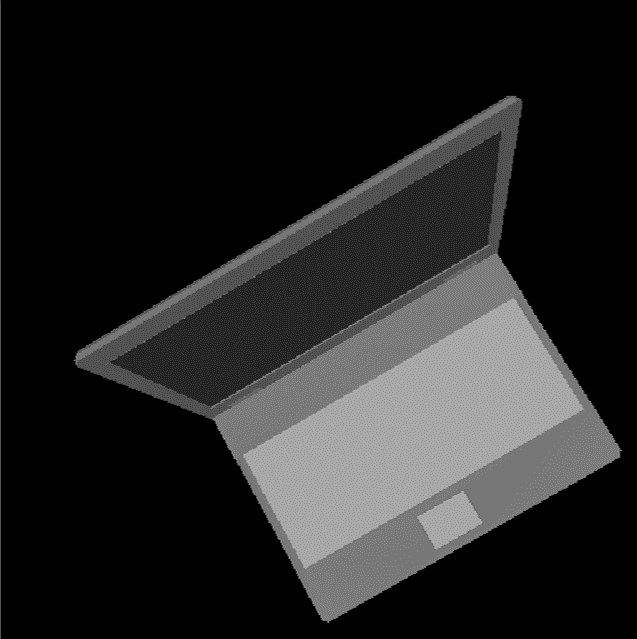}  \label{fig: ROI_3D_image_new}}	
		\subfloat[]{\includegraphics[width=0.45\hsize]{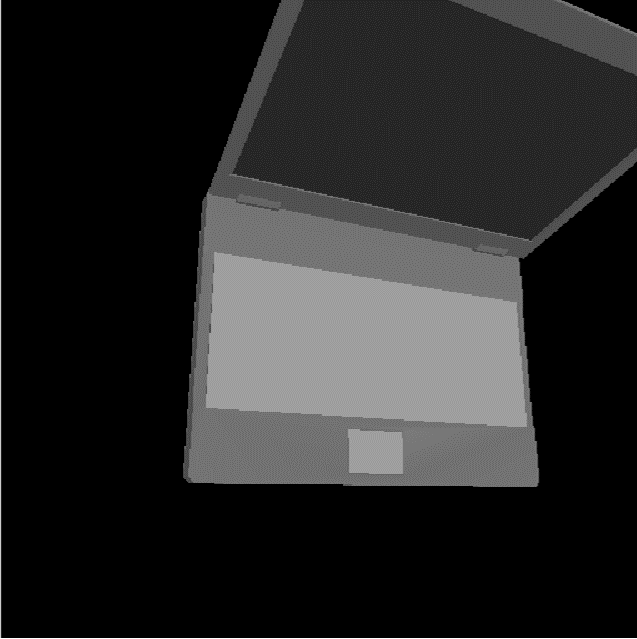}  \label{fig: ROI_3D_image_old}}		
		
		\subfloat[]{\includegraphics[width=0.49\hsize]{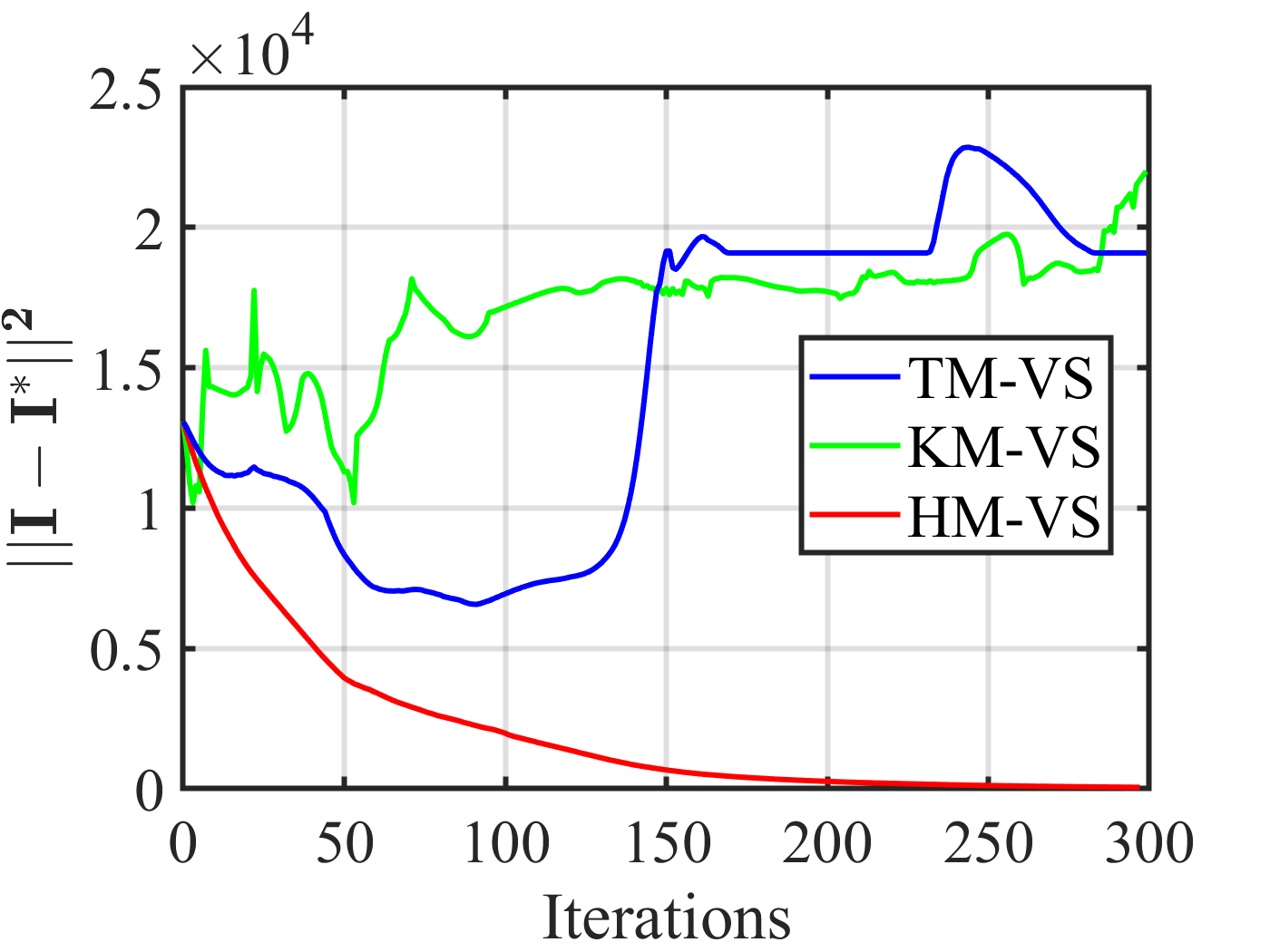}  \label{fig: ROI_3D_I}}	
		\subfloat[]{\includegraphics[width=0.49\hsize]{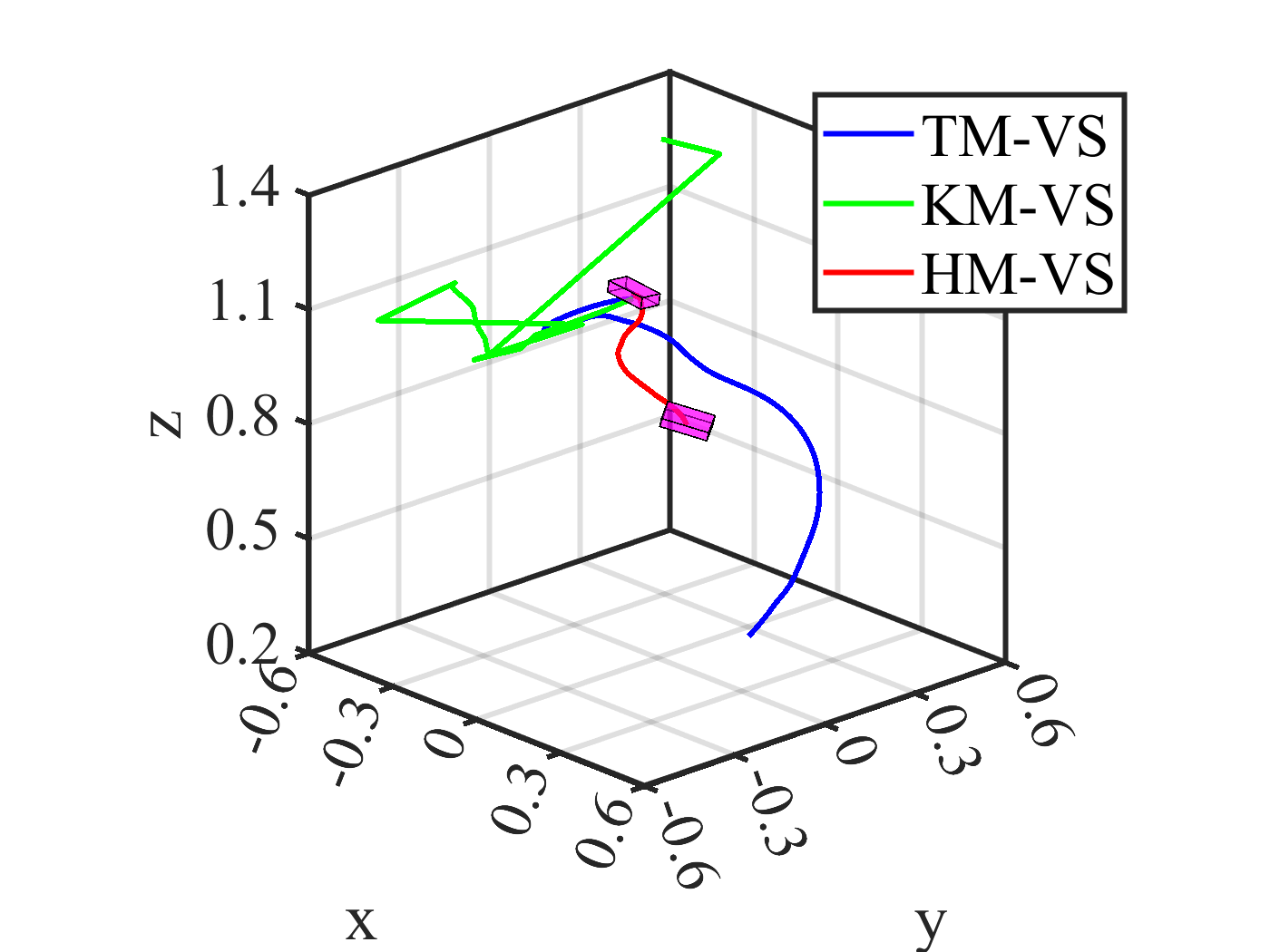}  \label{fig: ROI_3D_trajections}}		
		
		\caption{Experiment \#3: Comparison between TM-VS, KM-VS, and HM-VS in a 3-D virtual environment. (a) Initial image. (b) Desired image. (c) Pixel errors. (d) Camera trajectories  (in m).}
		
		\label{fig: ROI_3D}
	\end{figure}

	\subsubsection*{Experiment \#3 (see Fig. \ref{fig: ROI_3D})}
	
\begin{figure}
		\centering 
		
		\subfloat[]{\includegraphics[width=0.33\hsize]{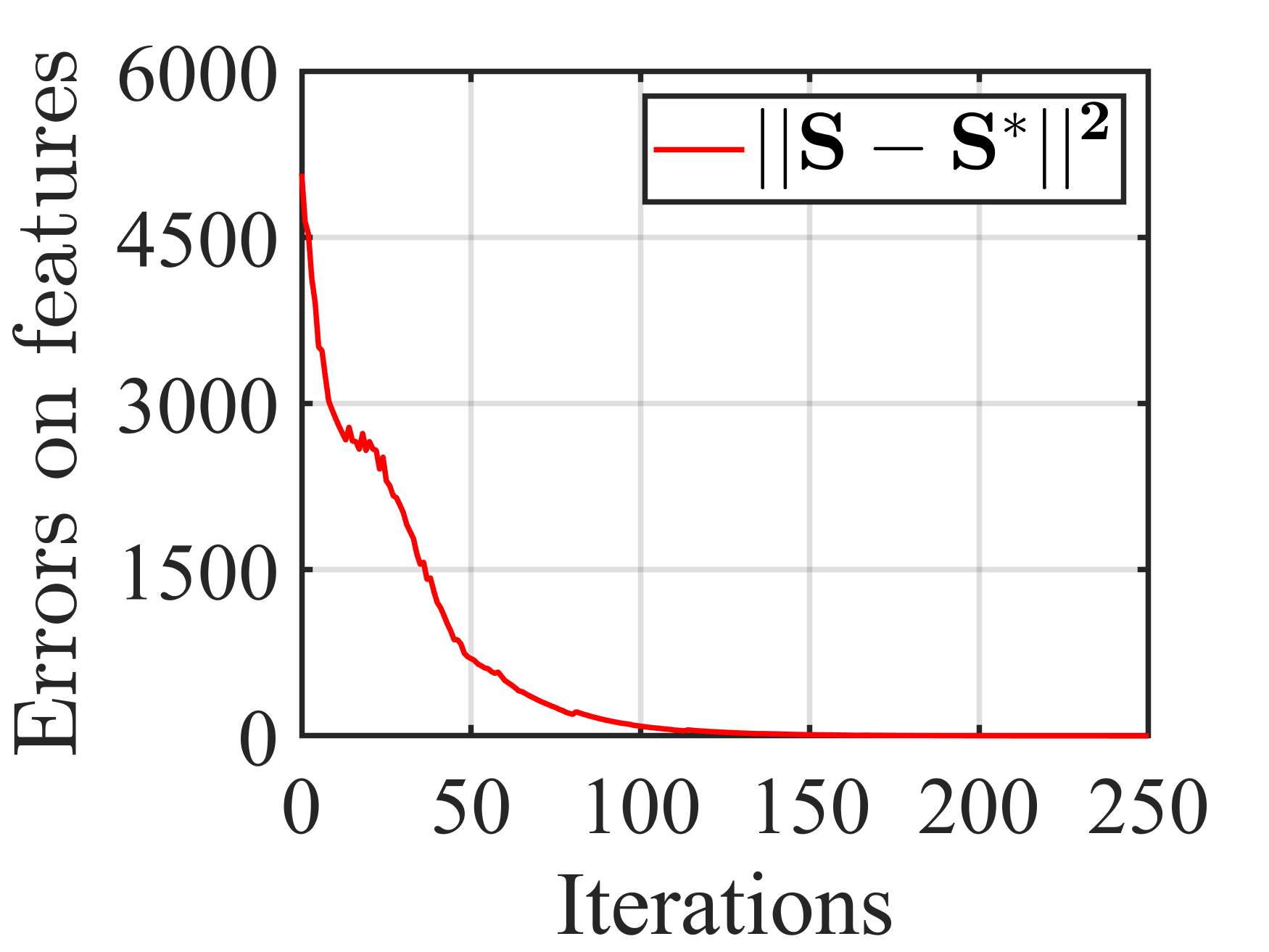}  \label{fig: ROI_3D_HMVS_feature_error}}	\qquad
		\subfloat[]{\includegraphics[width=0.33\hsize]{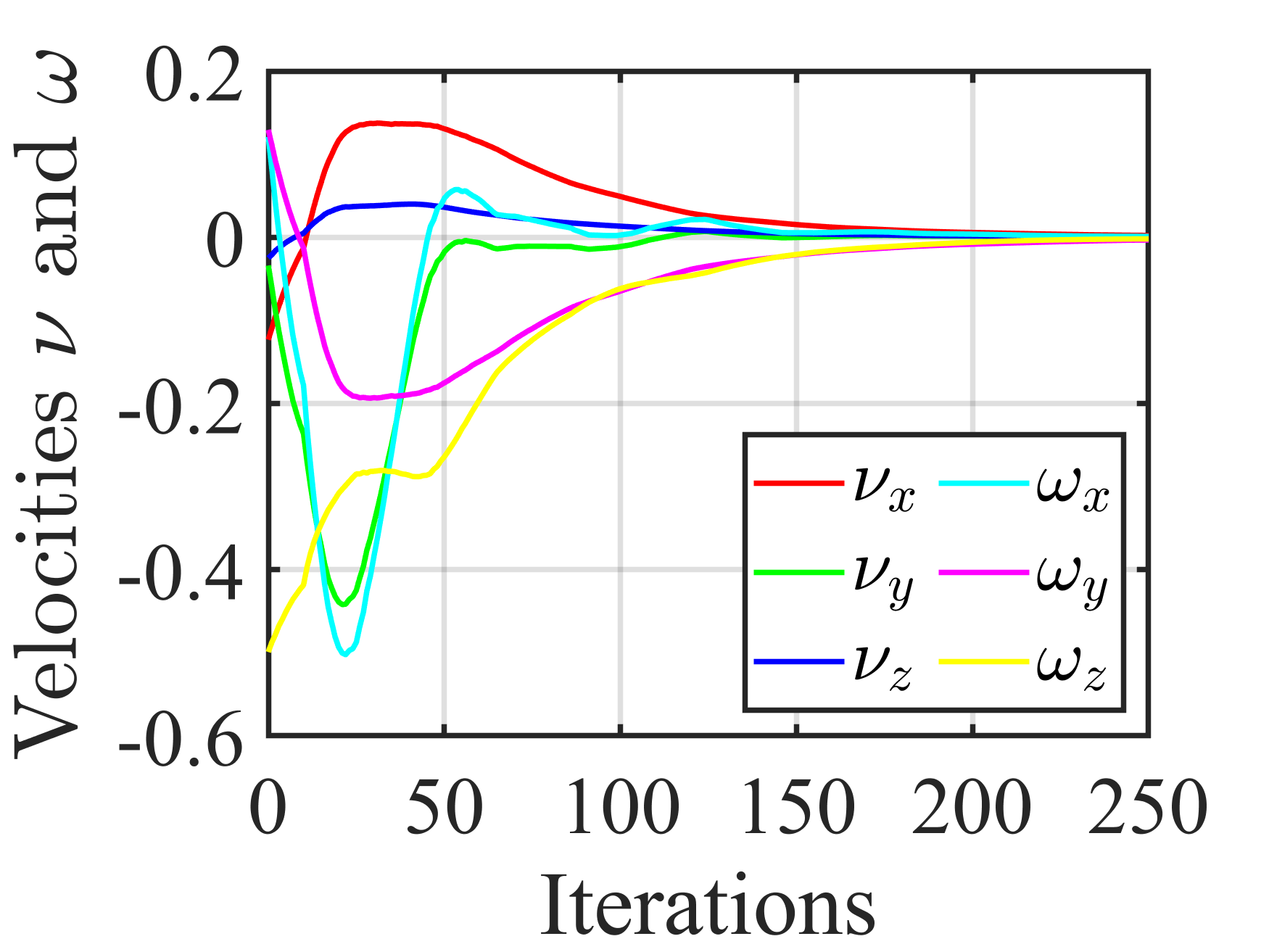}  \label{fig: ROI_3D_HMVS_velocity}}		
		
		\subfloat[]{\includegraphics[width=0.33\hsize]{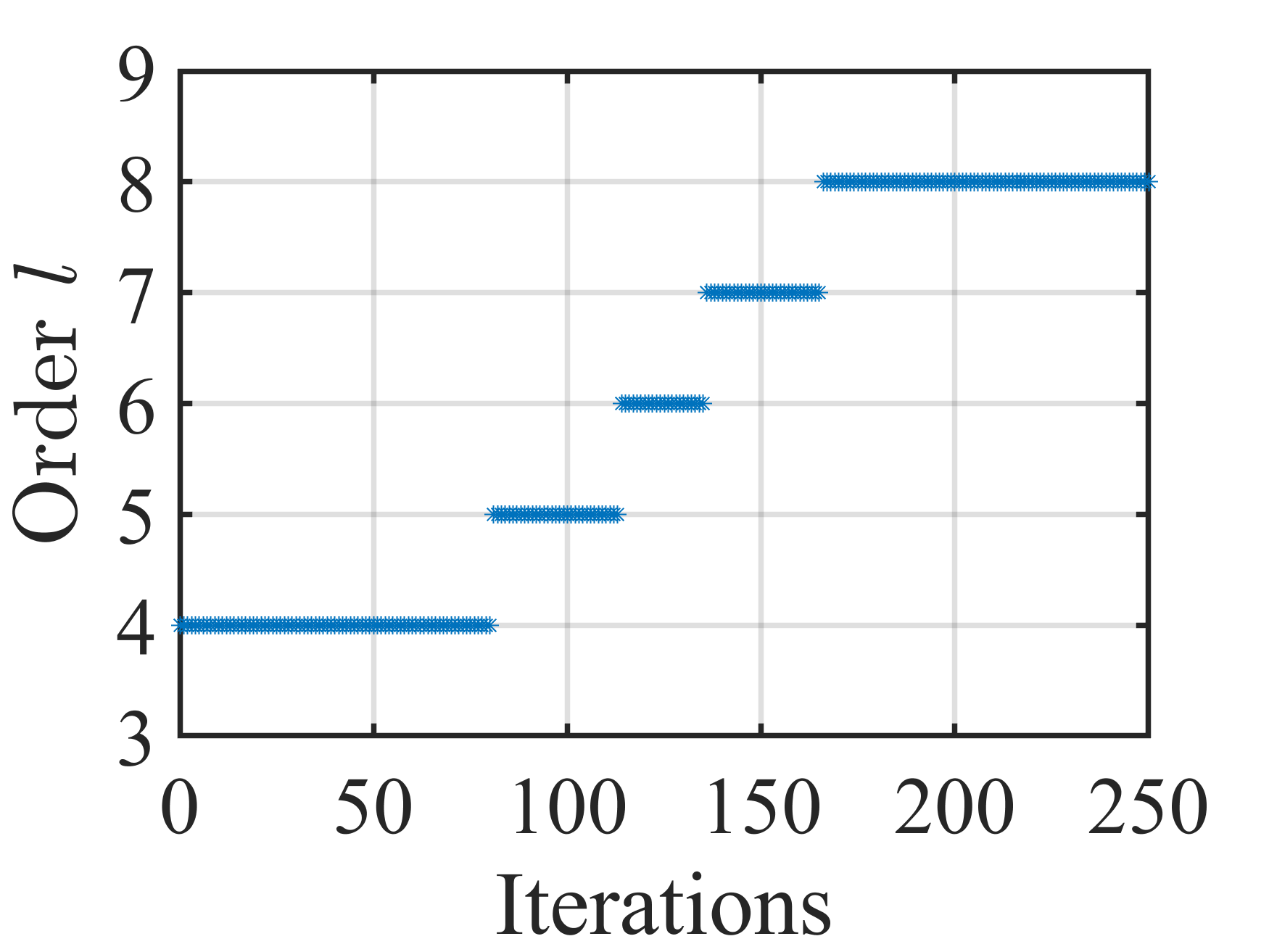}  \label{fig: ROI_3D_HMVS_order}}		\qquad
		\subfloat[]{\includegraphics[width=0.33\hsize]{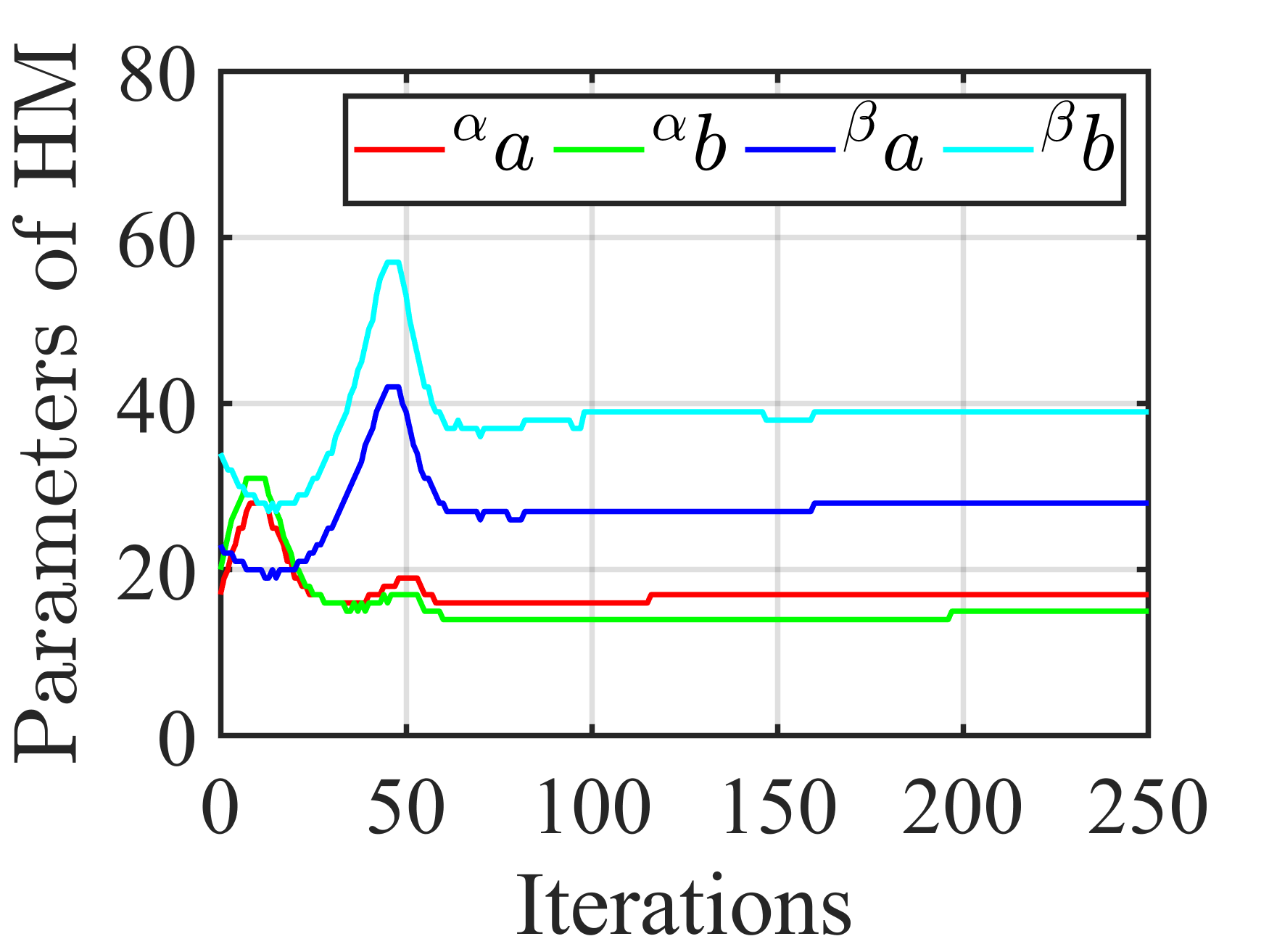}  \label{fig: ROI_3D_HMVS_ab}}				
		
		\caption{Results for HM-VS in Experiment \#3. (a) Errors on features. (b) Camera velocities (in m/s and rad/s). (c) Order of  DOMs as visual features. (d) Parameters of HMs.}
		
		\label{fig: ROI_3D_HMVS}
	\end{figure}

	In this experiment, we compare the proposed VS schemes in a 3-D virtual environment.
        We select the same depth value for each point in the current and desired images ($Z = Z^*$).
	Taking the "laptop" as an example, the initial and desired images are shown in Figs. \ref{fig: ROI_3D_image_new} and \ref{fig: ROI_3D_image_old}, respectively.
	The "laptop"  is partially outside the camera field-of-view in the desired image.
	Figs. \ref{fig: ROI_3D_I} and \ref{fig: ROI_3D_trajections} illustrate the pixel errors  and camera trajectories obtained from the TM-VS, KM-VS, and HM-VS methods.
	The displacement between the desired and the initial camera poses is ($-0.37\text{m}$, $-0.23\text{m}$, $-0.02\text{m}$, $15.41^{\circ}$, $22.63^{\circ}$, $35.13^{\circ}$).
	The orientations around the two axes orthogonal to the optical axis of the camera are of interest.
	Only the HM-VS  approach converges perfectly to  the desired pose with a final pose error equal to ($0.5\text{mm}$, $0.5\text{mm}$, $0.5\text{mm}$, $0.02^\circ$, $0.02^\circ$, $0.02^\circ$).
    Both TM-VS and KM-VS  fail because the "laptop" is all outside the camera field-of-view.
	The details of HM-VS are shown in Fig. \ref{fig: ROI_3D_HMVS}.
	The exponentially decreasing feature errors (see Fig. \ref{fig: ROI_3D_HMVS_feature_error}), "S"-shaped curve (see Fig. \ref{fig: ROI_3D_HMVS_order}), and real-time adjustment of HM parameters (see Fig. \ref{fig: ROI_3D_HMVS_ab}) validate the effectiveness of HM-VS for 3D environments.
	
	In conclusion, the HM-VS scheme is significantly better
	than the TM-VS and KM-VS schemes due to its
	flexible parameter tuning mechanism.
	
	\subsection{Evaluation of the Robustness with respect to Noise} \label{sec: Robustness}
\begin{figure*}
		\centering 
		
		\subfloat[]{\includegraphics[width=0.19\hsize]{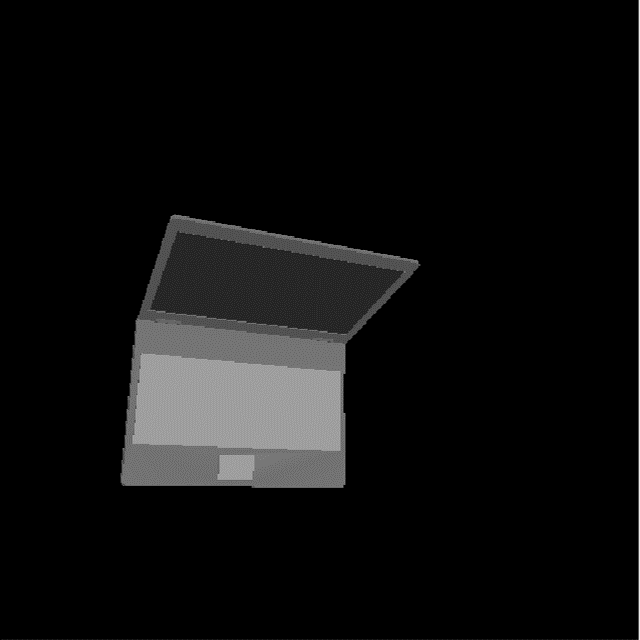}  \label{fig: image_noise_new_0}}	
		\subfloat[]{\includegraphics[width=0.19\hsize]{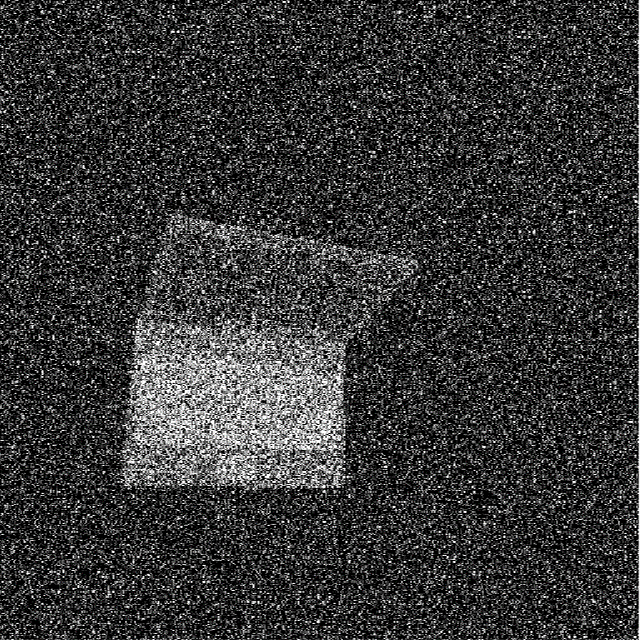}  \label{fig: image_noise_new_2}}		
		\subfloat[]{\includegraphics[width=0.19\hsize]{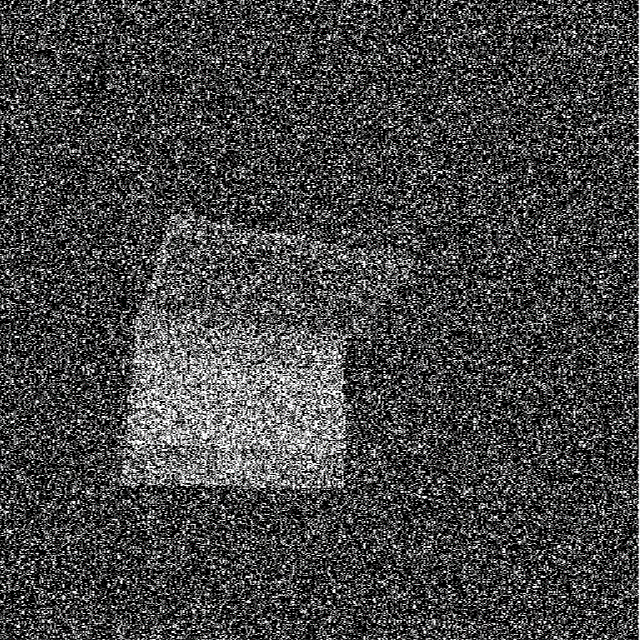}  \label{fig: image_noise_new_4}}		
		\subfloat[]{\includegraphics[width=0.19\hsize]{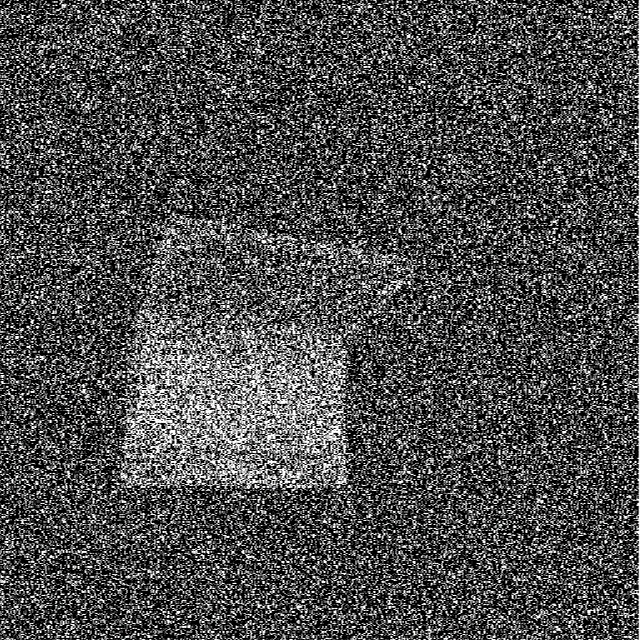}  \label{fig: image_noise_new_6}}		
		\subfloat[]{\includegraphics[width=0.19\hsize]{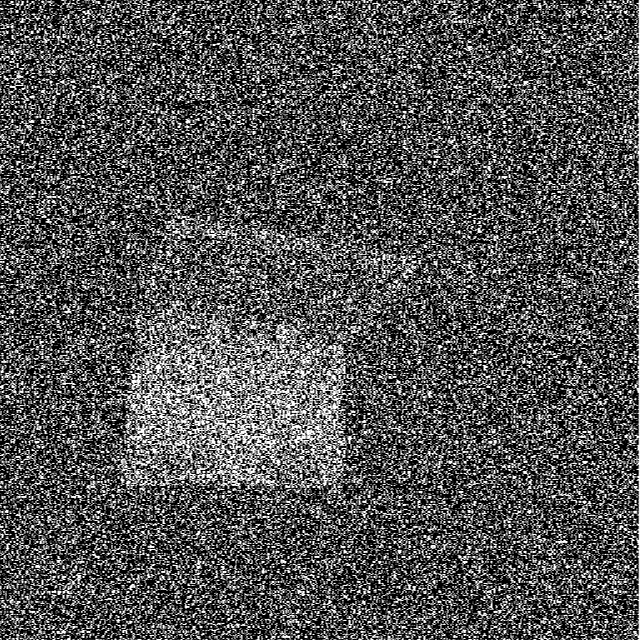}  \label{fig: image_noise_new_8}}				
		
		\subfloat[]{\includegraphics[width=0.19\hsize]{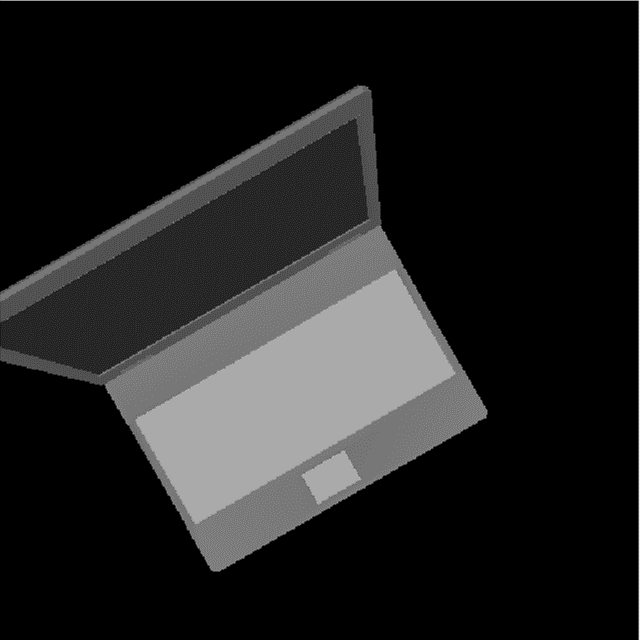}  \label{fig: image_noise_old_0}}	
		\subfloat[]{\includegraphics[width=0.19\hsize]{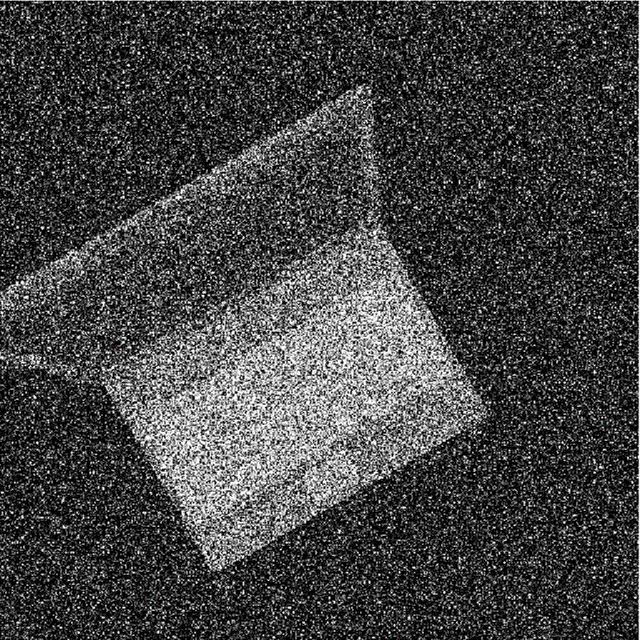}  \label{fig: image_noise_old_2}}		
		\subfloat[]{\includegraphics[width=0.19\hsize]{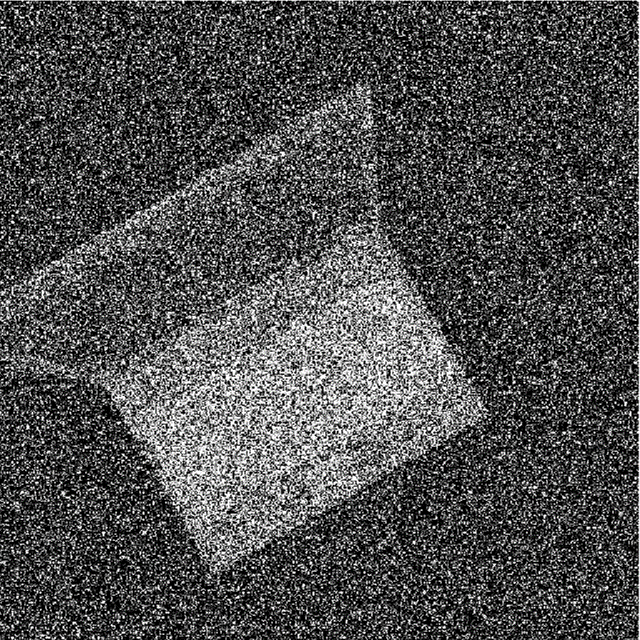}  \label{fig: image_noise_old_4}}		
		\subfloat[]{\includegraphics[width=0.19\hsize]{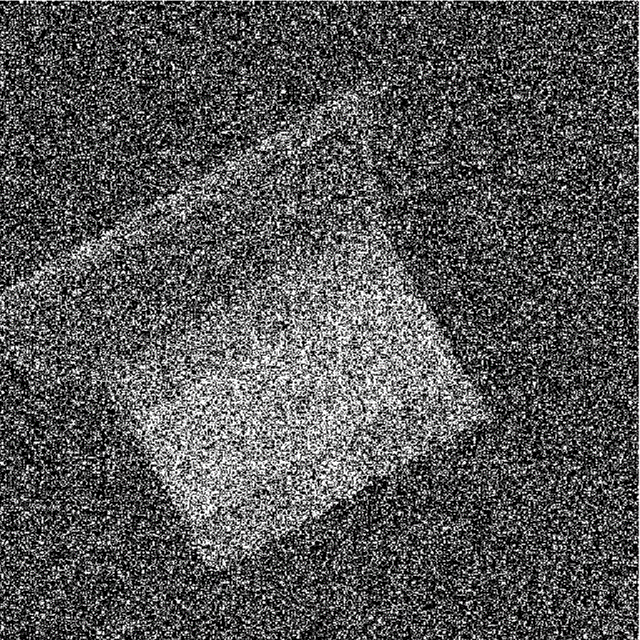}  \label{fig: image_noise_old_6}}		
		\subfloat[]{\includegraphics[width=0.19\hsize]{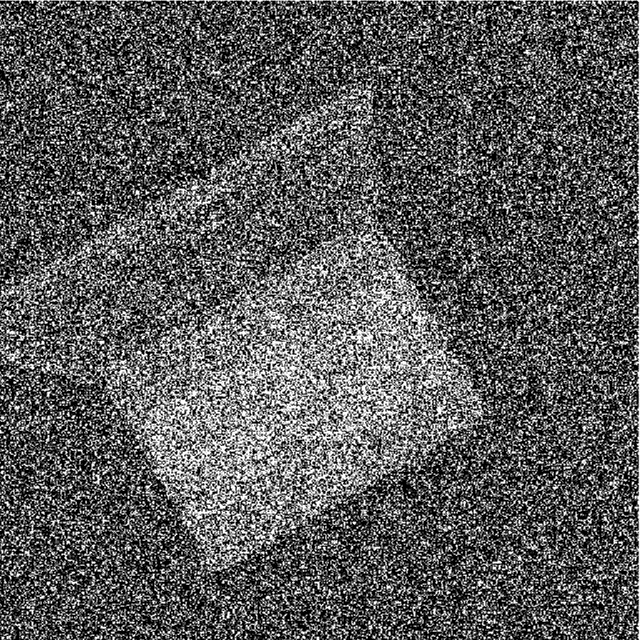}  \label{fig: image_noise_old_8}}		
		
		\caption{Experiment \#4: Gaussian noise robustness evaluation. (a)-(e) Initial images with variance $\sigma^2 = 0$, $0.2$, $0.4$, $0.6$ and $0.8$, respectively. (f)-(j) Desired images with variance $\sigma^2 = 0$, $0.2$, $0.4$, $0.6$ and $0.8$, respectively. }
		
		\label{fig: image_noise}
	\end{figure*}

	\subsubsection*{Experiment \#4 (see Fig. \ref{fig: image_noise})}
	
	The robustness of the proposed TM-VS, KM-VS, and HM-VS schemes was compared in this subsection.
  All VS utilizes the true depth value while we let $l=8$.
	Fig. \ref{fig: image_noise} exhibits an experiment evaluating the noise robustness.
	The displacement between the desired and the initial camera poses remains the same for each experiment: ($0.23\text{m}$, $-0.39\text{m}$, $-0.29\text{m}$, $-21.59^\circ$, $-19.39^\circ$, $-33.19^\circ$).
	First, the VS is performed without any image noise $\sigma^2=0$ (see Figs. \ref{fig: image_noise_new_0} and \ref{fig: image_noise_old_0}).
	Next, a stepwise increasing Gaussian noise is added to the initial image and the desired image (see Figs. \ref{fig: image_noise_new_2}-\ref{fig: image_noise_new_8} and \ref{fig: image_noise_old_2}-\ref{fig: image_noise_old_8}).
	Specifically, the noise intensity is enhanced between each experiment with a variance $\sigma^2 = 0.2$, $0.4$, $0.6$, and $0.8$.
	The average convergence error (position and orientation) are shown in Table \ref{tab: noise_result}.
	
	From Table \ref{tab: noise_result}, it can be easily seen that both TM-VS and HM-VS are remarkably robust against image noise.
	KM-VS is only available for $\sigma^2=0$ in this experiment.
	The accuracy at convergence decreases as the noise intensity increases, but it is still excellent for the added excessive noise, thanks to the filtering properties of the DOM.
	In addition, the HM-VS scheme has advantages over other methods.
	
	\begin{table}[]
		\begin{center}
			\caption{Results of the average convergence error (position and orientation) obtained by TM-VS, KM-VS, and HM-VS methods.}
			\label{tab: noise_result}
			\begin{tabular}{c|c|c|c|}
				\cline{2-4}
				& TM-VS &  KM-VS & HM-VS \\ \hline
				\multicolumn{1}{|c|}{$\sigma^2=0$ } &$0.05\text{mm}, 0.01^\circ$&$9.28 \text{mm}, 0.37^\circ$ &$\mathbf{0.03\text{mm}},  \mathbf{0.00^\circ}$\\
				\hline
				\multicolumn{1}{|c|}{$\sigma^2=0.2$}  &$2.20 \text{mm}, 0.24^\circ$ &N/A &$\mathbf{0.99mm, 0.20^\circ}$\\			 
				\hline
				\multicolumn{1}{|c|}{$\sigma^2=0.4$}  &$7.80\text{mm}, 0.67^\circ$ &N/A & $\mathbf{3.68mm, 0.22^\circ}$\\			 
				\hline
				\multicolumn{1}{|c|}{$\sigma^2=0.6$}  &$13.30 \text{mm}, 1.10^\circ$ &N/A &$\mathbf{4.09mm, 0.58^\circ}$\\			 
				\hline
				\multicolumn{1}{|c|}{$\sigma^2=0.8$}  & $32.28 \text{mm}, 2.29^\circ$&N/A &$\mathbf{12.10mm, 0.97^\circ}$\\			 
				\hline	
			\end{tabular}
		\end{center}
	\end{table}
	
	\subsection{Comparisons with  a Baseline Method and Two  State-of-the-Art Methods} \label{sec: state-of-art}
	In this section, simulations are performed to
	evaluate the proposed HM-VS scheme while allowing a fair
	comparison of a baseline method and two state-of-the-art methods: DVS \cite{collewet2008visual}, DCT-VS \cite{marchand2020direct}, and PGM-VS \cite{crombez2018visual}.
	HM-VS and DCT-VS methods use the method proposed in Section \ref{sec: order} to determine the order. The minimum and maximum orders are still $l_{\text{min}}=4, l_{\text{max}}=8$.
	The required parameter $\lambda_{gi}$  in the PGM-VS is determined to be $\lambda_{gi}=10$, $16$, and $60$. 
	To ensure the same conditions, all methods are based on the Gauss-Newton algorithm while every pixel's true depth value is leveraged.
	The image size is $128 \times 128$ in the following simulations.
     It's worth noting that the methods (DVS, DCT-VS, PGM-VS) are from my reproduction since the authors have not shared their source code.
	
\begin{figure}
		\centering 
		
		\subfloat[]{\includegraphics[width=0.45\hsize]{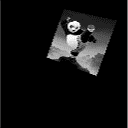}  \label{fig: Comparisons_image_new_panda}}	
		\subfloat[]{\includegraphics[width=0.45\hsize]{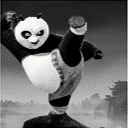}  \label{fig: Comparisons_image_old_panda}}				

		\subfloat[]{\includegraphics[width=0.48\hsize]{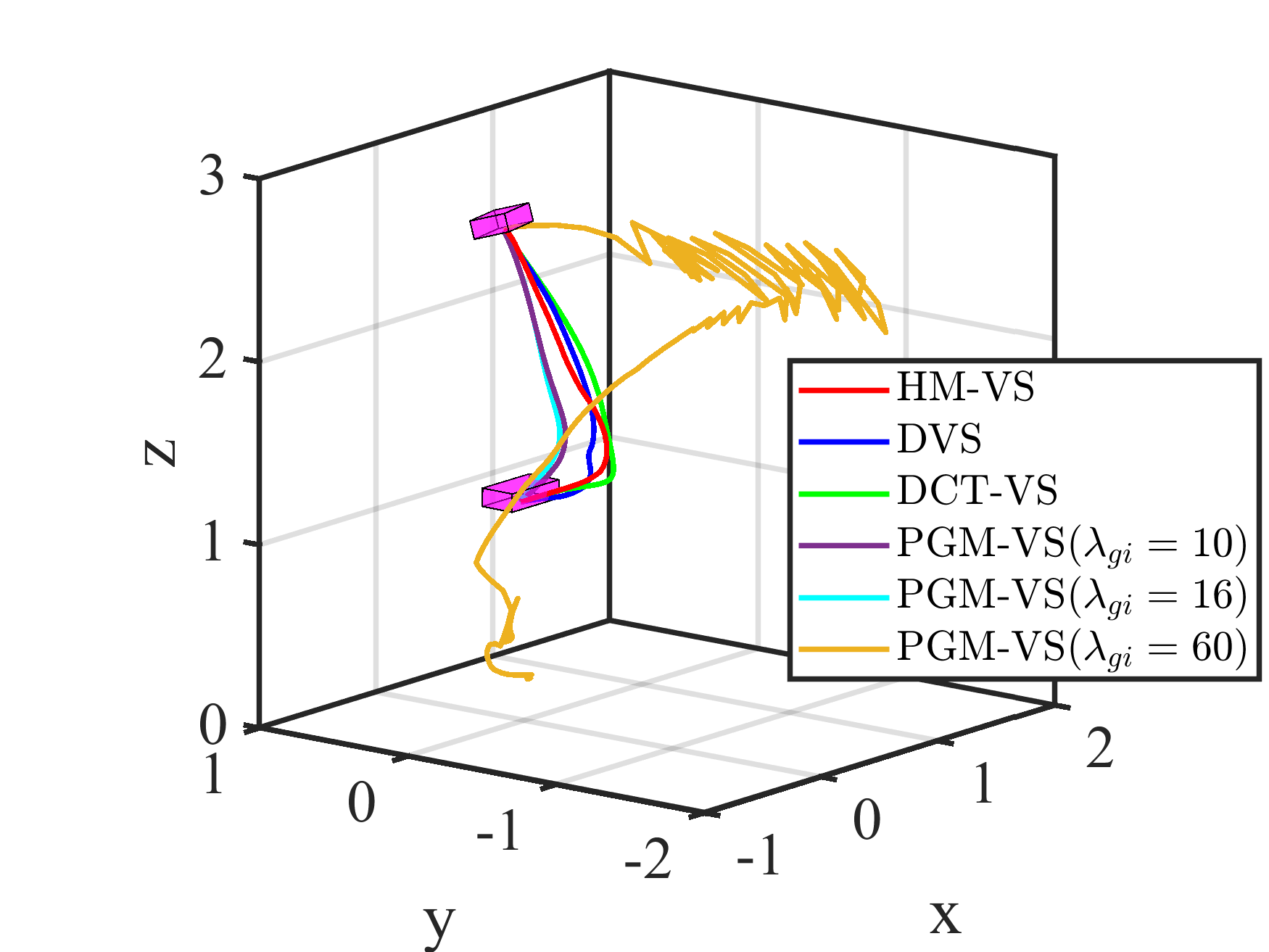}  \label{fig: Comparisons_trajections_panda}}	
		\subfloat[]{\includegraphics[width=0.48\hsize]{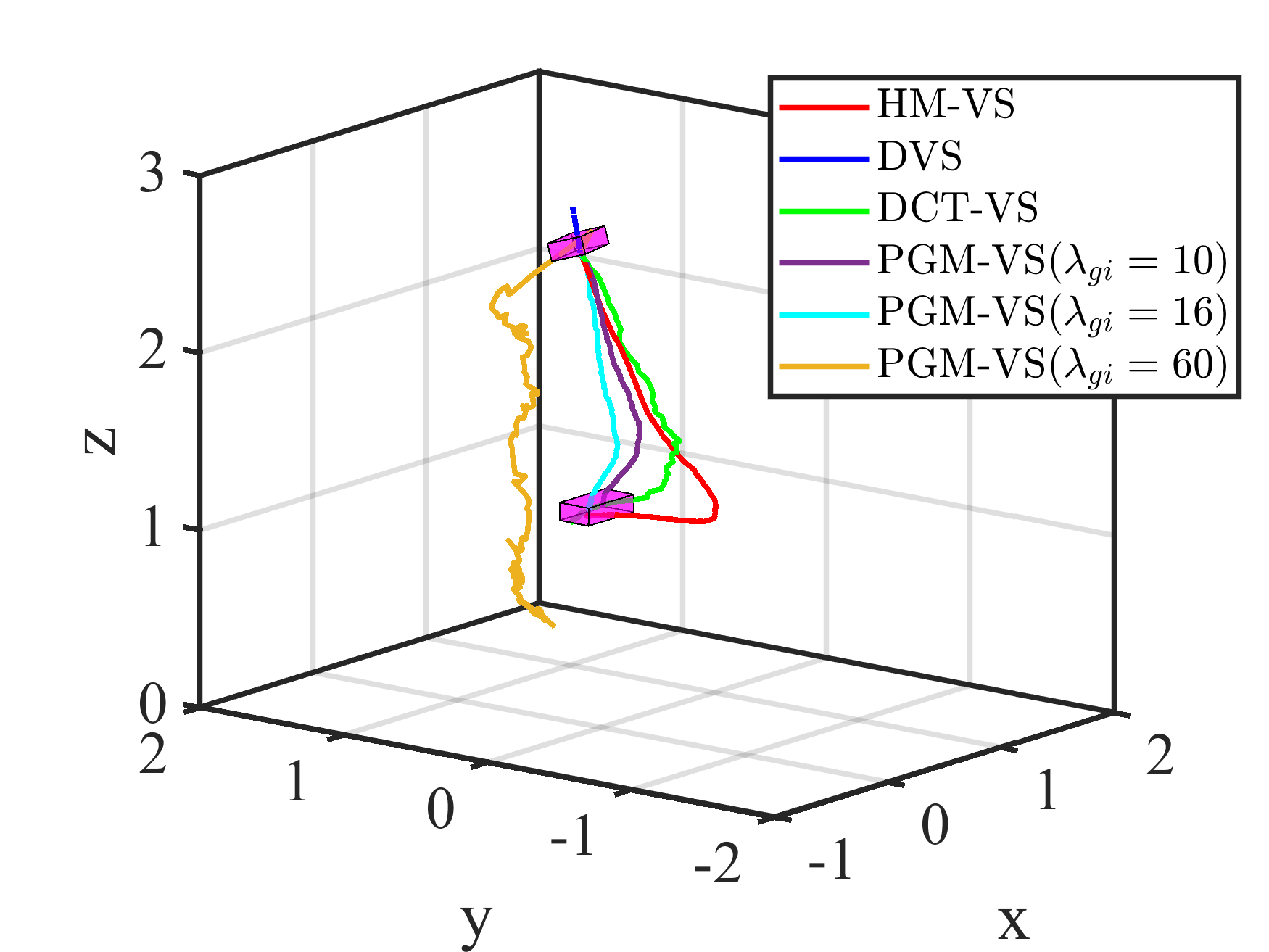}  \label{fig: Comparisons_trajections_noise_panda}}		
		\caption{Case \#1: Example with the classical scene and camera trajectories. (a) Initial image. (b) Desired image. (c) Camera trajectories with Gaussian noise $\sigma^2 = 0$   (in m). (d) Camera trajectories with Gaussian noise $\sigma^2 = 0.6$  (in m).} 
		
		\label{fig: Comparisons_result_panda}
	\end{figure}

	\subsubsection*{Case \#1 (see Fig. \ref{fig: Comparisons_result_panda})}
	
	In the case of a large displacement facing a 2-D scene, the noise-free initial and desired images are shown in Figs. \ref{fig: Comparisons_image_new_panda} and \ref{fig: Comparisons_image_old_panda}, respectively.
	Fig. \ref{fig: Comparisons_trajections_panda} illustrates  the obtained camera trajectories using these methods in the absence of Gaussian noise $\sigma^2=0$, where only PGM-VS ($\lambda_{gi}=60$) is a failure.
	For the convergence rate, HM-VS ($137$ iterations) is more satisfactory than  DVS ($1590$ iterations), DCT-VS ($148$ iterations), PGM-VS ($\lambda_{gi}=10$) ($192$ iterations), and PGM-VS ($\lambda_{gi}=16$) ($147$ iterations). 
	Fig. \ref{fig: Comparisons_trajections_noise_panda} shows  the obtained camera trajectories using these methods in Gaussian noise $\sigma^2=0.6$, where {HM-VS, DCT-VS, PGM-VS ($\lambda_{gi}=10$), and PGM-VS ($\lambda_{gi}=16$) can perform the task.
	However, the final error of HM-VS  ($43.4\text{mm}$, $2.6\text{mm}$, $5.5\text{mm}$, $0.2^\circ$, $2.7^\circ$, $1.3^\circ$) is better than that of DCT-VS
	($171.7\text{mm}$, $2.4\text{mm}$, $63.7\text{mm}$, $0.8^\circ$, $12.1^\circ$, $1.9^\circ$), PGM-VS ($\lambda_{gi}=10$) ($2.6\text{mm}$, $64.1\text{mm}$, $47.4\text{mm}$, $4.4^\circ$, $0.1^\circ$, $0.5^\circ$), and PGM-VS ($\lambda_{gi}=16$) ($40.4\text{mm}$, $1.9\text{mm}$, $27.0\text{mm}$, $0.5^\circ$, $3.0^\circ$, $0.3^\circ$).

	\subsubsection*{Case \#2 (see Fig. \ref{fig: Comparisons_result_nezha})}
	
	\begin{figure}
		\centering 
		
		\subfloat[]{\includegraphics[width=0.45\hsize]{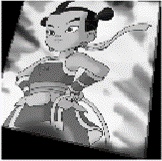}  \label{fig: Comparisons_result_new_nezha}}	
		\subfloat[]{\includegraphics[width=0.45\hsize]{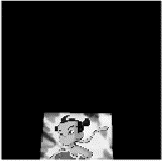}  \label{fig: Comparisons_image_old_nezha}}		
		
		\subfloat[]{\includegraphics[width=0.48\hsize]{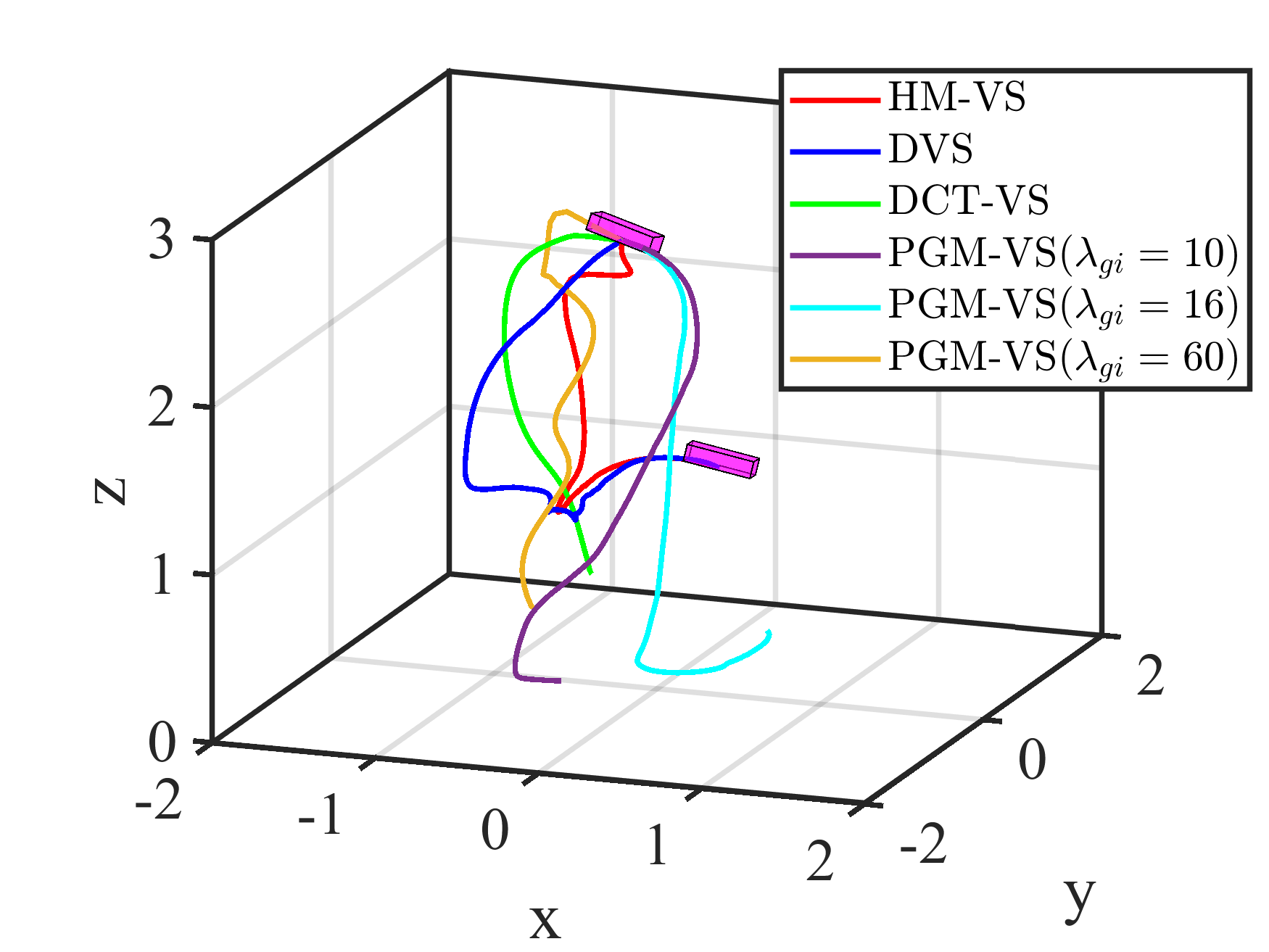}  \label{fig: Comparisons_trajections_nezha}}	
		\subfloat[]{\includegraphics[width=0.48\hsize]{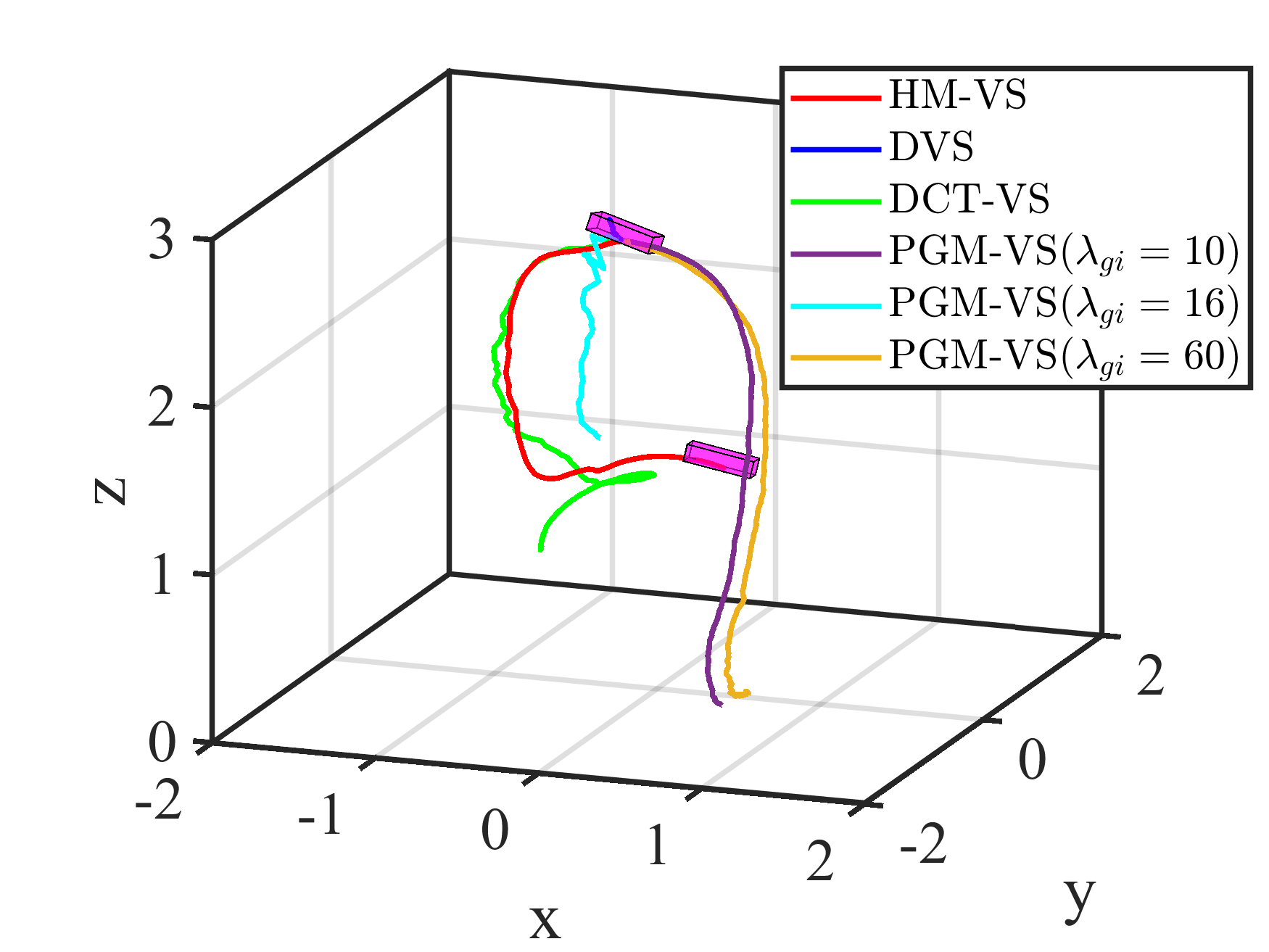}  \label{fig: Comparisons_trajections_noise_nezha}}					
		
		\caption{Case \#2: Example with a large difference between initial and desired images and camera trajectories. (a) Initial image. (b) Desired image. (c) Camera trajectories with Gaussian noise $\sigma^2 = 0$   (in m). (d) Camera trajectories with Gaussian noise $\sigma^2 = 0.2$  (in m). }
		
		\label{fig: Comparisons_result_nezha}
	\end{figure}
	
	A challenging case where a large part of the desired image is absent in the initial image is presented in Fig. \ref{fig: Comparisons_result_nezha}.
	Fig. \ref{fig: Comparisons_trajections_nezha} illustrates the camera trajectories obtained using these methods without Gaussian noise $\sigma^2=0$, of which only HM-VS and DVS methods can successfully drive the camera to the desired pose.
	When adding a Gaussian noise $\sigma^2=0.2$ on both the desired and the current images, only the HM-VS approach succeeds. All other methods fail because they fall into local minima or are out of  field-of-view.

	\subsubsection*{Case \#3 (see Fig. \ref{fig: Comparisons_result_building})}
	
	\begin{figure}
		\centering 
		
		\subfloat[]{\includegraphics[width=0.45\hsize]{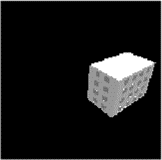}  \label{fig: Comparisons_result_new_building}}	
		\subfloat[]{\includegraphics[width=0.45\hsize]{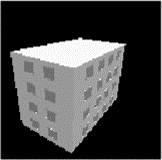}  \label{fig: Comparisons_image_old_building}}		
		
		\subfloat[]{\includegraphics[width=0.48\hsize]{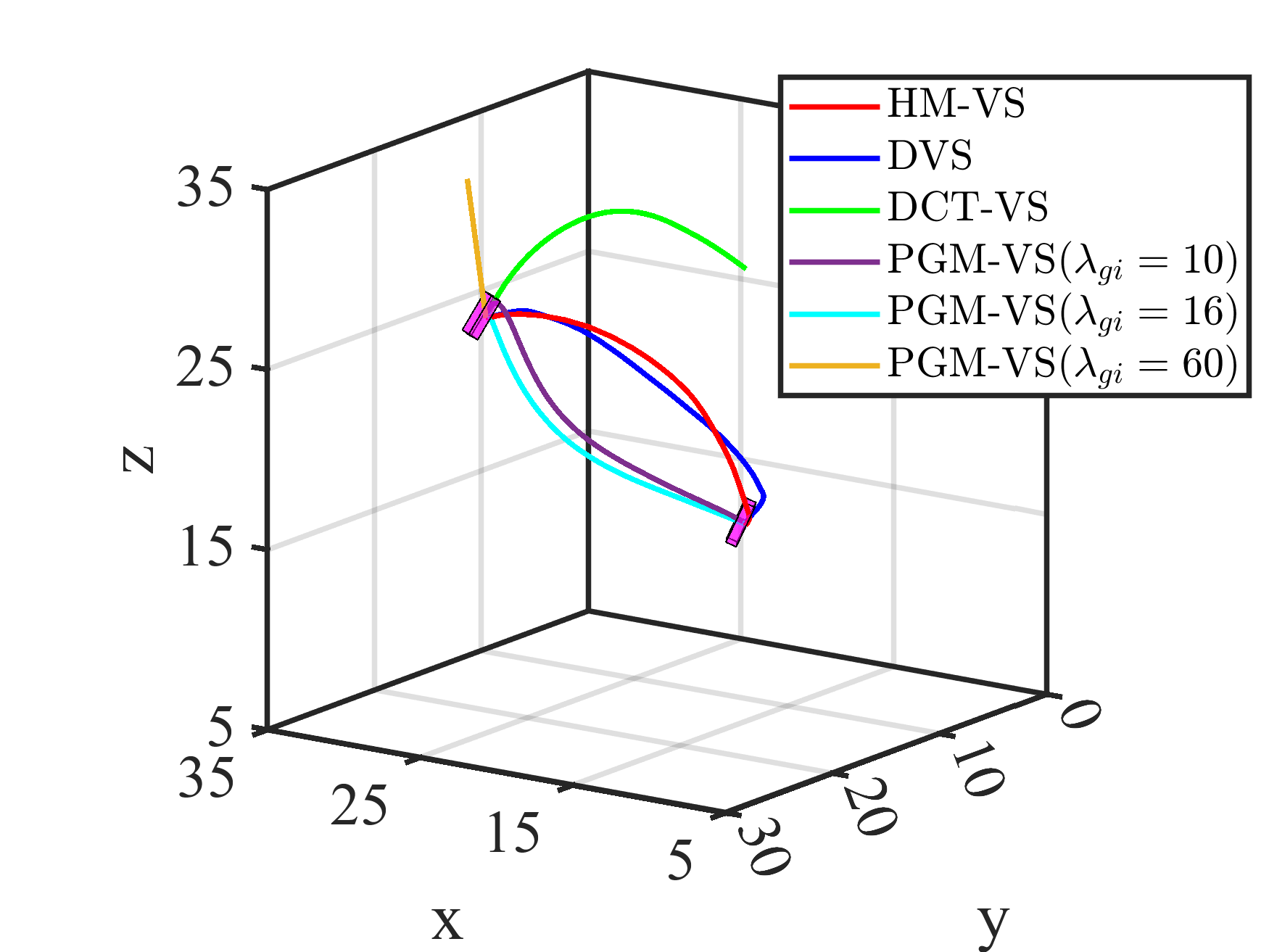}  \label{fig: Comparisons_trajections_building}}	
		\subfloat[]{\includegraphics[width=0.48\hsize]{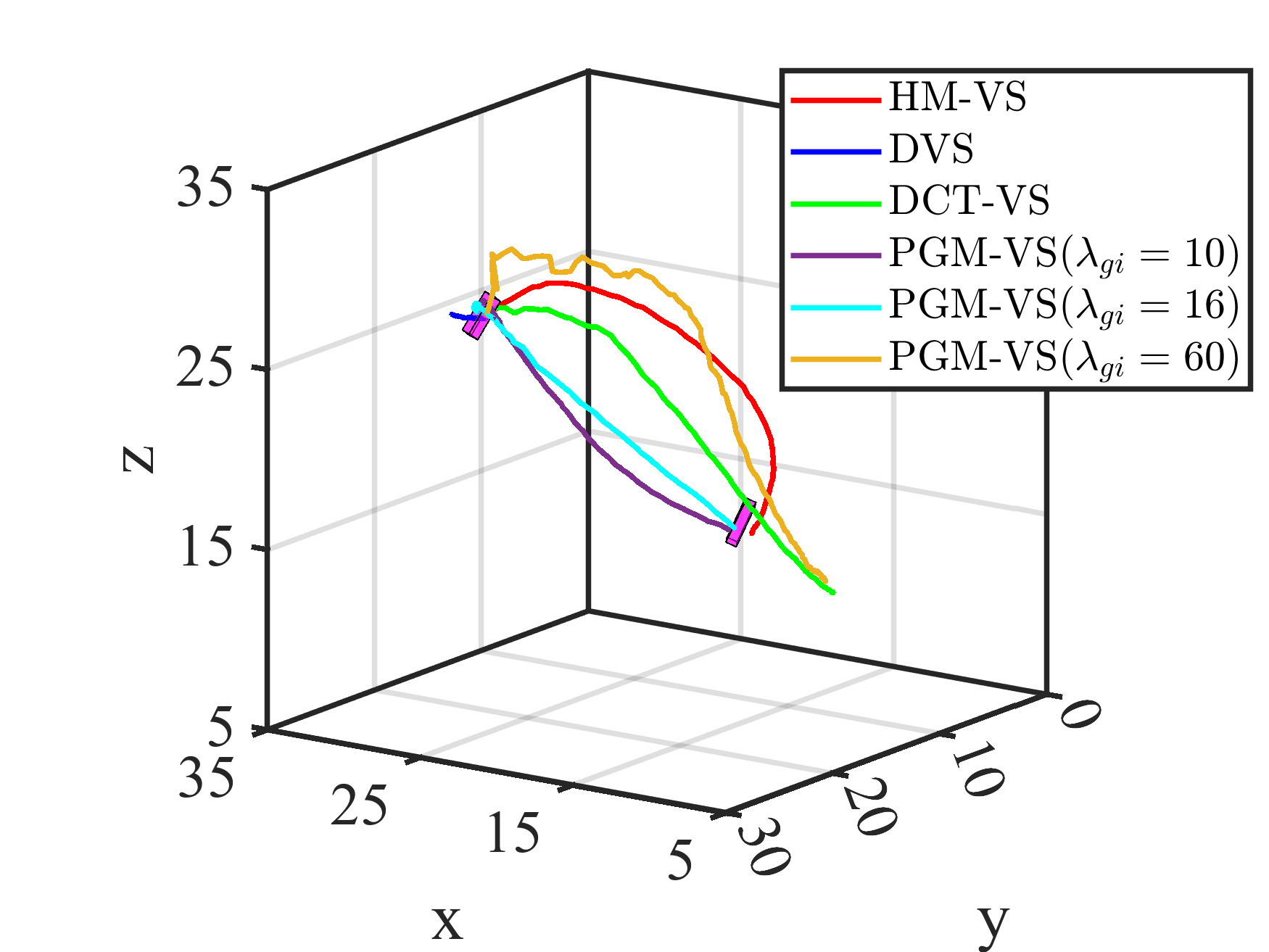}  \label{fig: Comparisons_trajections_noise_building}}					
		
		\caption{Case \#3: Example with a large displacement 3-D scene and camera trajectories. (a) Initial image. (b) Desired image. (c) Camera trajectories with Gaussian noise $\sigma^2 = 0$   (in m). (d) Camera trajectories with Gaussian noise $\sigma^2 = 0.5$  (in m).} 
		
		\label{fig: Comparisons_result_building}
	\end{figure}
	
	In the case of a large displacement facing a 3-D scene, the noise-free initial and desired images are shown in Figs. \ref{fig: Comparisons_result_new_building} and \ref{fig: Comparisons_image_old_building}, respectively.
	When Gaussian noise is absent, the HM-VS, DVS, PGM-VS ($\lambda_{gi}=10$), and PGM-VS ($\lambda_{gi}=16$) methods can successfully drive the camera to the desired pose (see Fig. \ref{fig: Comparisons_trajections_building}).
	When Gaussian noise $\sigma^2=0.5$ is present, only HM-VS, PGM-VS ($\lambda_{gi}=10$), and PGM-VS ($\lambda_{gi}=16$) methods cause the camera to converge next to the desired pose, but the final visual alignment of these methods is negligible (see Fig. \ref{fig: Comparisons_trajections_noise_building}).
	
	\begin{table*}
		\begin{center}
			\caption{HM-VS, DVS, DCT-VS, and PGM-VS methods comparison. }
			\label{tab: compare}
			\begin{tabular}{c|cc|cc|cc|}
				\cline{2-7}
				& \multicolumn{2}{c|}{Case \#1} & \multicolumn{2}{c|}{Case \#2} & \multicolumn{2}{c|}{Case \#3} \\ \cline{2-7} 
				& \multicolumn{1}{c|}{Noiseless}   & Noise & \multicolumn{1}{c|}{Noiseless}   & Noise & \multicolumn{1}{c|}{Noiseless}   & Noise  \\ \hline
				\multicolumn{1}{|c|}{HM-VS}  &  \multicolumn{1}{c|}{\color{green}{\CheckmarkBold}}   & \color{green}{\CheckmarkBold}  & \multicolumn{1}{c|}{\color{green}{\CheckmarkBold}}   &  \color{green}{\CheckmarkBold}& \multicolumn{1}{c|}{\color{green}{\CheckmarkBold}}   &  \color{green}{\CheckmarkBold} \\ \hline
				\multicolumn{1}{|c|}{DVS}    & \multicolumn{1}{c|}{\color{green}{\CheckmarkBold}}   &  \color{blue}{\CircleShadow}  & \multicolumn{1}{c|}{\color{green}{\CheckmarkBold}}   & \color{blue}{\CircleShadow}   & \multicolumn{1}{c|}{\color{green}{\CheckmarkBold}}   &\color{blue}{\CircleShadow}  \\ \hline
				\multicolumn{1}{|c|}{DCT-VS} & \multicolumn{1}{c|}{\color{green}{\CheckmarkBold}}   &  \color{green}{\CheckmarkBold} & \multicolumn{1}{c|}{\color{blue}{\CircleShadow} }   &  \color{blue}{\CircleShadow}  & \multicolumn{1}{c|}{\color{red}{\XSolidBold}}   &  \color{red}{\XSolidBold} \\ \hline
				\multicolumn{1}{|c|}{PGM-VS($\lambda_{gi}=10$)} & \multicolumn{1}{c|}{\color{green}{\CheckmarkBold}}   & \color{green}{\CheckmarkBold}  & \multicolumn{1}{c|}{\color{red}{\XSolidBold}}    & \color{red}{\XSolidBold} & \multicolumn{1}{c|}{\color{green}{\CheckmarkBold} }   & \color{green}{\CheckmarkBold} \\ \hline
				\multicolumn{1}{|c|}{PGM-VS($\lambda_{gi}=16$)} & \multicolumn{1}{c|}{\color{green}{\CheckmarkBold}}   & \color{green}{\CheckmarkBold}  & \multicolumn{1}{c|}{\color{red}{\XSolidBold}}   & \color{red}{\XSolidBold}  & \multicolumn{1}{c|}{\color{green}{\CheckmarkBold} }   &  \color{green}{\CheckmarkBold} \\ \hline
				\multicolumn{1}{|c|}{PGM-VS($\lambda_{gi}=60$)} & \multicolumn{1}{c|}{\color{red}{\XSolidBold}}   & \color{red}{\XSolidBold}  & \multicolumn{1}{c|}{\color{blue}{\CircleShadow}}   & \color{red}{\XSolidBold}  & \multicolumn{1}{c|}{\color{red}{\XSolidBold} }   &  \color{red}{\XSolidBold} \\ \hline    
			\end{tabular}
		\end{center}
		Successful ({\color{green}{\CheckmarkBold}}) and failed ({\color{red}{\XSolidBold}}) convergences for the three cases (see Figs. \ref{fig: Comparisons_result_panda}-\ref{fig: Comparisons_result_building}). A blue marker ({\color{blue}{\CircleShadow}}) means that the camera has converged next to the desired pose and that the final visual alignment is not negligible. 
	\end{table*}
	
	Table \ref{tab: compare} provides the successful and failed convergences for these cases.
	Only the HM-VS method is available for all cases.
	The experiments show that the robustness of the DVS method is not satisfactory, as it does not have the ability to filter.
	Strictly speaking, DCT-VS is also a DOMs-based VS method, but HM-VS has more advantages due to its flexible parameter tuning mechanism.
	Finally, the PGM-VS method also performs well, but the parameter $\lambda_{gi}$ required for this method need to be adjusted empirically for different scenarios. 
	
	\subsection{Experimental Results on a 7-DoF Robot} \label{sec: 3D_exp}
	This subsection aims to demonstrate that the proposed method works well even in  real environments for both 2-D and 3-D objects.
	The experiments were conducted on a 7-DoF Franka Emika  robotic arm with an Intel RealSense L515  LiDAR camera.
	The LiDAR camera simultaneously acquires color and depth images with a ($640 \times 480$) resolution.
	The camera calibration, as well as the hand-eye calibration, have been done in an offline step.
	The depths are not estimated but are available from the camera truth data.
	The image processing and the control law computation are performed on a PC equipped with a 14-core 2.3 GHz Intel Core i7-12700H.
	This allows a frequency for the servo loop around 2 Hz.
	The required parameters are
	$l_{\text{min}}=6$ and $l_{\text{max}}=16$ in the following experiences.
	
	\subsubsection*{Experiment \#5 (see Figs. \ref{fig: Experiment_2D_case} and \ref{fig: Experiment_2D_result})}
	
	\begin{figure}
		\centering 
		
		\includegraphics[width=0.49\hsize]{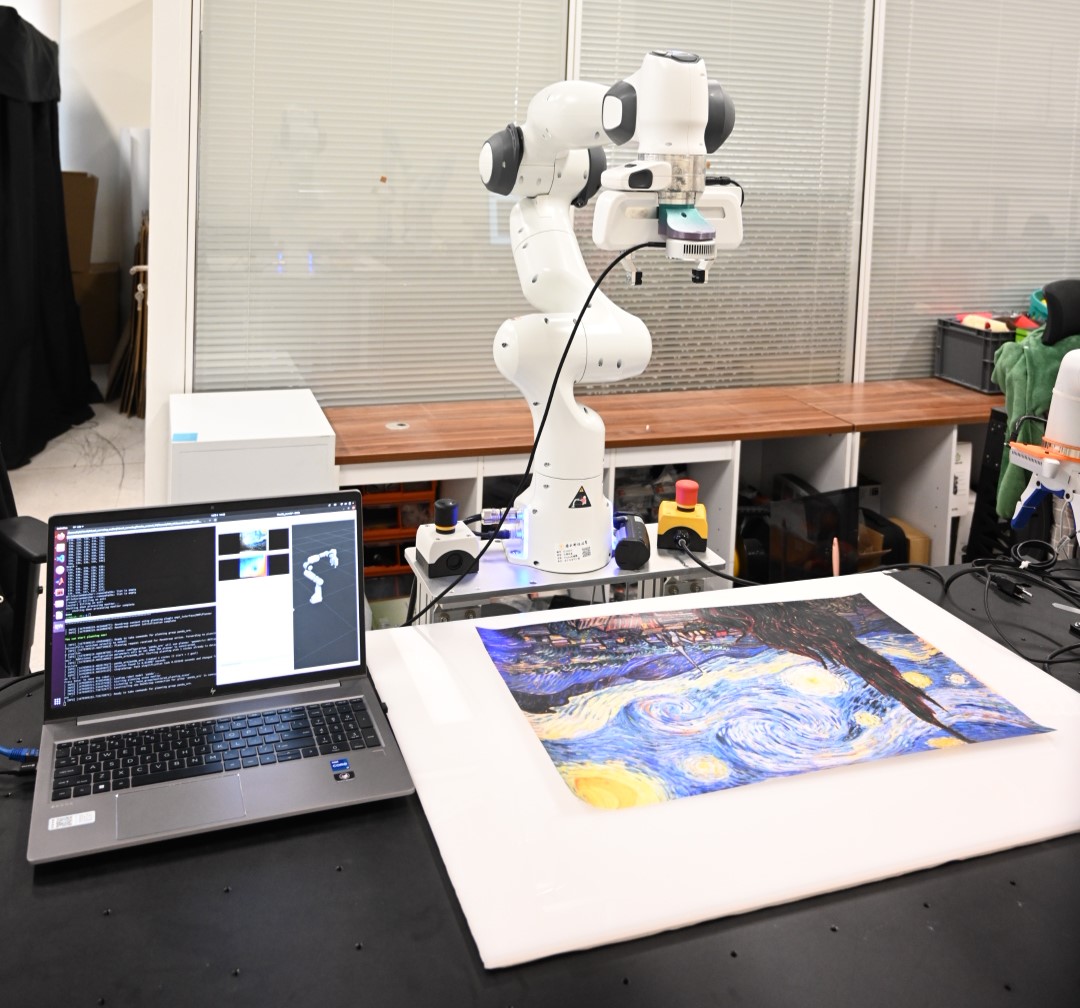} 		
		
		\caption{The real 2-D experimental environment.}
		
		\label{fig: Experiment_2D_case}
	\end{figure}
	
	\begin{figure}
		\centering 
		
		\subfloat[]{\includegraphics[width=0.45\hsize]{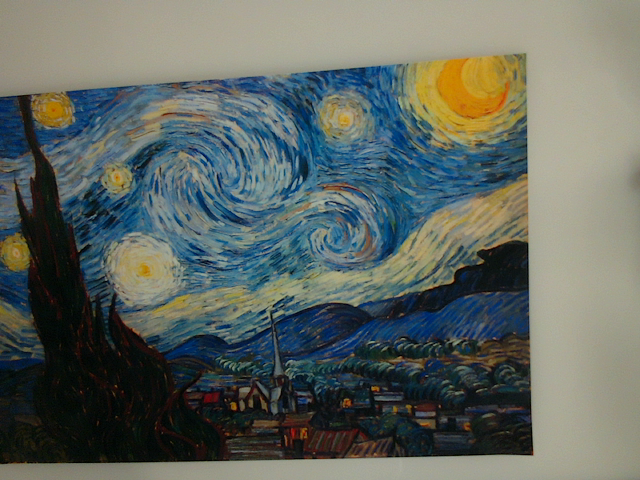}  \label{fig: Experiment_2D_image_new}}	
		\subfloat[]{\includegraphics[width=0.45\hsize]{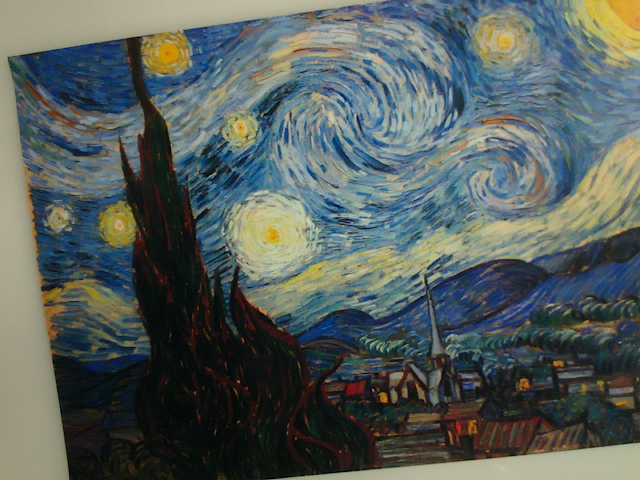}  \label{fig: Experiment_2D_image_old}}				
		
		\subfloat[]{\includegraphics[width=0.49\hsize]{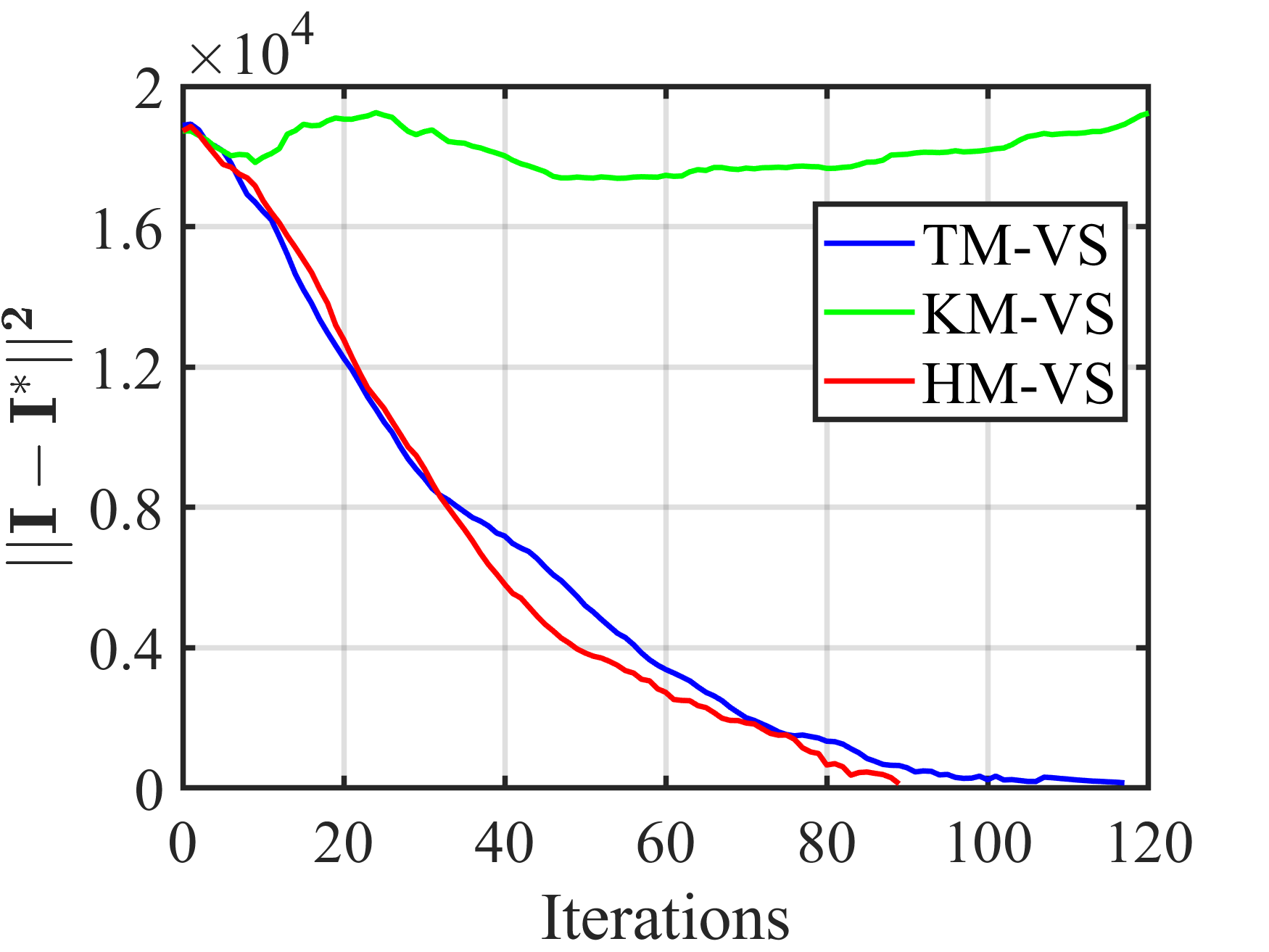}  \label{fig: Experiment_2D_I}}	
		\subfloat[]{\includegraphics[width=0.49\hsize]{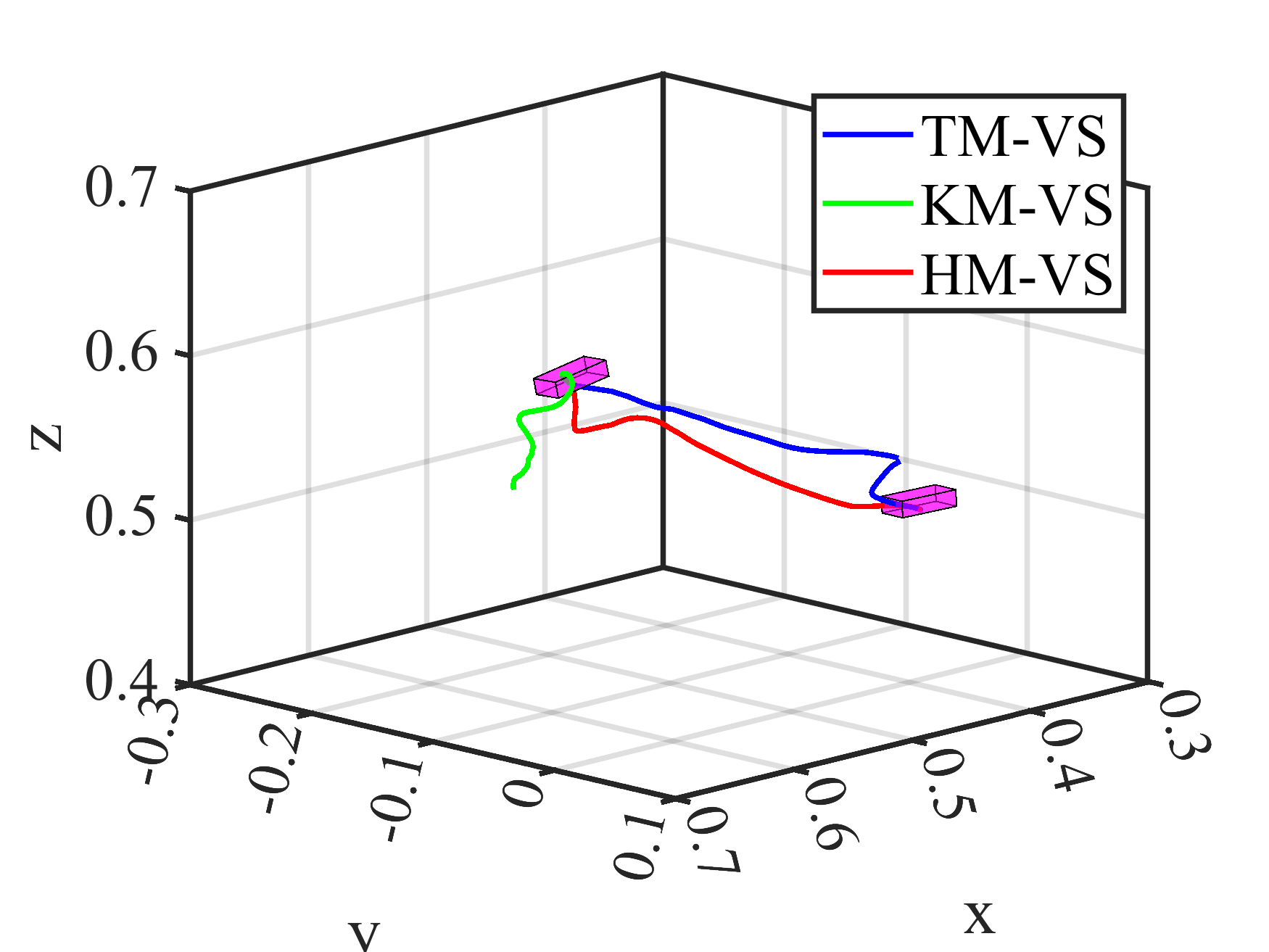}  \label{fig: Experiment_2D_trajections}}		
		
		\caption{Experiment \#5: Comparison between TM-VS, KM-VS, and HM-VS in a real 2-D environment. (a) Initial image. (b) Desired image. (c) Pixel errors. (d) Camera trajectories (in m).}
		
		\label{fig: Experiment_2D_result}
	\end{figure}
	
	Fig. \ref{fig: Experiment_2D_case} illustrates the 2-D real experimental environment.  
	The scene contains a 2-D object under common lighting conditions.
	Figs. \ref{fig: Experiment_2D_image_new} and \ref{fig: Experiment_2D_image_old} show the initial and desired images, respectively, where the complex plane is partially outside the camera field-of-view in the desired image. 
	The displacement between the initial and the desired camera poses is given by  ($0.15\text{m}$, $-0.13\text{m}$, $-0.09\text{m}$, $-8.68^{\circ}$, $-1.13^{\circ}$, $-3.00^{\circ}$).
	All three methods, TM-VS, KM-VS, and HM-VS, perform VS control.
	The camera trajectories obtained from these methods are presented in Fig. \ref{fig: Experiment_2D_trajections}.
	Both TM-VS and  HM-VS schemes can converge the pose error to less than $(2\text{mm}$, $2\text{mm}$, $2\text{mm}$,  $0.5^{\circ}$, $0.5^{\circ}$, $0.5^{\circ})$.
	Unfortunately, the KM-VS scheme fails in the present case.
	The pixel errors $||\mathbf{I} - \mathbf{I}^*||^2$ obtained from these three methods are shown in Fig. \ref{fig: Experiment_2D_I}.
	Specifically, the TM-VS and  HM-VS require $118$ and $90$ iterations, respectively.
	The HM-VS scheme still has the fastest convergence rate.
	The control results of the TM-VS and HM-VS methods are illustrated in Figs. \ref{fig: Experiment_2D_TMVS} and  \ref{fig: Experiment_2D_HMVS} , respectively.
	It is worth explaining that the variation of the HM parameter in Fig. \ref{fig: Experiment_2D_HMVS_ab} is much smaller than in the simulations since the background is not black in the real experiment, making the ROI of the current and desired images less variable.

	\begin{figure}
		\centering 
		
		\subfloat[]{\includegraphics[width=0.33\hsize]{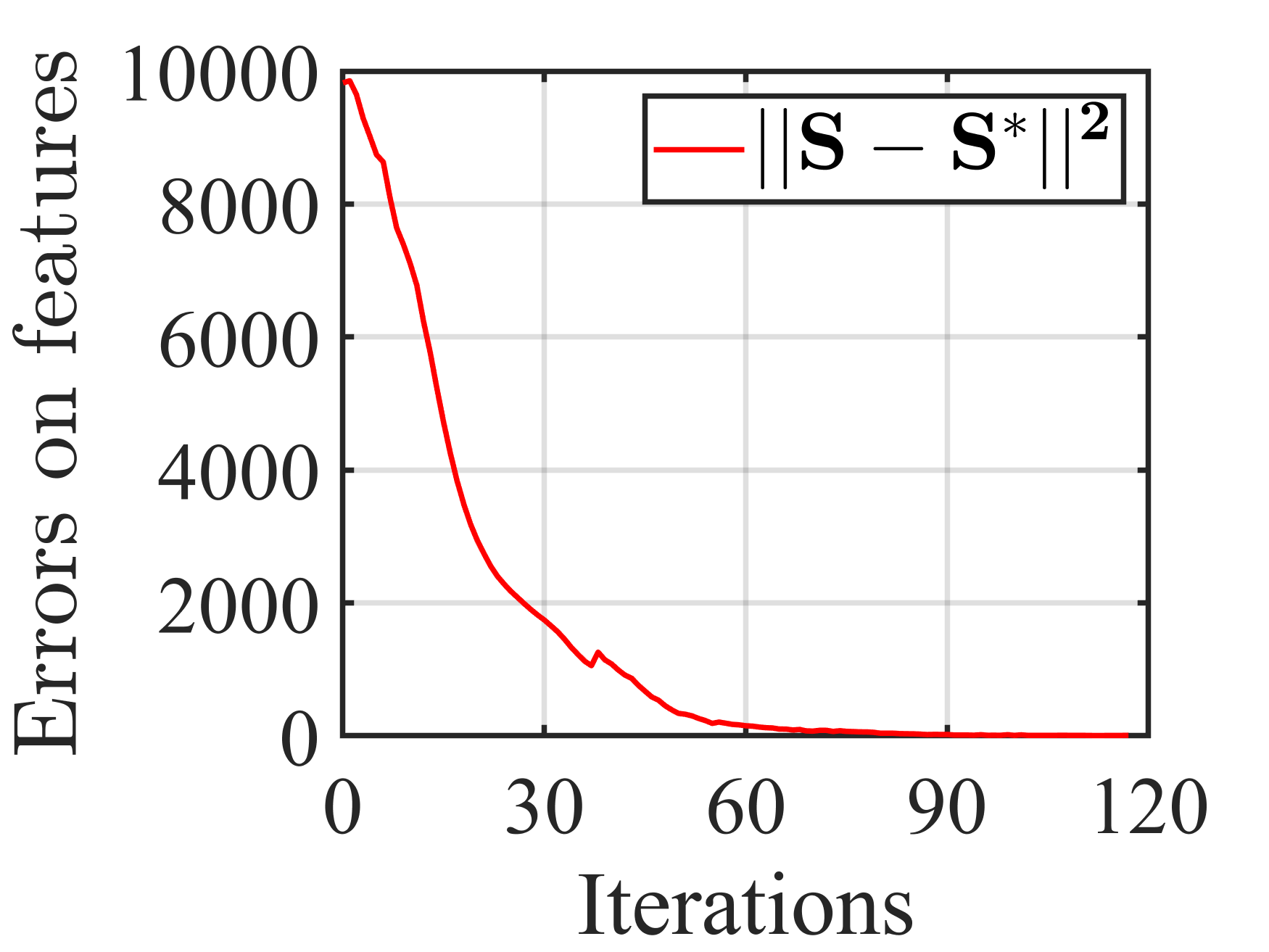}  \label{fig: Experiment_2D_TMVS_feature_error}}	
		\subfloat[]{\includegraphics[width=0.33\hsize]{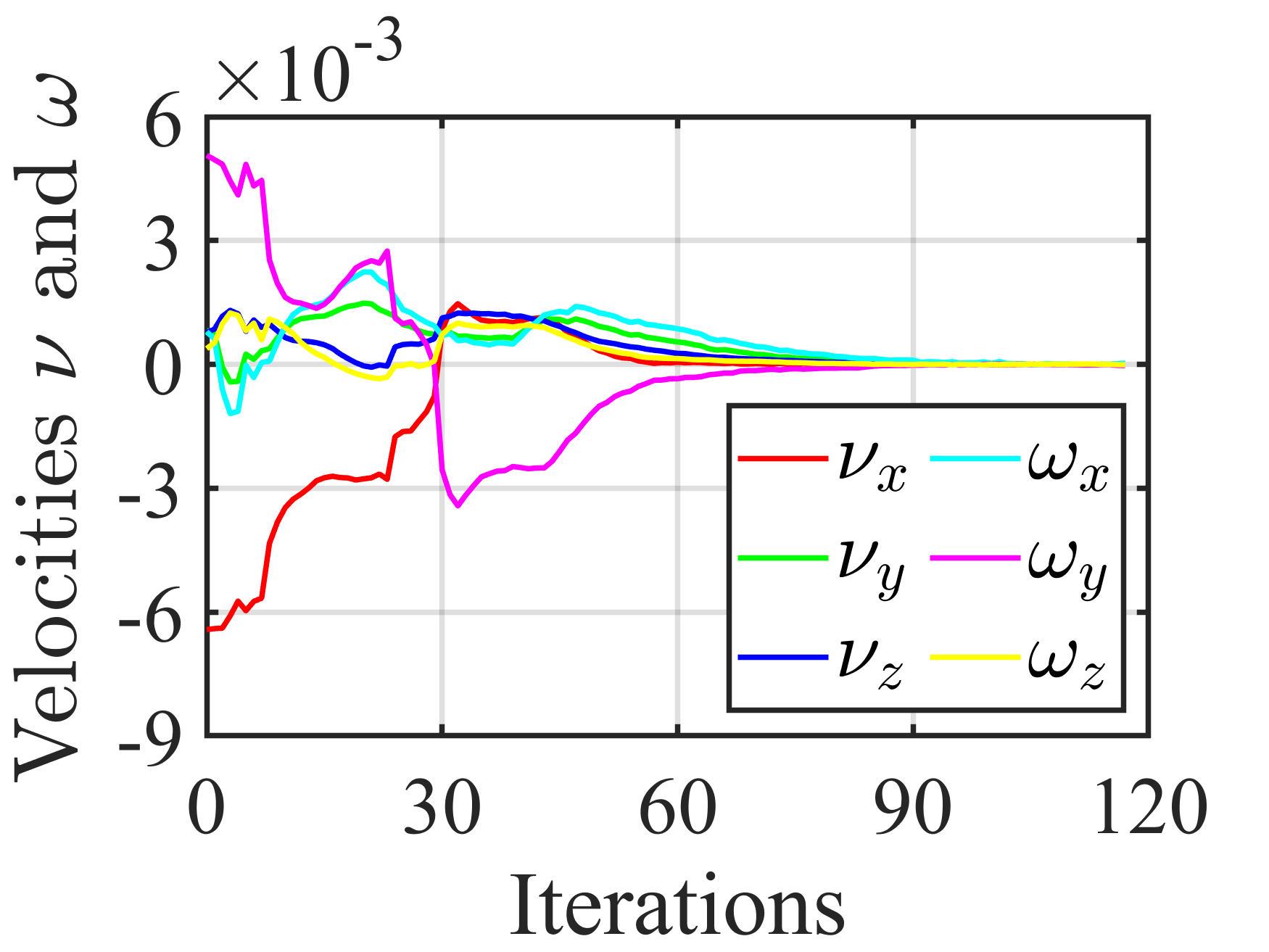}  \label{fig: Experiment_2D_TMVS_velocity}}		
		\subfloat[]{\includegraphics[width=0.33\hsize]{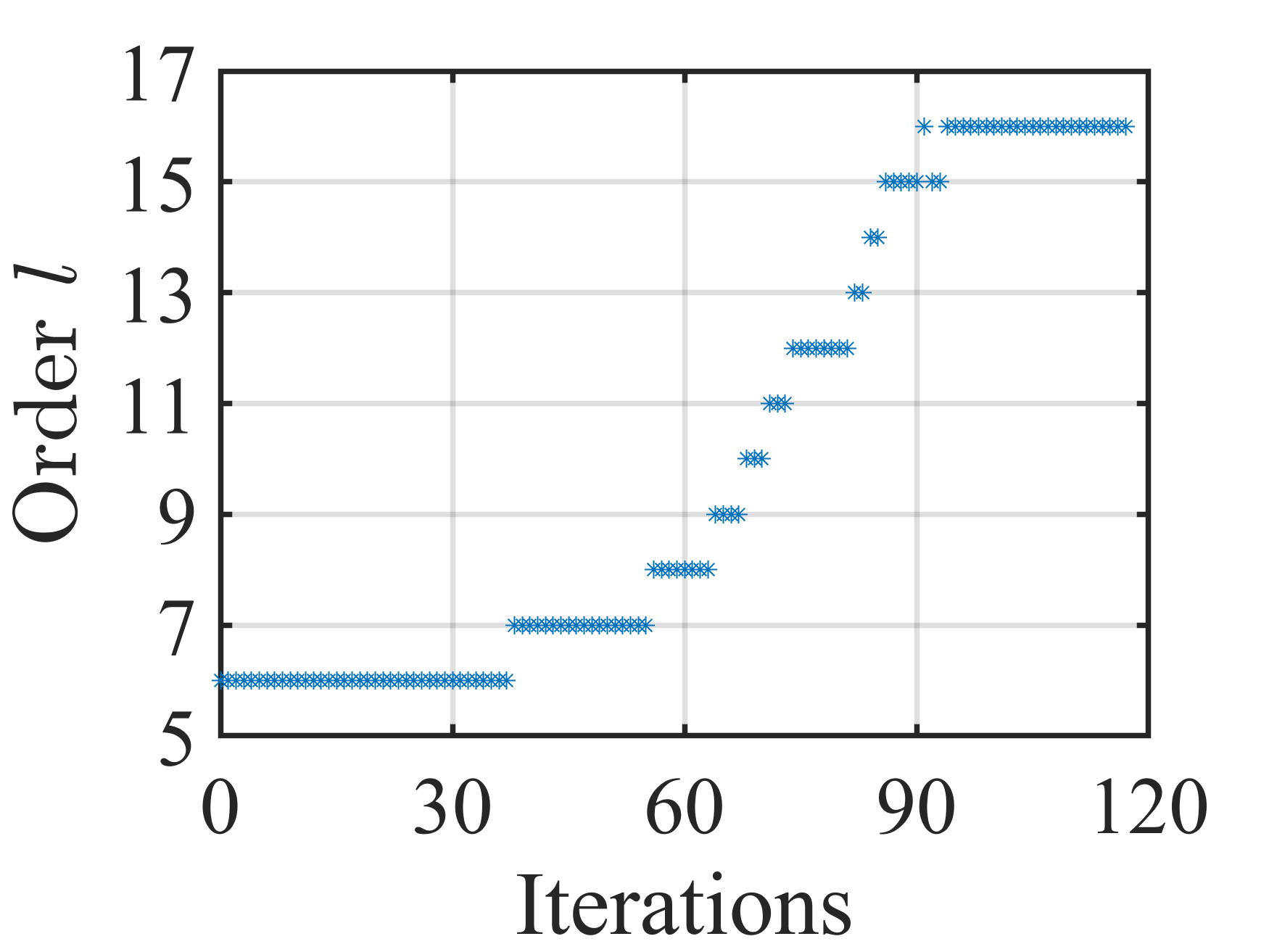}  \label{fig: Experiment_2D_TMVS_order}}		
		
		\caption{Results for TM-VS in Experiment \#5. (a) Errors on features. (b) Camera velocities (in m/s and rad/s). (c) Order of  DOMs as visual features.}
		
		\label{fig: Experiment_2D_TMVS}
	\end{figure}

	\begin{figure}
		\centering 
		
		\subfloat[]{\includegraphics[width=0.33\hsize]{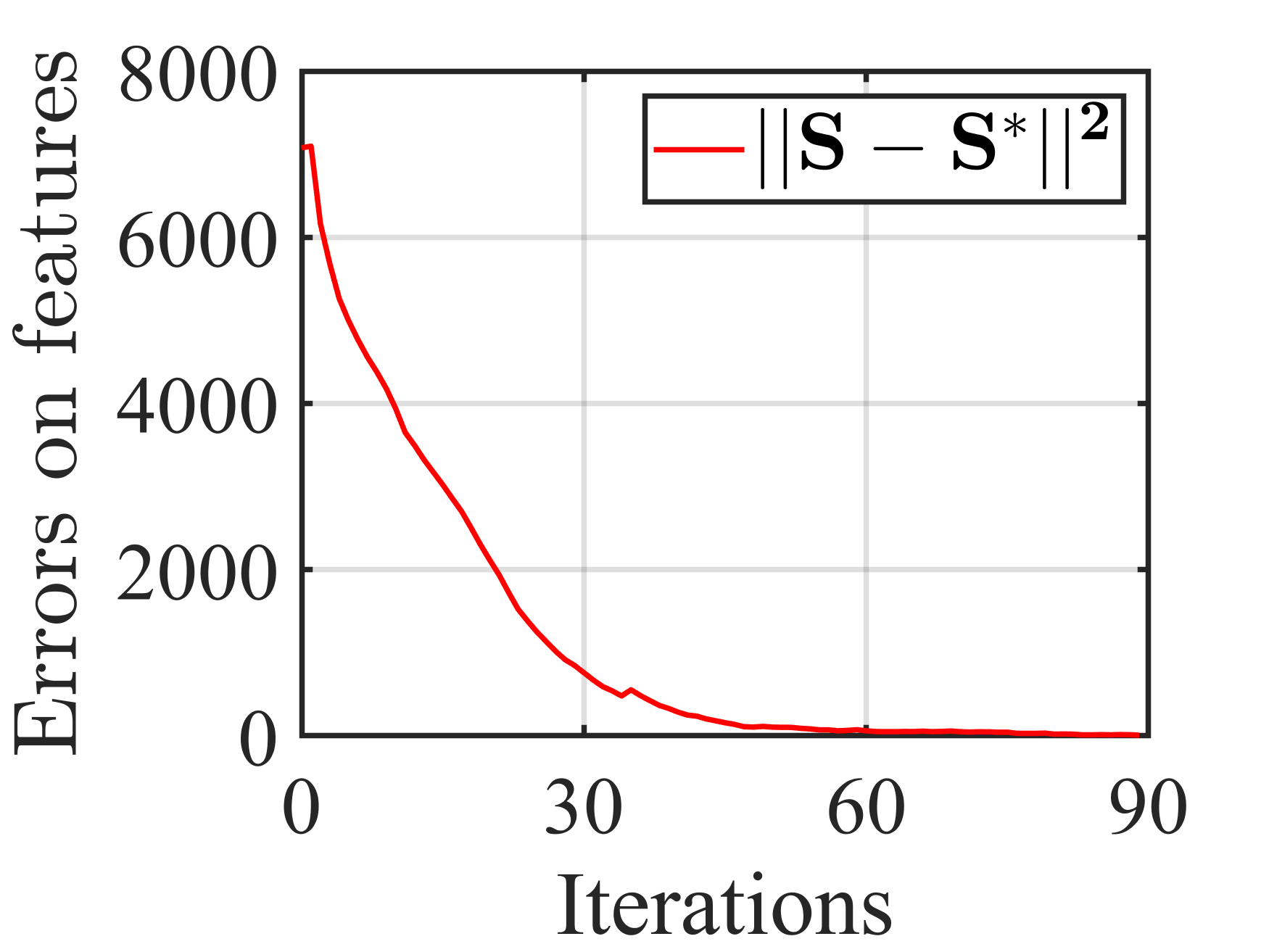}  \label{fig: Experiment_2D_HMVS_feature_error}}	\qquad
		\subfloat[]{\includegraphics[width=0.33\hsize]{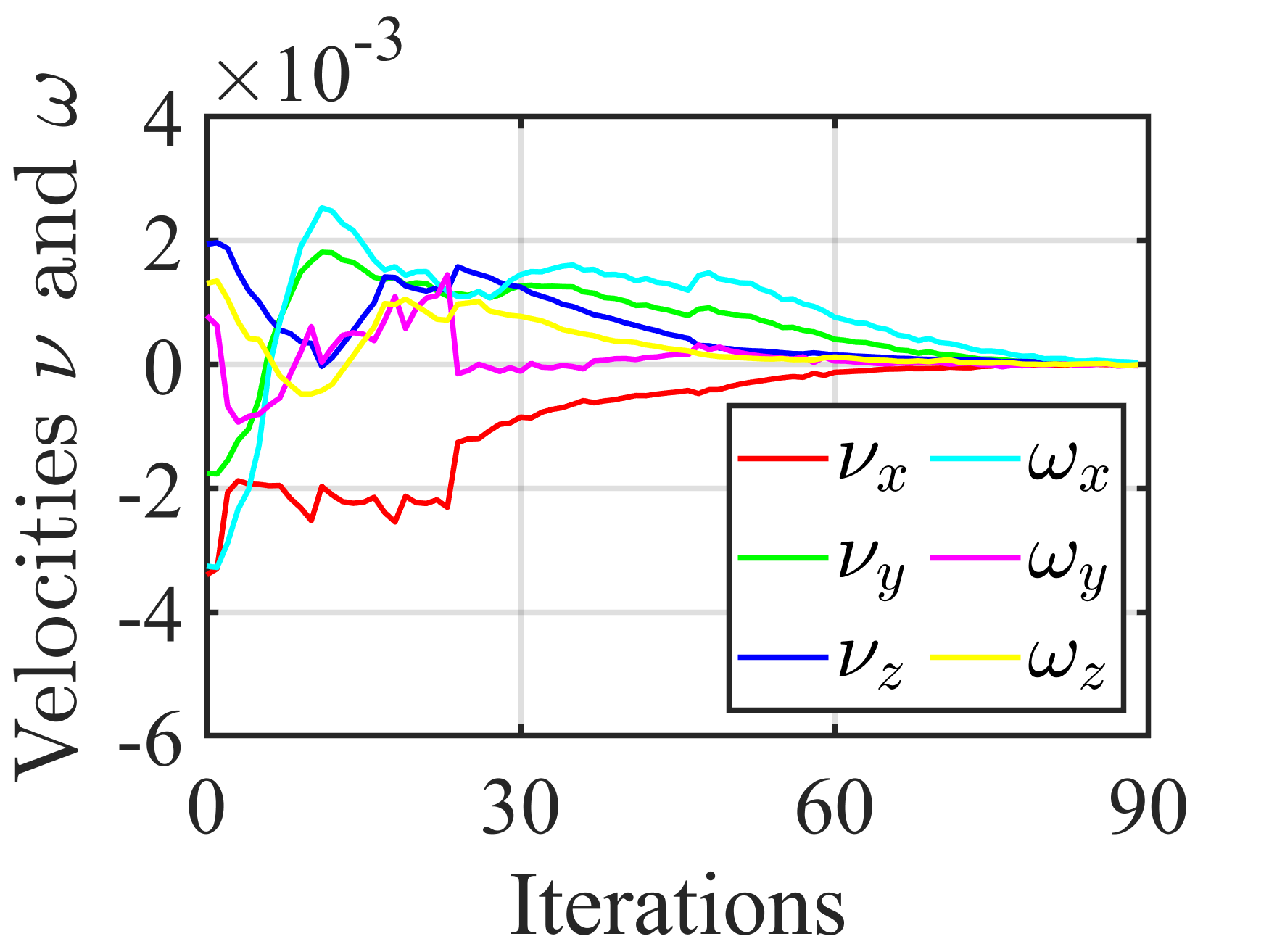}  \label{fig: Experiment_2D_HMVS_velocity}}		
		
		\subfloat[]{\includegraphics[width=0.33\hsize]{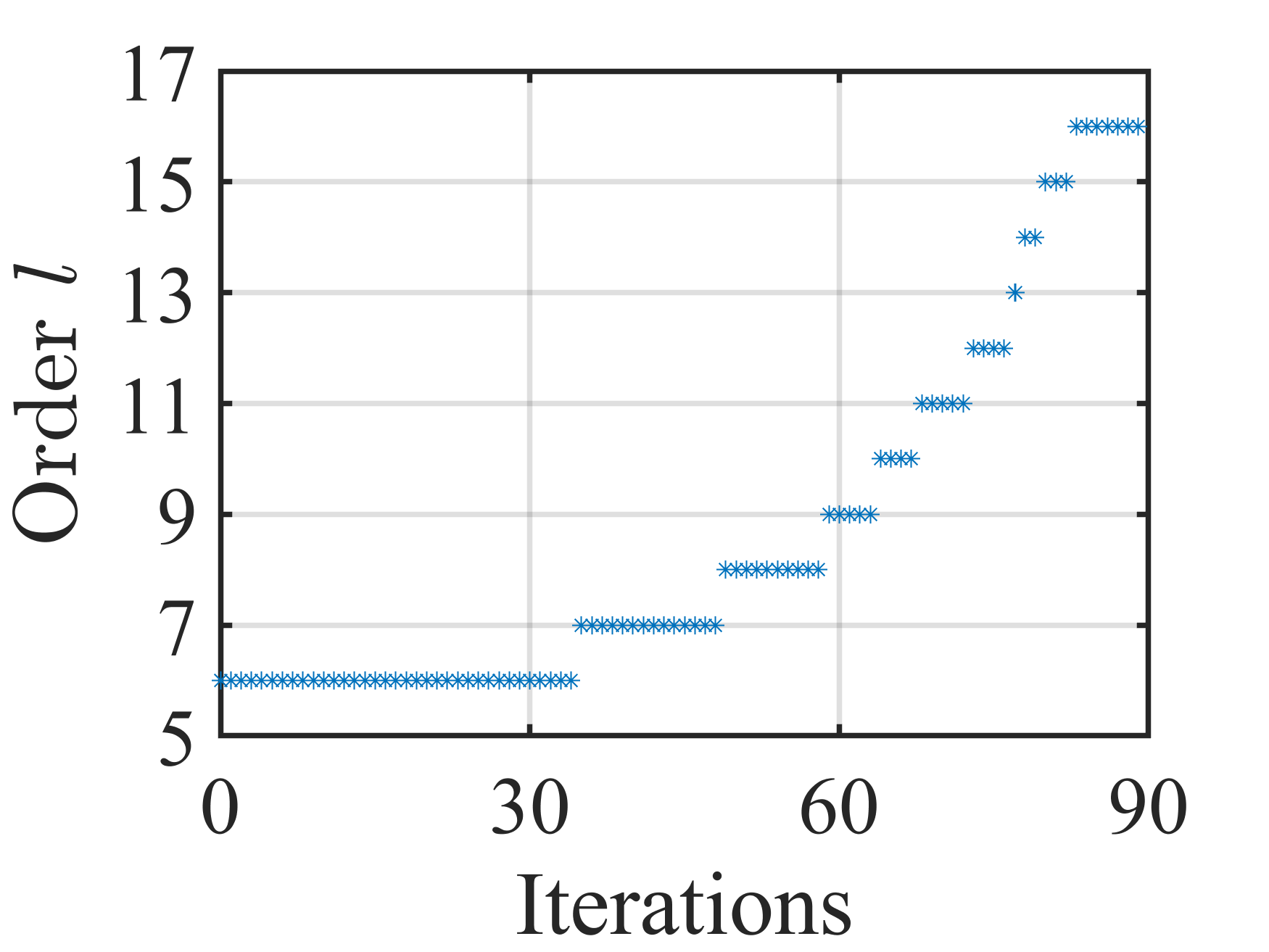}  \label{fig: Experiment_2D_HMVS_order}}		\qquad
		\subfloat[]{\includegraphics[width=0.33\hsize]{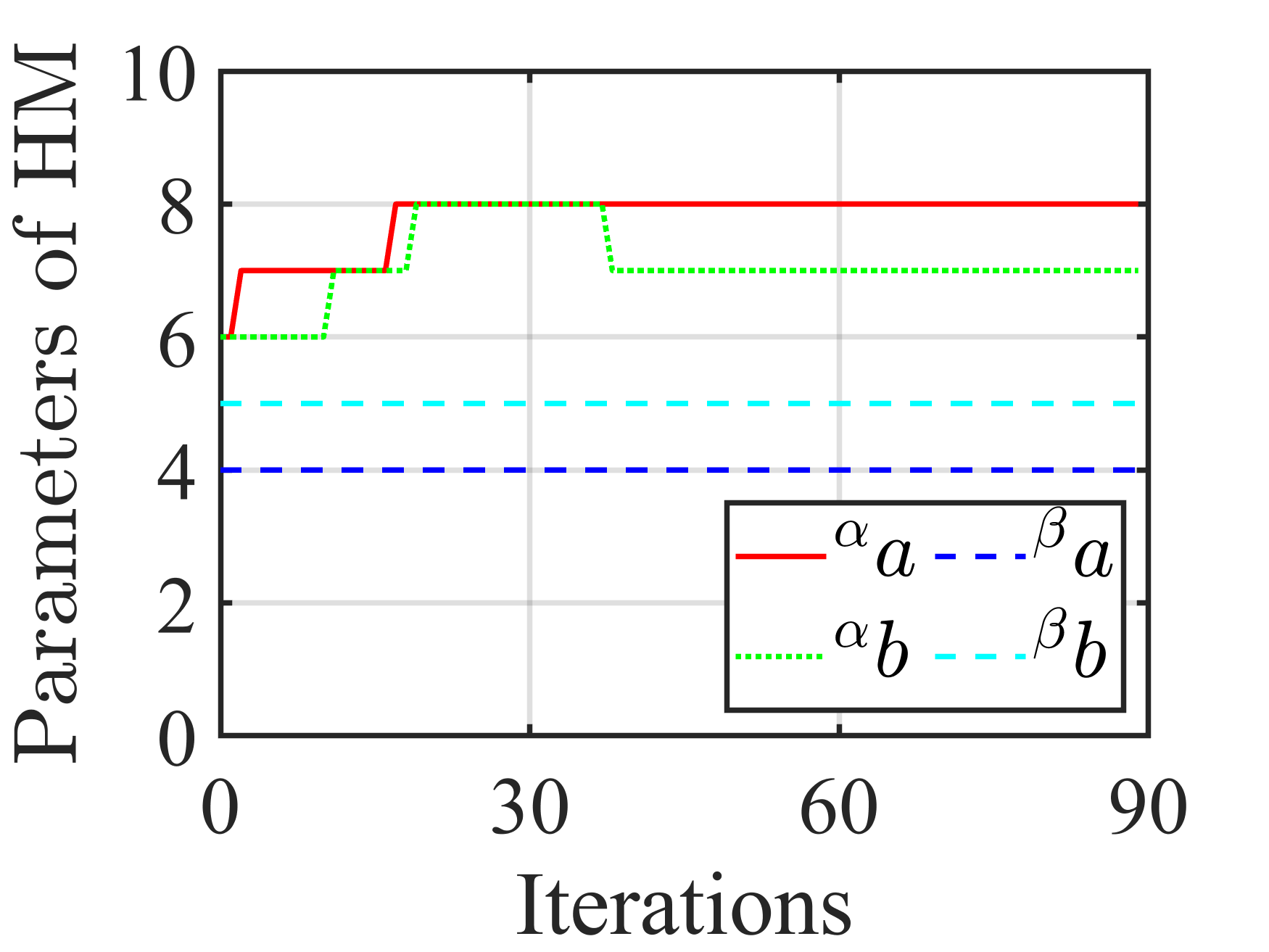}  \label{fig: Experiment_2D_HMVS_ab}}				
		
		\caption{Results for HM-VS in Experiment \#5. (a) Errors on features. (b) Camera velocities (in m/s and rad/s). (c) Order of  DOMs as visual features. (d) Parameters of HMs.}
		
		\label{fig: Experiment_2D_HMVS}
	\end{figure}
	
	\subsubsection*{Experiment \#6 (see Figs. \ref{fig: Experiment_3D_case} and \ref{fig: Experiment_3D_result})}
	
	\begin{figure}
		\centering 
		
		\includegraphics[width=0.49\hsize]{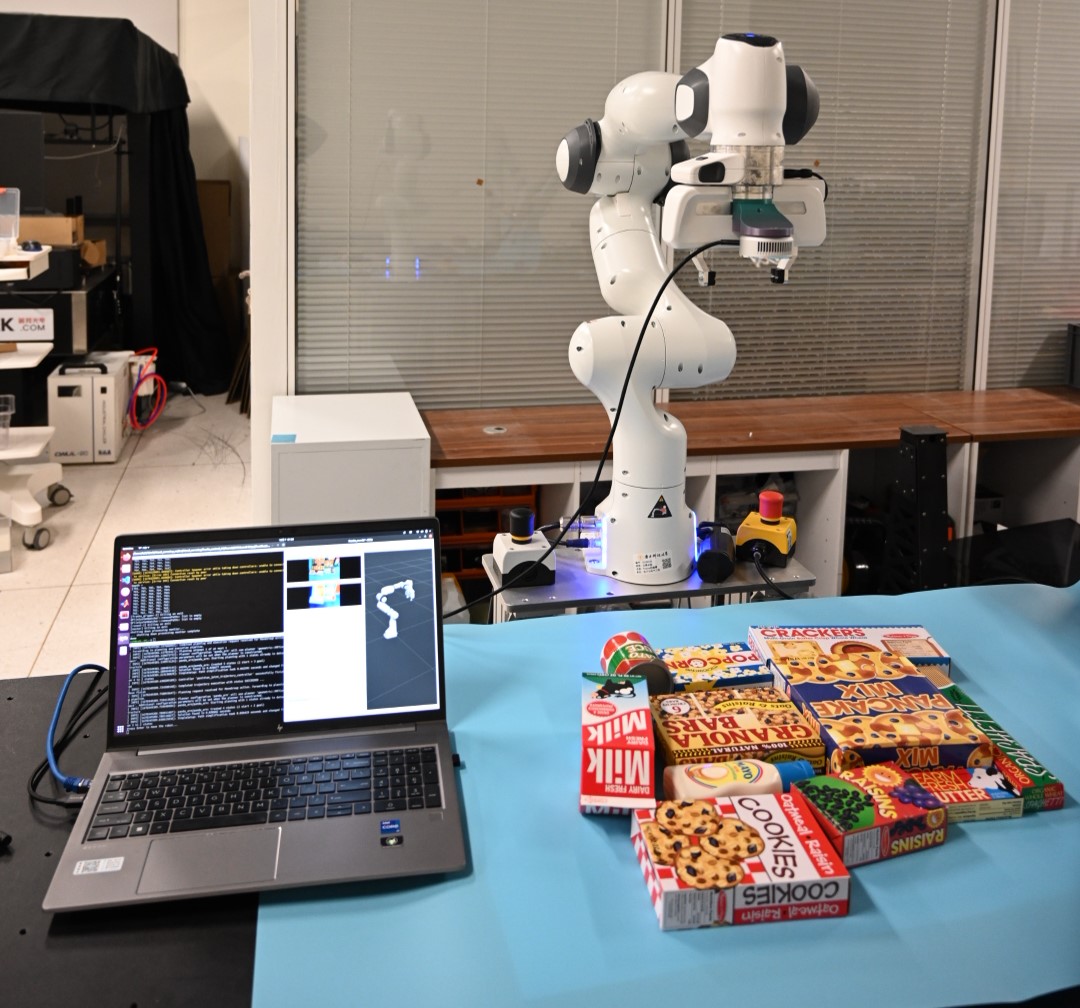}  
		
		\caption{The real 3-D experimental environment.}
		
		\label{fig: Experiment_3D_case}
	\end{figure}
	
	\begin{figure}
		\centering 
		
		\subfloat[]{\includegraphics[width=0.49\hsize]{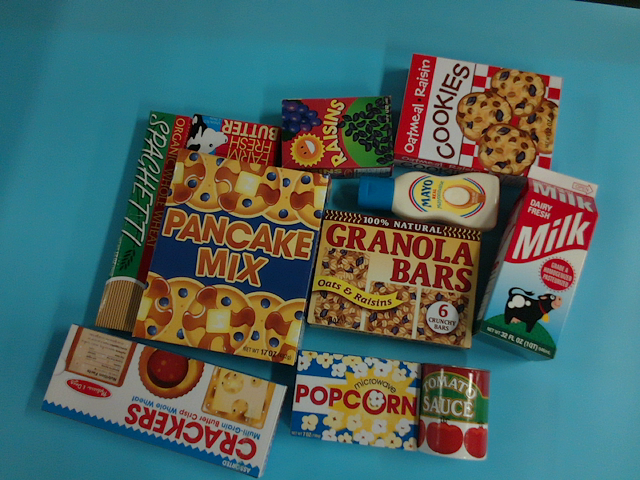}  \label{fig: Experiment_3D_image_new}}	
		\subfloat[]{\includegraphics[width=0.49\hsize]{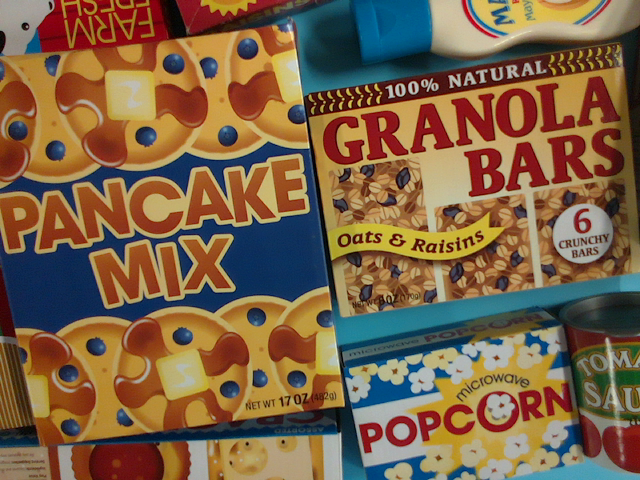}  \label{fig: Experiment_3D_image_old}}		
		
		\subfloat[]{\includegraphics[width=0.49\hsize]{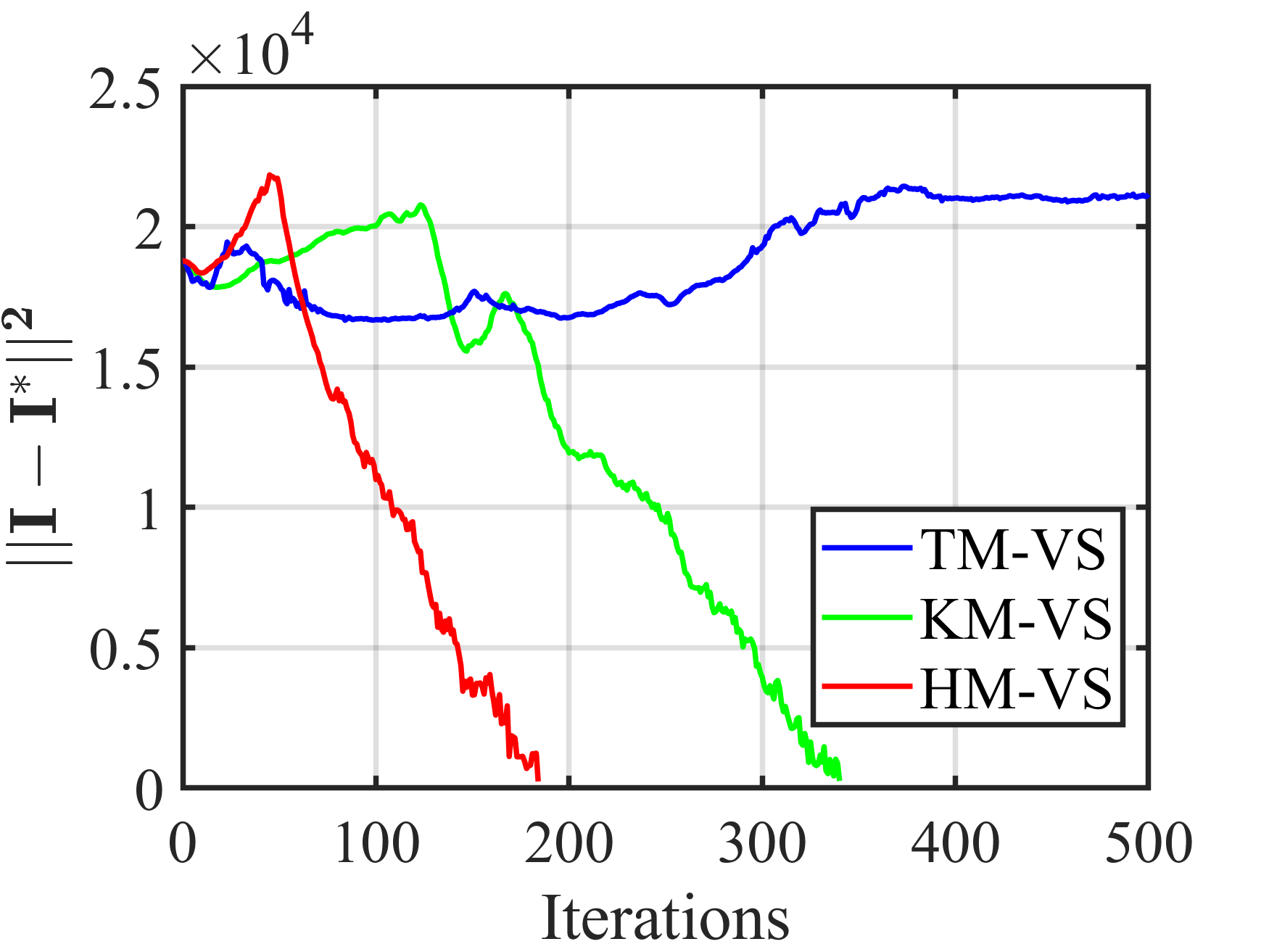}  \label{fig: Experiment_3D_I}}	
		\subfloat[]{\includegraphics[width=0.49\hsize]{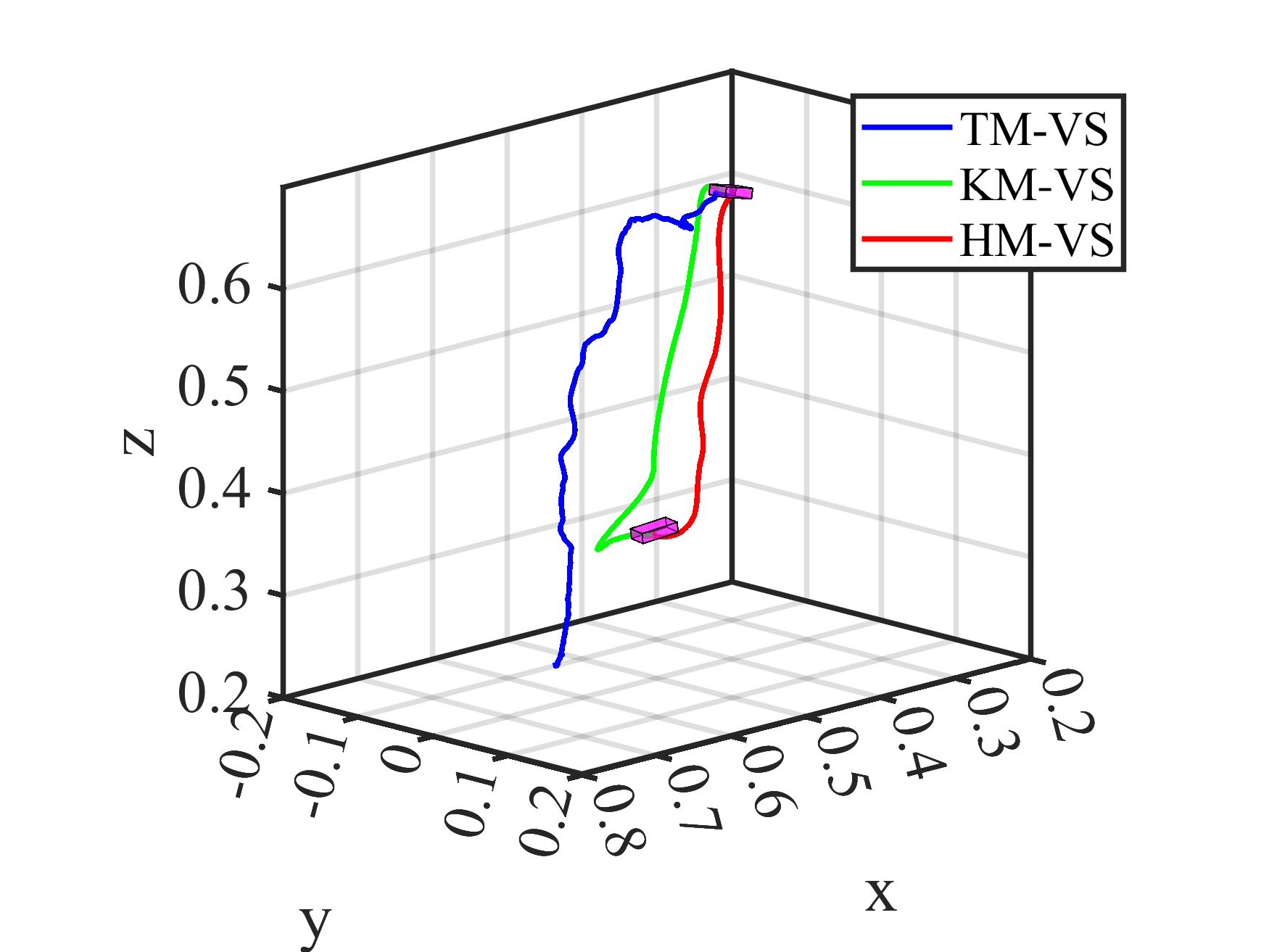}  \label{fig: Experiment_3D_trajections}}				
		
		\caption{Experiment \#6: Comparison between TM-VS, KM-VS, and HM-VS in a real 3-D  environment. (a) Initial image. (b) Desired image. (c) Pixel errors. (d) Camera trajectories (in m).}
		
		\label{fig: Experiment_3D_result}
	\end{figure}
	
	The experiment is implemented in a real 3-D environment.
	A realistic scenario is set up by placing several 3-D objects of varying shapes, sizes, and colors in the scene, as shown in Fig. \ref{fig: Experiment_3D_case}.
	The desired image is given in Fig. \ref{fig: Experiment_3D_image_old} while the initial one is shown in Fig. \ref{fig: Experiment_3D_image_new}.
	The challenge of this experiment is that some objects are outside  the camera field-of-view in the desired image.
	The corresponding displacement is ($0.04\text{m}$, $0.13\text{m}$, $-0.31\text{m}$, $20.88^{\circ}$, $-3.76^{\circ}$, $-17.70^{\circ}$).
	The pixel errors $||\mathbf{I} - \mathbf{I}^*||^2$ and the camera trajectories obtained from these three methods (TM-VS, KM-VS, and HM-VS)  are shown in Figs. \ref{fig: Experiment_3D_I} and \ref{fig: Experiment_3D_trajections}, respectively.
	It can be easily seen that only the KM-VS and HM-VS methods succeed in the VS task, while the TM-VS approach fails because it falls into local minima in the VS.
	Both KM-VS and HM-VS schemes can converge the pose error to less than $(1\text{mm}$, $1\text{mm}$, $1\text{mm}$,  $0.3^{\circ}$, $0.3^{\circ}$, $0.3^{\circ})$.
	However, the convergence rate of the HM-VS method is better than that of KM-VS.
	The details of the KM-VS and HM-VS methods are illustrated in Figs. \ref{fig: Experiment_3D_KMVS} and \ref{fig: Experiment_3D_HMVS}, respectively.

	\begin{figure}
		\centering 
		
		\subfloat[]{\includegraphics[width=0.33\hsize]{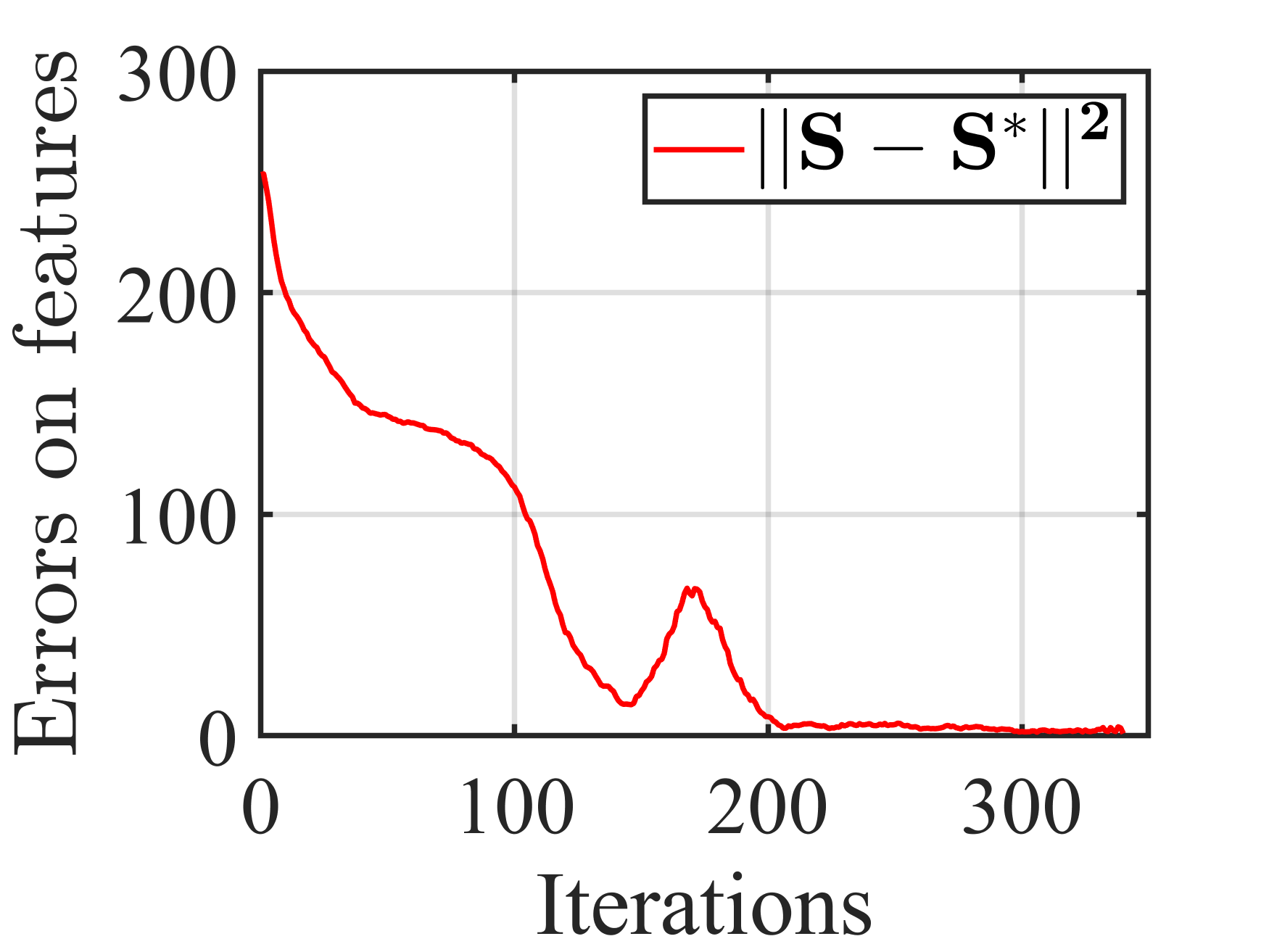}  \label{fig: Experiment_3D_KMVS_feature_error}}	\qquad
		\subfloat[]{\includegraphics[width=0.33\hsize]{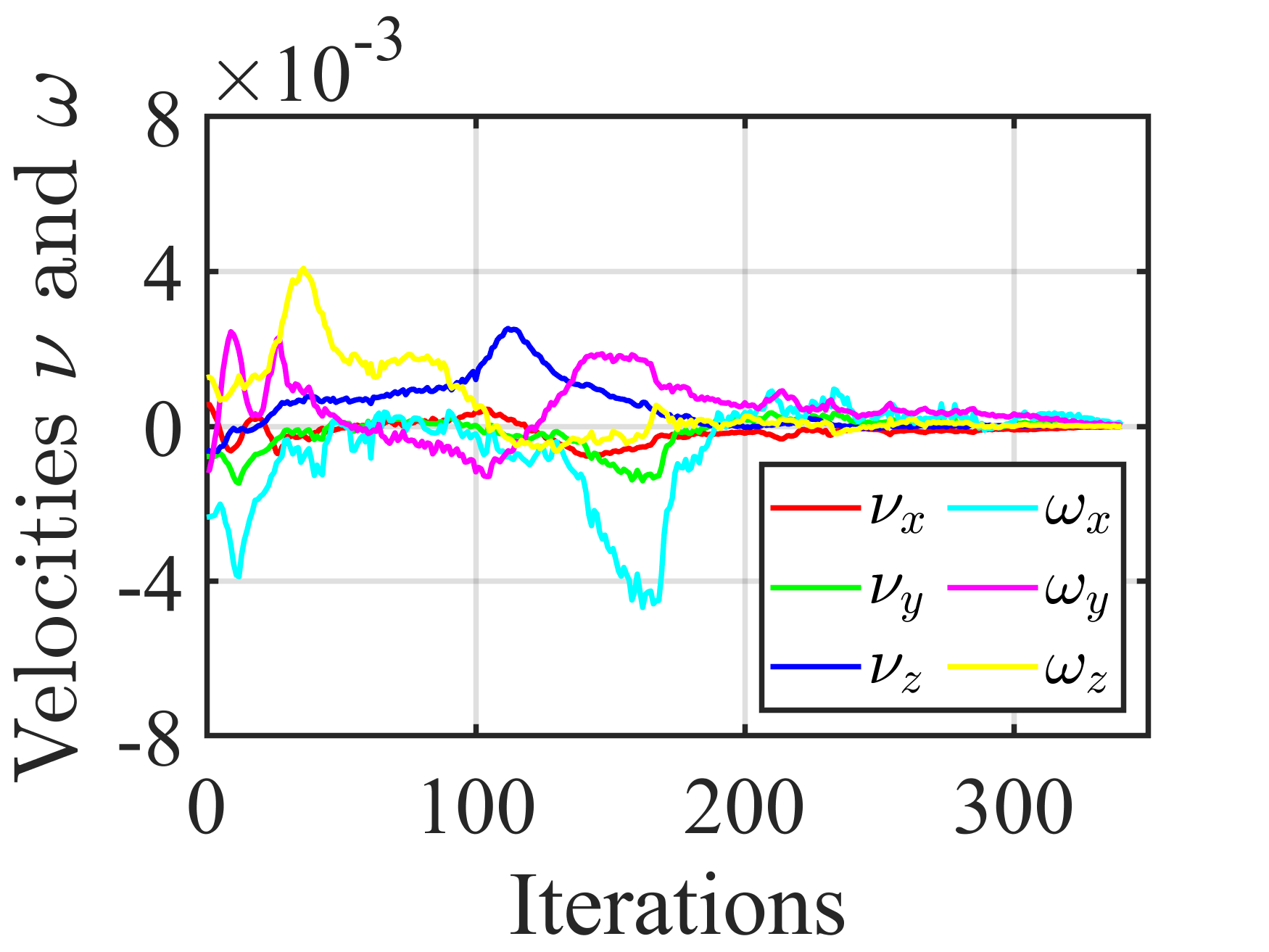}  \label{fig: Experiment_3D_KMVS_velocity}}		
		
		\subfloat[]{\includegraphics[width=0.33\hsize]{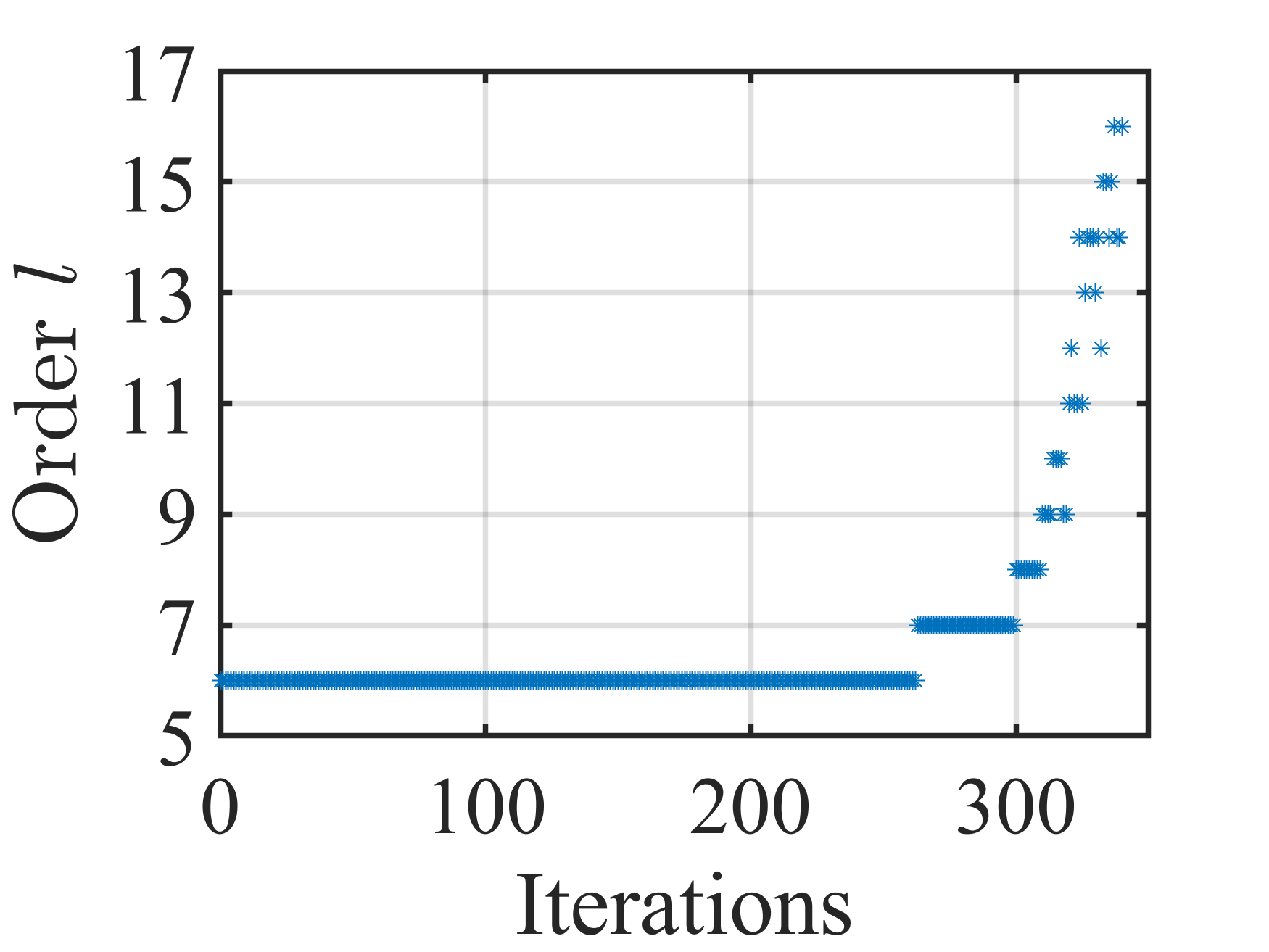}  \label{fig: Experiment_3D_KMVS_order}}		\qquad
		\subfloat[]{\includegraphics[width=0.33\hsize]{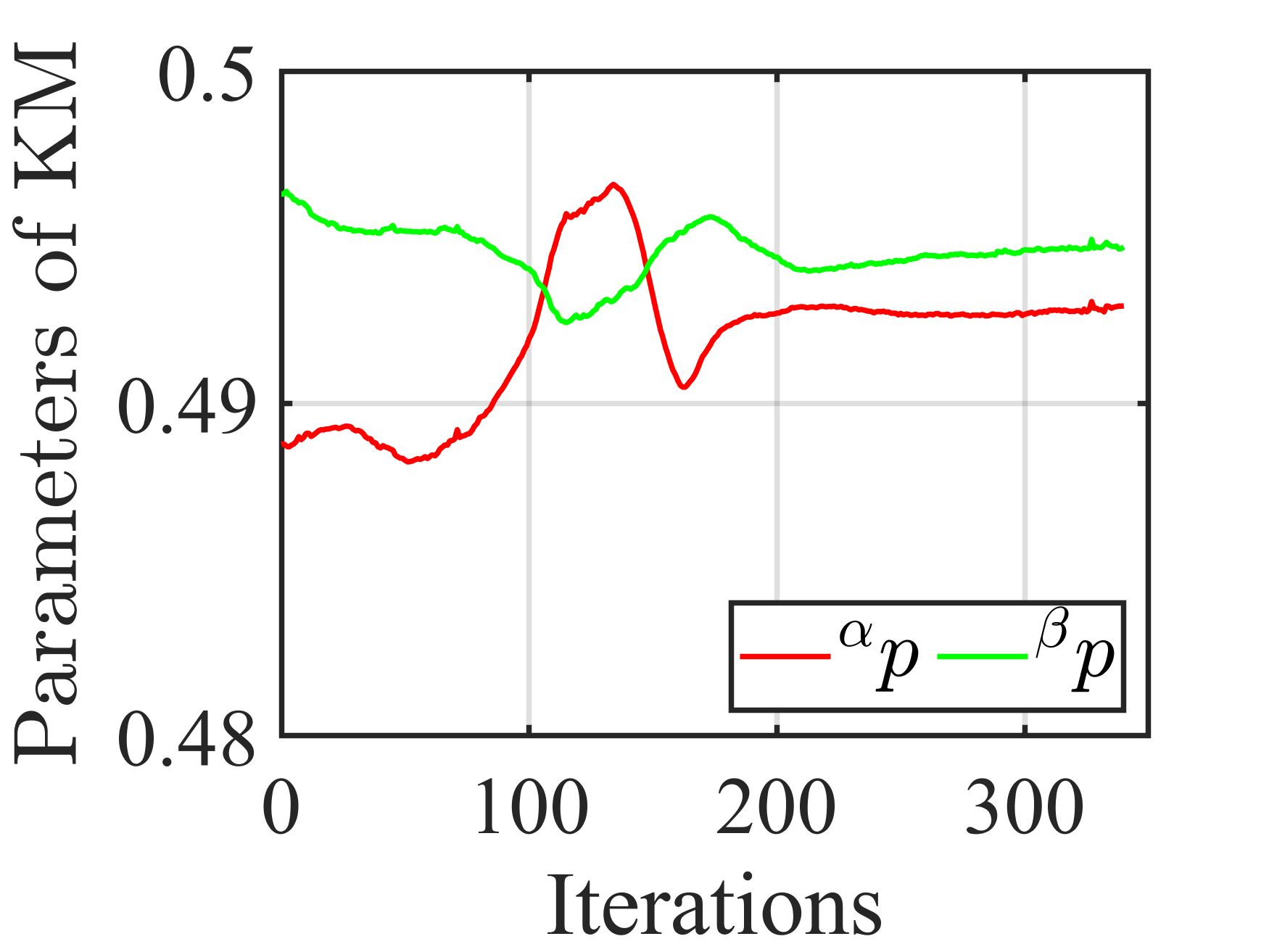}  \label{fig: Experiment_3D_KMVS_p}}				
		
		\caption{Results for KM-VS in Experiment \#6. (a) Errors on features.  (b) Camera velocities (in m/s and rad/s). (c) Order of  DOMs as visual features. (d) Parameters of KMs.}
		
		\label{fig: Experiment_3D_KMVS}
	\end{figure}
	
	\begin{figure}
		\centering 
		
		\subfloat[]{\includegraphics[width=0.33\hsize]{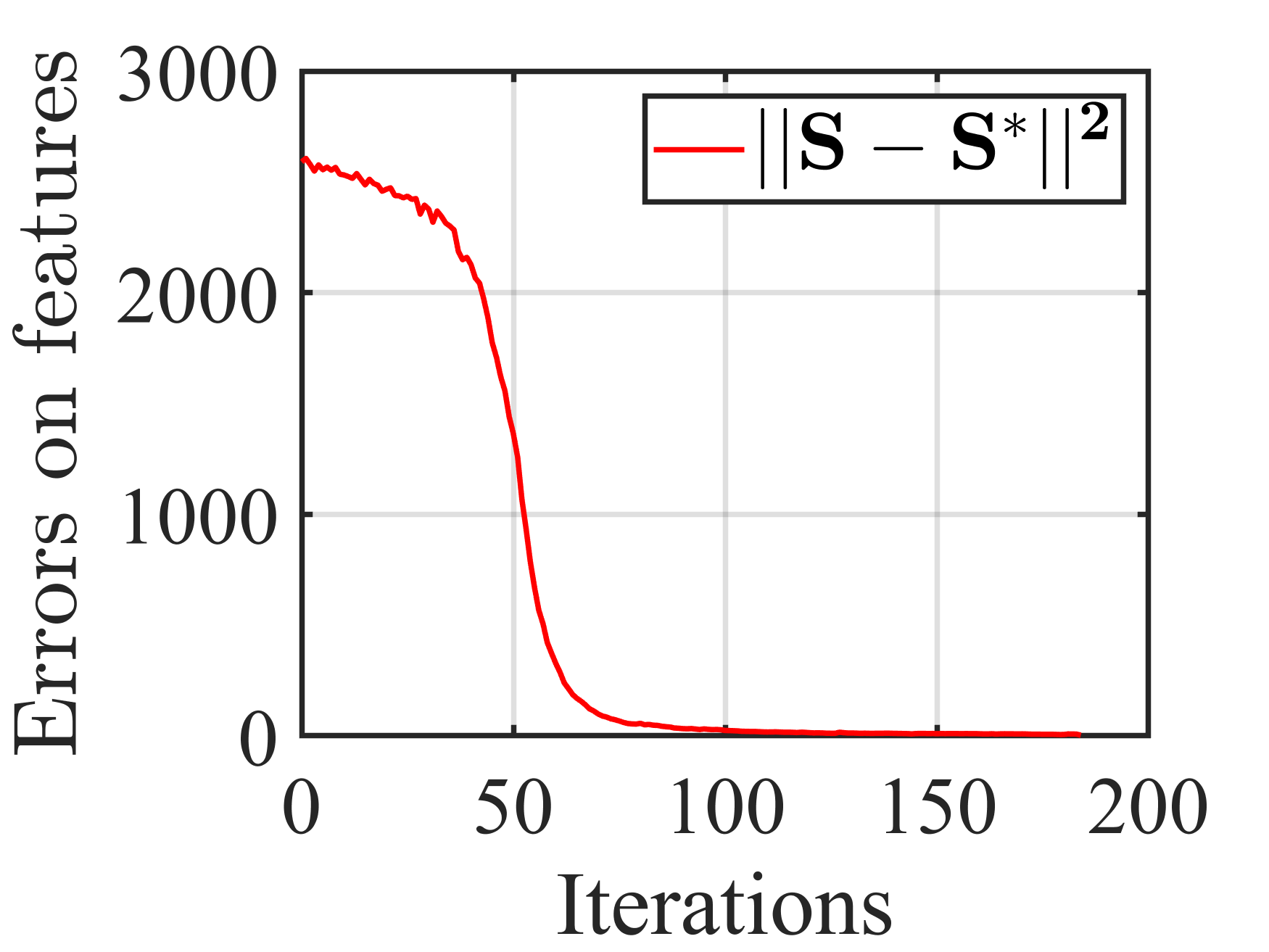}  \label{fig: Experiment_3D_HMVS_feature_error}}	\qquad
		\subfloat[]{\includegraphics[width=0.33\hsize]{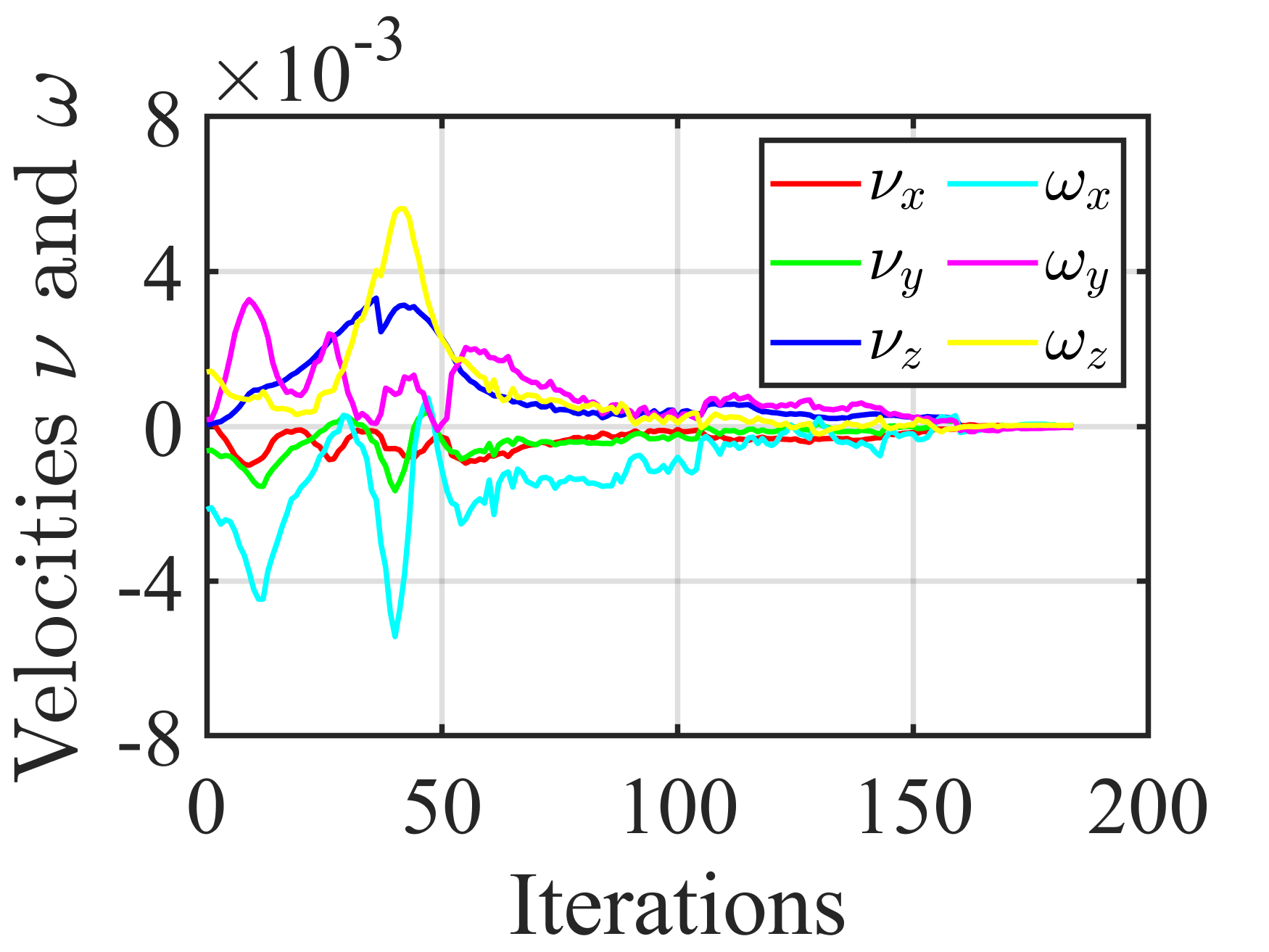}  \label{fig: Experiment_3D_HMVS_velocity}}		
		
		\subfloat[]{\includegraphics[width=0.33\hsize]{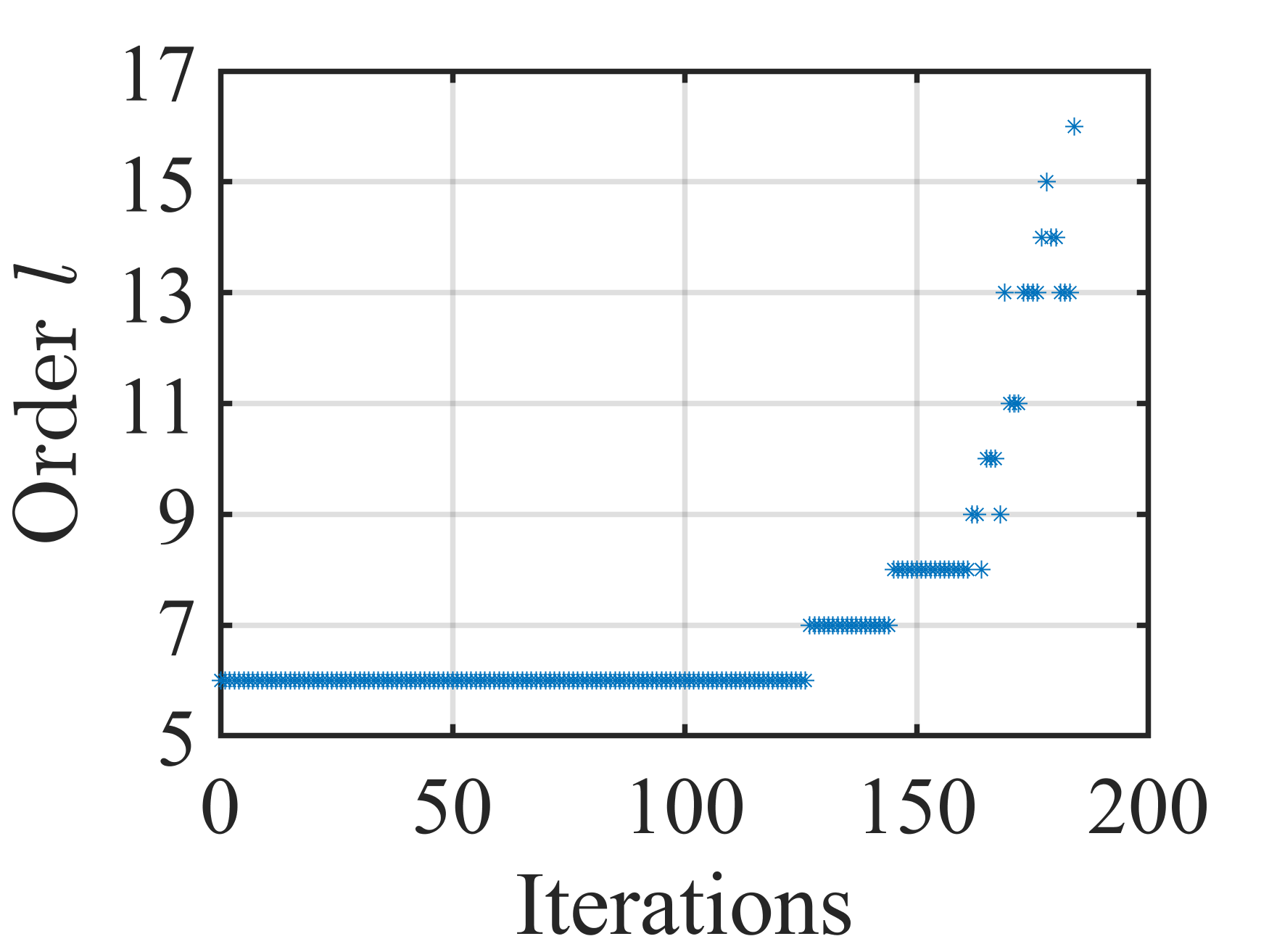}  \label{fig: Experiment_3D_HMVS_order}}		\qquad
		\subfloat[]{\includegraphics[width=0.33\hsize]{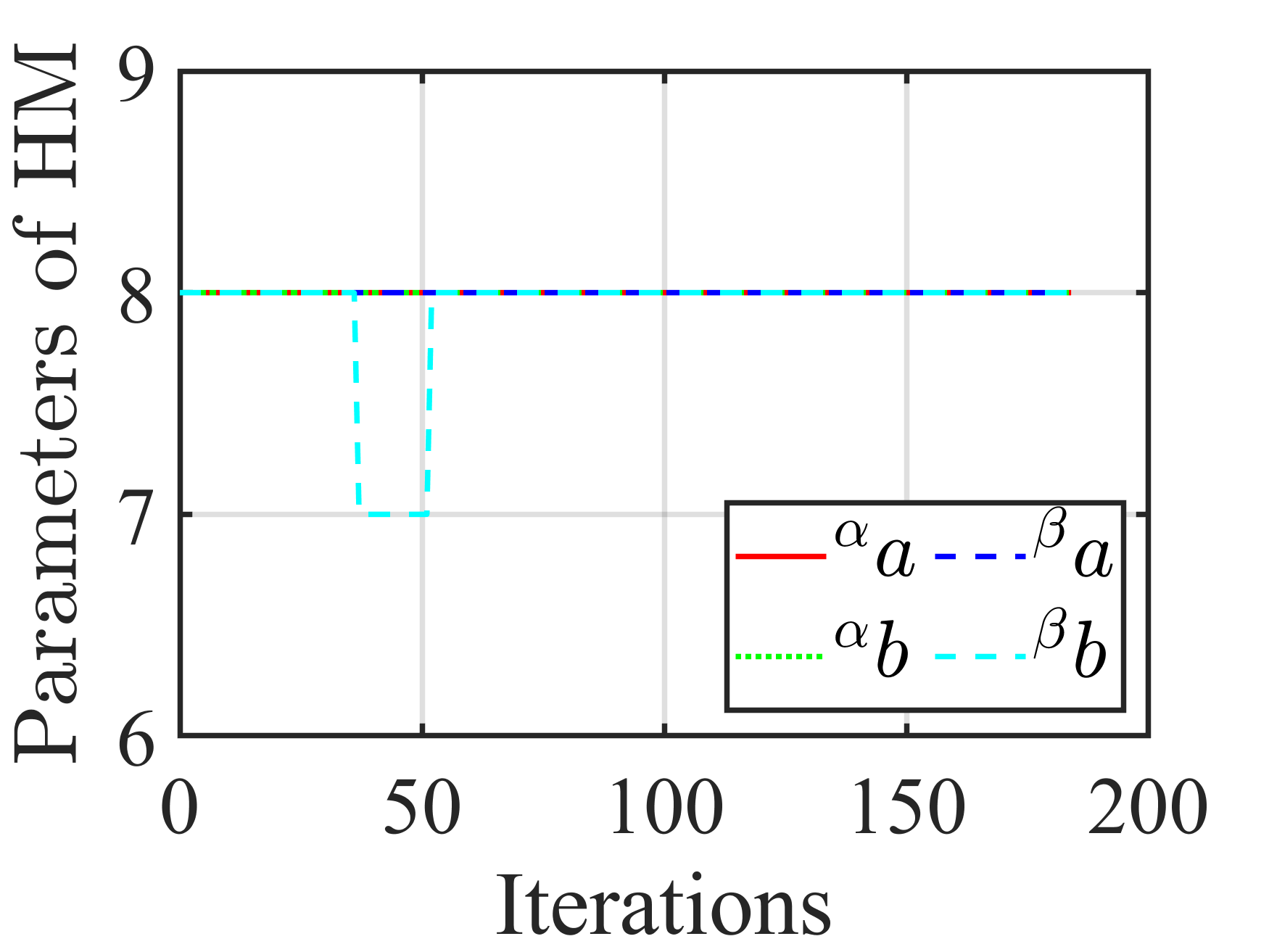}  \label{fig: Experiment_3D_HMVS_ab}}				
		
		\caption{Results for HM-VS in Experiment \#6. (a) Errors on features. (b) Camera velocities (in m/s and rad/s). (c) Order of  DOMs as visual features. (d) Parameters of HMs.}
		
		\label{fig: Experiment_3D_HMVS}
	\end{figure}
	
	\subsubsection*{Experiment \#7 (see Figs. \ref{fig: Experiment_3D_Complex_case} and \ref{fig: Experiment_3D_Complex_result})}
	
	\begin{figure}
		\centering 
		
		\includegraphics[width=0.49\hsize]{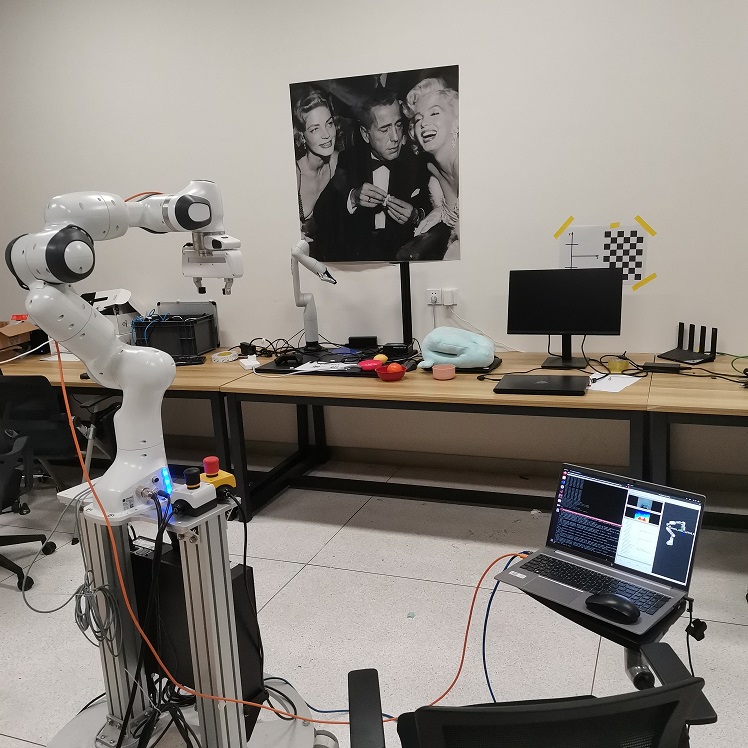}  
		
		\caption{The real complex 3-D experimental environment.}
		
		\label{fig: Experiment_3D_Complex_case}
	\end{figure}
	
	\begin{figure}
		\centering 
		
		\subfloat[]{\includegraphics[width=0.49\hsize]{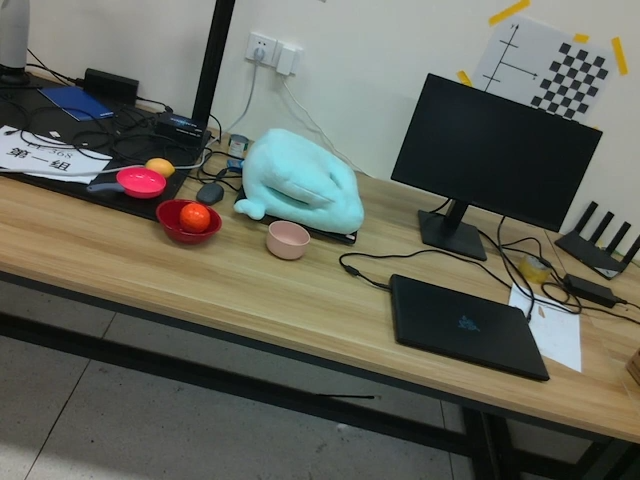}  \label{fig: Experiment_3D_Complex_image_new}}	
		\subfloat[]{\includegraphics[width=0.49\hsize]{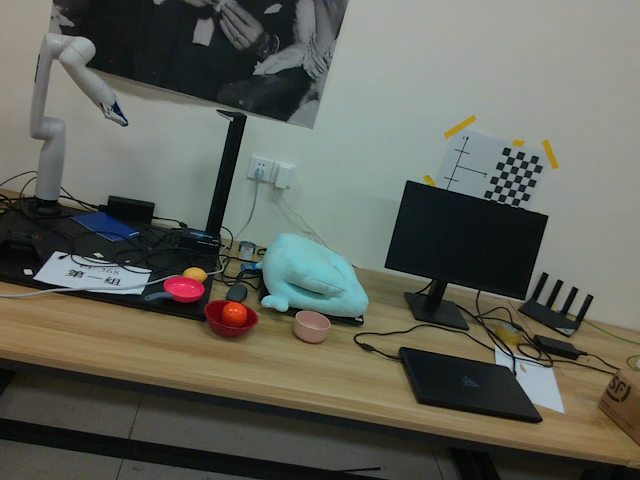}  \label{fig: Experiment_3D_Complex_image_old}}		
		
		\subfloat[]{\includegraphics[width=0.49\hsize]{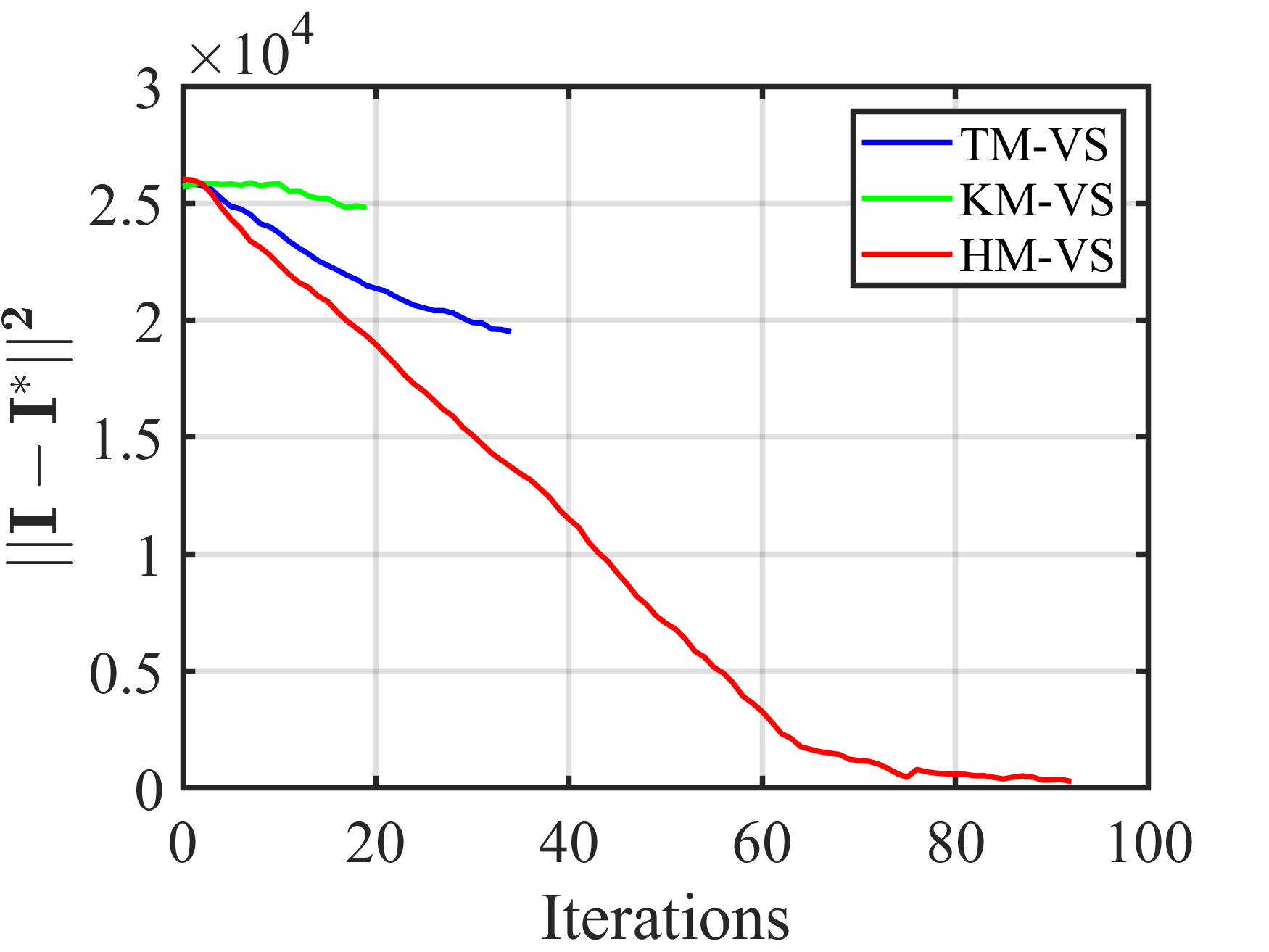}  \label{fig: Experiment_3D_Complex_I}}	
		\subfloat[]{\includegraphics[width=0.49\hsize]{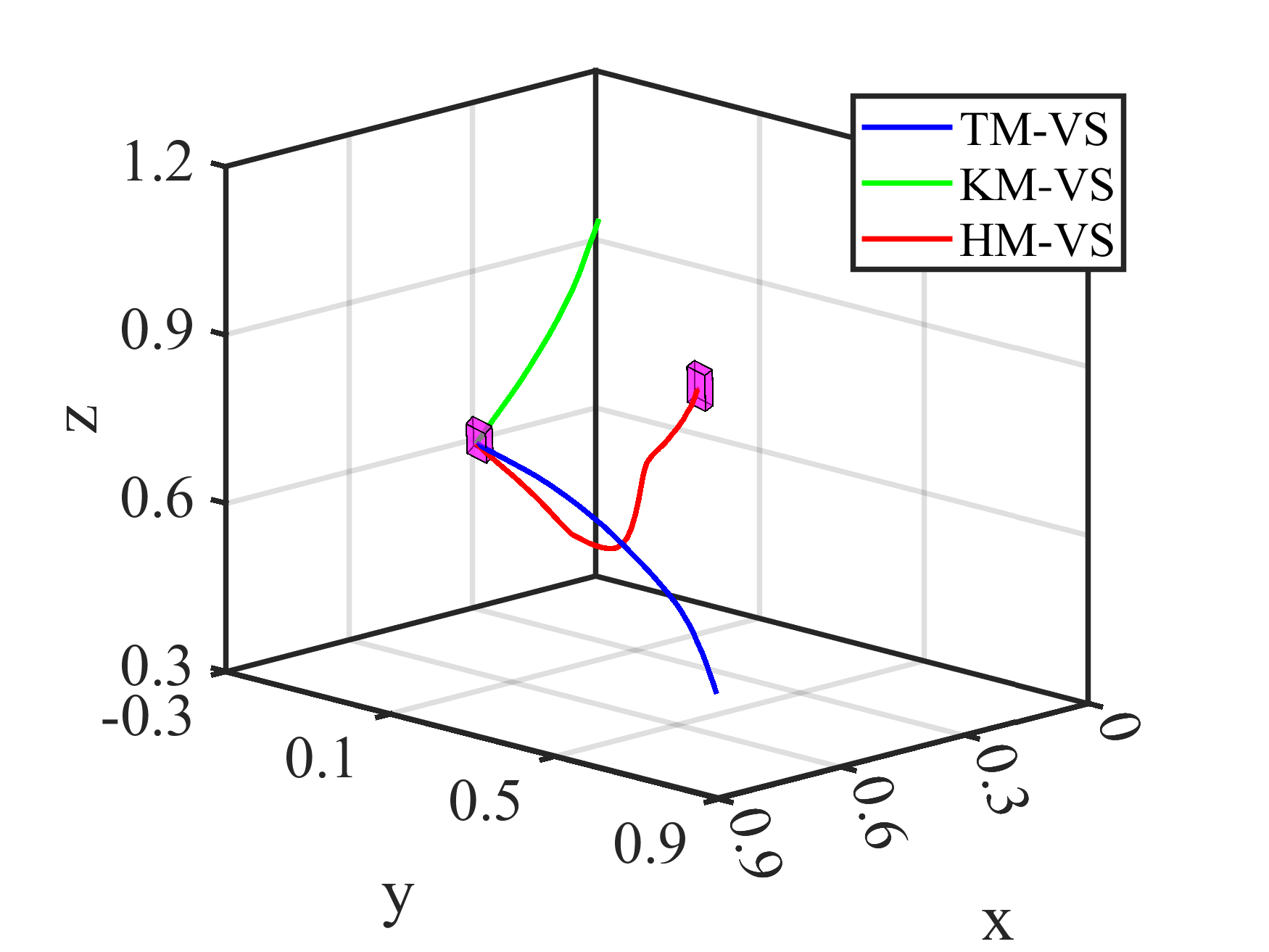}  \label{fig: Experiment_3D_Complex_trajections}}				
		
		\caption{Experiment \#7: Comparison between TM-VS, KM-VS, and HM-VS in a real complex 3-D environment. (a) Initial image. (b) Desired image. (c) Pixel errors. (d) Camera trajectories (in m).}
		
		\label{fig: Experiment_3D_Complex_result}
	\end{figure}
	
	\begin{figure}
		\centering 
		
		\subfloat[]{\includegraphics[width=0.33\hsize]{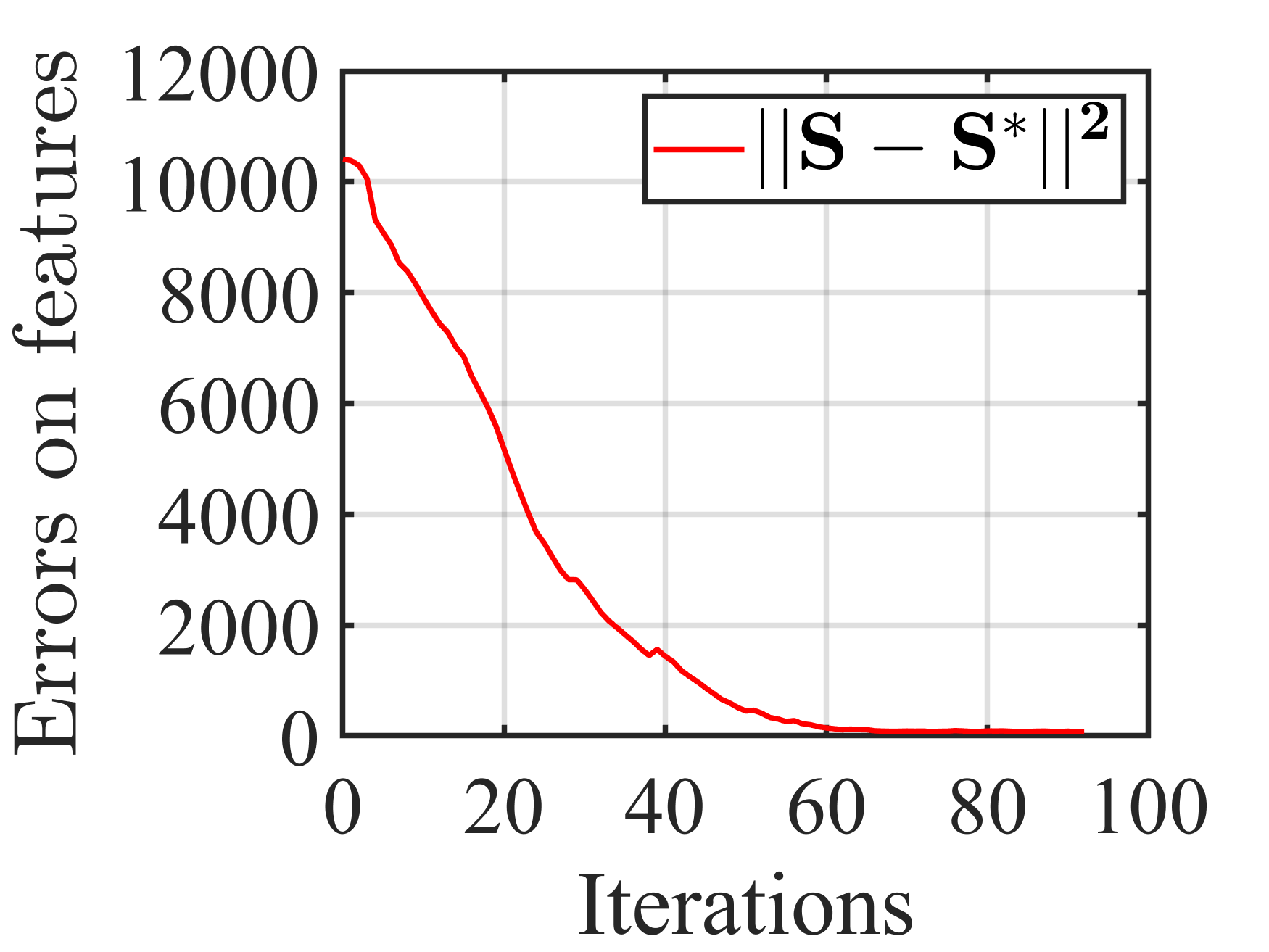}  \label{fig: Experiment_3D_Complex_HMVS_feature_error}}	\qquad
		\subfloat[]{\includegraphics[width=0.33\hsize]{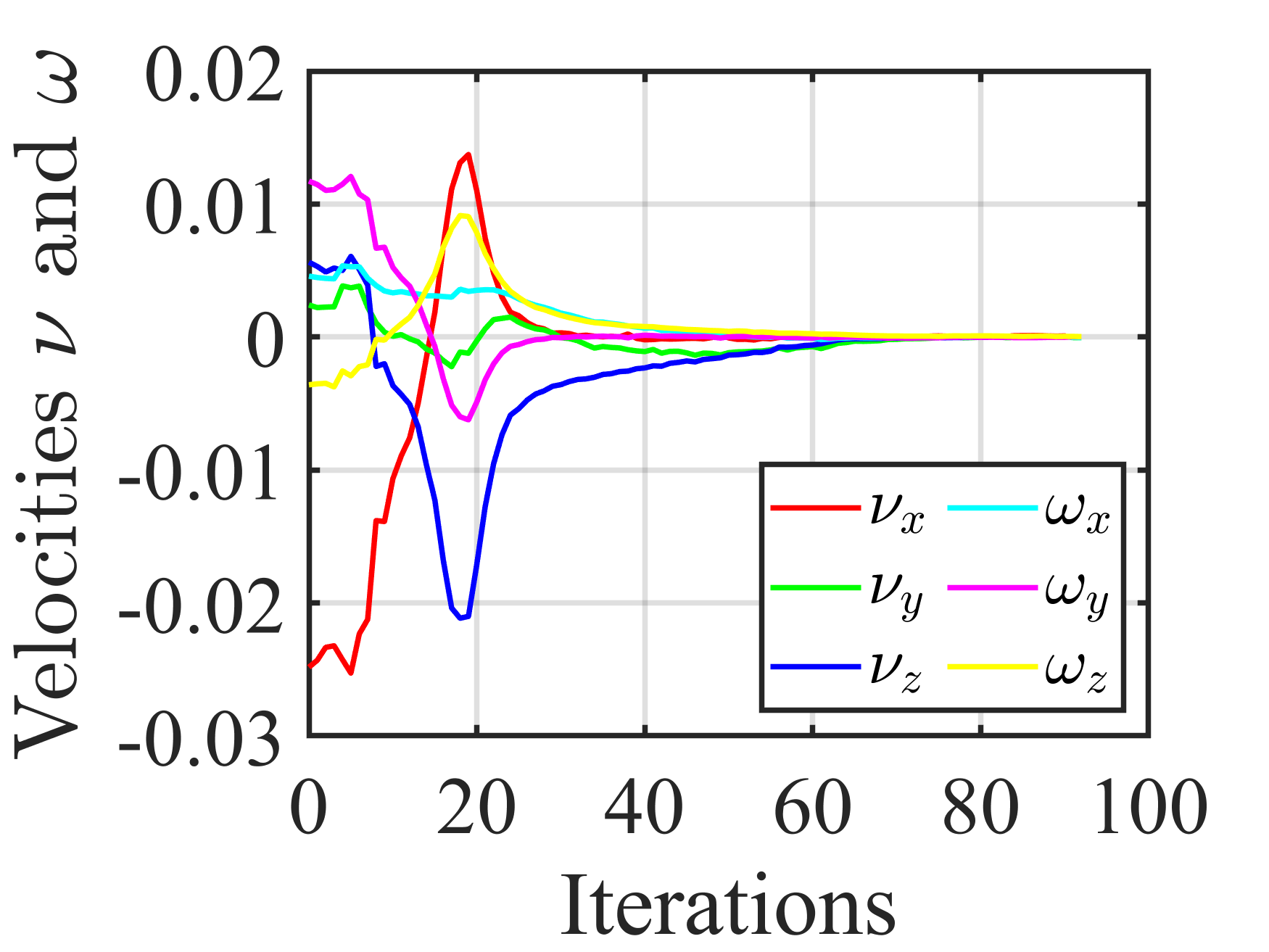}  \label{fig: Experiment_3D_Complex_HMVS_velocity}}		
		
		\subfloat[]{\includegraphics[width=0.33\hsize]{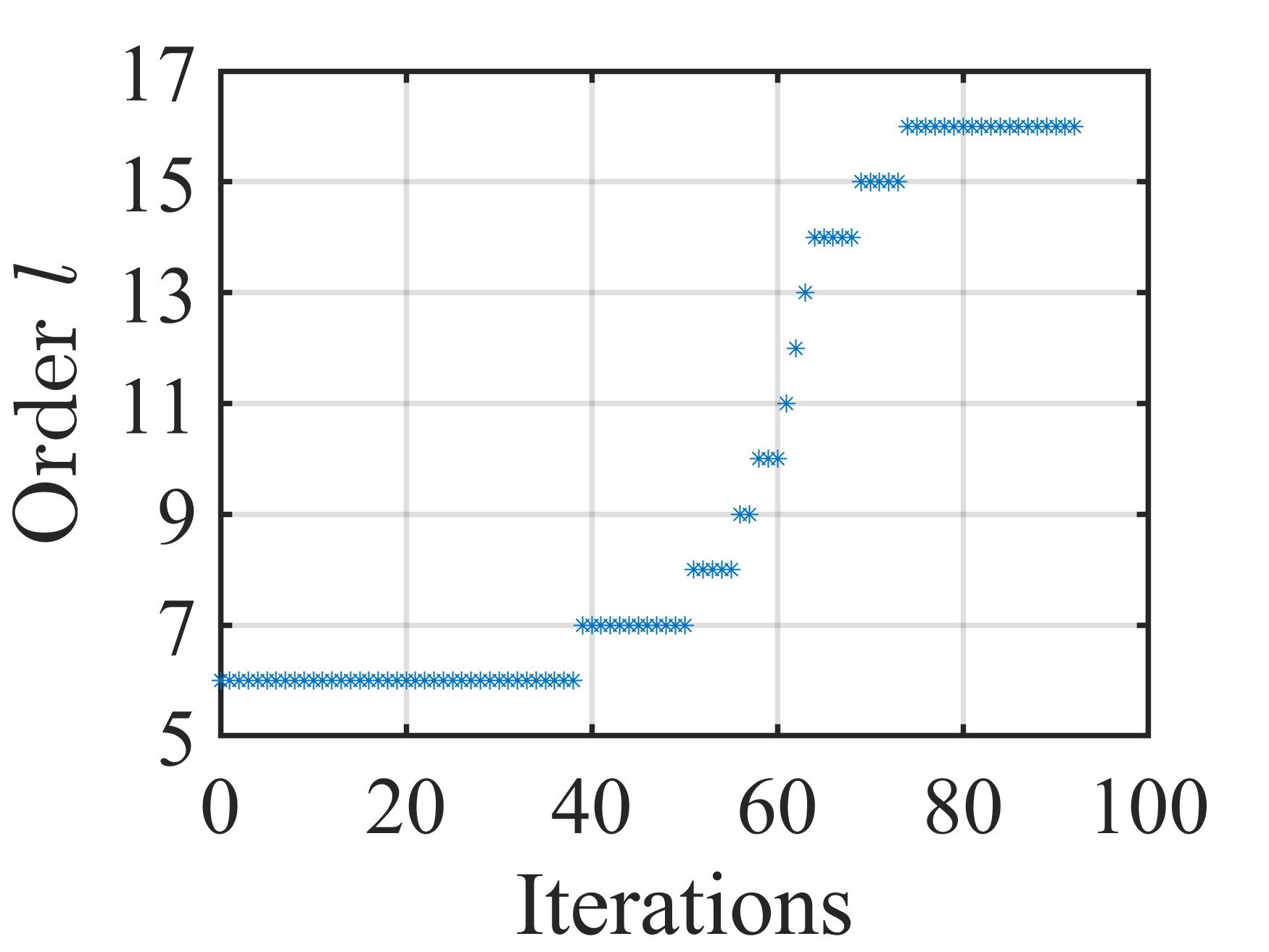}  \label{fig: Experiment_3D_Complex_HMVS_order}}		\qquad
		\subfloat[]{\includegraphics[width=0.33\hsize]{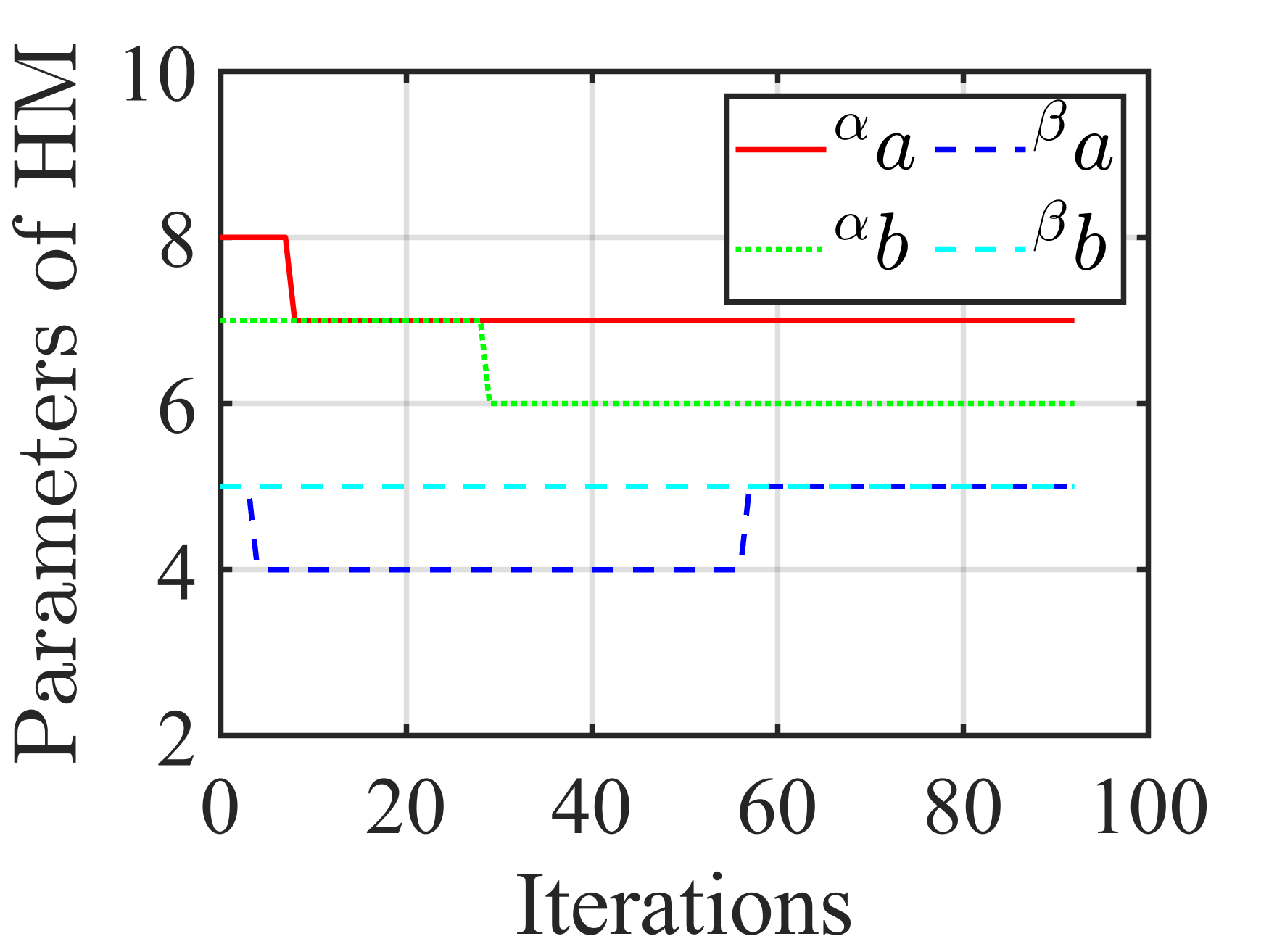}  \label{fig: Experiment_3D_Complex_HMVS_ab}}				
		
		\caption{Results for HM-VS in Experiment \#7. (a) Errors on features. (b) Camera velocities (in m/s and rad/s). (c) Order of  DOMs as visual features. (d) Parameters of HMs.}
		
		\label{fig: Experiment_3D_Complex_HMVS}
	\end{figure}
	
	This last experiment was performed in a real complex 3-D environment containing objects of various shapes and different colors, as can be seen in Fig. \ref{fig: Experiment_3D_Complex_case}.
	The corresponding displacement between the initial and the desired camera poses is given by ($0.23\text{m}$, $0.01\text{m}$, $0.32\text{m}$, $-8.65^{\circ}$, $-5.94^{\circ}$, $-5.41^{\circ}$).
	The visual difference between the initial and desired images is also a challenge (see Figs. \ref{fig: Experiment_3D_Complex_image_new} and \ref{fig: Experiment_3D_Complex_image_old}).
	The pixel errors $||\mathbf{I} - \mathbf{I}^*||^2$ and the camera trajectories obtained from these three methods (TM-VS, KM-VS, and HM-VS)  are shown in Figs. \ref{fig: Experiment_3D_Complex_I} and \ref{fig: Experiment_3D_Complex_trajections}, respectively.
	Both methods, TM-VS and KM-VS, fail because the camera reaches the maximum workspace of the robotic arm.
	Only the HM-VS method can successfully converge the pose error to less than $(1\text{mm}$, $1\text{mm}$, $1\text{mm}$,  $0.5^{\circ}$, $0.5^{\circ}$, $0.5^{\circ})$.		
	The control results of the HM-VS method are illustrated in Fig. \ref{fig: Experiment_3D_Complex_HMVS}.
	Additionally, we repeat the above experiment, with the difference that the lighting changes were introduced during the VS process (see Fig. \ref{fig: Experiment_3D_Complex_Change_HMVS_Process}).
	Specifically, we introduce four lighting changes in the 6th, 19th, 30th, and 51st iterations, which explain why the large perturbations in the results for the HM-VS method (see Fig. \ref{fig: Experiment_3D_Complex_Change_HMVS}).
	Finally, the HM-VS method still converges and remains stable.
	
	\begin{figure}
		\centering 
		
		\subfloat[]{\includegraphics[width=0.23\hsize]{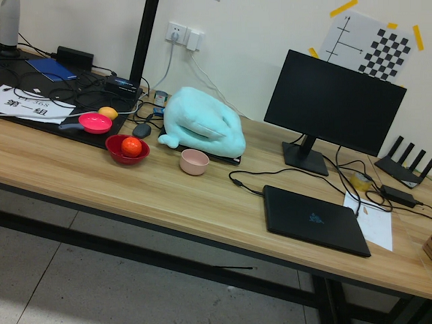}  \label{fig: Experiment_3D_Complex_Change_1}}		
		\subfloat[]{\includegraphics[width=0.23\hsize]{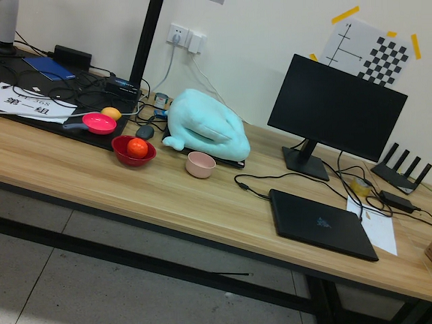}  \label{fig: Experiment_3D_Complex_Change_3}}		\subfloat[]{\includegraphics[width=0.23\hsize]{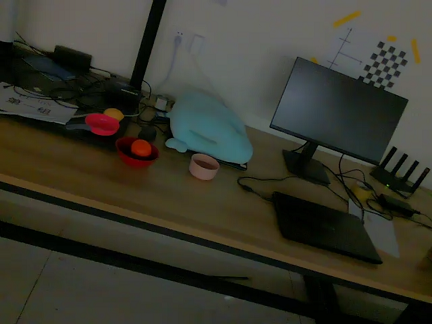}  \label{fig: Experiment_3D_Complex_Change_6}}
		\subfloat[]{\includegraphics[width=0.23\hsize]{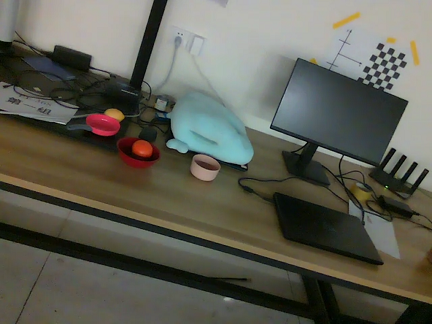}  \label{fig: Experiment_3D_Complex_Change_8}}	
		
		\subfloat[]{\includegraphics[width=0.23\hsize]{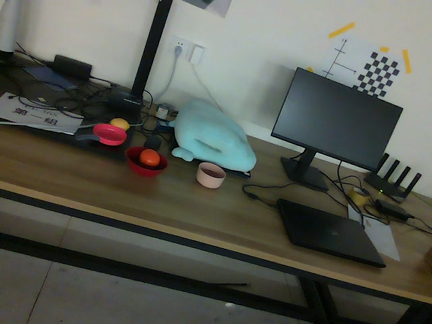}  \label{fig: Experiment_3D_Complex_Change_13}}	
		\subfloat[]{\includegraphics[width=0.23\hsize]{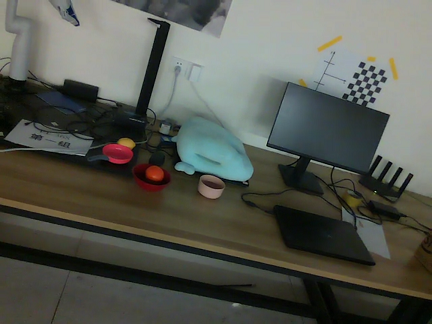}  \label{fig: Experiment_3D_Complex_Change_19}}	
		\subfloat[]{\includegraphics[width=0.23\hsize]{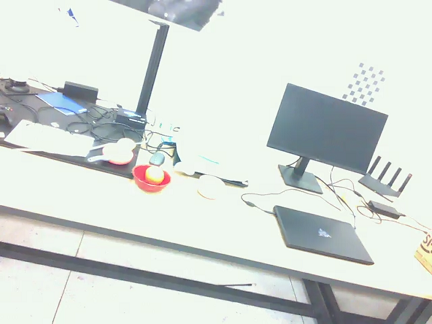}  \label{fig: Experiment_3D_Complex_Change_20}}	
		\subfloat[]{\includegraphics[width=0.23\hsize]{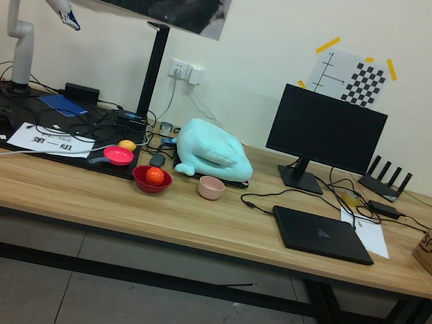}  \label{fig: Experiment_3D_Complex_Change_21}}	
		
		\subfloat[]{\includegraphics[width=0.23\hsize]{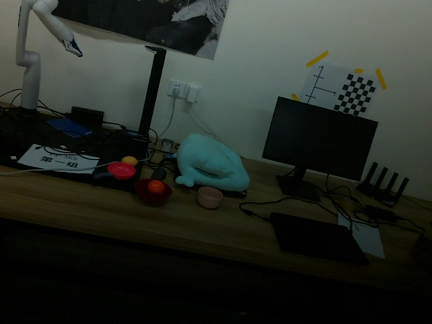}  \label{fig: Experiment_3D_Complex_Change_30}}		\subfloat[]{\includegraphics[width=0.23\hsize]{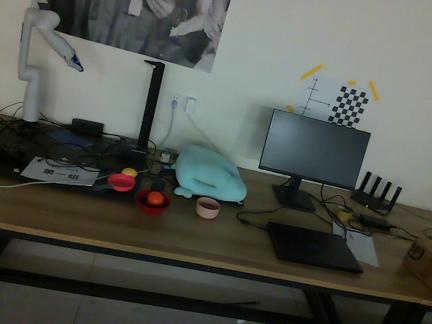}  \label{fig: Experiment_3D_Complex_Change_40}}		
		\subfloat[]{\includegraphics[width=0.23\hsize]{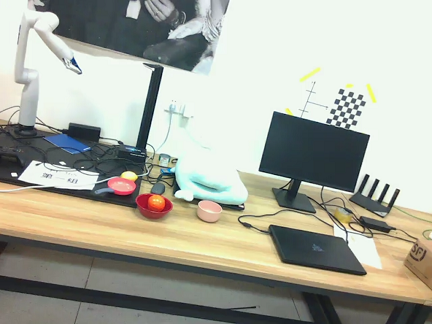}  \label{fig: Experiment_3D_Complex_Change_51}}	
		\subfloat[]{\includegraphics[width=0.23\hsize]{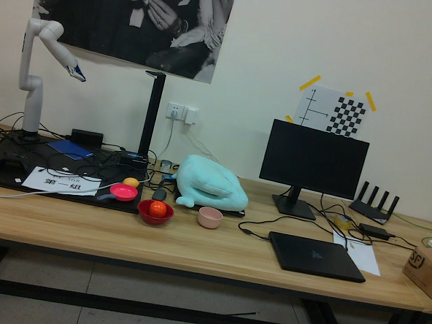}  \label{fig: Experiment_3D_Complex_Change_97}}		
		
		\caption{The VS process for HM-VS under the changing light condition in Experiment \#7. (a-l) 1st, 3rd, 6th, 8th, 13th, 19th, 20th, 21st, 30th, 40th, 51st, and 97th iterations.}
		
		\label{fig: Experiment_3D_Complex_Change_HMVS_Process}
	\end{figure}

	\begin{figure}
		\centering 
		
		\subfloat[]{\includegraphics[width=0.33\hsize]{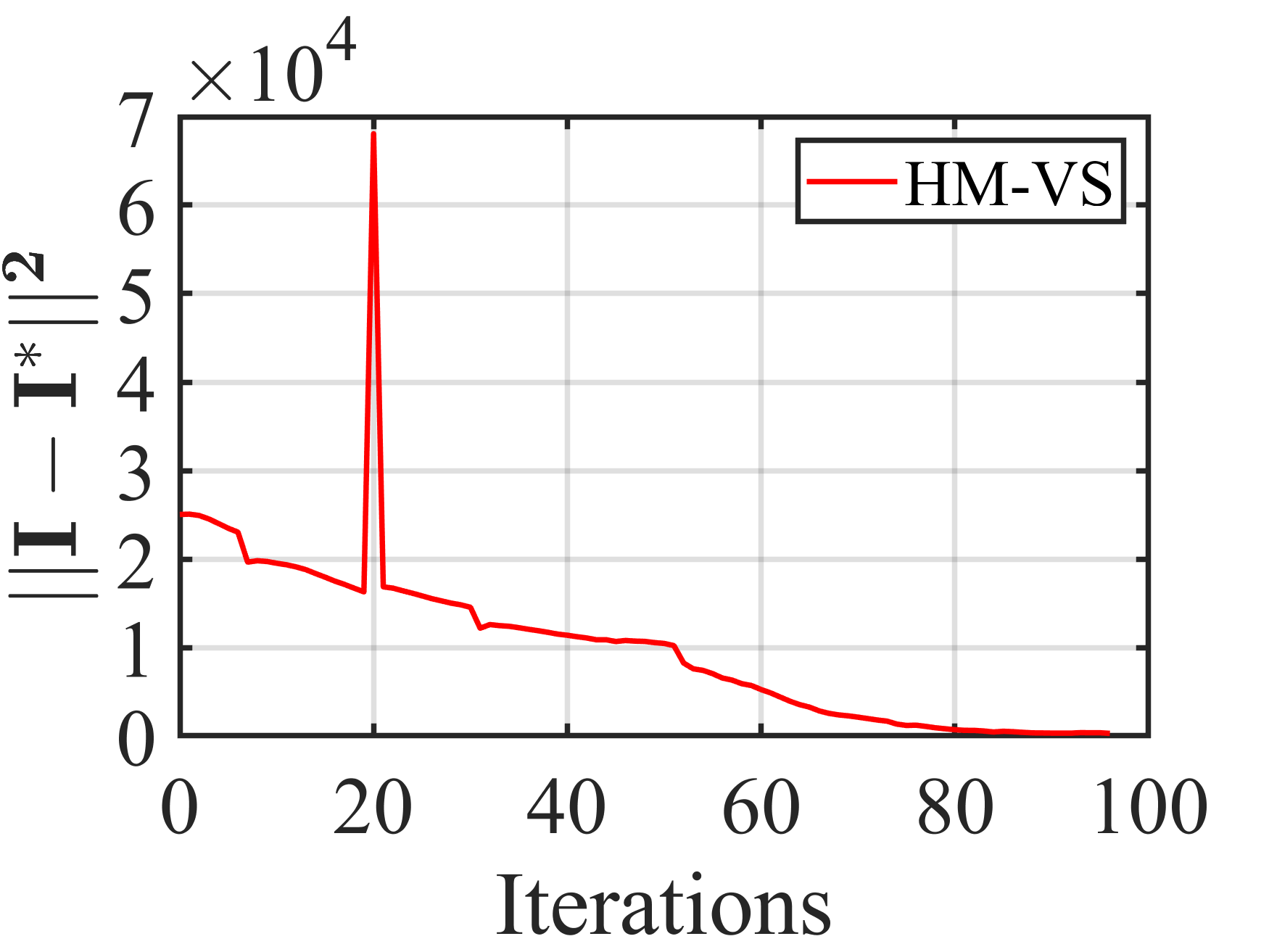}  \label{fig: Experiment_3D_Complex_Change_I}}	
		\subfloat[]{\includegraphics[width=0.33\hsize]{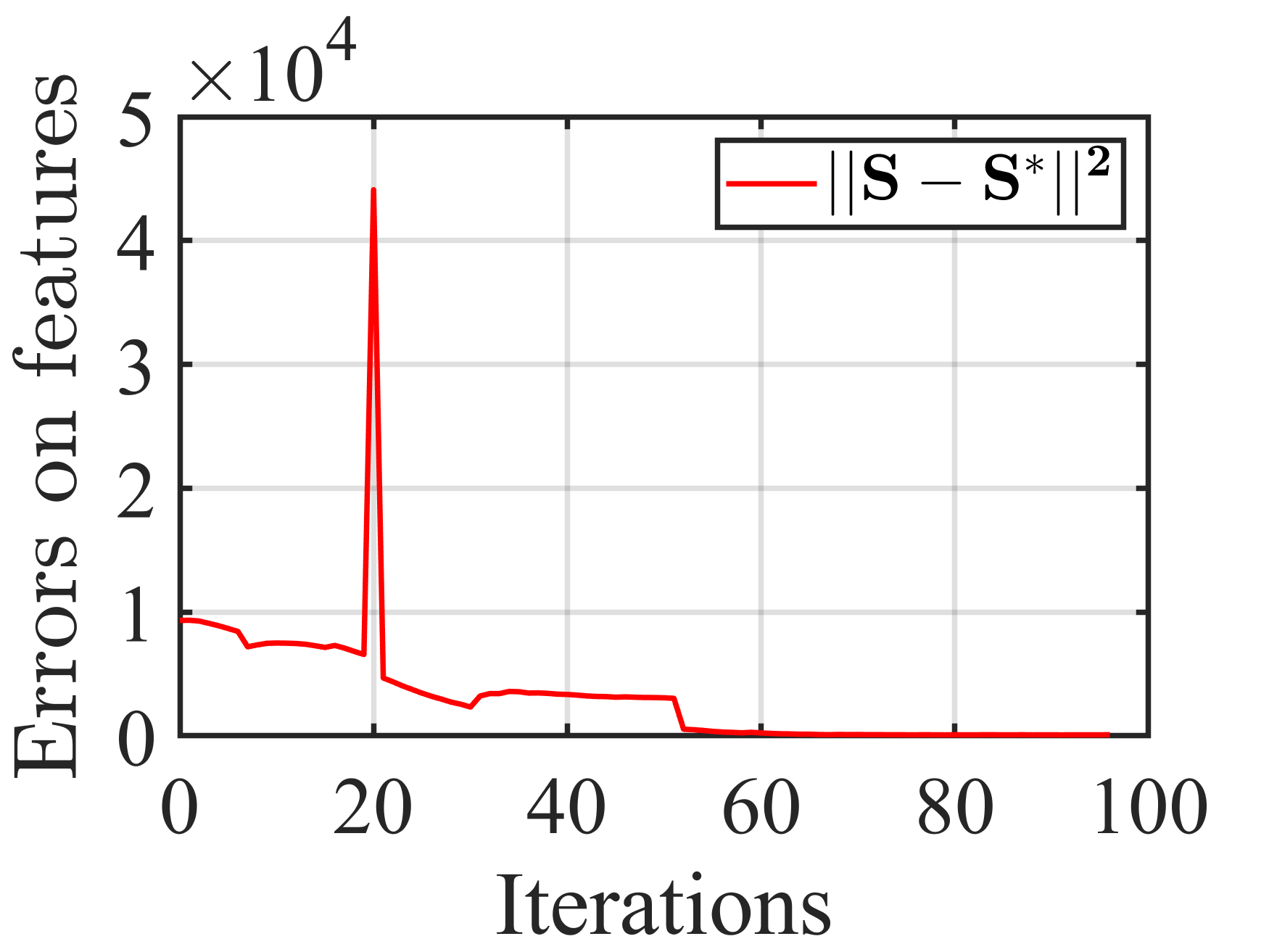}  \label{fig: Experiment_3D_Complex_Change_HMVS_feature_error}}				
		\subfloat[]{\includegraphics[width=0.33\hsize]{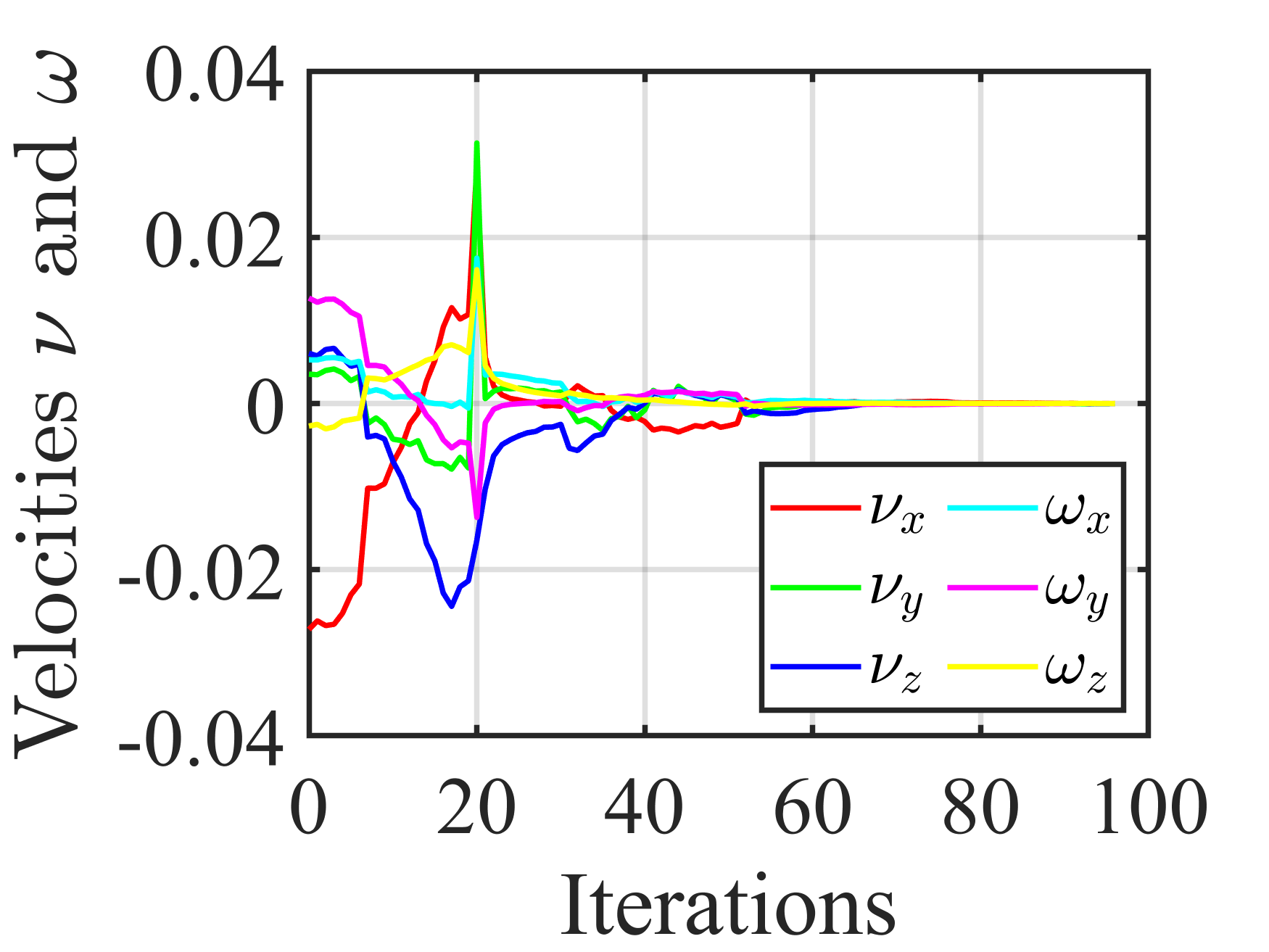}  \label{fig: Experiment_3D_Complex_Change_HMVS_velocity}}		
		
		\caption{Results for HM-VS under the changing light condition in Experiment \#7. (a) Pixel errors. (b) Errors on features. (c) Camera velocities (in m/s and rad/s).}
		
		\label{fig: Experiment_3D_Complex_Change_HMVS}
	\end{figure}

	Overall, convergence and stability are still achieved despite VS in the real environment, which validates the efficacy of our method.

	\subsection{Discussion} \label{sec: Discussion}
	The previous experiments have demonstrated the effectiveness of our method even in complex environments.
	However, as explained in Section \ref{sec: order}, two parameters are involved in our approach: the minimum order of DOMs ($l_{\text{min}}$) and the maximum order of DOMs ($l_{\text{max}}$).
	In general, the choice of these parameters depends on the convergence rate, the convergence error, and many other factors, which is still an open question.
	This section, therefore, discusses qualitatively the effect of the DOM order on the VS.
	
	We add a Gaussian noise $\sigma^2=0.4$ to the image and then perform the HM-VS with $l = 3, 6, 9, 12, 15$, respectively.
	The results are presented in Fig. \ref{fig: Discussion}.
	The initial and desired images are shown in Figs. \ref{fig: Discussion_image_new_noise_4} and \ref{fig: Discussion_image_old_noise_4}, respectively.
	It is clear from Fig. \ref{fig: Discussion_trajectories} that HM-VS can perform the task with different $l$.
	The simple relationship between the DOM order, the convergence rate, and the convergence error can be obtained from the position and orientation errors (see Figs. \ref{fig: Discussion_Position_error} and \ref{fig: Discussion_orientation_error}):
	\begin{itemize}
		\item{for the convergence rate, the smaller the DOM order, the faster the convergence;}
		\item{for convergence errors, the higher the order, the better the accuracy.}
	\end{itemize}
	In addition, we found that VS can fail due to falling into local minima when the order $l$ is too large.
	
	We may therefore give the following advice for the choice of $l_{\text{min}}$ and $l_{\text{max}}$:
	\begin{itemize}
		\item{$l_{\text{min}}$ should be as small as possible in the case of convergence;}
		\item{$l_{\text{max}}$ should be as large as possible in the case of a suitable convergence rate.}
	\end{itemize}

	\begin{figure}
		\centering 
		
		\subfloat[]{\includegraphics[width=0.45\hsize]{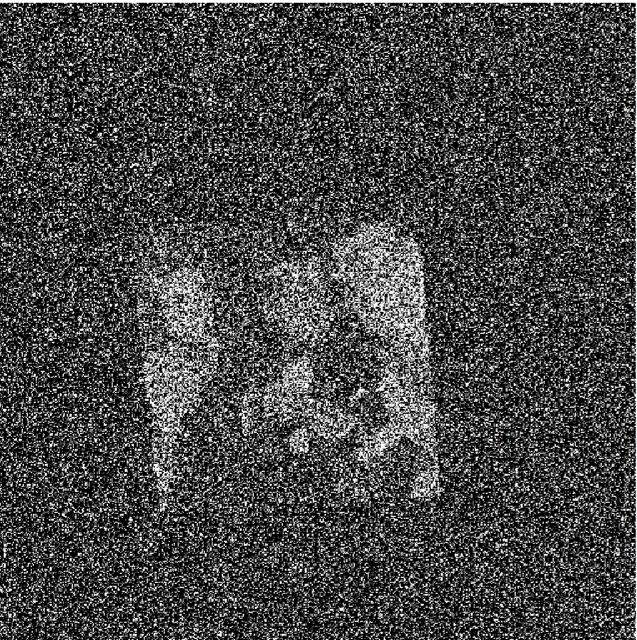}  \label{fig: Discussion_image_new_noise_4}}	
		\subfloat[]{\includegraphics[width=0.45\hsize]{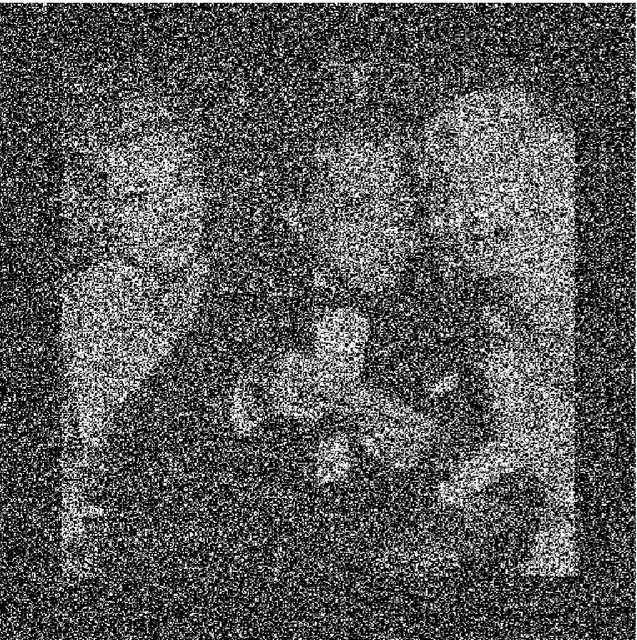}  \label{fig: Discussion_image_old_noise_4}}
		
		\subfloat[]{\includegraphics[width=0.49\hsize]{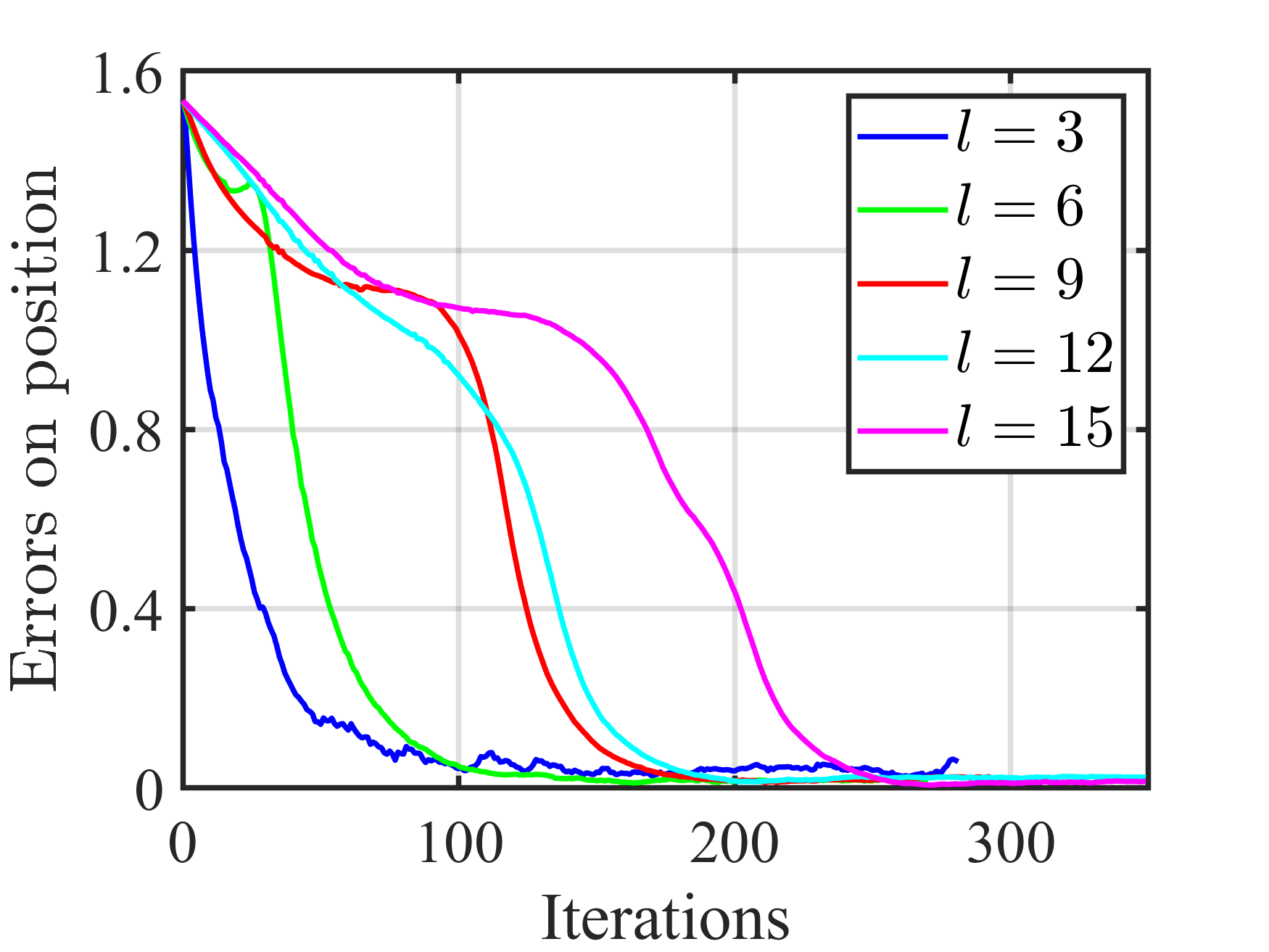}  \label{fig: Discussion_Position_error}}	
		\subfloat[]{\includegraphics[width=0.49\hsize]{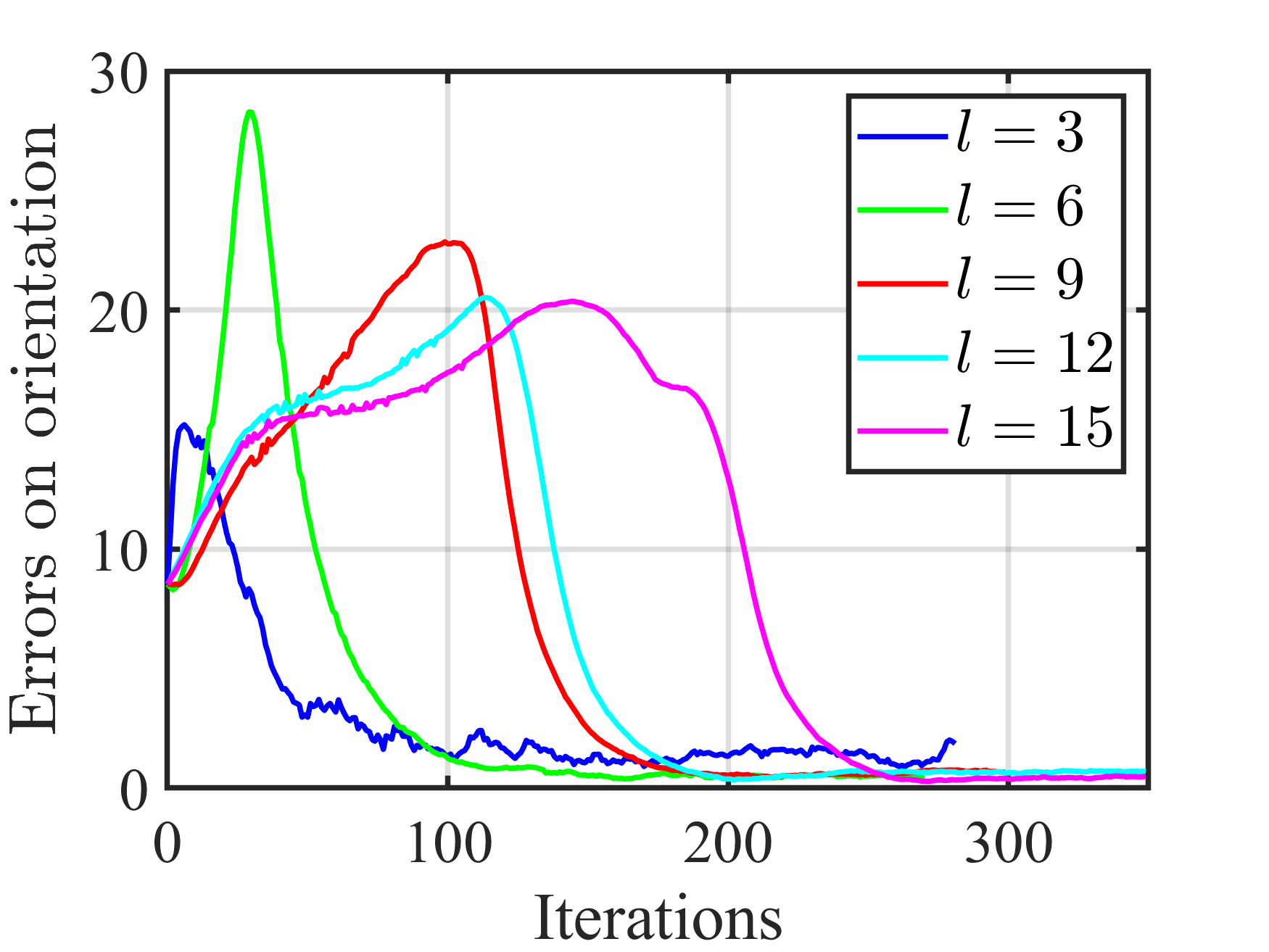}  \label{fig: Discussion_orientation_error}}		
		
		\subfloat[]{\includegraphics[width=0.49\hsize]{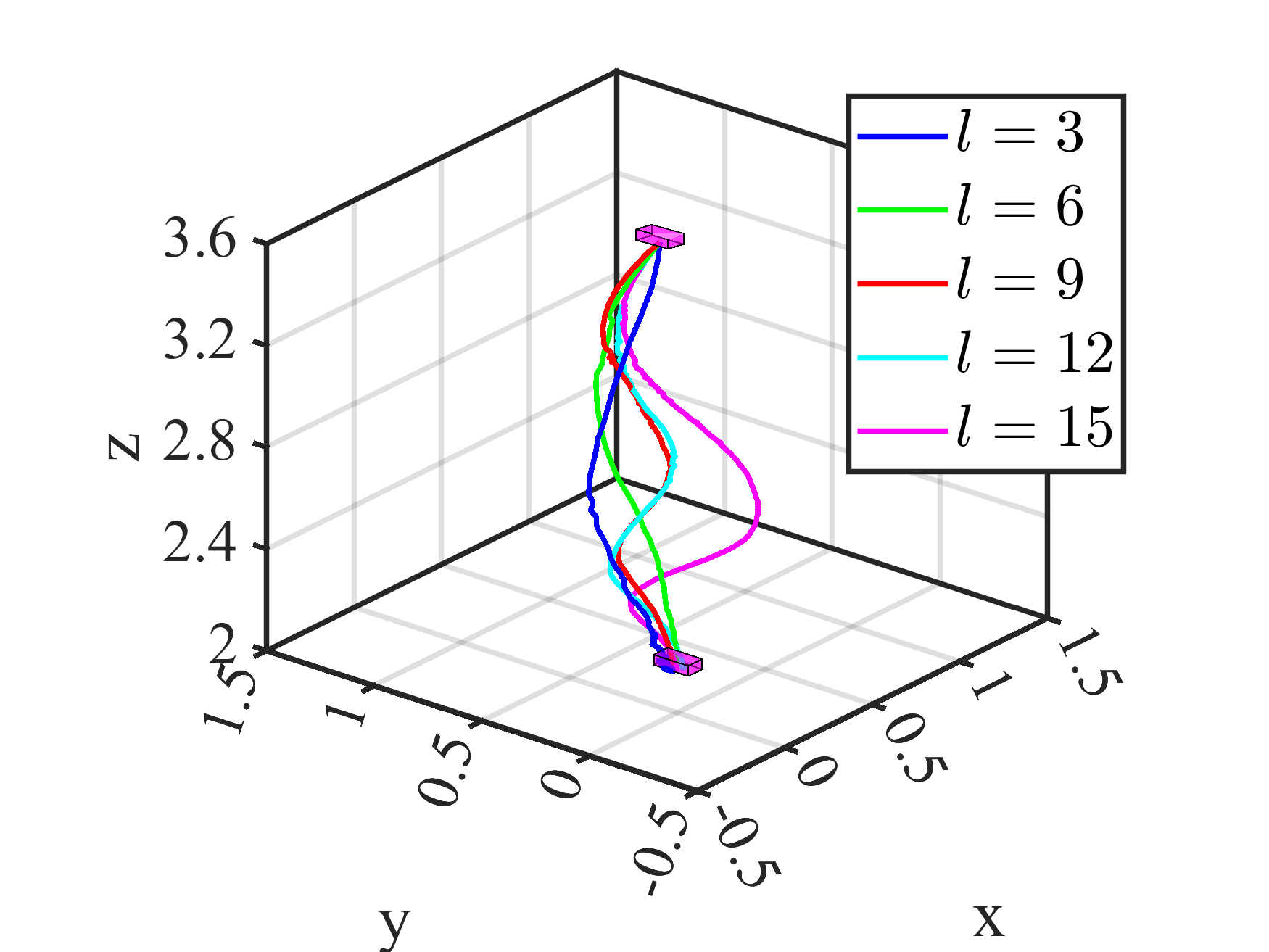}  \label{fig: Discussion_trajectories}}					
		
		\caption{Influence of  the DOM order in case of Gaussian noise ($\sigma^2 = 0.4$). (a) Initial image. (b) Desired image. (c) Position  errors (in m). (d) Orientation errors (in $^{\circ}$). (e) Camera trajectories (in m).}
		\label{fig: Discussion}
	\end{figure}

	\section{Conclusion} \label{sec: Conclusion}
	In this paper, for the first time, we proposed a generic framework to consider DOMs as visual features for DVS.
	Moreover, it was shown that the interaction matrix related to DOMs can be calculated explicitly.
	Taking TMs, KMs, and HMs as examples, three DVS schemes, TM-VS, KM-VS, and HM-VS, are proposed, and adaptive estimation methods for the associated parameters are also introduced.
	The experimental results indicated that our proposed control schemes are effective and robust for VS of both 2-D and 3-D objects. 
	This is due to the image compression and filtering properties of the DOM.
	Note that the HM-VS method outperforms state-of-the-art methods regarding convergence rate and robustness.
	
	Future work will be devoted to designing combinations of DOMs as visual features that can be used to control the camera's trajectory in Cartesian space.
	Additionally, we intend to investigate more flexible DOMs that can be used as visual features, such as Racah moments, etc.
	It is worth noting that the DOM-VS method is relatively time-consuming. 
	The main reason is that discrete orthogonal polynomials can only be computed by recurrence methods.
	Therefore, improving the computational efficiency of the proposed method is also our future work.

	\section*{Acknowledgments}
	Thanks to Shuo Wang for providing language help and all our colleagues for providing all types of help during the preparation of this manuscript.

	\bibliographystyle{References/IEEEtran}
	\bibliography{References/reference}

	\section{Biography Section}
	\begin{IEEEbiography}[{\includegraphics[width=1in,height=1.25in,clip,keepaspectratio]{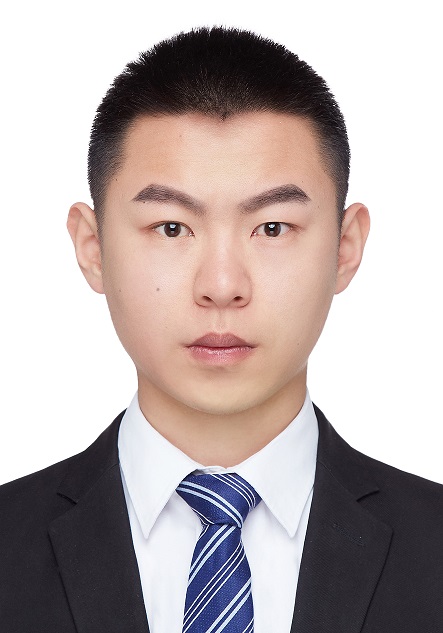}}]{Yuhan Chen}
		received the Ph.D. degree in mechanical engineering from the Beijing Institute of Technology, Beijing, China, in 2022 and the B.S. degree in mechanical engineering from the Taiyuan University of Technology, Taiyuan, China, in 2016. His research interests include visual servoing, robotics, mobile manipulation, and robot dynamics.
	\end{IEEEbiography}
	\begin{IEEEbiography}[{\includegraphics[width=1in,height=1.25in,clip,keepaspectratio]{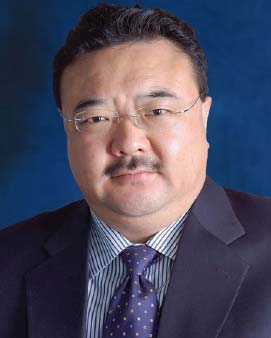}}]{Max Q.-H. Meng}
		(Fellow, IEEE) received the Ph.D. degree in electrical and computer engineering from the University of Victoria, Victoria, BC, Canada,
		in 1992.
		
		He is with Shenzhen Key Laboratory of Robotics Perception and Intelligence and the Department of Electronic and Electrical Engineering at Southern University of Science and Technology in Shenzhen, China. He is a Professor Emeritus in the Department of Electronic Engineering at The Chinese University of Hong Kong in Hong Kong and was a Professor in the Department of Electrical and Computer Engineering at the University of Alberta in Canada.		
		His research interests include robotics, medical robotics and devices, perception, and scenario intelligence.
		
		Dr. Meng is a recipient of the IEEE Millennium Medal. He has served as an editor for several journals and also as the General and Program Chair for many conferences, including the General Chair of IROS 2005 and the General Chair of ICRA 2021. He is an Elected Member of the Administrative Committee of the IEEE Robotics and Automation Society. He is a Fellow of the Canadian Academy of Engineering and HKIE.
	\end{IEEEbiography}
	
	\begin{IEEEbiography}[{\includegraphics[width=1in,height=1.25in,clip,keepaspectratio]{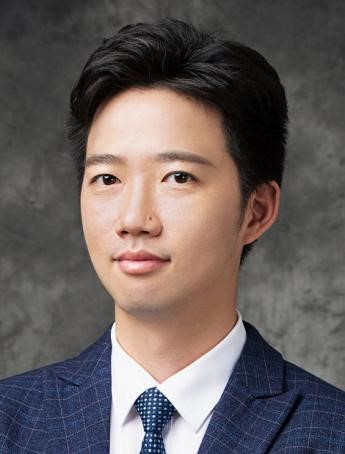}}]{Li Liu}
		received his Ph.D. degree in Biomedical Engineering from the University of Bern, Switzerland, and then worked as a postdoctoral fellow at the University of Bern and the Chinese University of Hong Kong. He held the position of assistant professor in the School of Biomedical Engineering at Shenzhen University since 2016. In 2019 he joined CUHK as a faculty member. He joined the Southern University of Science and Technology as a research associate professor in 2023. He has served as Program and Publication Chair of many international conferences including Publication Chair of IEEE ICIA 2017, 2018 and Program Chair of ROBIO 2019, Video Chair of IEEE ICRA 2021. He served as Associate Editor of Biomimetic Intelligence and Robotics (BIROB) since 2021. He is a recipient of the Distinguished Doctorate Dissertation Award, Swiss Institute of Computer Assisted Surgery (2016), the MICCAI Student Travel Award (2014), and the Best Paper Award of IEEE ICIA (2009). His current interests focus on the interface of surgical robotics, in-situ sensing, and medical imaging, and to develop robotic-enabled medical imaging as well as image-guided robotic surgical systems, where ultrasound, photoacoustic sensing/imaging, and endoscopic OCT are three major modalities to be investigated and incorporated with minimally invasive surgical robotics.
	\end{IEEEbiography}

\end{document}